\documentclass[10pt,twocolumn,letterpaper]{article}

\usepackage{iccv}
\usepackage{times}
\usepackage{epsfig}
\usepackage{graphicx}
\usepackage{amsmath}
\usepackage{amssymb}
\usepackage{braket}
\usepackage{makecell}
\usepackage{lipsum}
\usepackage{subcaption}
\usepackage{multirow}
\usepackage{stackengine}
\usepackage{authblk}

\newcommand{\rulesep}{\unskip\ \vrule\ }

\makeatletter
\newcommand{\email}[1]{\def\@email{#1}}
\renewcommand\AB@affilsepx{, \protect\Affilfont}
\let\orig@author\@author
\renewcommand\@author{\orig@author \\ \vspace*{0.1em}\small\textit{\@email}}
\makeatother

\usepackage[pagebackref=true,breaklinks=true,letterpaper=true,colorlinks,bookmarks=false]{hyperref}
\usepackage[accsupp]{axessibility}  

\iccvfinalcopy 

\def\iccvPaperID{3149} 

\title{ME-PCN: Point Completion Conditioned on Mask Emptiness}

\author[1]{Bingchen Gong}
\author[2]{Yinyu Nie}
\author[3]{Yiqun Lin}
\author[3]{Xiaoguang Han\thanks{Corresponding author}}
\author[1]{Yizhou Yu\protect\footnotemark[1]}
\affil[1]{The University of Hong Kong}
\affil[2]{Technical University of Munich}
\affil[3]{SSE, CUHK(SZ)}
\email{bccs@connect.hku.hk, yinyu.nie@tum.de, lyq211003@gmail.com, hanxiaoguang@cuhk.edu.cn, yizhouy@acm.org}

\begin{document}
	
\maketitle
\ificcvfinal\thispagestyle{empty}\fi

\begin{abstract}
Point completion refers to completing the missing geometries of an object from incomplete observations. Main-stream methods predict the missing shapes by decoding a global feature learned from the input point cloud, which often leads to deficient results in preserving topology consistency and surface details. In this work, we present ME-PCN, a point completion network that leverages \textbf{emptiness} in 3D shape space. Given a single depth scan, previous methods often encode the occupied partial shapes while ignoring the empty regions (e.g. holes) in depth maps. In contrast, we argue that these `emptiness' clues indicate shape boundaries that can be used to improve topology representation and detail granularity on surfaces. Specifically, our ME-PCN encodes both the occupied point cloud and the neighboring `empty points'. It estimates coarse-grained but complete and reasonable surface points in the first stage, followed by a refinement stage to produce fine-grained surface details. Comprehensive experiments verify that our ME-PCN presents better qualitative and quantitative performance against the state-of-the-art. Besides, we further prove that our `emptiness' design is lightweight and easy to embed in existing methods, which shows consistent effectiveness in improving the CD and EMD scores.
\end{abstract}

\section{Introduction}

Capturing 3D data for objects around us is as easy as taking a picture with cell phones thanks to the popularity of
common 3D scanning sensors like LIDAR and depth cameras. Such availability has greatly enriched the practical applications in vision and robotics communities.

Different from image sensors, data from 3D scanners usually come incompletely with a much lower resolution, e.g., depth maps with missing values. To recover complete shapes from the partial inputs, volumetric and view-based projection methods leverage 3D convolutions to represent shapes into voxel grids. However, those 3D convolutions, suffering from expensive memory and computational cost, are bottlenecked by the resolution-computation balance. It becomes even disadvantaged when the inputs are unordered and sparse. Implicit methods learn a signed distance function (SDF) to represent shape surfaces and are capable of reaching any high resolution. However, they still rely on voxel grids, and an extra computation-intensive post-processing step is required to extract the final shape surface. In contrast, the point cloud is a more compact representation of 3D shapes. Compared with voxels, it is more scalable and computation-efficient to express shapes with different granularity. Several deep neural networks have been proposed to directly take advantage of point cloud representation, such as PCN~\cite{yuan2018pcn}, TopNet~\cite{tchapmi2019topnet}, MSN~\cite{liu2020morphing}.

\begin{figure}[t]
	\centering
	\begin{subfigure}[t]{0.115\textwidth}
		\includegraphics[width=\textwidth]
		{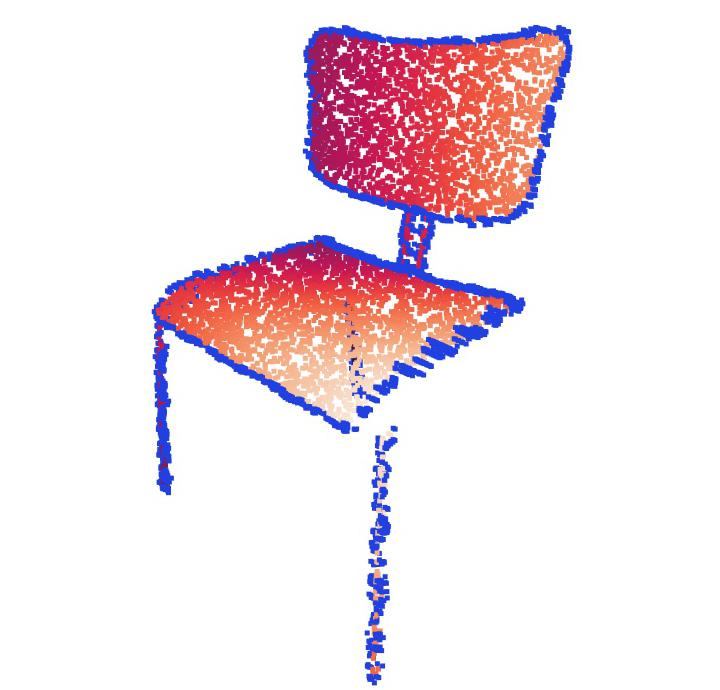}
		\includegraphics[width=\textwidth]
		{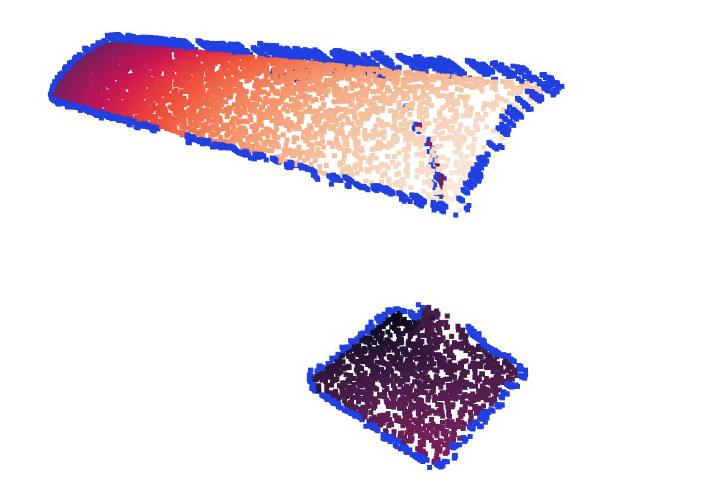}
		\includegraphics[width=\textwidth]
		{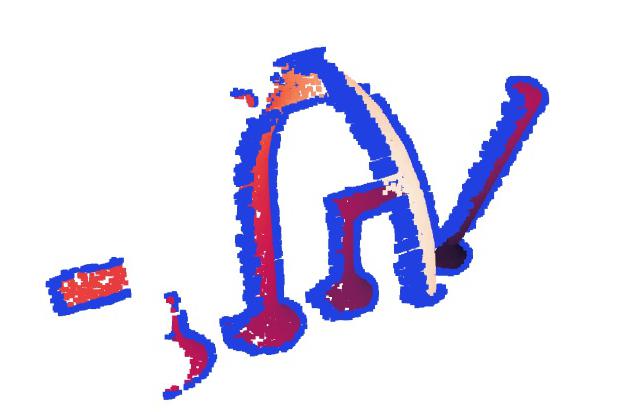}
		\caption{Input}
	\end{subfigure}
	\begin{subfigure}[t]{0.115\textwidth}
		\includegraphics[width=\textwidth]
		{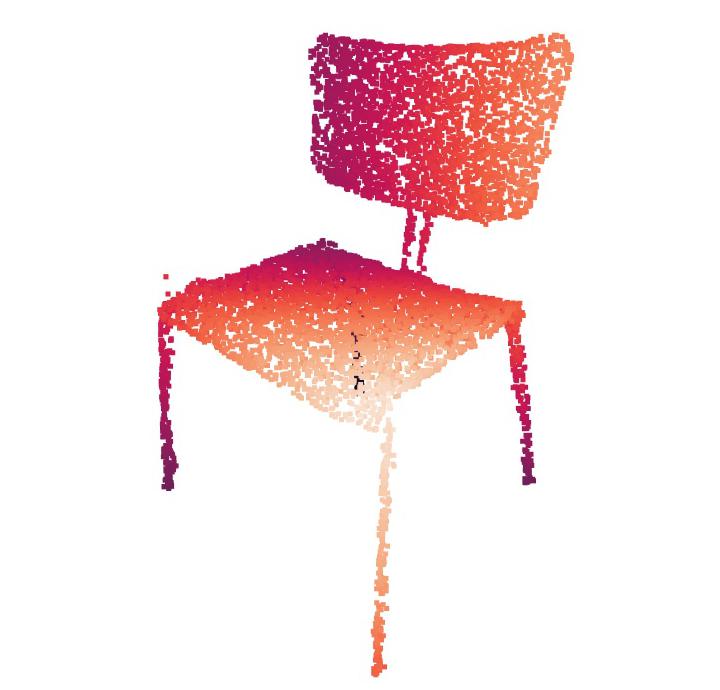}
		\includegraphics[width=\textwidth]
		{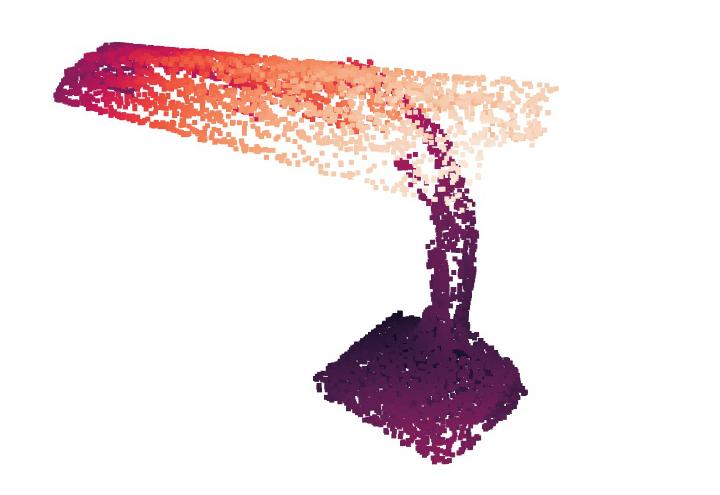}
		\includegraphics[width=\textwidth]
		{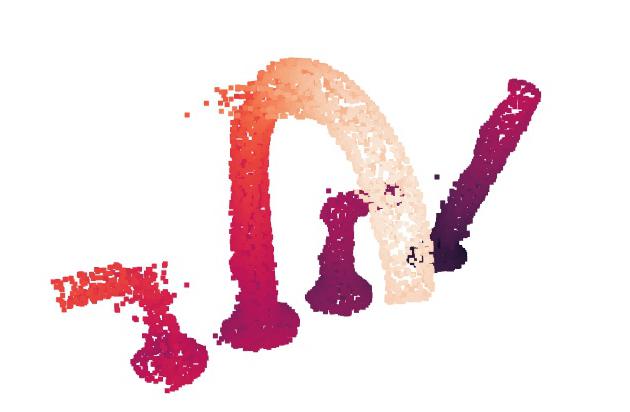}
		\caption{Ours}
	\end{subfigure}
	\begin{subfigure}[t]{0.115\textwidth}
		\includegraphics[width=\textwidth]
		{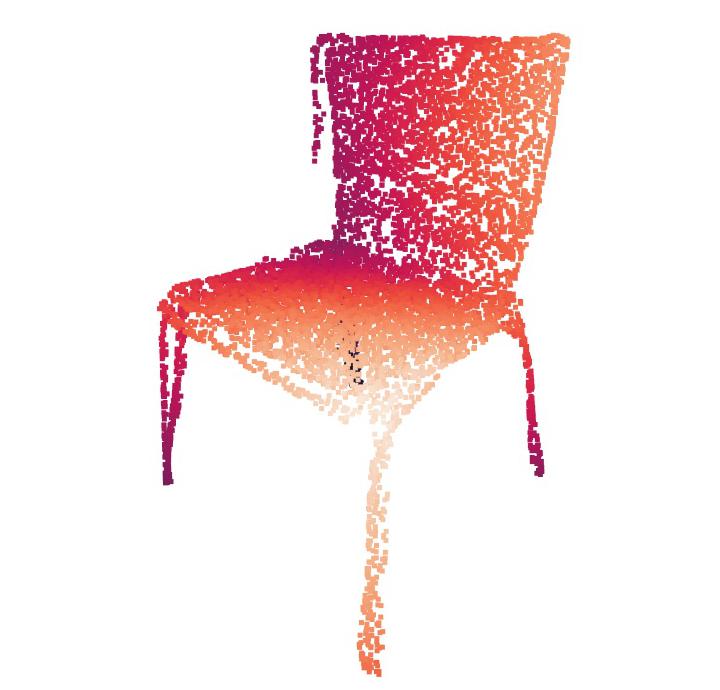}
		\includegraphics[width=\textwidth]
		{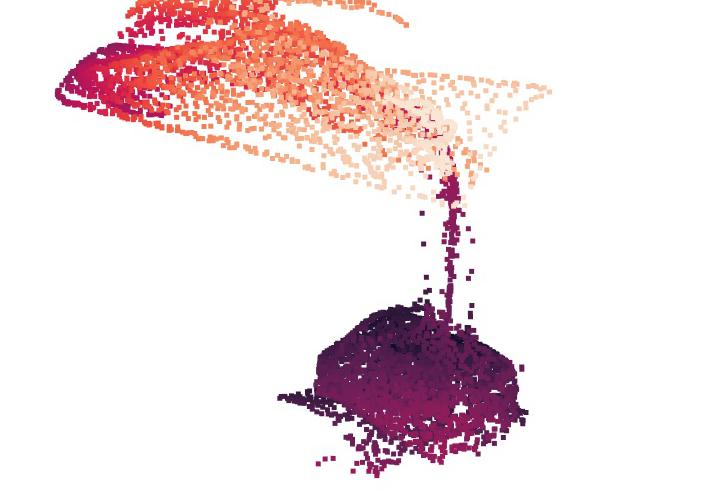}
		\includegraphics[width=\textwidth]
		{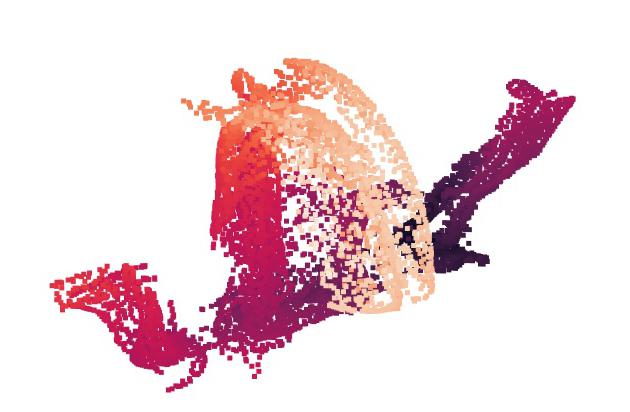}
		\caption{MSN}
	\end{subfigure}
	\begin{subfigure}[t]{0.115\textwidth}
		\includegraphics[width=\textwidth]
		{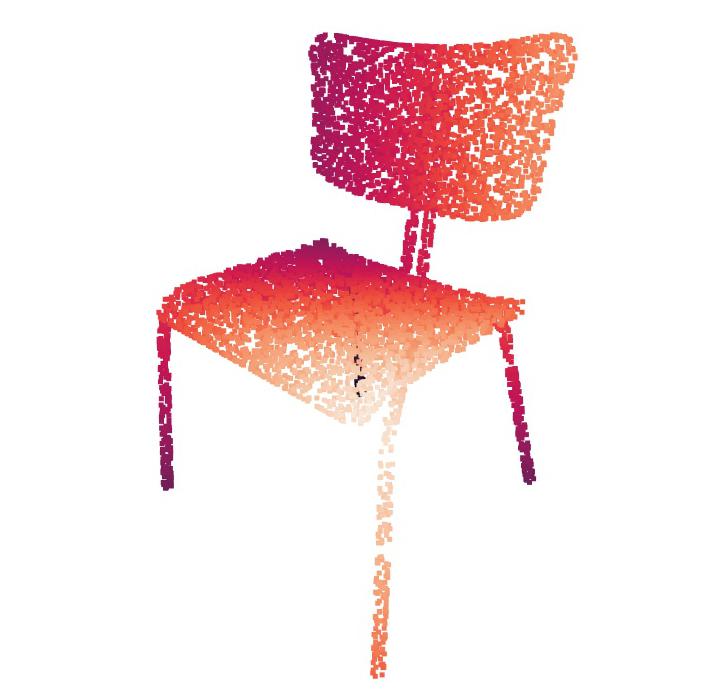}
		\includegraphics[width=\textwidth]
		{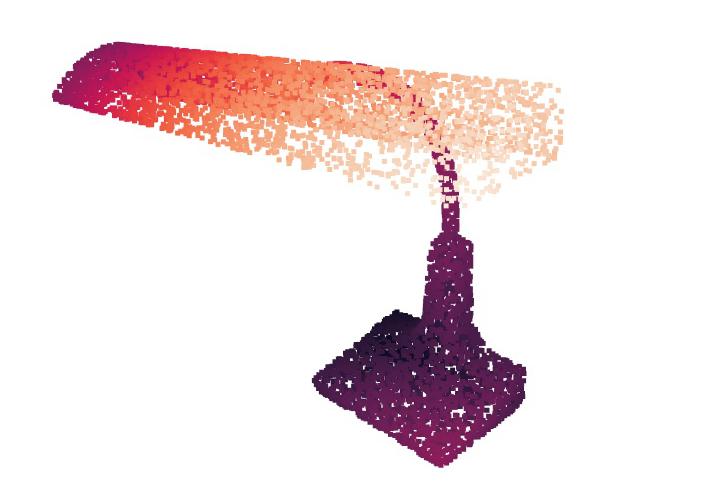}
		\includegraphics[width=\textwidth]
		{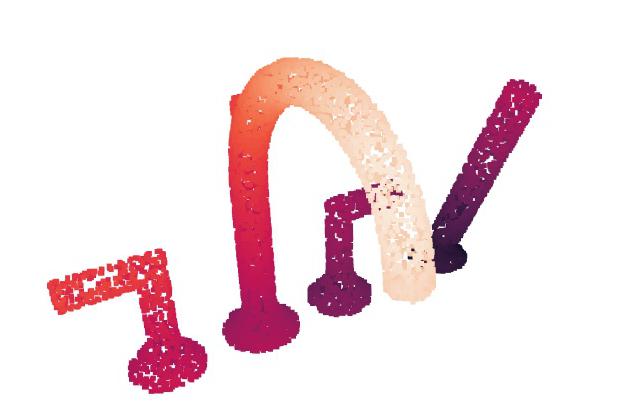}
		\caption{GT}
	\end{subfigure}
	\vspace{-10pt}
	\caption{Given a partial scan (a), our method encodes the spatial emptiness (blue points in (a)) neighboring to observations and predicts complete and topology consistent surfaces (b) compared with MSN \cite{liu2020morphing} on the ground-truth (d).}
	\label{fig:teaser}
	\vspace{-12pt}
\end{figure}


A key concept in existing point completion methods is an encoder-decoder architecture followed by an optional refining process, where the completed point cloud is generated from an encoded global feature. The widely used encoders for point cloud are PointNet~\cite{qi2017pointnet} and its variant PointNet++~\cite{qi2017pointnet++}. With the encoder-decoder design, current methods directly predict the complete points from the visible, occupied input points, while ignoring the unoccupied, empty regions in the inputs. In our view, the unoccupied regions are the complement of shape occupancy, thus also indicating the topology of 3D objects. Compared with learning from observable shapes, learning the emptiness presents extra significance, especially for complex shapes, such as non-convex surfaces with holes. In other words, the input scan tells not only the shape occupancy but also `where should not be occupied'. However, current methods predict shapes in the whole 3D space, which could be insensitive to subtle topologies. In our method, the emptiness in the input can inform our network `\textit{which regions should \textbf{not} be occupied}', and helps to keep consistent shape topology. Such emptiness information can be encoded by a mask given a viewpoint and can be easily obtained by thresholding the input depth scan or extracted from RGB images.


Inspired by the above, we introduce ME-PCN, a novel point completion network informed with mask emptiness. To encode the emptiness clues on a mask, 3D rays are radiated from the viewpoint towards the empty regions of the mask. All points along the rays will be encoded as empty points. In ME-PCN, only empty points that are in the neighborhood of visible points are processed by neural networks in addition to the input point cloud. Since visible points and empty points have totally different semantics, two separate networks are used to encode them into two global features.

For surface completion, directly decoding shapes from global features can predict plausible structures but usually results in coarse and over-smoothed surfaces \cite{liu2020morphing,yuan2018pcn}. It also neglects subtle structures on the boundary between visible and empty points. To this end, a final refining stage is performed after the coarse decoder. Local features are learned from neighboring empty and visible points for each coarse point, which augments the coarse input with on-surface details. The effectiveness of the emptiness inputs is verified both qualitatively and quantitatively. In summary, the main contributions of our work are:
\begin{itemize}
    \item We provide a novel encoding modality for point completion. Prior arts learn complete shapes only from visible points, while our method involves \textit{emptiness} learning to represent consistent shape topology and improves surface details.
    \item We propose ME-PCN to learn the shape emptiness from depth masks. Given a depth scan, 3D rays radiated from the viewpoint to empty regions on the mask are encoded to represent the emptiness in 3D shape space. It informs ME-PCN of the shape boundaries and improves the completion performance.
    \item Extensive experiments verify that our emptiness learning strategy can be easily embedded into modern point-based shape completion pipelines to improve the CD and EMD scores, which further makes our method outperform the state-of-the-art.

\end{itemize}


\begin{figure*}[t]
	\includegraphics[width=\textwidth]{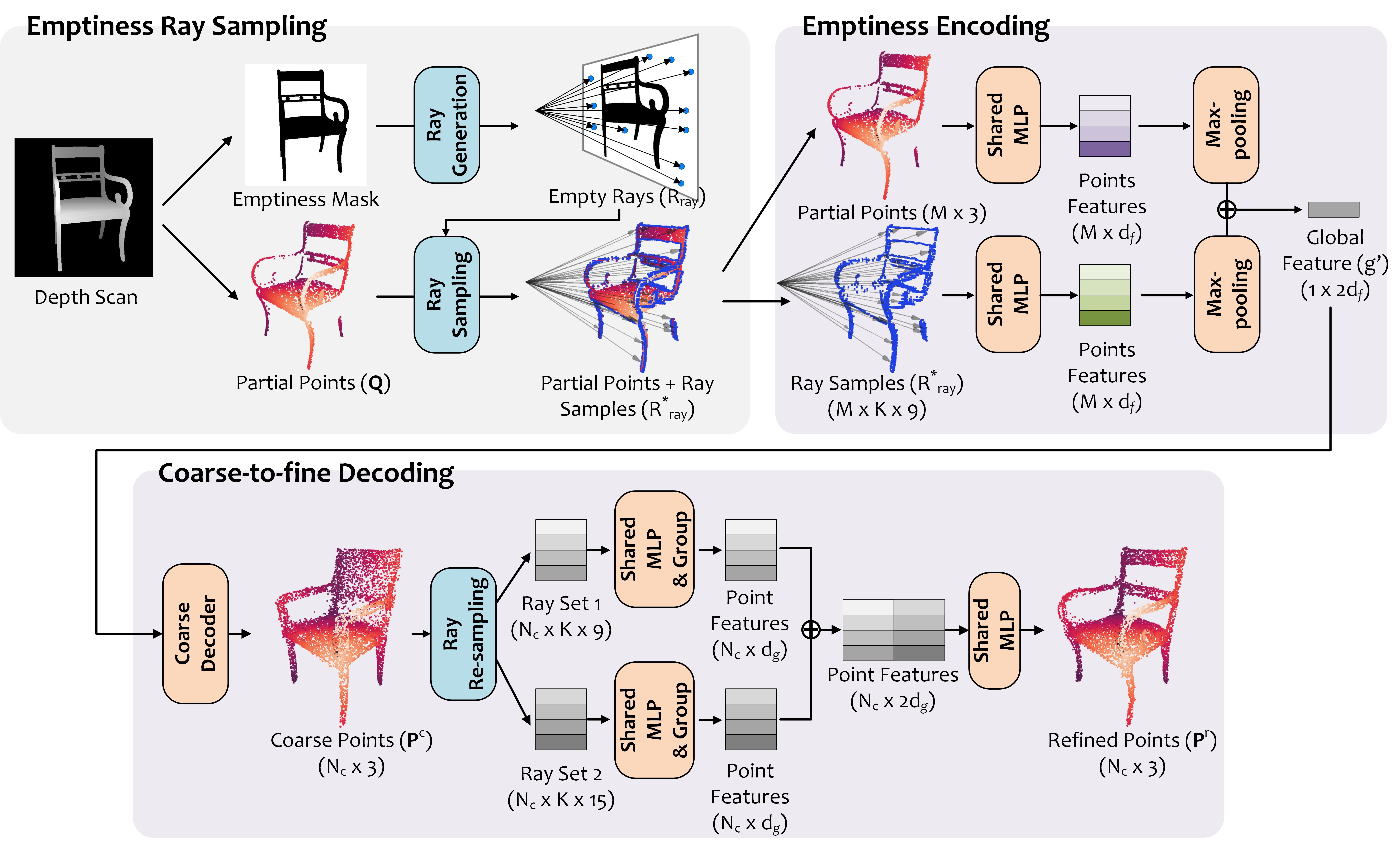}
	\caption{Architecture of ME-PCN. Given a partial scan, our network learns the spatial occupancy and emptiness from the observable points $\mathbf{Q}$ and 3D empty rays $\mathcal{R}_{ray}$ from the object mask. After sampling the neighboring empty rays $\mathcal{R}^{*}_{ray}$ queried by $\mathbf{Q}$, an \textit{Emptiness Encoder} is adopted to learn a global shape feature $g'$ by encoding the shape occupancy $\mathbf{Q}$ and spatial emptiness $\mathcal{R}^{*}_{ray}$ separately. We firstly predict a coarse shape $\mathbf{P}^{c}$ to obtain a rough structure. To recover surface details. A \textit{Ray Re-sampling} strategy is adopted to obtain two sets of empty rays from $\mathcal{R}_{ray}$ and $\mathcal{R}^{*}_{ray}$ respectively queried by $\mathbf{P}^{c}$. Two separate MLPs are used to respectively learn the point features before concatenation to predict the refined points $\mathbf{P}^{r}$.}
	\label{fig:architecture}
	\vspace{-5pt}
\end{figure*}

\section{Related Work}
\label{sec:rel}
\paragraph{Shape Representation and Reconstruction} Given an object observation (e.g., images, depth maps, point clouds), shape reconstruction aims to predict a plausible geometry and recover the shape surfaces. Early works extend the advantages of 2D convolutions in image perception to 3D, and adopt 3D convolutions to reconstruct shapes with discretized voxels \cite{firman2016structured,choy20163d,dai2017shape,han2017high,stutz2018learning,sharma2016vconv}. These works pioneered the 3D shape analysis modality but the expensive 3D convolutions make them bottlenecked by the resolution-efficiency problem, demanding an extra Octree to improve local details \cite{riegler2017octnet,tatarchenko2017octree,wang2018adaptive}. On this top, some works represent shapes with SDFs \cite{mescheder2019occupancy,chibane2020implicit,liao2018deep}, which theoretically can achieve any high resolution. However, the SDF methods still rely on voxel grids and require time-consuming post-processing to extract mesh surfaces. Besides, both the voxel and SDF methods can not well express shape boundaries thus leading to inferior surface details. Some other works directly generate surface meshes as the output \cite{wang2018pixel2mesh,groueix2018papier}. These methods approximate the target surfaces by deforming template meshes (e.g., planes or spheres), where the shape topology is usually restrained by the original templates. To this end, other works \cite{pan2019deep,nie2020total3dunderstanding} learn to modify the topology of template meshes, but it requires massive computations and often results in open boundaries. Besides, the kernel in the above methods is an encoder-decoder structure, where shapes are decoded or deformed from a global feature, decoding from which would be insensitive to boundary details and produce over-smoothed results. In our method, we leverage the emptiness information close to the shape boundary, and demonstrate its effectiveness in improving surface details.
\vspace{-5pt}
\paragraph{Shape Completion} Different from shape reconstruction, shape completion focuses on predicting a complete shape from a partial, observable surface. Similar to reconstruction methods, many works also adopt an encoder-decoder structure backboned by 3D CNNs or MLPs \cite{firman2016structured,dai2017shape,fan2017point,xie2020grnet} to represent shapes with voxels or points. Since such a structure cannot produce fine details, \cite{yuan2018pcn,liu2020morphing,xie2021style} adopt a coarse-to-fine completion strategy to firstly predict coarse points with MLPs and subsequently generate dense and refined results. Besides, \cite{wang2020cascaded, wen2020point} propose skip-connections or cascaded blocks to revisit shallow-end point features to complement surface details. \cite{huang2020pf} provides a multi-resolution encoder to perceive shape details under different granularity, and deploys a pyramid network to recover complete points by increasing the resolution. However, the key concept in these methods is how to improve point features to encode and decode more enriched shape signals. There are no explicit constraints to keep topology consistency with the target shape. On this point, \cite{nie2020skeleton} provides a skeleton-bridged method to predict surface points by first learning shape skeletons. However, the skeletal points of objects are extremely sparse. Any skeletal errors would directly influence the structure and surface quality. Besides, \cite{xie2021style} involves an extra adversarial point rendering by minimizing the depth map distance with the ground-truth under different views. In our work, we provide a lightweight approach to encode topological information by learning the `empty points' close to the observed input points. It informs our network that which regions are unoccupied thus helps to predict consistent sub-structures.
\section{Approach}
\label{sec:approach}

Our method consumes unordered ray sets of points as the input. A ray set is denoted as a set of 3D vector pairs $\mathcal{R}_{ray}=\{ (p_{i}, v_{i}) | i = 1, ..., N\}$, where
each start point $p_{i}\in\mathbf{R}^3$ is a vector of 3D coordinates. $v_{i}\in\mathbf{R}^3$ is the normalized vector of its ray direction from the viewpoint to $p_{i}$.

\subsection{Ray Generation from Masks}
Our method is illustrated in Figure~\ref{fig:architecture}. Given a depth map with the corresponding viewpoint, we define a 2D `emptiness mask' as the following:
\vspace{-3pt}
\begin{equation}
	mask_{s,t} =
	\begin{cases}
	1, & \text{if position}\ s, t\ \text{is empty}, \\
	0, & \text{otherwise}.
	\end{cases}
	\label{eq:01}
\end{equation}
A mask can be easily obtained by thresholding the depth map or extracted from the corresponding RGB image. Once we have the $mask$, an empty ray set $\mathcal{R}_{ray}$ can be calculated by back-projecting the $mask$ into 3D space:
\vspace{-3pt}
\begin{align}
	p_{i} &= \textit{Back-project}(s,t,d_{far}) \\
	l \cdot v_{i} &= p_{i} - \textit{Back-project}(s,t,d_{near})
\end{align}
$\textit{Back-project}$ is the reverse process of projection, which lifts a 2D image or depth map back to 3D space. Position $(s, t)$ is the $i$-th non-zero elements in $mask$, which indicates an empty pixel by Equation~\ref{eq:01}. $l$ is the normalization factor to ensure $v_{i}$ is a unit vector. $d_{far}$ and $d_{near}$ are the depth values of the farest plane and the nearest plane, respectively. Therefore, we let $\mathcal{R}_{ray}$ to represent the 3D empty points back-projected from $mask$. $d_{far}$ and $d_{near}$ are kept consistent among all shapes.




\subsection{Ray Points Sampling}
\label{sec:raysample}
Emptiness only has meanings when there are subjects in its neighborhood. It means that both the empty rays $\mathcal{R}_{ray}$ and visible points are not isolated. Our network should learn the local structures from the nearby empty points and the combinatorial interactions among local structures.

However, in $\mathcal{R}_{ray}$, many rays are actually too far from the subject and therefore convey little information. In this section, we provide a sampling strategy to obtain informative rays from $\mathcal{R}_{ray}$ as the input for ME-PCN. Those rays should be close to shape surfaces to convey local details. Specifically, given a visible point $q_{j}\in\mathbf{R}^3$ on the shape surface, we sample a subset of rays from $\mathcal{R}_{ray}$ in the neighborhood of $q_{j}$, where we choose $K$ nearest rays for each $q_{j}$. On each neighboring ray $\mathbf{r}_{k}\in\mathcal{R}_{ray}$, we select the nearest point as an empty point candidate $p^{e}_{k}\in\mathbf{R}^3$. Thus each visible point $q_{j}$ has $K$ candidates of empty points $\{p^{e}_{k}\}$, $k=1,2,...,K$. The Euclidean distance $\|D_{\mathbf{r},q}\|$ between a ray $\mathbf{r}=(p, v)\in\mathcal{R}_{ray}$ and a visible point $q$ is defined by:
%
%
\vspace{-10pt}
\begin{align}
	p^{e} &= p-\left[(p-q) \cdot v\right] v, \\
	D_{\mathbf{r},q} &= p^{e} - q,
	\label{eq:ray_distance}
\end{align}
where $p^{e}$ is the nearest point from ray $\mathbf{r}$ to visible point $q$. $D_{\mathbf{r},q}\in\mathbf{R}^{3}$ is the offset vector from $q$ to $p^{e}$.

After sampling, for each visible point $q$, we combine its $K$ nearest empty points $\{p^{e}_{k}\}\in\mathbf{R}^{K,3}$ with the corresponding ray direction vectors $\{v_{k}\}\in\mathbf{R}^{K,3}$ and the offset vectors $\{D_{k}\}\in\mathbf{R}^{K,3}$.
Denote that there are $M$ visible points $\mathbf{Q}\in\mathbf{R}^{M,3}$, we input our network with the processed rays:
\vspace{-3pt}
\begin{equation}
   \mathcal{R}^{*}_{ray}=\{\{p^{e}_{k}\}, \{D_{k}\},\{v_{k}\}\}\in\mathbf{R}^{M\times K\times9}.
   \label{eq:sampled_rays}
\end{equation}
Both $p^{e}_{k}$ and $D_{k}$ indicate the spatial neighborhood to visible point $q$ in Euclidean space. This could provide explicit cues for our network to capture local structures from nearby rays.

\subsection{Emptiness Encoding}
\label{sec:globalfeat}
We illustrate the architecture in Figure~\ref{fig:architecture}. Our approach takes a partial point
cloud $\mathbf{Q}$ and sampled rays $\mathcal{R}^{*}_{ray}$ as inputs and encodes them into a global feature vector (GFV) with emptiness semantics, which will be used to predict complete point cloud with a coarse-to-fine strategy.

In the encoding stage, since visible points $\mathbf{Q}$ and empty points in $\mathcal{R}^{*}_{ray}$ have totally different semantics with non-identical scale, we use two separate networks to encode them into two global feature vectors, respectively.

The encoder part consists of two Feature Encoding (FE) layers to respectively process visible points $\mathbf{Q}$ and sampled rays $\mathcal{R}^{*}_{ray}$. The first FE layer consumes the coordinates of visible points $\mathbf{Q}$ as the input. A shared multi-layer perceptron (MLP) consisting of two linear layers with
ReLU activation is used to transform points $\{q_{i}|q_{i}\in\mathbf{Q}\}$ into point features $\{f_{i}\}\in\mathbf{R}^{M,d_{f}}$. The second FE layer takes the sampled rays $\mathcal{R}^{*}_{ray}$ as the input and produces point features $\{g_{i}\}\in\mathbf{R}^{N,d_{f}}$, similar to the first FE layer.

The two FE layers output two feature matrices $F=\{f_{i}\}$, $G=\{g_{i}\}$. A point-wise
max-pooling is respectively performed on $F,G$ to obtain $d_{f}$-dimensional
global features $f$ and $g$. Lastly, $f$ and $g$ are concatenated together to form a single global feature vector $g'=[f,g]\in\mathbf{R}^{2d_{f}}$.




\subsection{Coarse-to-Fine Decoding by Ray Resampling}

From the global feature $g'$, we can directly decode the complete point cloud that captures the overall shape following \cite{yuan2018pcn, liu2020morphing}. However, as discussed in Section~\ref{sec:rel}, decoding shapes merely from a global feature would neglect local details and result in over-smoothed structures. To this end, a refining stage that operates on generated coarse points is usually preferred. In this part, we build our decoder with a coarse-to-fine strategy. A coarse decoder from \cite{liu2020morphing} is adopted to firstly predict a coarse-grained but structure completed points $\mathbf{P}^{c}\in\mathbf{R}^{N_{c}\times 3}$. $N_{c}$ is the number of coarse points. However, unlike \cite{yuan2018pcn,wang2020cascaded} where the local features are complemented by skip-connecting the shallow-end layer responses, we provide an explicit approach to revisit the emptiness information and decode fine-grained surface points. Our design is based on insights: 1) coarse points decoded from a global feature are still not accurate to preserve a consistent boundary compared to the ground-truth due to its roughness; 2) the point information in $\mathcal{R}^{*}_{ray}$ conveys the surface clues that can improve shape detail recovery.

\begin{figure}[!b]
	\centering
	\begin{subfigure}[t]{0.23\textwidth}
		\includegraphics[height=0.15\textheight]
		{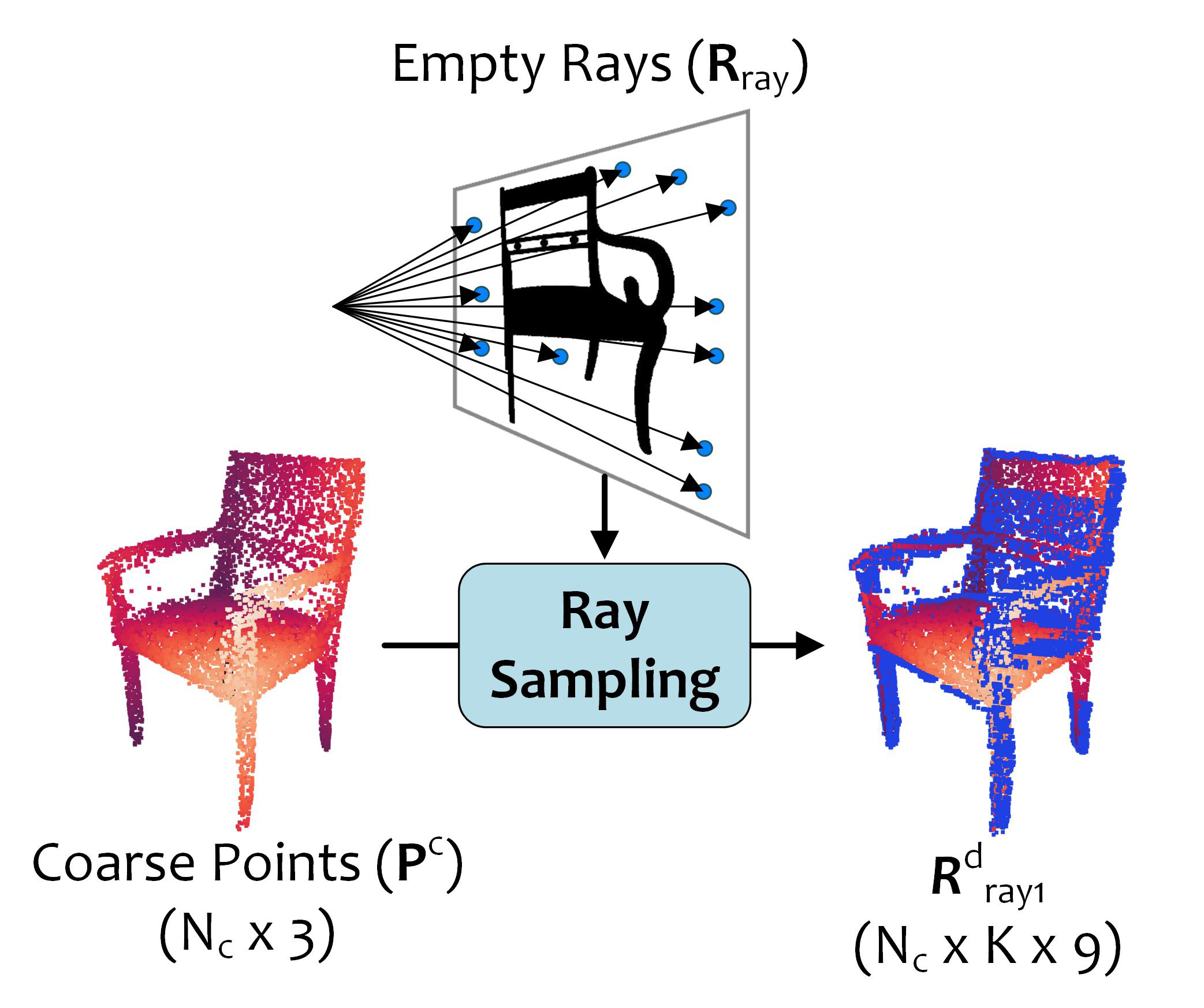}
	\end{subfigure}
	\rulesep
	\begin{subfigure}[t]{0.23\textwidth}
		\includegraphics[height=0.15\textheight]
		{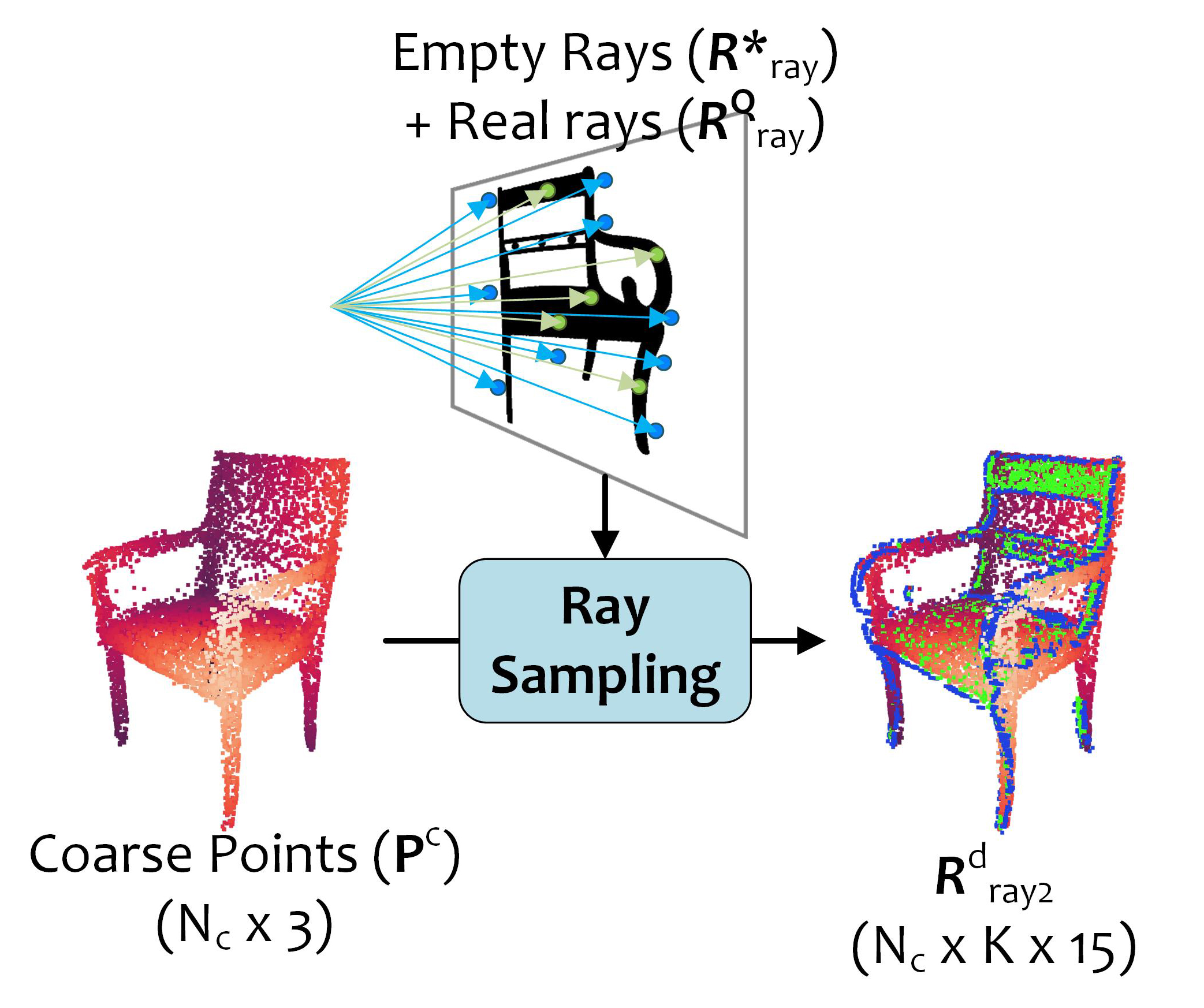}
	\end{subfigure}
	\caption{Resampling rays queried by coarse points $\mathbf{P}^{c}$ into $\mathcal{R}^{d}_{ray1}$ (left) and $\mathcal{R}^{d}_{ray2}$ (right). Blue points denote the sampled rays from empty rays $\mathcal{R}_{ray}$ (left) or $\mathcal{R}^{*}_{ray}$ (right). Green points represent the rays sampled from visible points. Blue points imply whether coarse points are in empty regions. Green points reveal the position of visible points.}
	\label{fig:resampling}
	\vspace{-5pt}
\end{figure}

For the first part, we inform our decoder with the emptiness information (i.e., `emptiness mask' in Equation~\ref{eq:01}). It tells our decoder `\textit{whether the coarse points are in empty regions}'. For the second part, we inform our decoder with the shape information. It tells our decoder \textit{'what the real surface looks like'}. To realize these, given the coarse points set $\mathbf{P}^{c}$, we respectively resample two sets of empty rays $\mathcal{R}^{d}_{ray1}$ and $\mathcal{R}^{d}_{ray2}$ as the input for the surface refining decoder, which is illustrated in Figure~\ref{fig:resampling}.
\vspace{-5pt}
\paragraph{Sampling $\mathcal{R}^{d}_{ray1}$:} We sample $\mathcal{R}^{d}_{ray1}$ with the same method of sampling $\mathcal{R}^{*}_{ray}$ (see Section \ref{sec:raysample}).
The only difference is that the visible points $\mathbf{Q}$ in Section~\ref{sec:raysample} is replaced with the coarse points $\mathbf{P}^{c}$. Then the sampled empty rays $\mathcal{R}^{d}_{ray1}\in\mathbf{R}^{N_{c}\times K\times 9}$ will tell whether a coarse point in $\mathbf{P}^{c}$ is in an `empty' region.
\vspace{-5pt}
\paragraph{Sampling $\mathcal{R}^{d}_{ray2}$:} For $\mathcal{R}^{d}_{ray2}$, we adopt the same sampling method and use coarse points $\mathbf{P}^{c}$ as the query to collect neighboring rays into $\mathcal{R}^{d}_{ray2}$.
The difference is that, we no longer sample rays from $\mathcal{R}_{ray}$ but from the union of its subset $\mathcal{R}^{*}_{ray}$ and the ray set $\mathcal{R}^{Q}_{ray}$ (generated from visible points $\mathbf{Q}$).
$\mathcal{R}^{*}_{ray}$ presents the boundary rays in $\mathcal{R}_{ray}$ (see Section~\ref{sec:raysample}). With the definition in Equation~\ref{eq:sampled_rays}, the rays generated from $\mathbf{Q}$ are defined by:
\vspace{-3pt}
\begin{align}
	\mathcal{R}^{Q}_{ray} &= \{(q_{i}, \vec{0}, v_{i})|i=1,...,M\} \\
	l \cdot v_{i} &= q_{i} - \textit{Back-project}(s,t,d_{near}),
\end{align}
where $q_{i}$ is the 3D coordinates of a visible point in $\mathbf{Q}$. $(s,t)$ is its projection on the depth map. $v_{i}$ is the unit directional vector from camera viewpoint to $q_{i}$. Since $q_{i}$ is a visible point, so the offset between ray $(q_{i}, v_{i})$ and a visible point is $\vec{0}$ (see Equation~\ref{eq:ray_distance}). $l$ is a factor to ensure $v_{i}$ is a unit vector. Note that to satisfy the assumption: all points on rays are empty points. We only consider those rays $\{(q_{i}, \vec{0}, v_{i})\}\subset\mathcal{R}^{Q}_{ray}$ whose depth values at $\{(s,t)\}$ is
larger than the corresponding depths of coarse points $\mathbf{P}^{c}$ on the image plane.

Note that we sample $\mathcal{R}^{d}_{ray2}$ from $\{\mathcal{R}^{*}_{ray}, \mathcal{R}^{Q}_{ray}\}$. Following Section~\ref{sec:raysample}, for each point $p^{c}\in\mathbf{P}^{c}$ we sample $K$ nearest rays and concatenate the empty points $\{p^{c,e}_{k}\}$, offset vectors $\{D^{c}_{k}\}$ with the corresponding empty ray informtion $\{\mathbf{r}^{c}\}\subset\mathcal{R}^{*}_{ray}\cup\mathcal{R}^{Q}_{ray}$. Therefore, $\mathcal{R}^{d}_{ray2}=\{\{p^{c,e}_{k}\}, \{D^{c}_{k}\},\{\mathbf{r}^{c}\}\} \in \mathbf{R}^{N_{c}\times K\times 15}$
\vspace{-5pt}
\paragraph{Decoding Refined Shape:}
Rays in $\mathcal{R}^{d}_{ray1}$ represent empty space neighboring to coarse points, while $\mathcal{R}^{d}_{ray2}$ informs the coarse points with the real shape boundary. Two FE layers are respectively used to encode $\mathcal{R}^{d}_{ray1}$ and $\mathcal{R}^{d}_{ray2}$.
For $\mathcal{R}^{d}_{ray1}$, a shared MLP consisting of two linear layers with ReLU activation are used to transform points in $\mathcal{R}^{d}_{ray1}\in\mathbf{R}^{N_{c}\times K\times 9}$ into a grouped point feature vector $\{f^{d}_{i}\}\in\mathbf{R}^{N_{c}\times d_{g}}$. The second FE layer consumes $\mathcal{R}^{d}_{ray2}$ and produce a grouped point feature vector $\{g^{d}_{i}\}\in\mathbf{R}^{N_{c}\times d_{g}}$, similar to the first FE layer. The grouping operation is shown as ($\{f^{d}_{i}\}$ is taken as an example):
\vspace{-3pt}
\begin{equation}
	f^{d}_{i} = \sum_{k=1}^{K} \sum_{d=1}^{d_{g}} w(k,d)\mathbf{r}(i, k,d), \mathbf{r} \in\mathcal{R}^{d}_{ray1},
\end{equation}
where $\{w\}$ are the weights calculated following \cite{wu2019pointconv}. We concatenate $\{f^{d}_{i}\}$ and $\{g^{d}_{i}\}$ to regress the coordinates of complete surface points as the refined output $\mathbf{P}^{r}$.


\subsection{Loss Function}
We design our loss function via the Earth Mover's Distance (EMD) and
the regularizer $\mathcal{L}_{expansion}$ from \cite{liu2020morphing}:
\vspace{-3pt}
\begin{equation}
\begin{split}
	\mathcal{L} = & \mathrm{EMD}(\mathbf{P}^{c}, \mathbf{P}^{gt}) + \lambda_{1} \cdot \mathcal{L}_{expansion} \\
	& + \lambda_{2} \cdot \mathrm{EMD}(\mathbf{P}^{r}, \mathbf{P}^{gt}).
\end{split}
\end{equation}
The EMD measures the distance from the coarse prediction $\mathbf{P}^{c}$ (or the refined prediction $\mathbf{P}^{r}$) to the ground-truth surface points $\mathbf{P}^{gt}$.
The regularizer ensures point patches in $\mathbf{P}^{c}$ fit in local areas and not overlap too much. $\lambda_{1}, \lambda_{2}$ are two weights of importance that balance different losses.

\section{Experiment and Evaluation}


\noindent\textbf{Network Specifications.} Our network architecture is illustrated in Figure~\ref{fig:architecture}, where the coarse decoder is adopted following \cite{liu2020morphing}. The parameters and layer information in Section~\ref{sec:approach} are detailed in the supp. material.

\noindent\textbf{Dataset.} We evaluate our methods on a subset of the ShapeNet dataset. 14 categories that contain a large number of models are selected where 29,795 CAD models are included. We report our evaluation on six categories: faucet, cabinet, table, chair, vase and lamp. The others are included in the supp. material. The complete point clouds are created by uniformly sampling $n_{gt} = 8192$ points on the mesh surfaces and the partial point clouds are generated by back-projecting 2.5D depth images into 3D. All CAD models are normalized into $[-1,1]^{3}$ and located at the origin. For each category, we sample 9,000 pairs of partial and complete point clouds from different models using random viewpoint, resulting in 9,000$\times$14 pairs of point clouds, where $10\%$ of them for testing, and the rest are for training.

\noindent\textbf{Benchmark.} To validate our performance, we extensively compare our method with the state-of-the-art including PCN \cite{yuan2018pcn}, PF-Net \cite{huang2020pf}, P2P-Net \cite{yin2018p2p}, SoftPoolNet\cite{wang2020softpoolnet}, CRN \cite{wang2020cascaded}, GR-Net \cite{xie2020grnet}, MSN \cite{liu2020morphing} and SK-PCN \cite{nie2020skeleton}.
All methods are inputted with 5,000 points. Two resolutions of output points (2048 and 8192) are compared considering some methods support upsampling while the others do not. Besides, we also embed our emptiness learning into PCN (i.e. PCN+ray) and MSN (i.e. MSN+ray) to verify its effectiveness to different backbones. The network details of PCN+ray are illustrated in the supp. material.


\noindent\textbf{Model Training.} We trained all models on 2$\times$NVIDIA GeForce RTX 2080 Ti GPUs for 25 epochs with a batch size of 16. The initial learning rate is $10^{-3}$ and is decayed by 0.1 per 10 epochs. Adam is used as the optimizer.

\begin{figure*}[t]
\centering
\renewcommand{\arraystretch}{0}
\renewcommand{\arrayrulewidth}{0pt}%
\setlength{\tabcolsep}{2pt}
\begin{tabular}{ccccccccccc}
%
Input & \begin{subfigure}{0.07\textwidth}\centering\includegraphics[trim=300 150 200 150,clip,width=\textwidth]{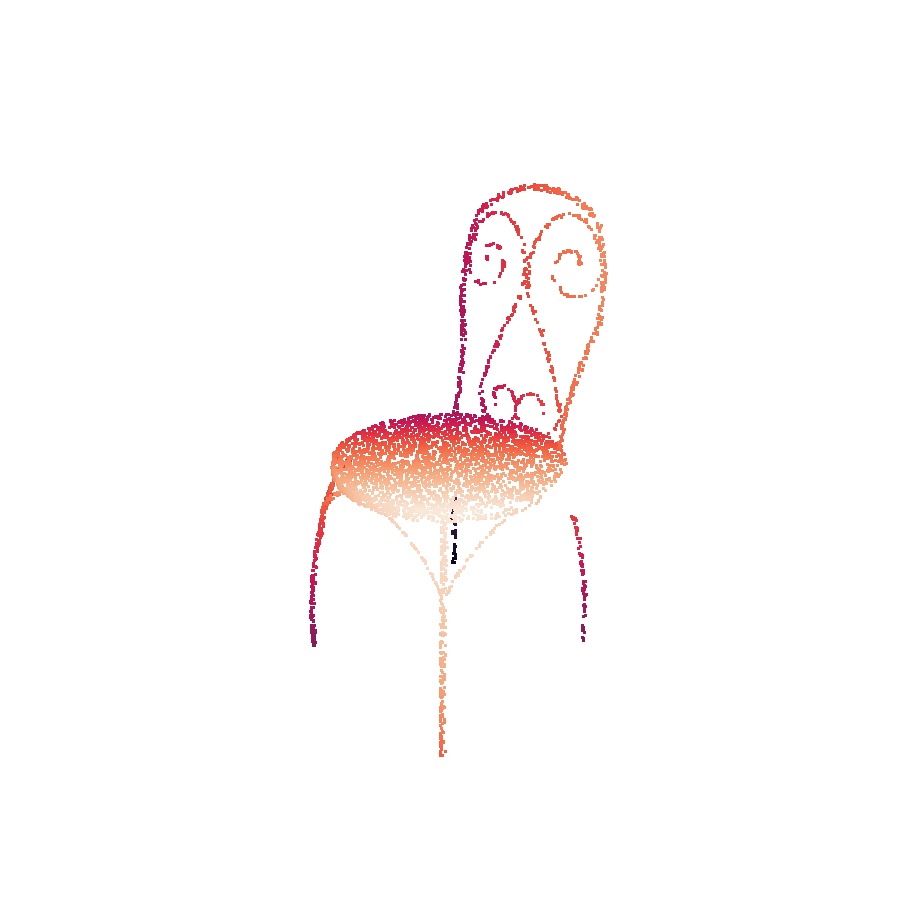}\end{subfigure} & \begin{subfigure}{0.077\textwidth}\centering\includegraphics[trim=250 150 200 150,clip,width=\textwidth]{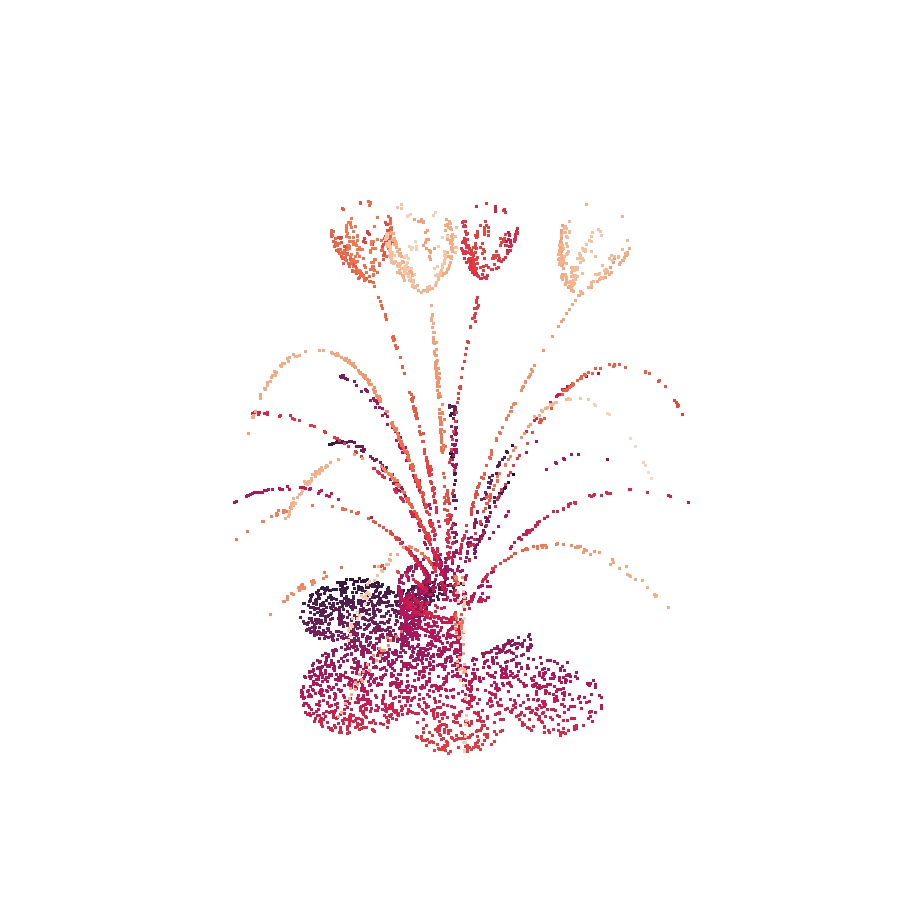}\end{subfigure} & \begin{subfigure}{0.071\textwidth}\centering\includegraphics[trim=300 150 190 140,clip,width=\textwidth]{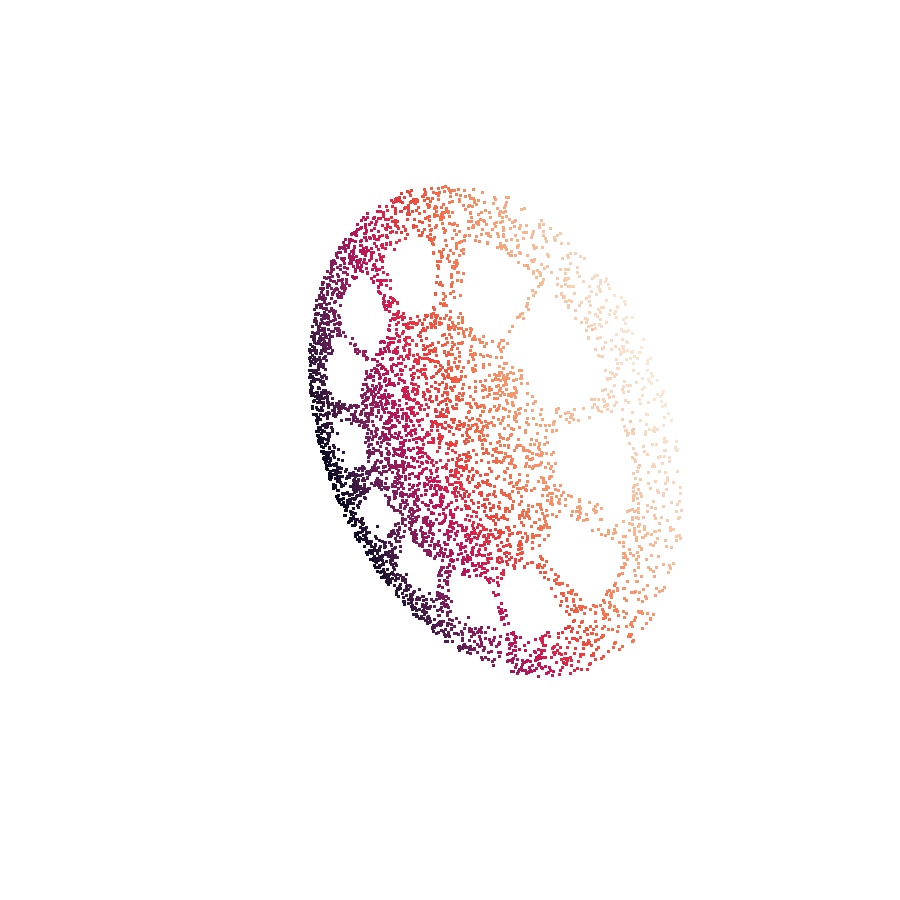}\end{subfigure} & \begin{subfigure}{0.074\textwidth}\centering\includegraphics[trim=220 170 230 110,clip,width=\textwidth]{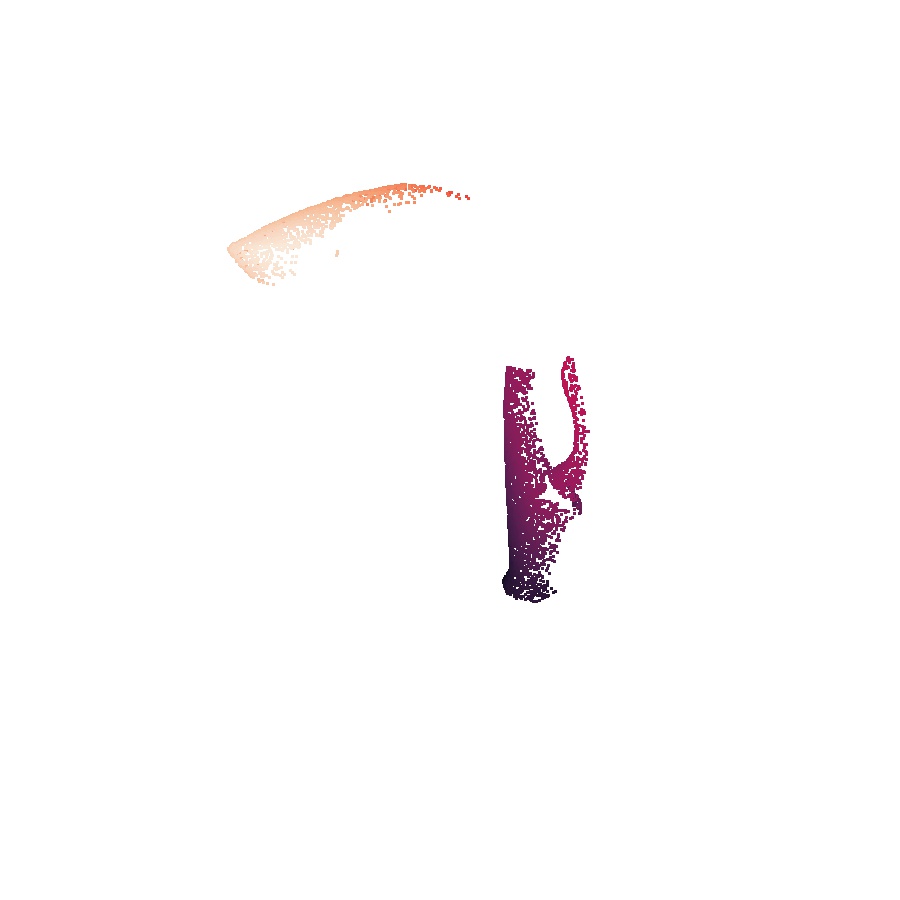}\end{subfigure} & \begin{subfigure}{0.12\textwidth}\centering\includegraphics[trim=270 280 200 370,clip,width=\textwidth]{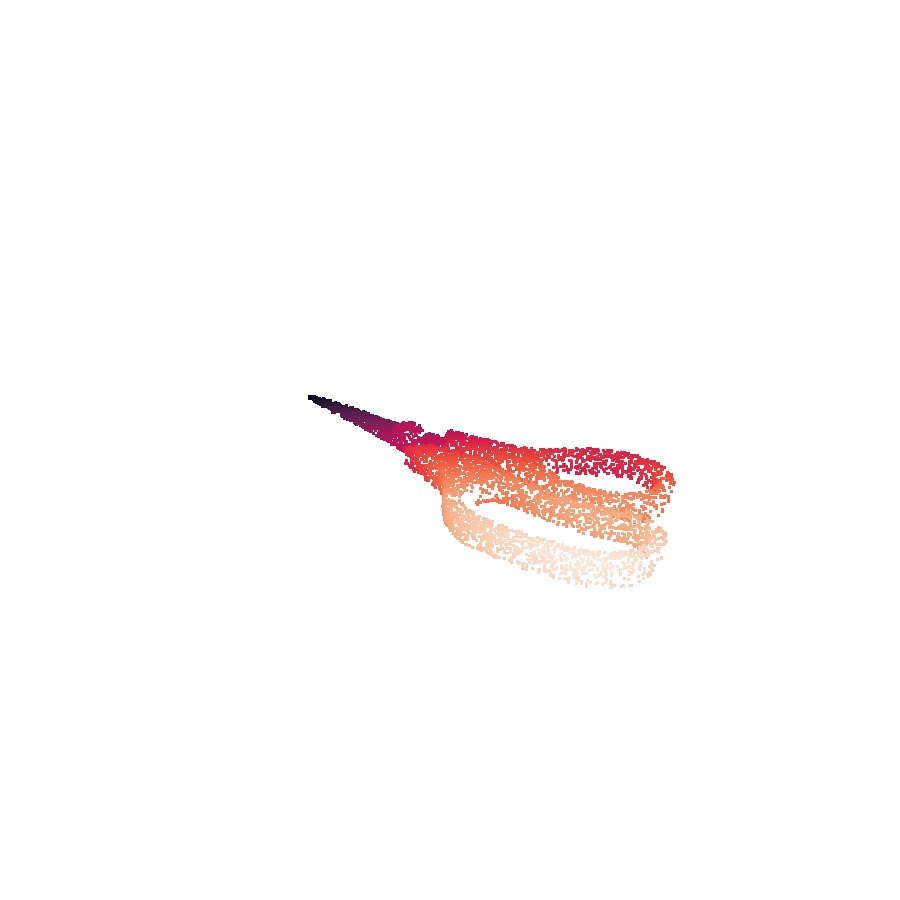}\end{subfigure} & \begin{subfigure}{0.082\textwidth}\centering\includegraphics[trim=150 200 250 250,clip,width=\textwidth]{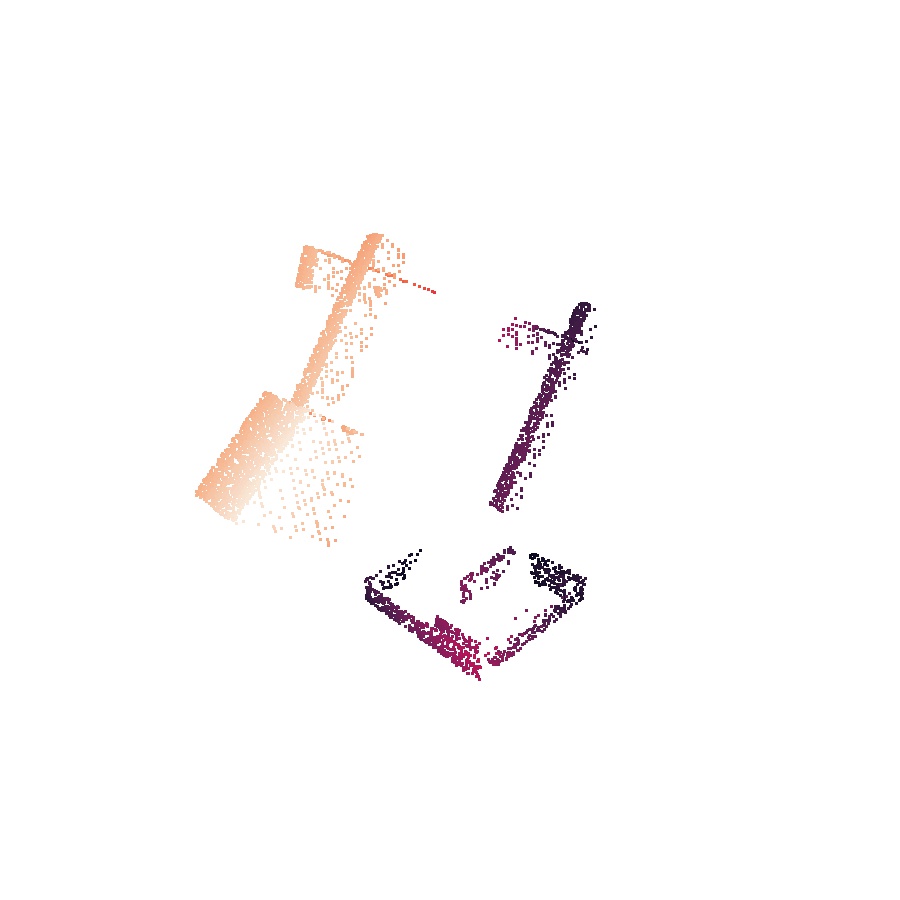}\end{subfigure} & \begin{subfigure}{0.05\textwidth}\centering\includegraphics[trim=300 200 300 200,clip,width=\textwidth]{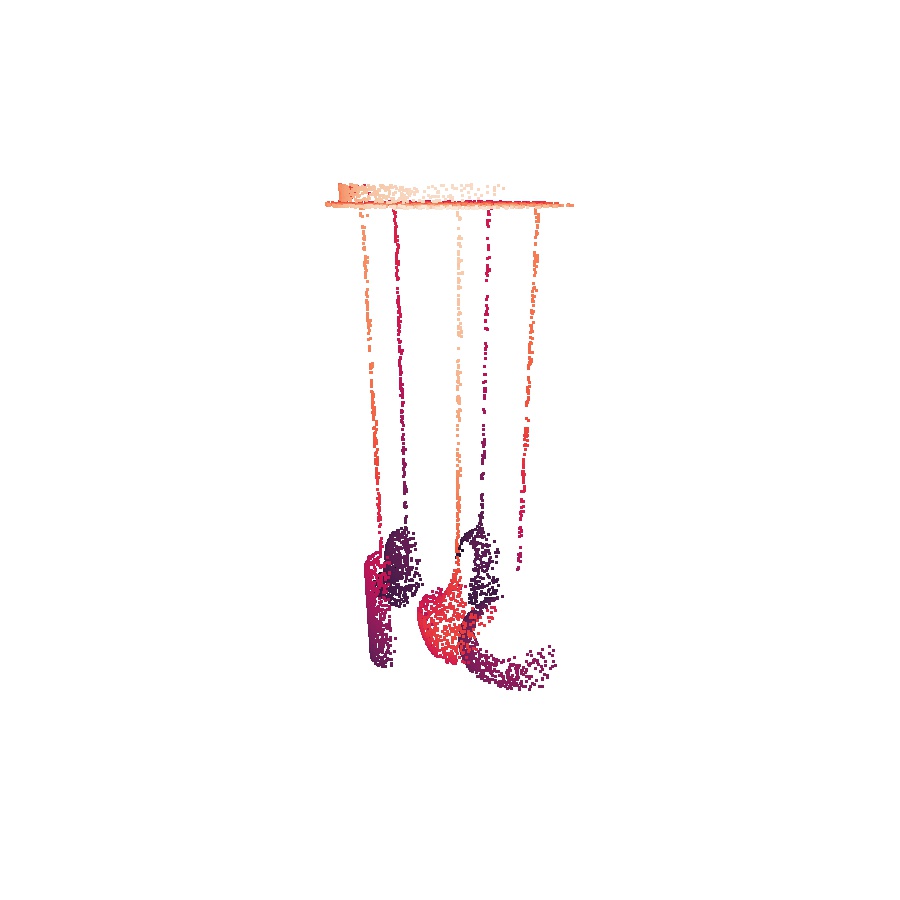}\end{subfigure} & \begin{subfigure}{0.095\textwidth}\centering\includegraphics[trim=200 250 200 250,clip,width=\textwidth]{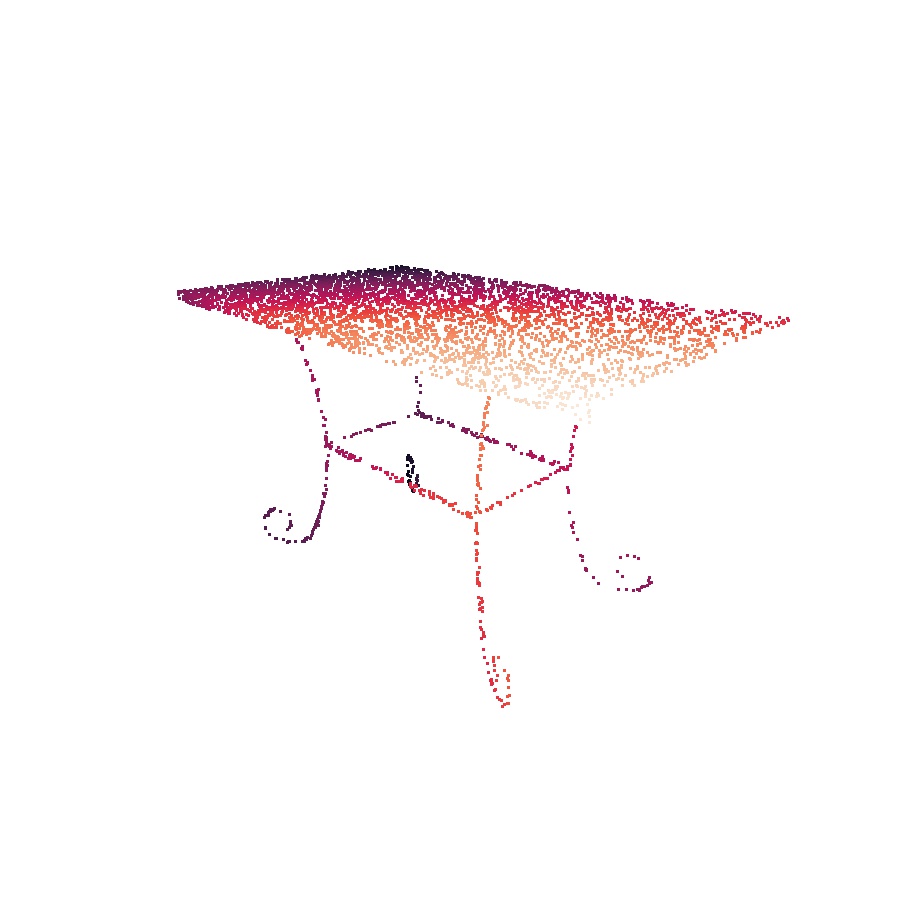}\end{subfigure} & \begin{subfigure}{0.086\textwidth}\centering\includegraphics[trim=100 150 200 100,clip,width=\textwidth]{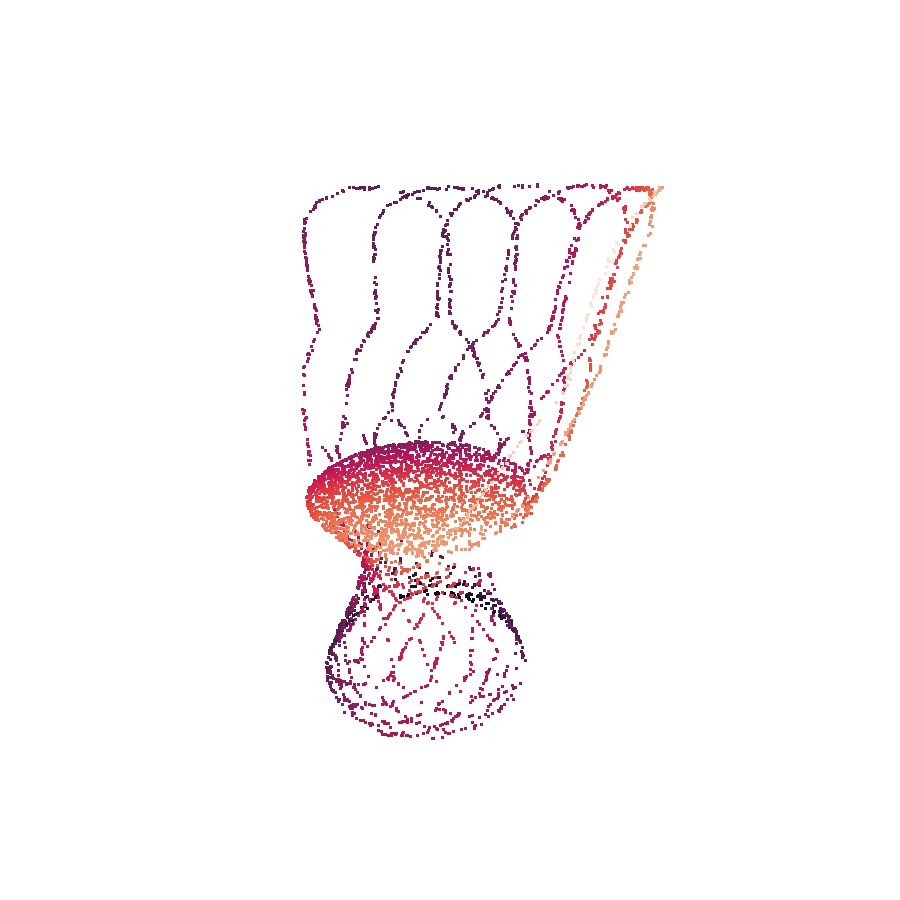}\end{subfigure} & \begin{subfigure}{0.11\textwidth}\centering\includegraphics[trim=200 300 100 250,clip,width=\textwidth]{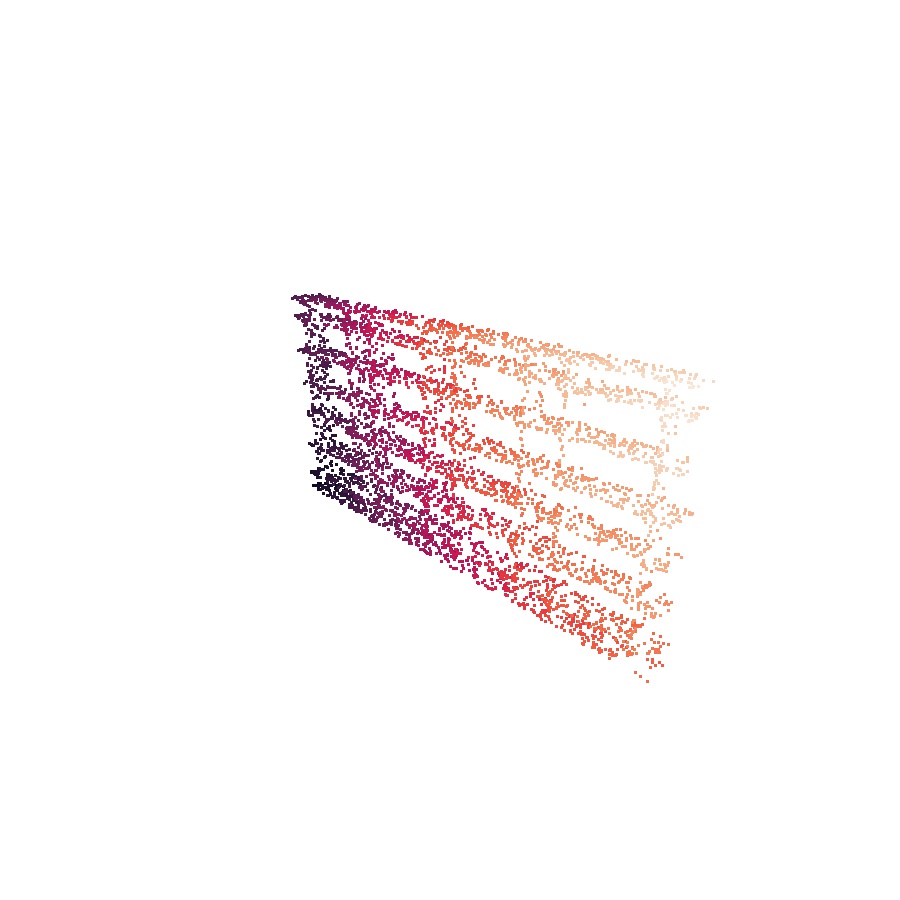}\end{subfigure} \\
PCN & \begin{subfigure}{0.07\textwidth}\centering\includegraphics[trim=300 150 200 150,clip,width=\textwidth]{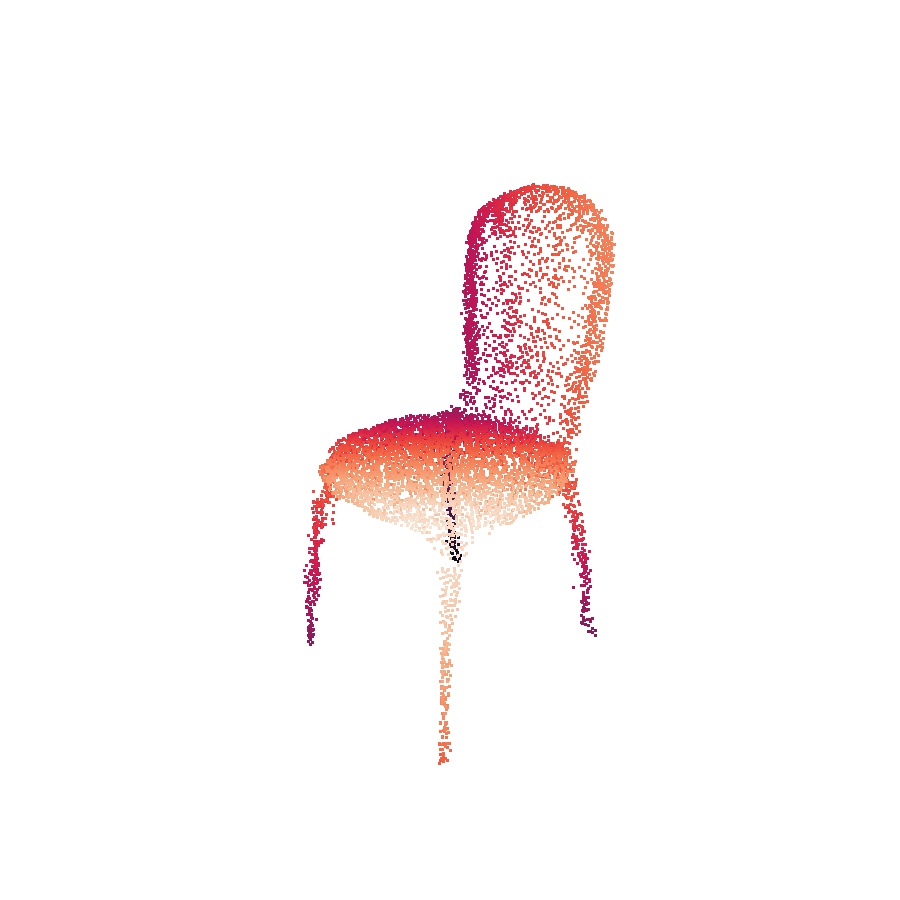}\end{subfigure} & \begin{subfigure}{0.077\textwidth}\centering\includegraphics[trim=250 150 200 150,clip,width=\textwidth]{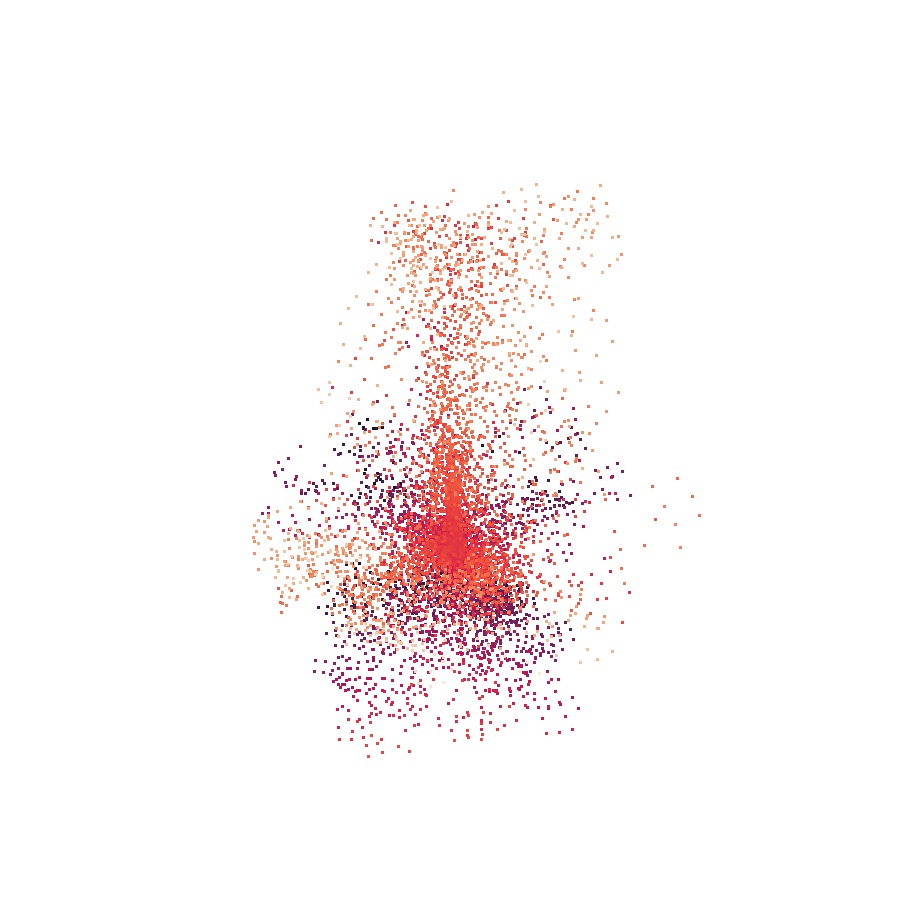}\end{subfigure} & \begin{subfigure}{0.071\textwidth}\centering\includegraphics[trim=300 150 190 140,clip,width=\textwidth]{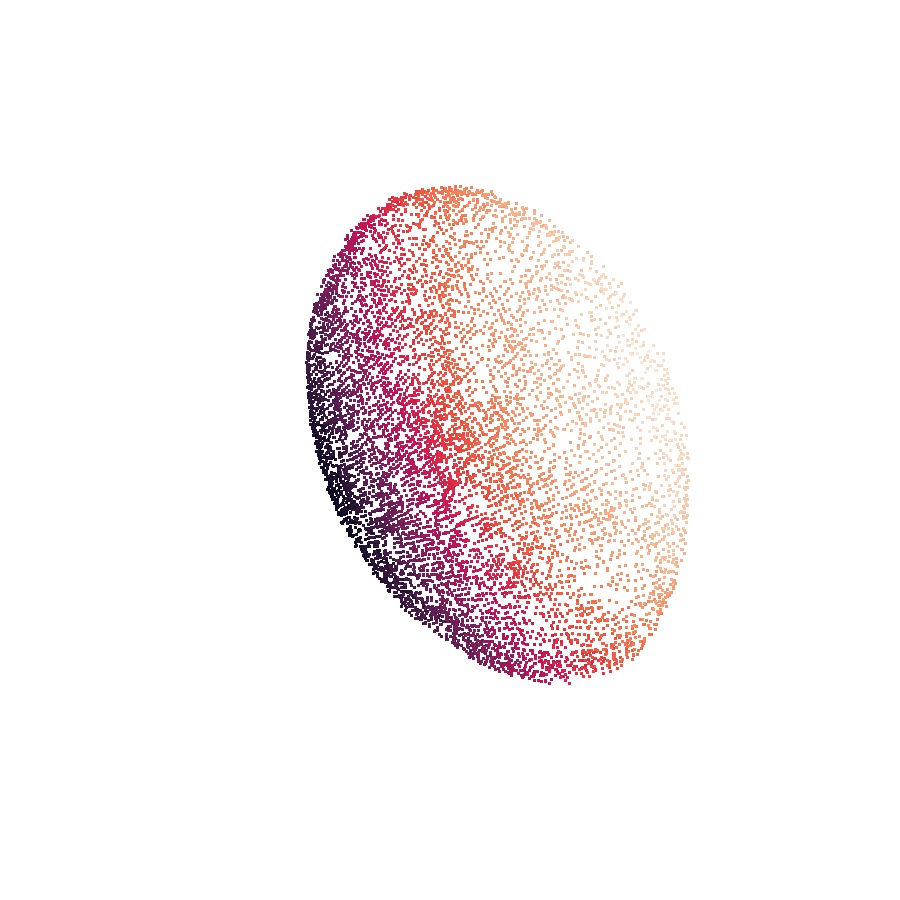}\end{subfigure} & \begin{subfigure}{0.074\textwidth}\centering\includegraphics[trim=220 170 230 110,clip,width=\textwidth]{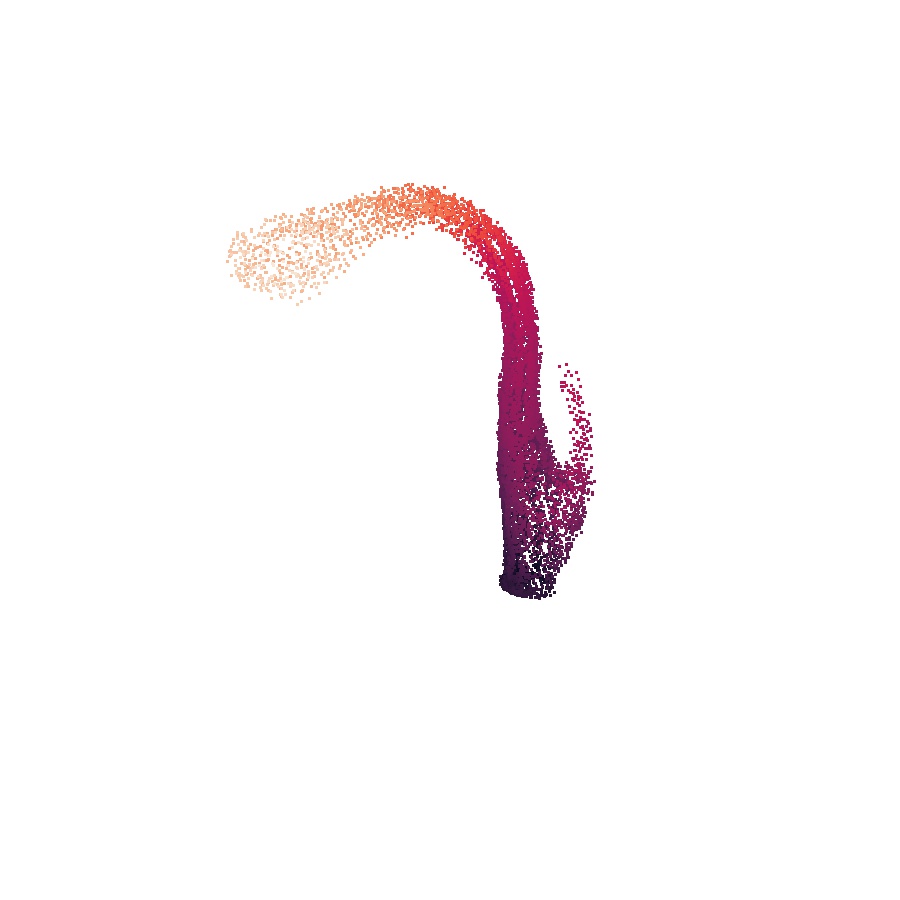}\end{subfigure} & \begin{subfigure}{0.12\textwidth}\centering\includegraphics[trim=270 280 200 370,clip,width=\textwidth]{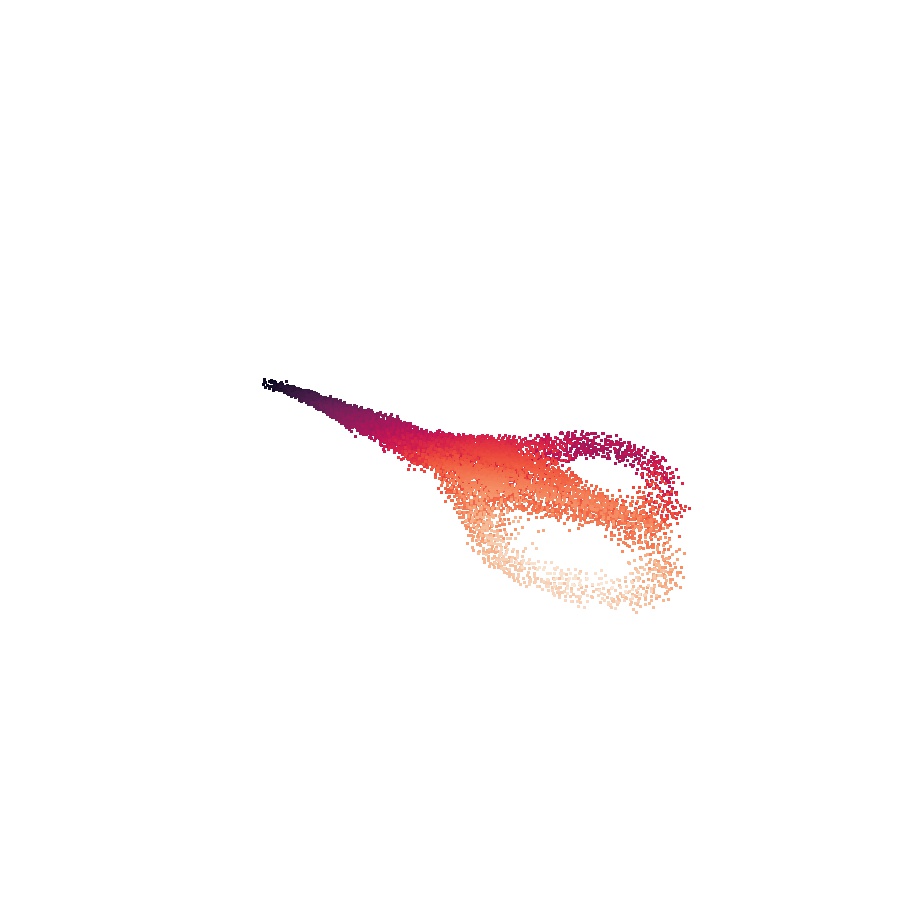}\end{subfigure} & \begin{subfigure}{0.082\textwidth}\centering\includegraphics[trim=150 200 250 250,clip,width=\textwidth]{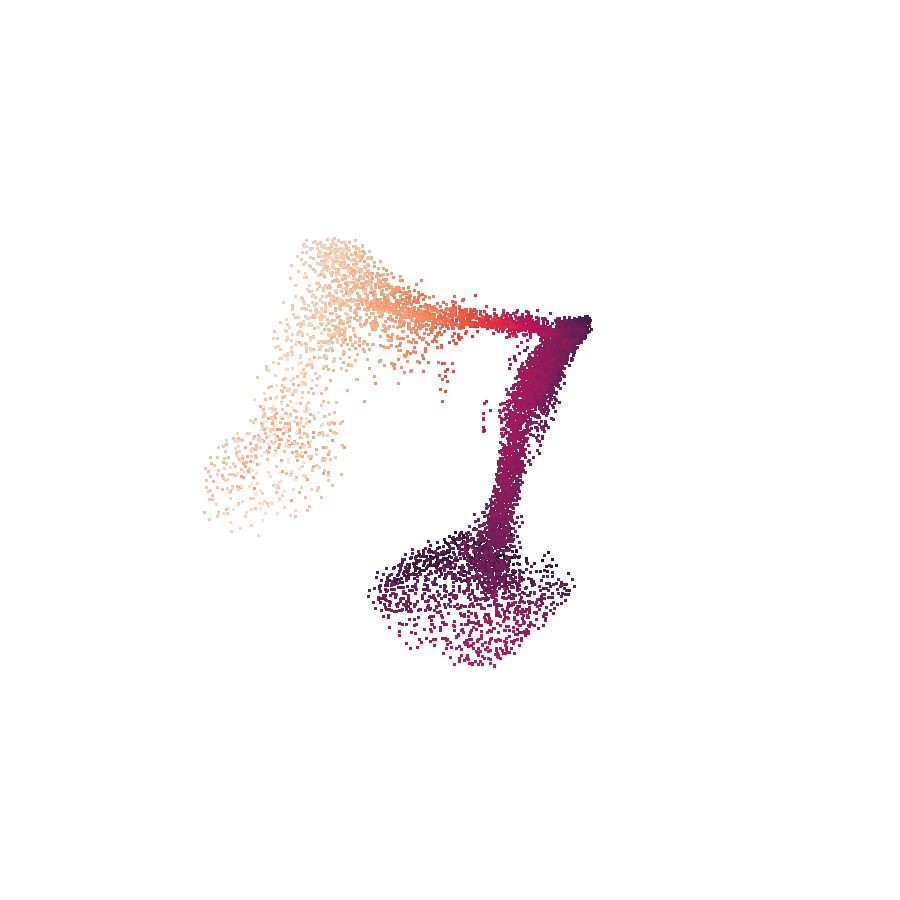}\end{subfigure} & \begin{subfigure}{0.05\textwidth}\centering\includegraphics[trim=300 200 300 200,clip,width=\textwidth]{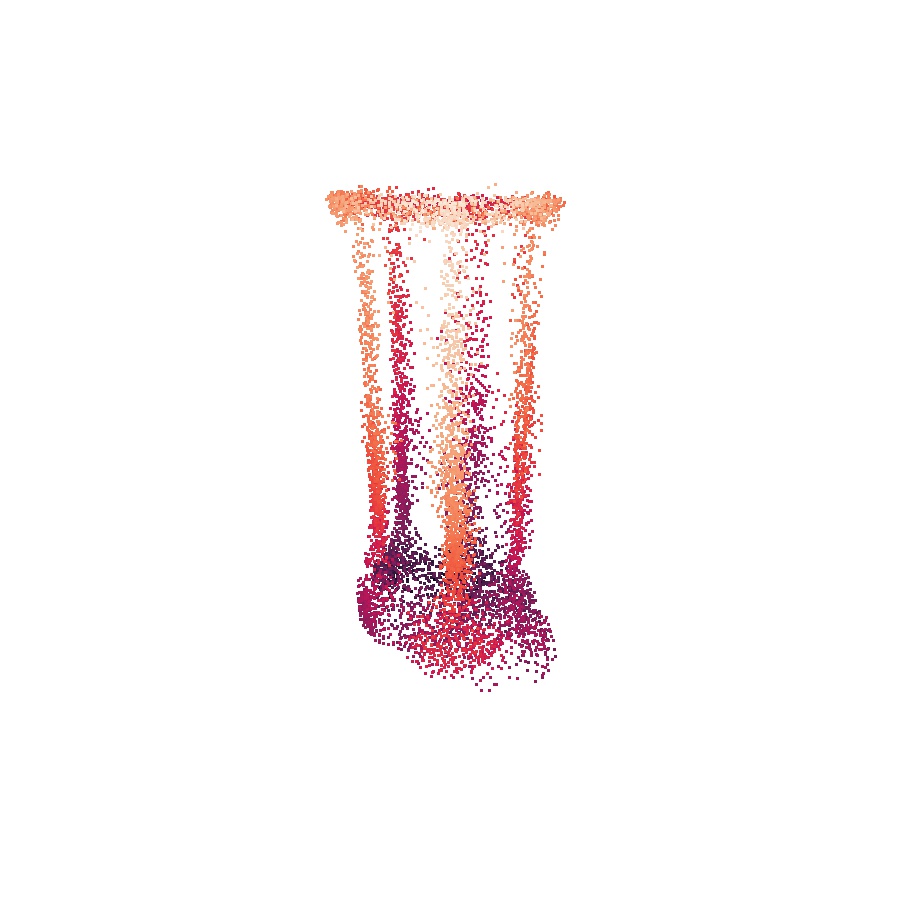}\end{subfigure} & \begin{subfigure}{0.095\textwidth}\centering\includegraphics[trim=200 250 200 250,clip,width=\textwidth]{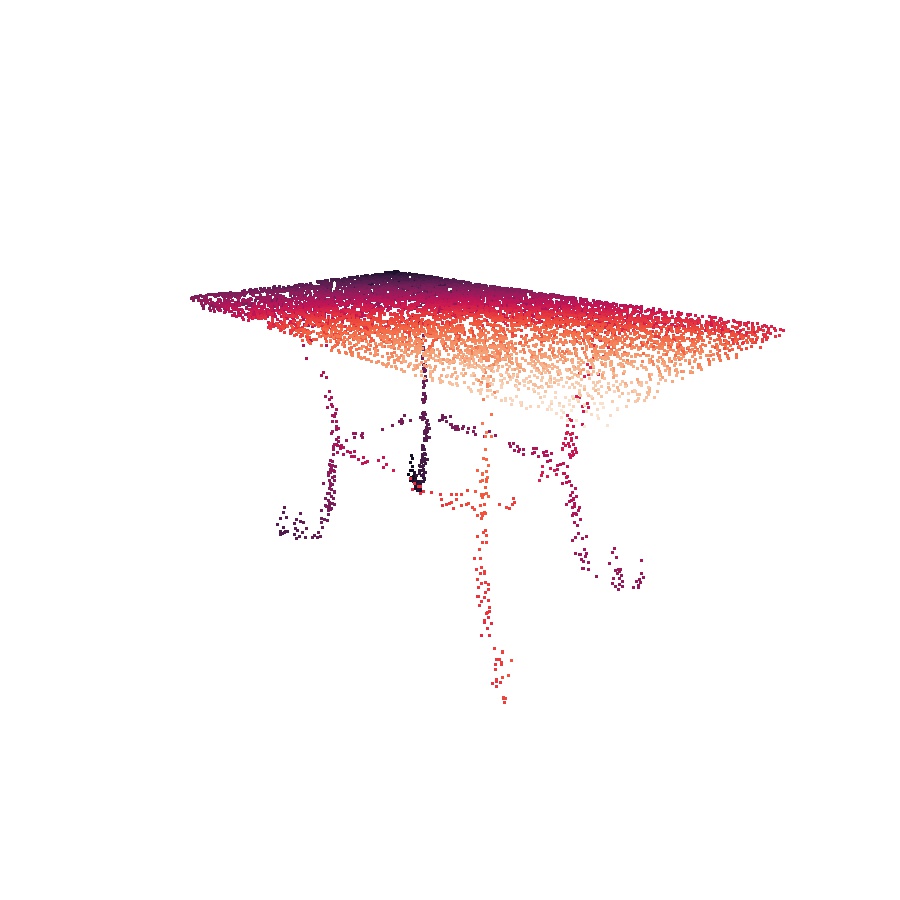}\end{subfigure} & \begin{subfigure}{0.086\textwidth}\centering\includegraphics[trim=100 150 200 100,clip,width=\textwidth]{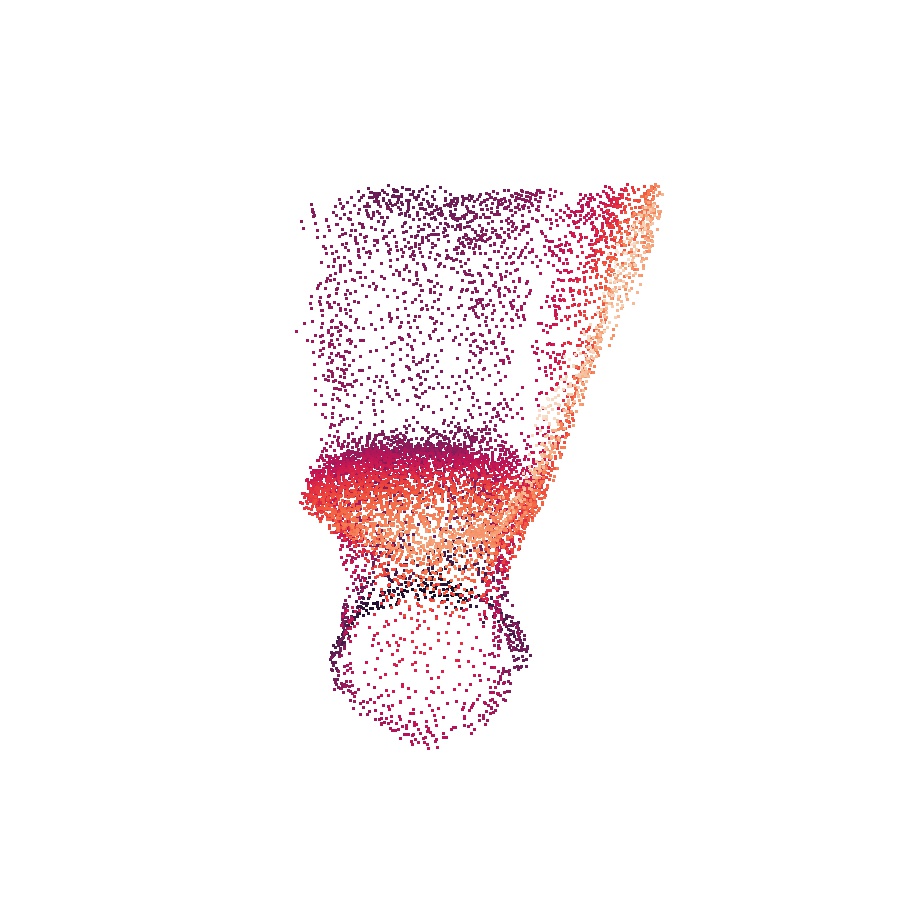}\end{subfigure} & \begin{subfigure}{0.11\textwidth}\centering\includegraphics[trim=200 300 100 250,clip,width=\textwidth]{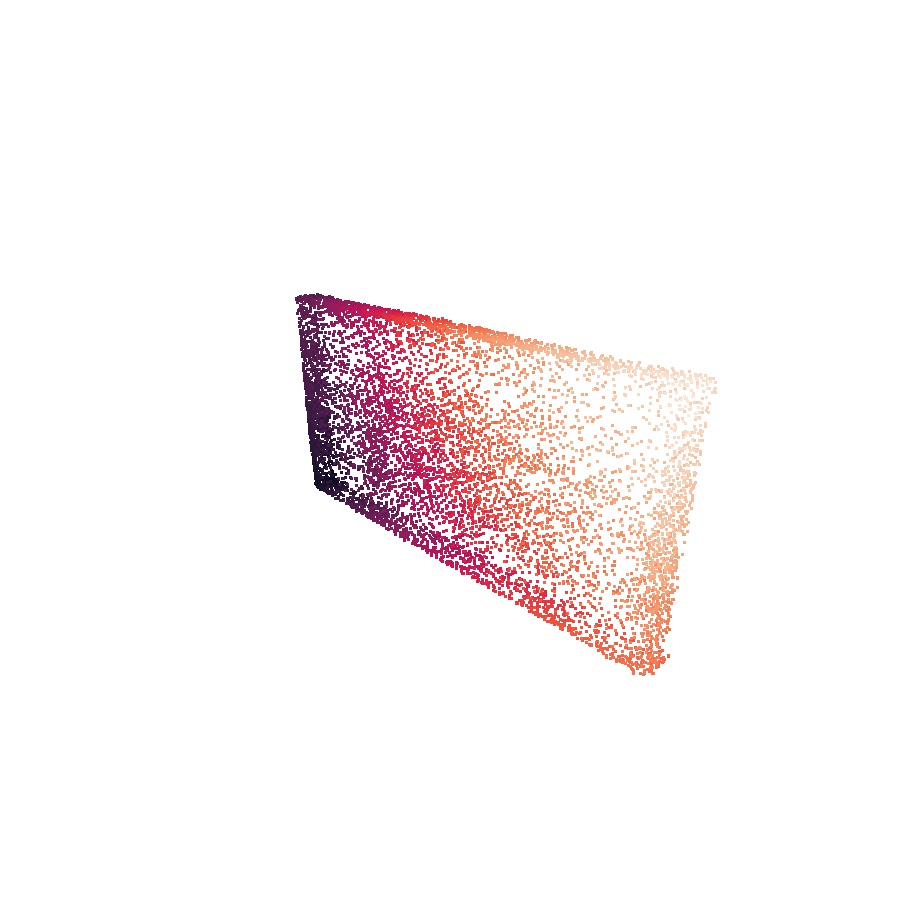}\end{subfigure} \\
SoftPool & \begin{subfigure}{0.07\textwidth}\centering\includegraphics[trim=270 150 230 150,clip,width=\textwidth]{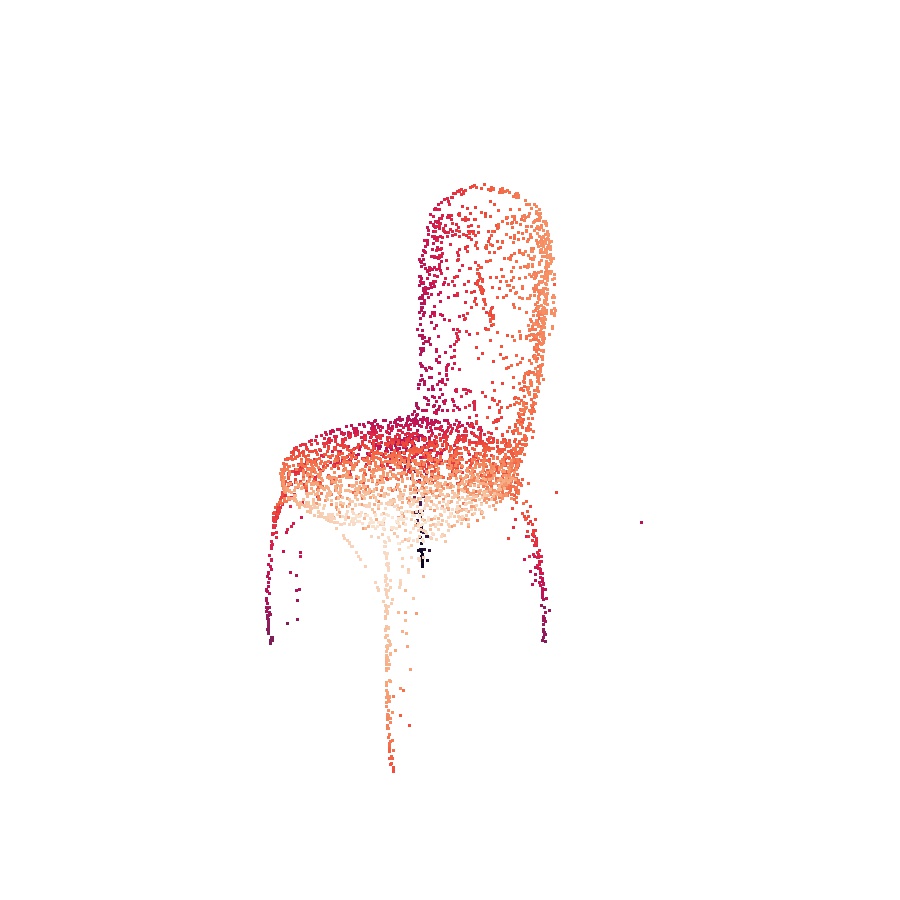}\end{subfigure} & \begin{subfigure}{0.077\textwidth}\centering\includegraphics[trim=250 150 200 150,clip,width=\textwidth]{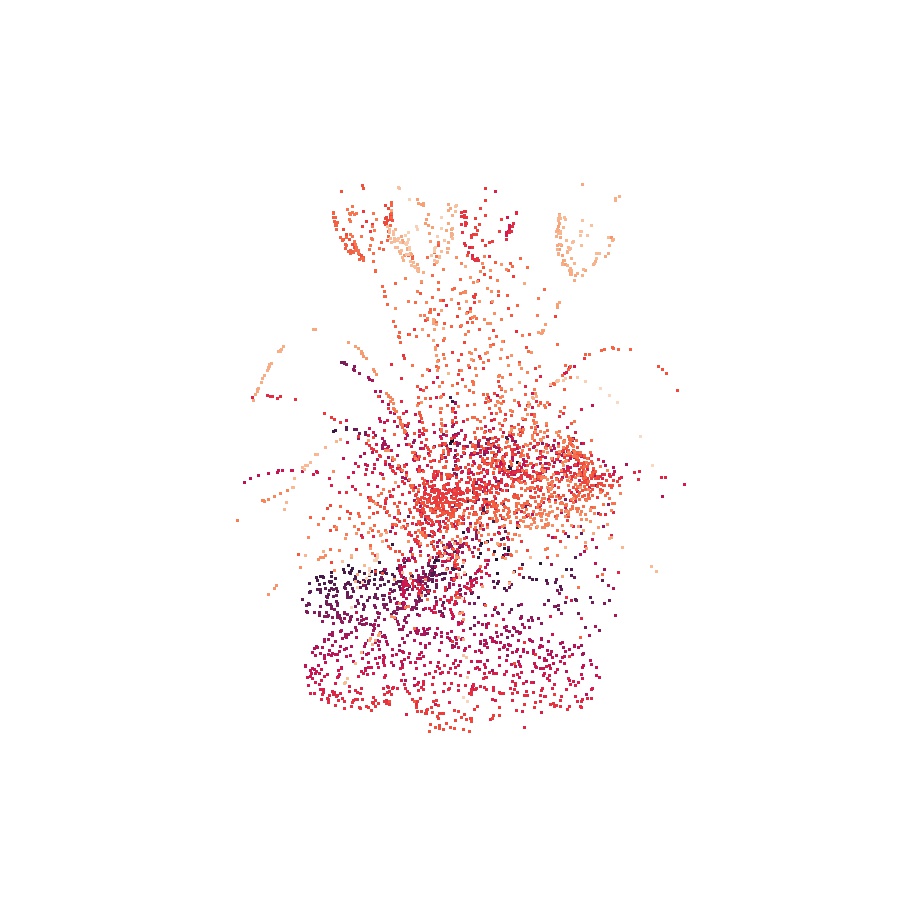}\end{subfigure} & \begin{subfigure}{0.071\textwidth}\centering\includegraphics[trim=300 150 190 140,clip,width=\textwidth]{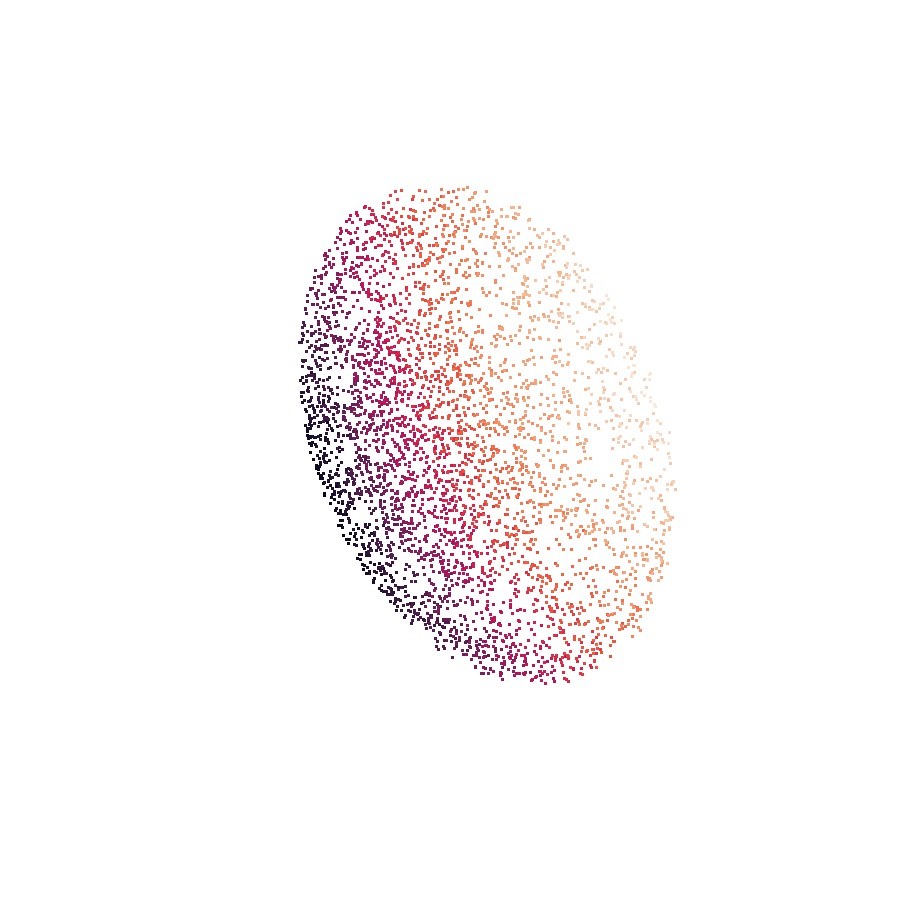}\end{subfigure} & \begin{subfigure}{0.074\textwidth}\centering\includegraphics[trim=190 170 260 110,clip,width=\textwidth]{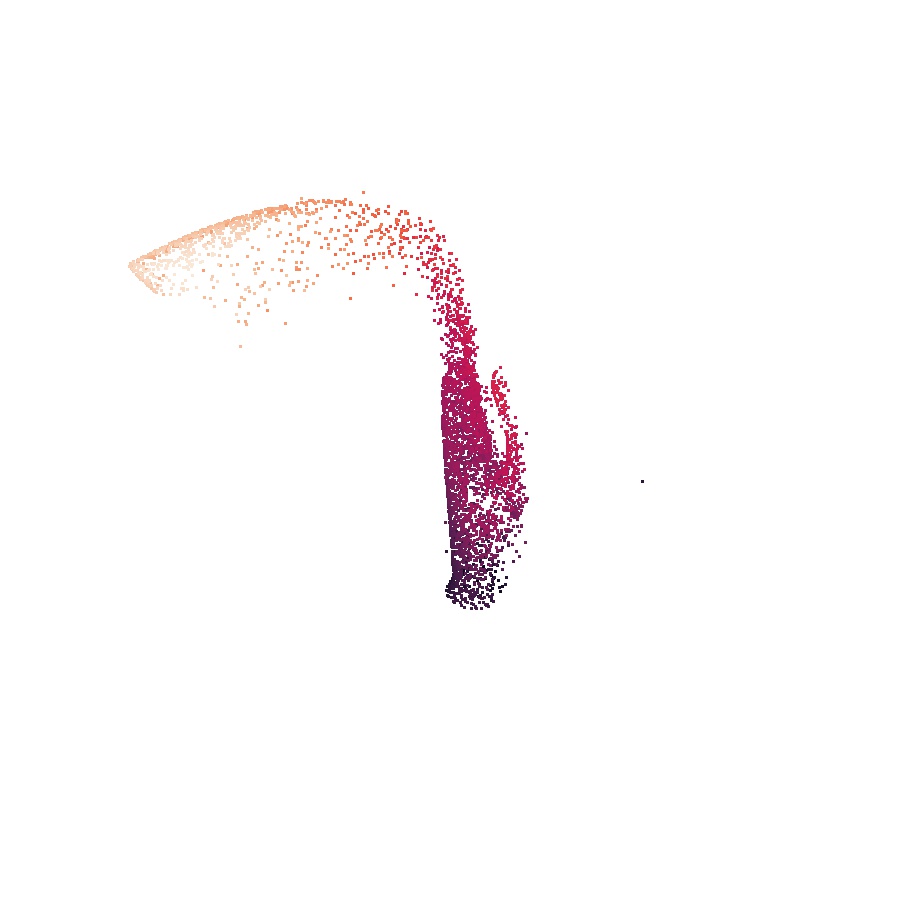}\end{subfigure} & \begin{subfigure}{0.12\textwidth}\centering\includegraphics[trim=270 280 200 370,clip,width=\textwidth]{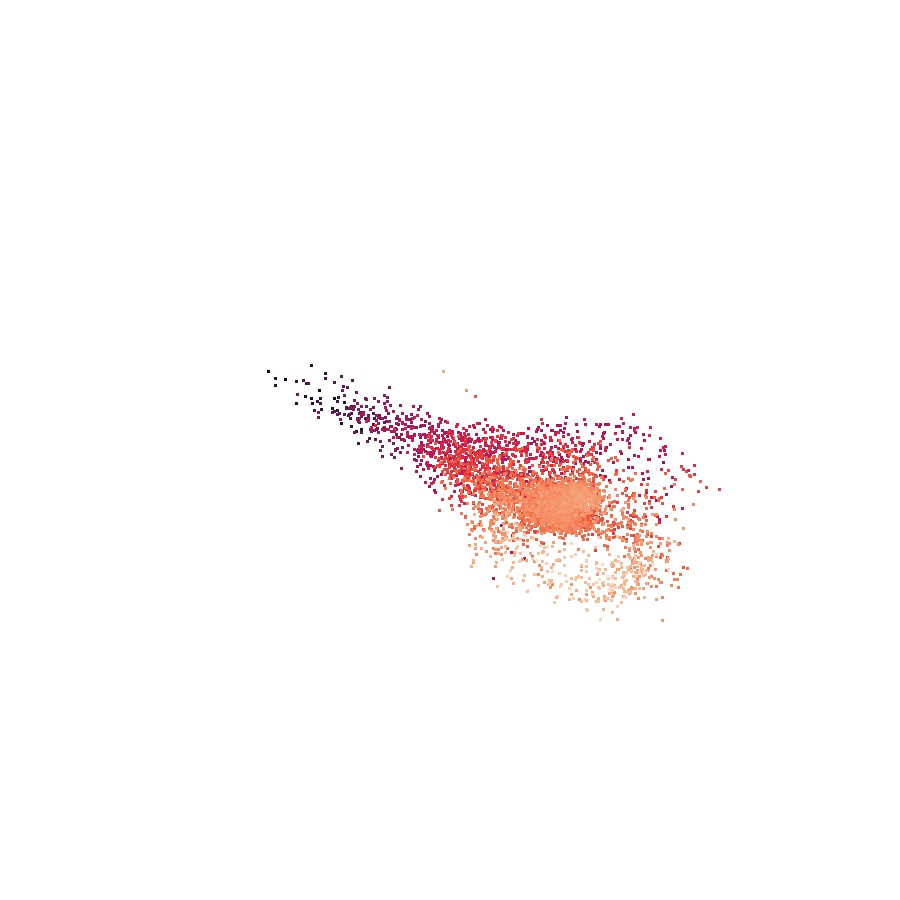}\end{subfigure} & \begin{subfigure}{0.084\textwidth}\centering\includegraphics[trim=130 200 280 250,clip,width=\textwidth]{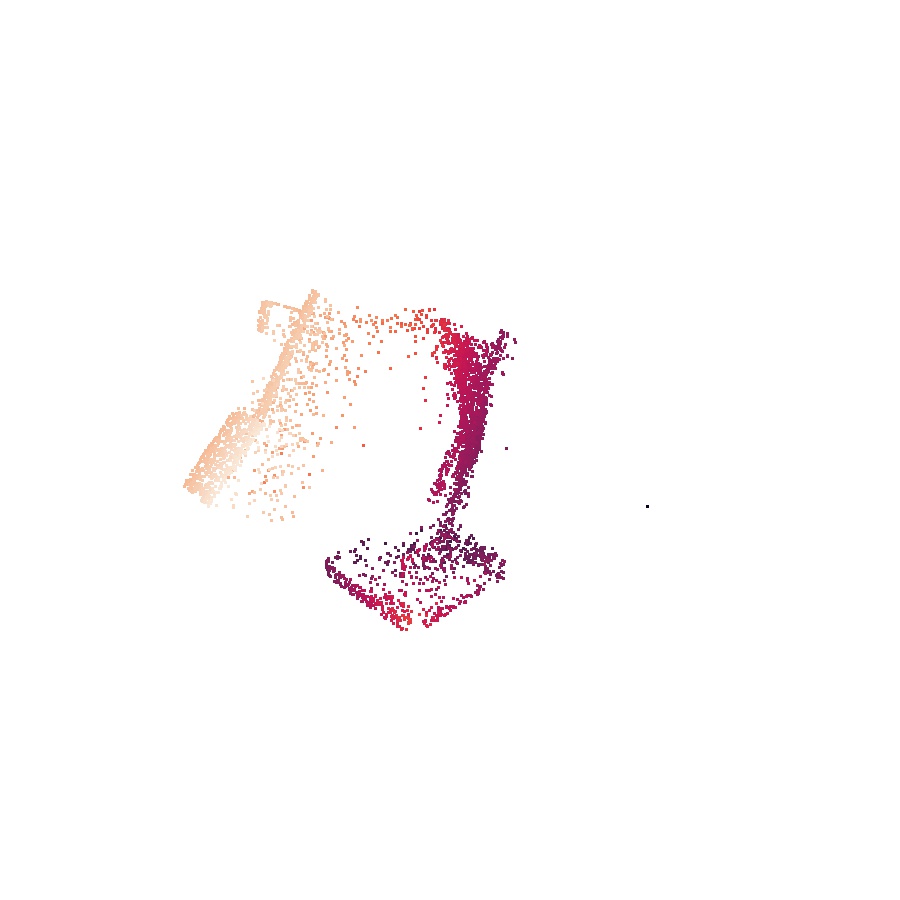}\end{subfigure} & \begin{subfigure}{0.05\textwidth}\centering\includegraphics[trim=300 200 350 200,clip,width=\textwidth]{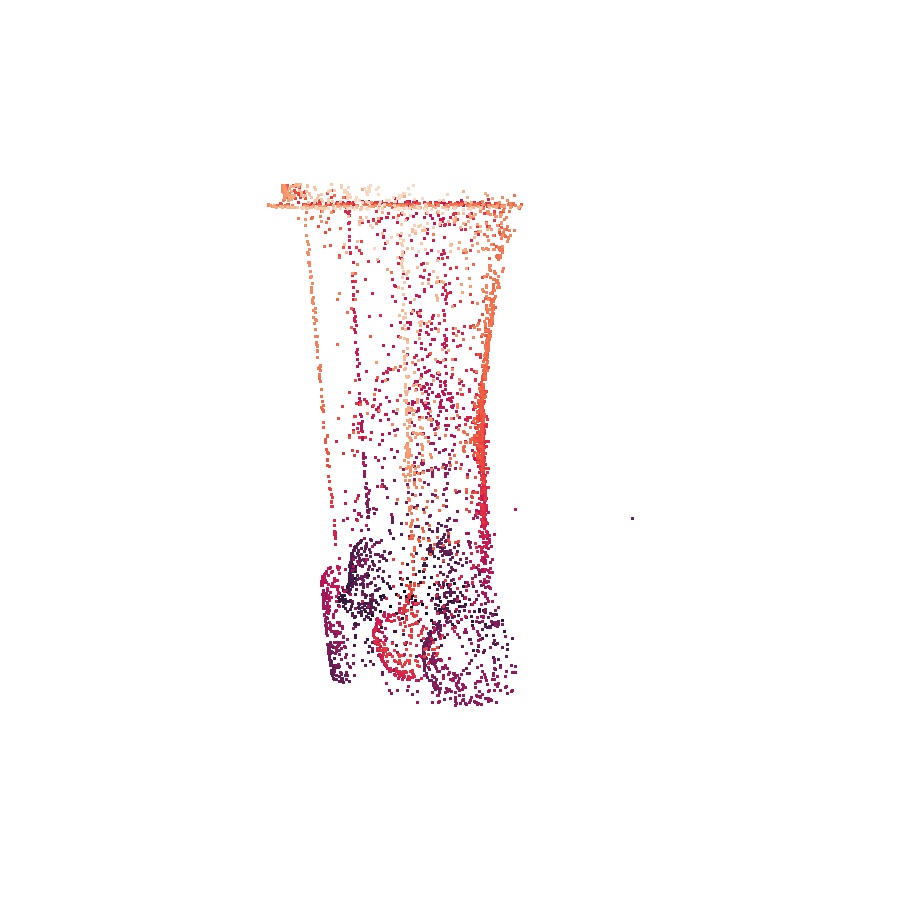}\end{subfigure} & \begin{subfigure}{0.095\textwidth}\centering\includegraphics[trim=200 250 200 250,clip,width=\textwidth]{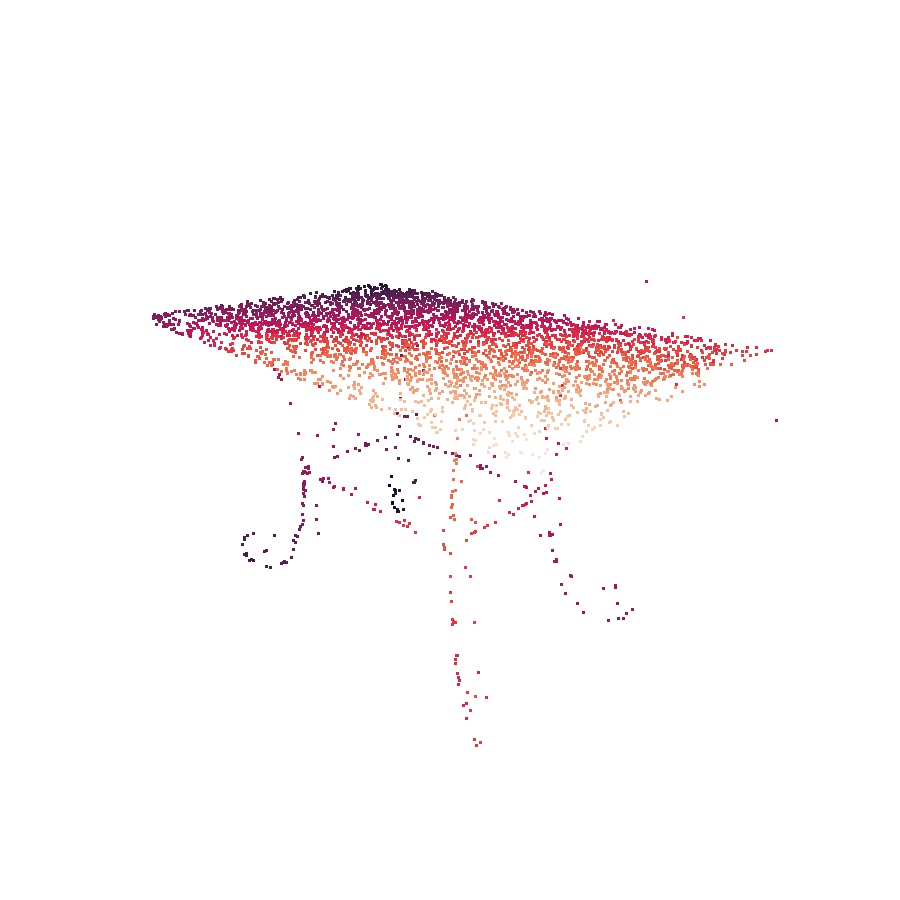}\end{subfigure} & \begin{subfigure}{0.086\textwidth}\centering\includegraphics[trim=100 150 200 100,clip,width=\textwidth]{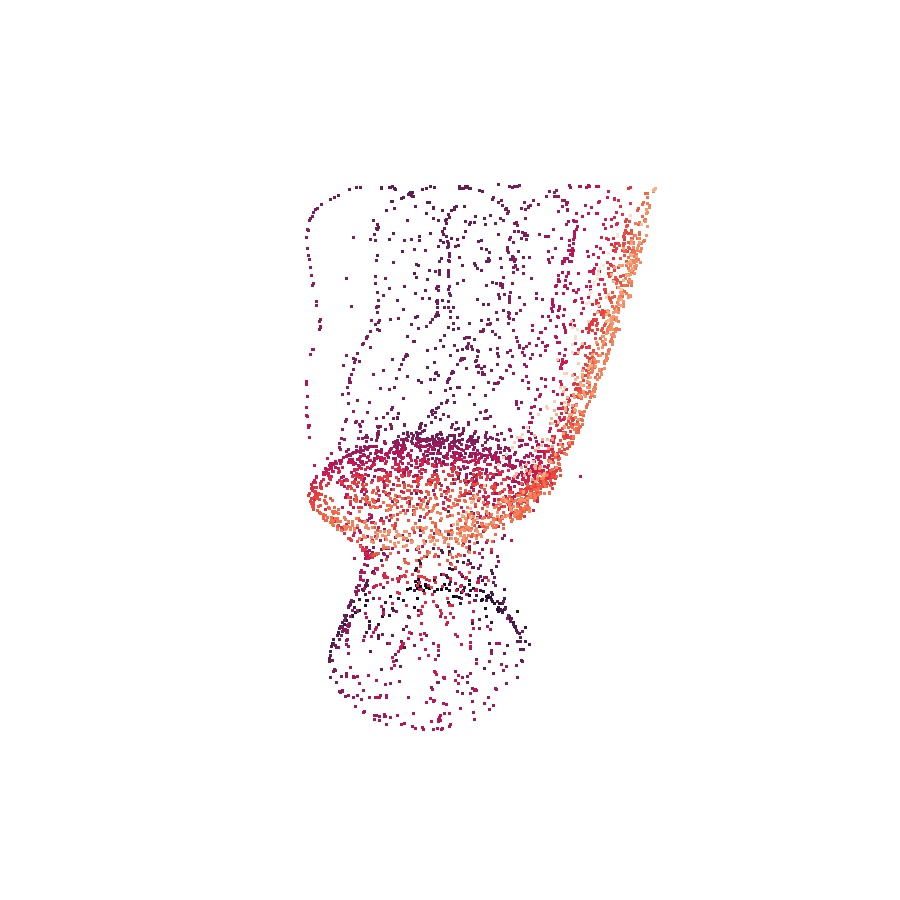}\end{subfigure} & \begin{subfigure}{0.11\textwidth}\centering\includegraphics[trim=200 300 100 250,clip,width=\textwidth]{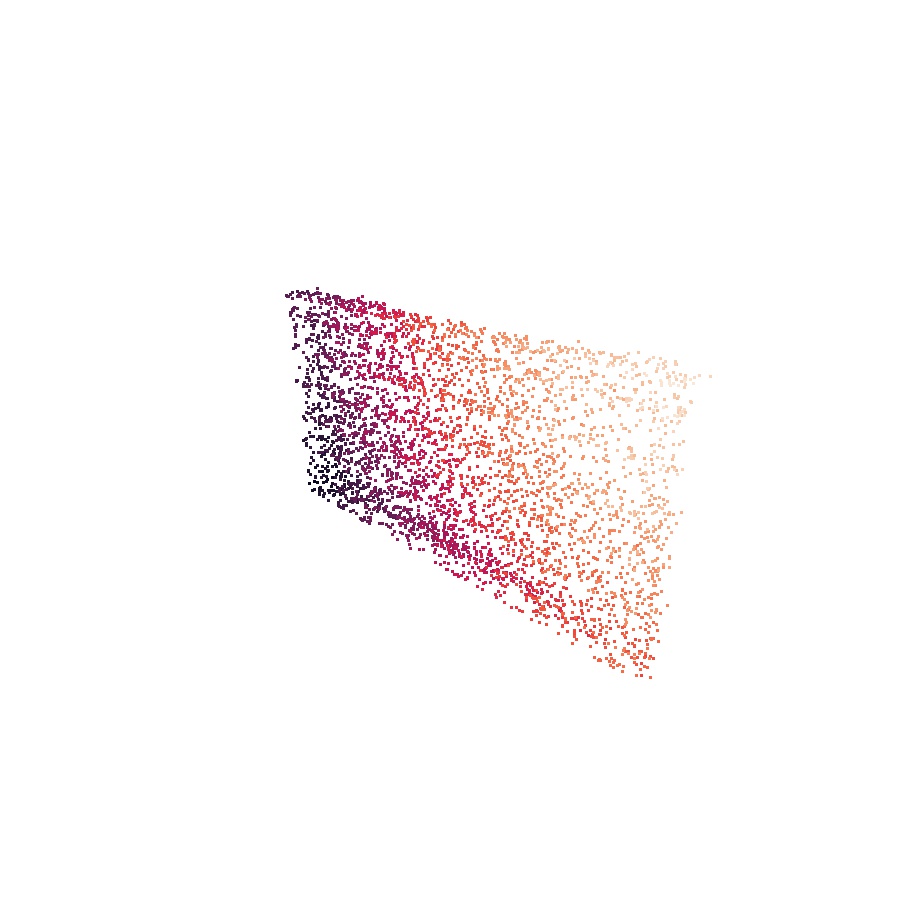}\end{subfigure} \\
MSN & \begin{subfigure}{0.07\textwidth}\centering\includegraphics[trim=300 150 200 150,clip,width=\textwidth]{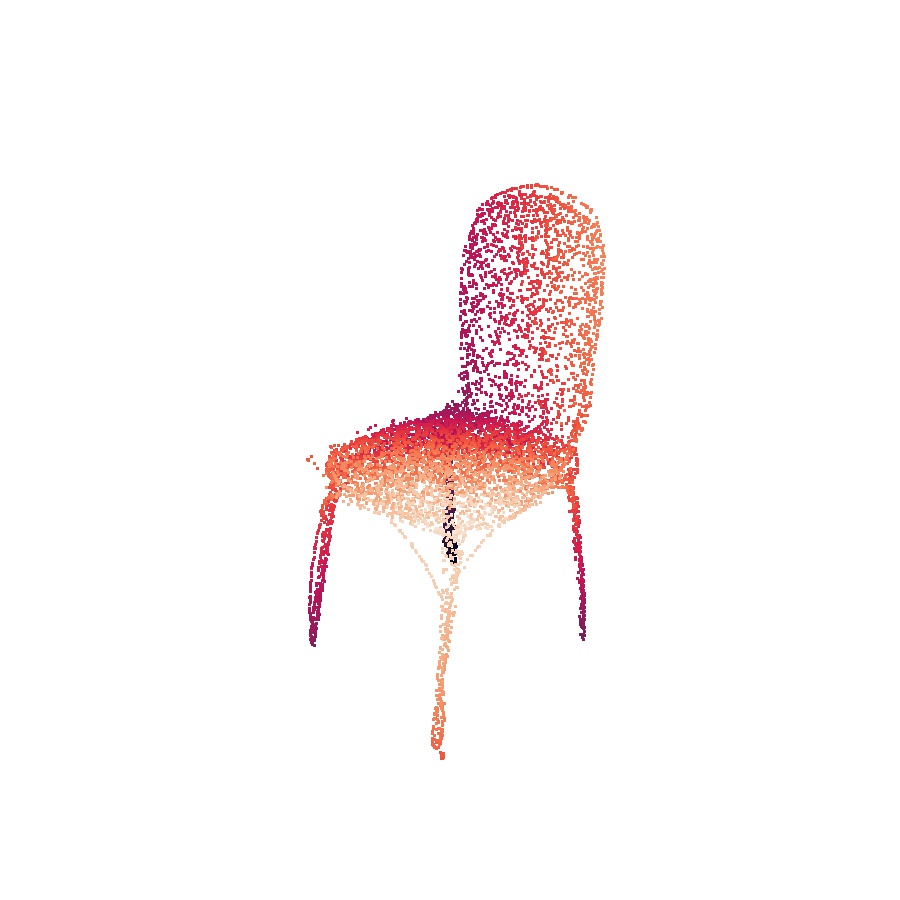}\end{subfigure} & \begin{subfigure}{0.077\textwidth}\centering\includegraphics[trim=250 150 200 150,clip,width=\textwidth]{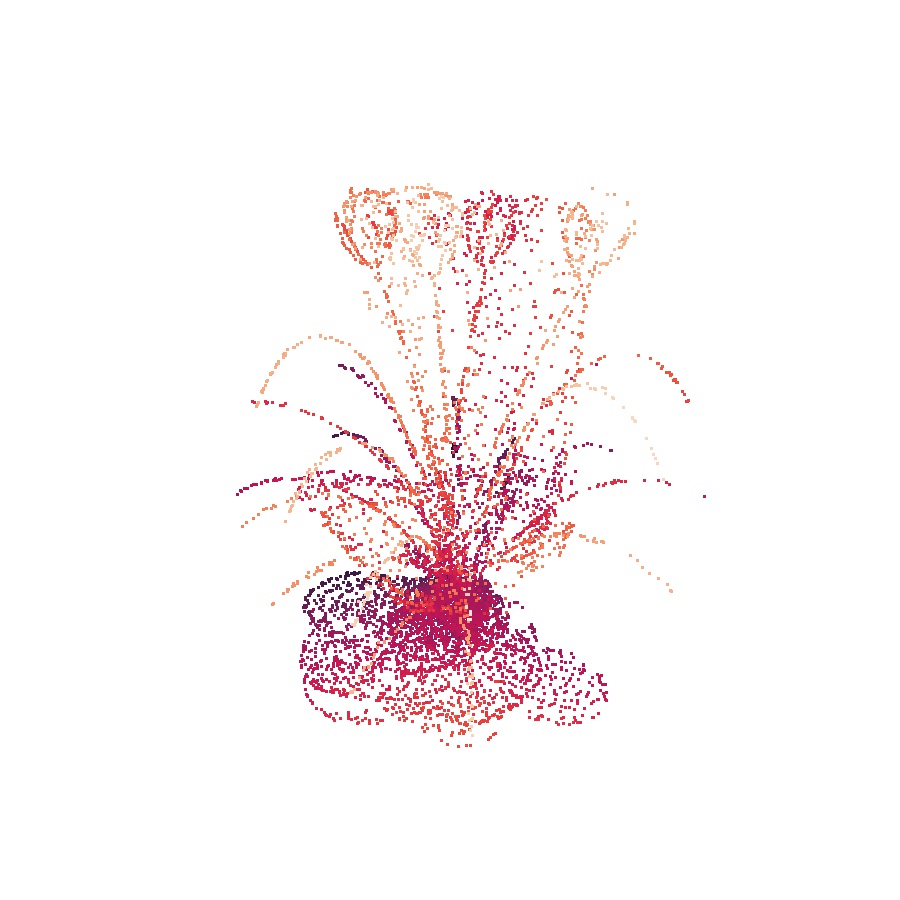}\end{subfigure} & \begin{subfigure}{0.071\textwidth}\centering\includegraphics[trim=300 150 190 140,clip,width=\textwidth]{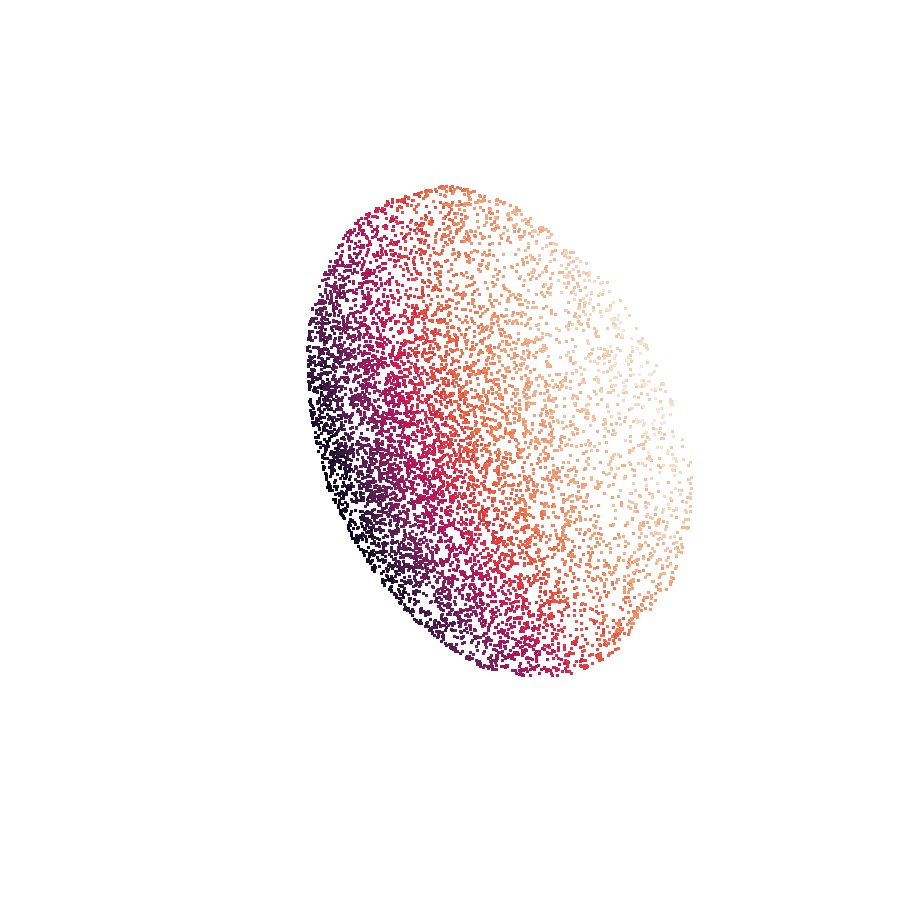}\end{subfigure} & \begin{subfigure}{0.074\textwidth}\centering\includegraphics[trim=220 170 230 110,clip,width=\textwidth]{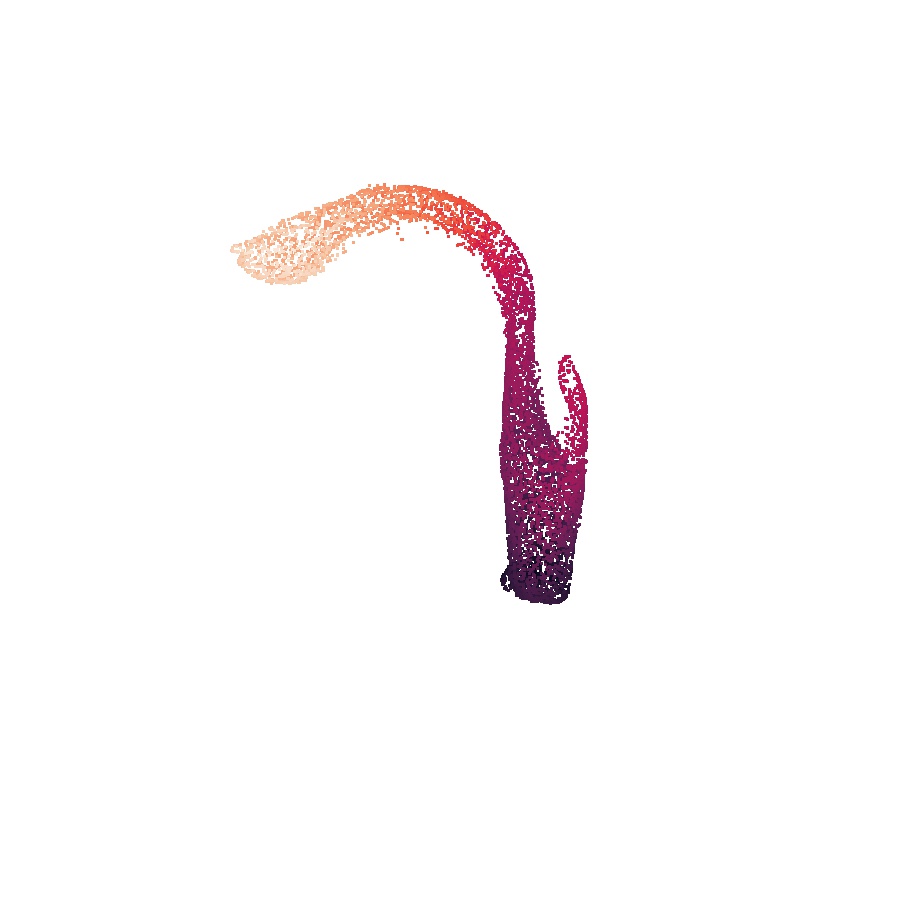}\end{subfigure} & \begin{subfigure}{0.12\textwidth}\centering\includegraphics[trim=270 280 200 370,clip,width=\textwidth]{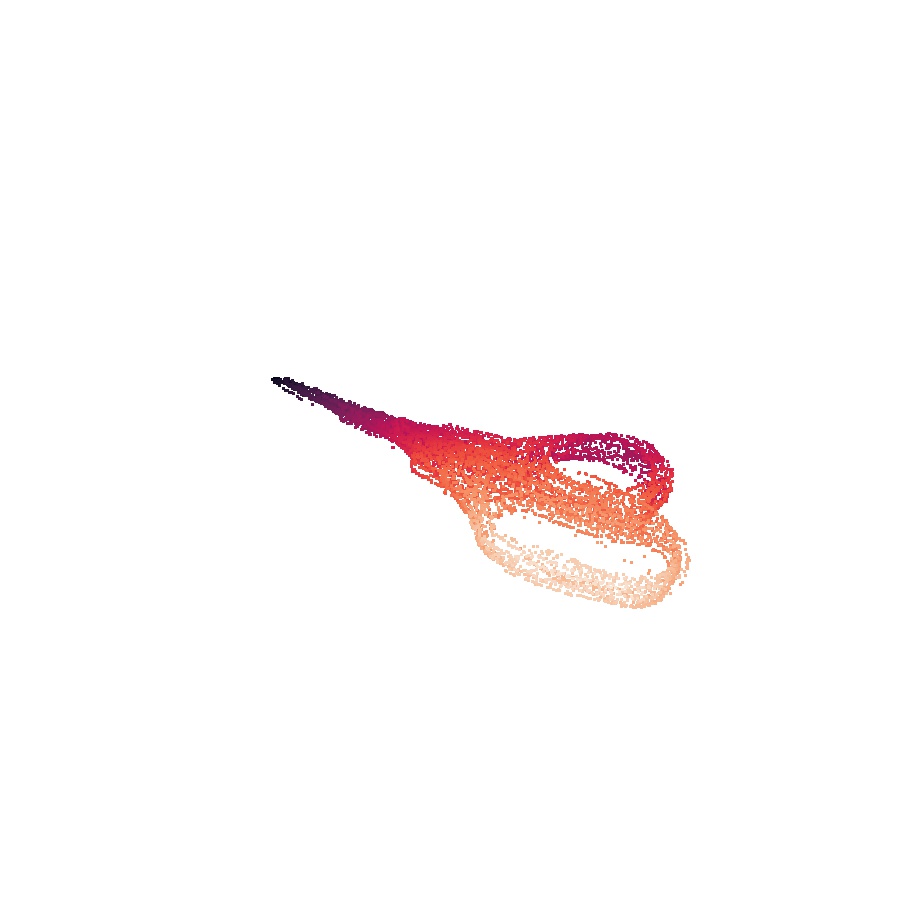}\end{subfigure} & \begin{subfigure}{0.082\textwidth}\centering\includegraphics[trim=150 200 250 250,clip,width=\textwidth]{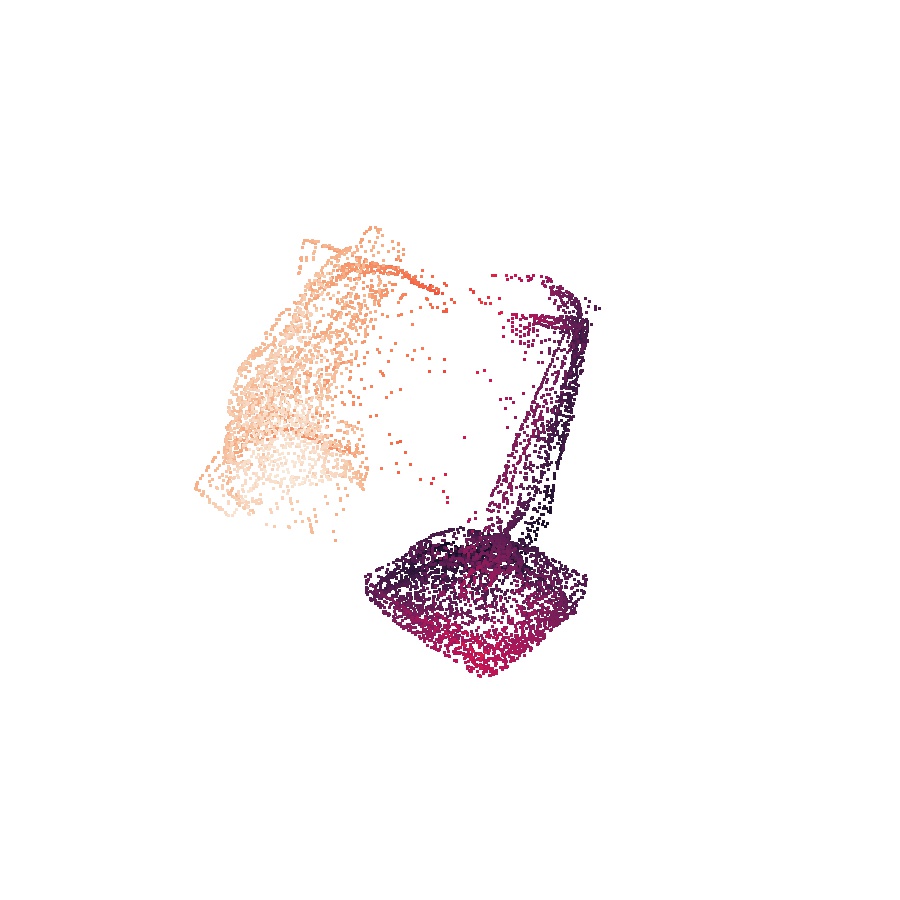}\end{subfigure} & \begin{subfigure}{0.05\textwidth}\centering\includegraphics[trim=300 200 300 200,clip,width=\textwidth]{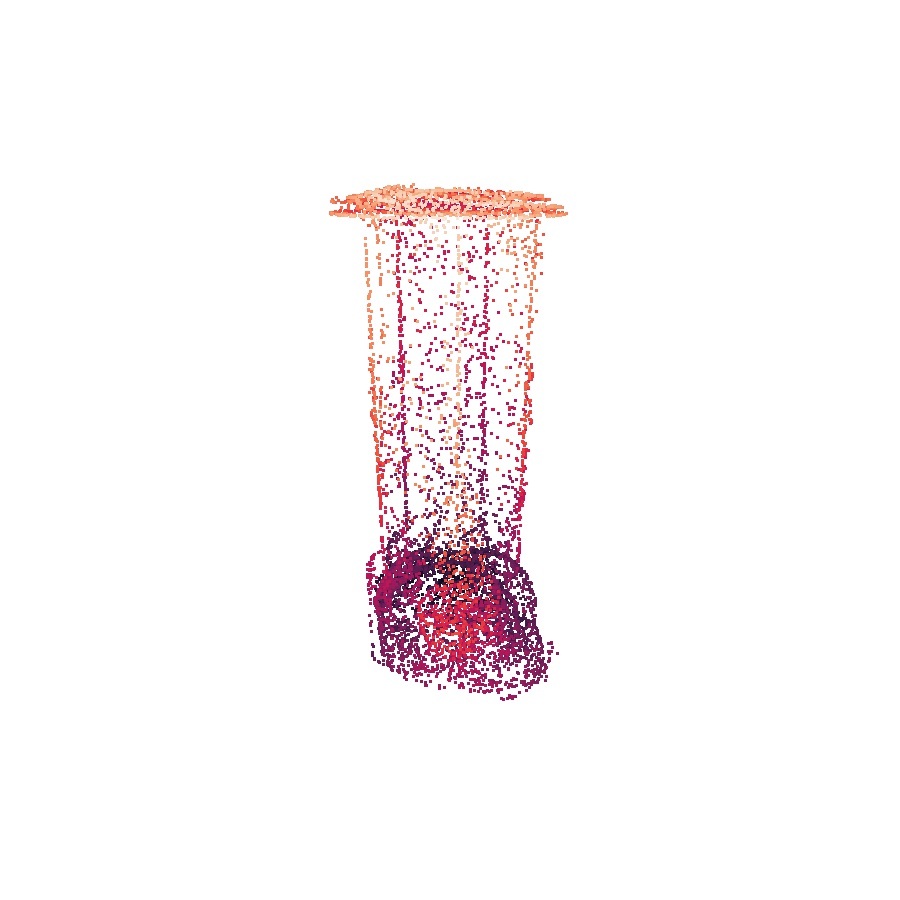}\end{subfigure} & \begin{subfigure}{0.095\textwidth}\centering\includegraphics[trim=200 250 200 250,clip,width=\textwidth]{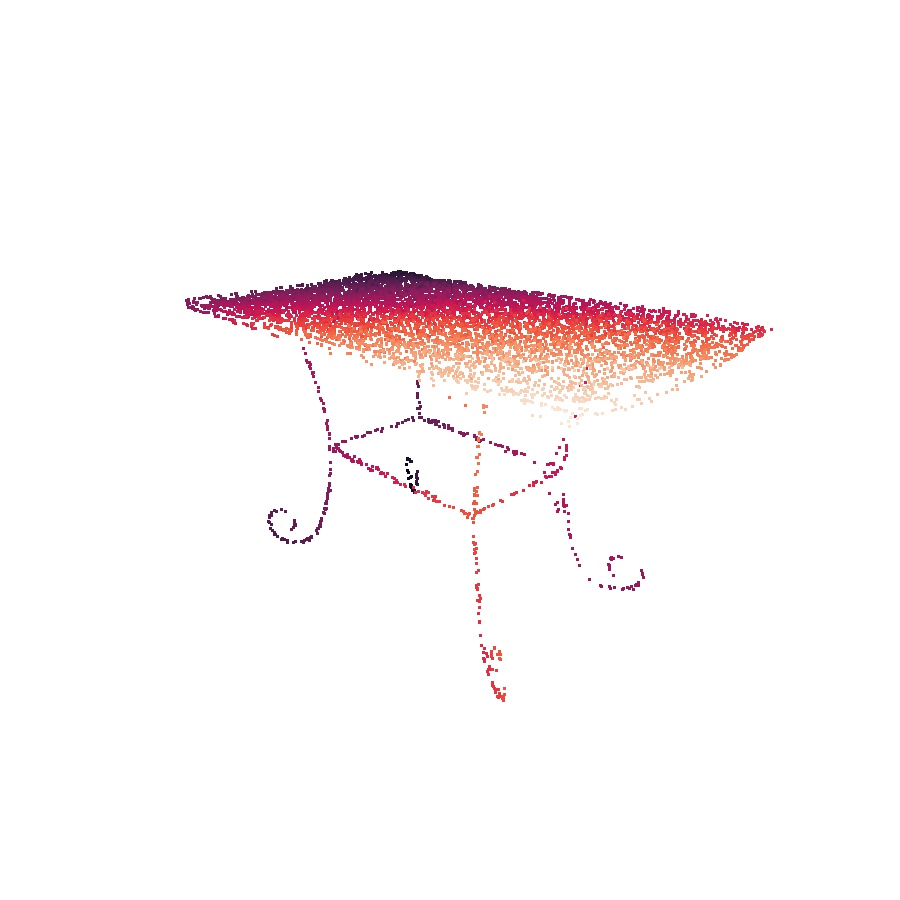}\end{subfigure} & \begin{subfigure}{0.086\textwidth}\centering\includegraphics[trim=100 150 200 100,clip,width=\textwidth]{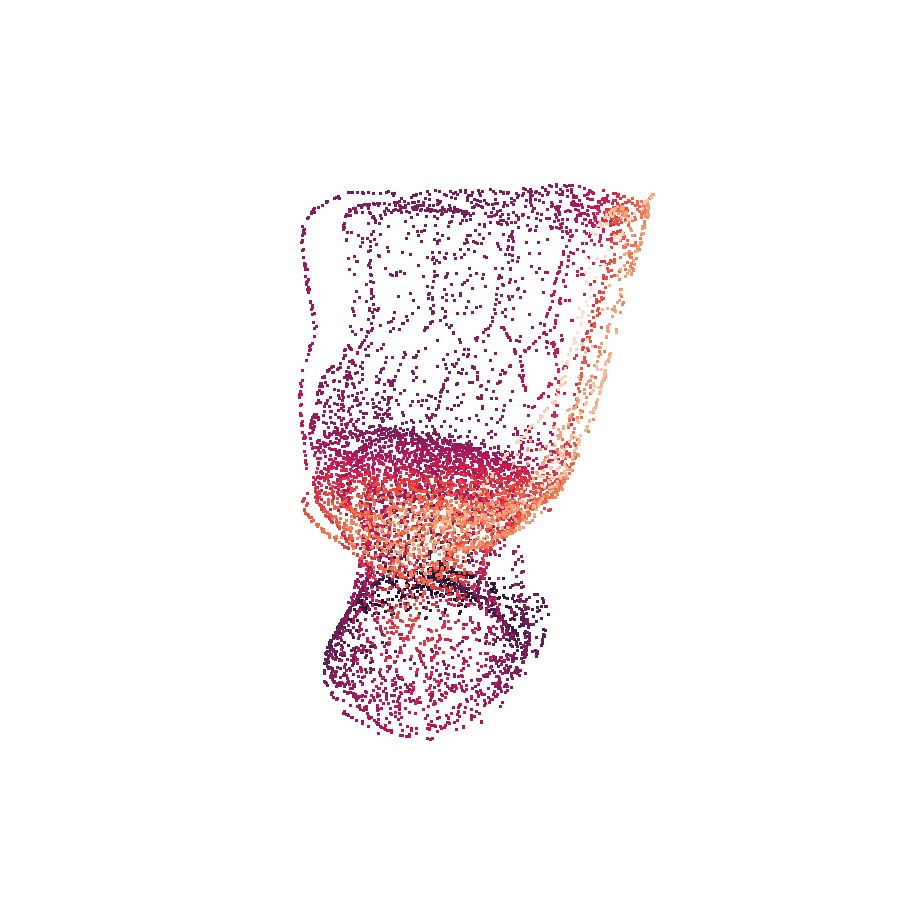}\end{subfigure} & \begin{subfigure}{0.11\textwidth}\centering\includegraphics[trim=200 300 100 250,clip,width=\textwidth]{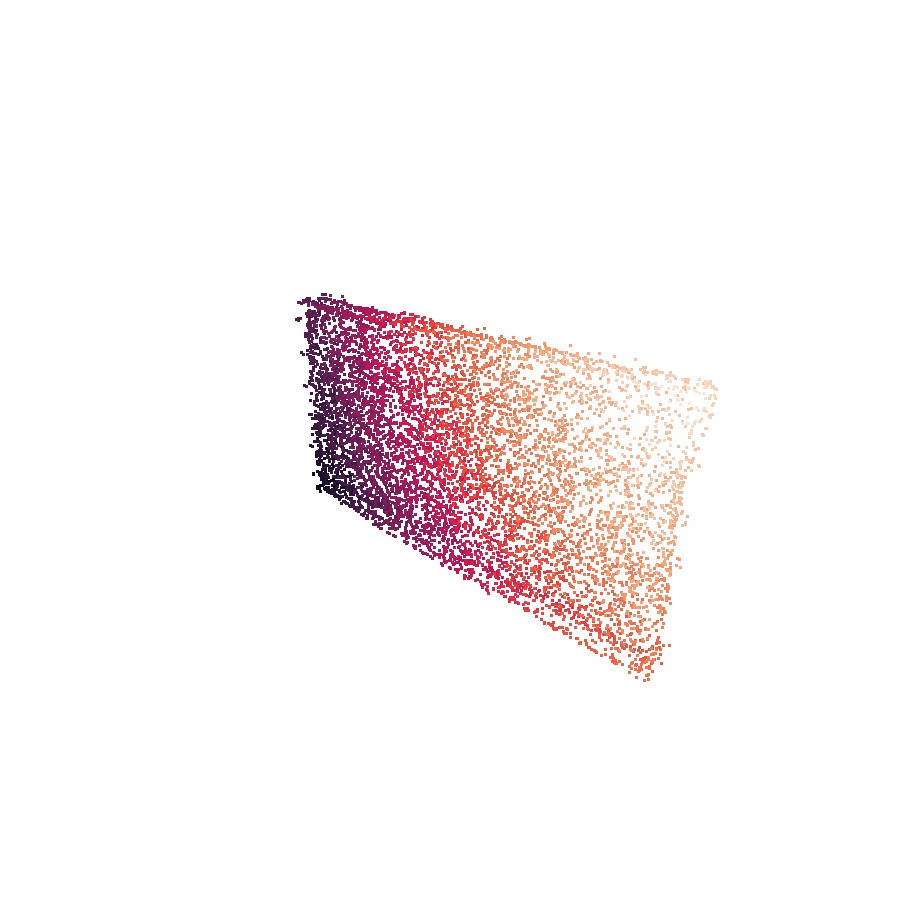}\end{subfigure} \\
CRN & \begin{subfigure}{0.07\textwidth}\centering\includegraphics[trim=300 150 200 150,clip,width=\textwidth]{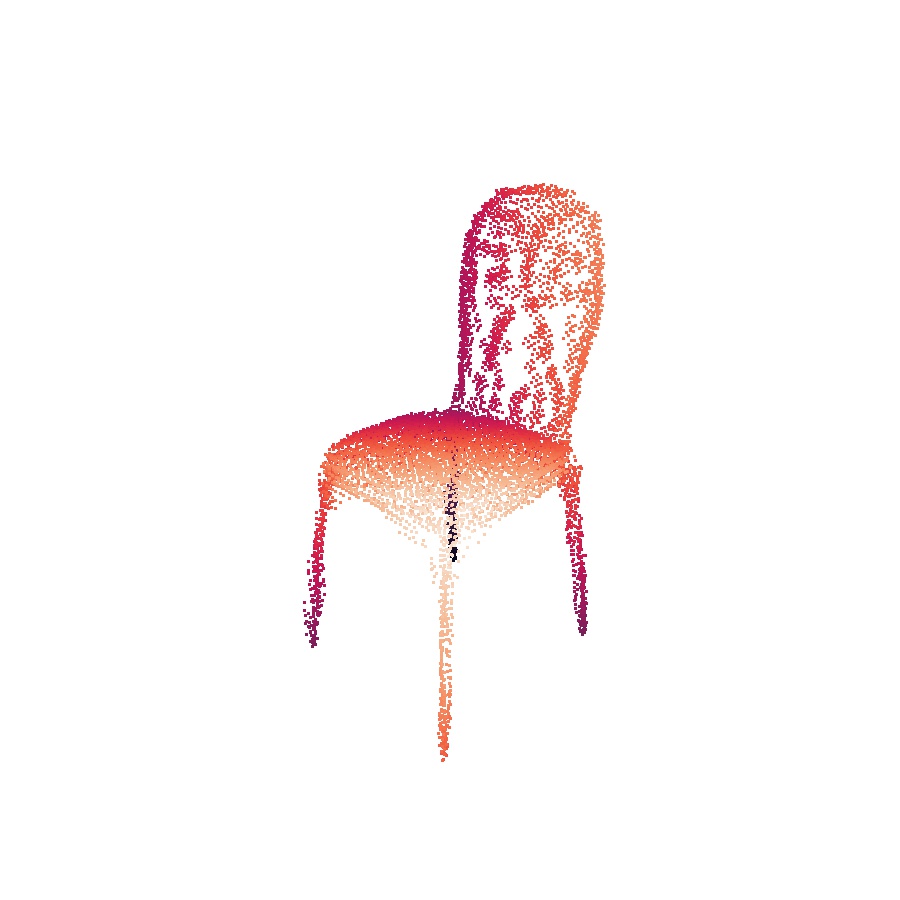}\end{subfigure} & \begin{subfigure}{0.077\textwidth}\centering\includegraphics[trim=250 150 200 150,clip,width=\textwidth]{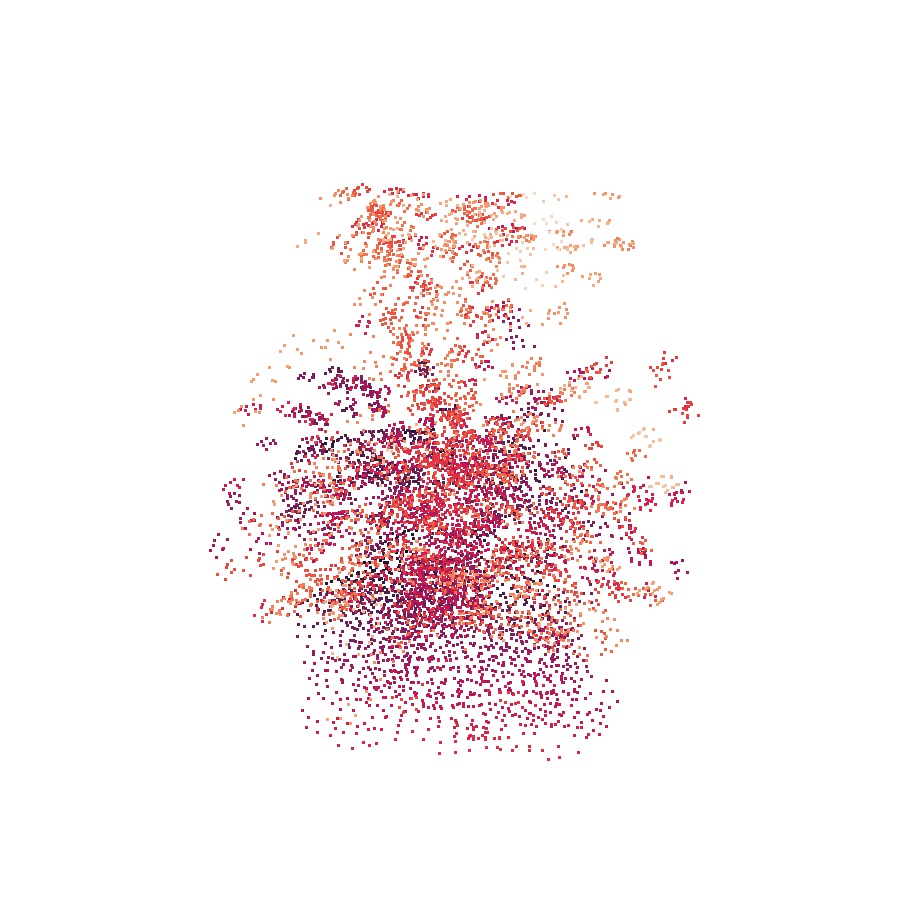}\end{subfigure} & \begin{subfigure}{0.071\textwidth}\centering\includegraphics[trim=300 150 190 140,clip,width=\textwidth]{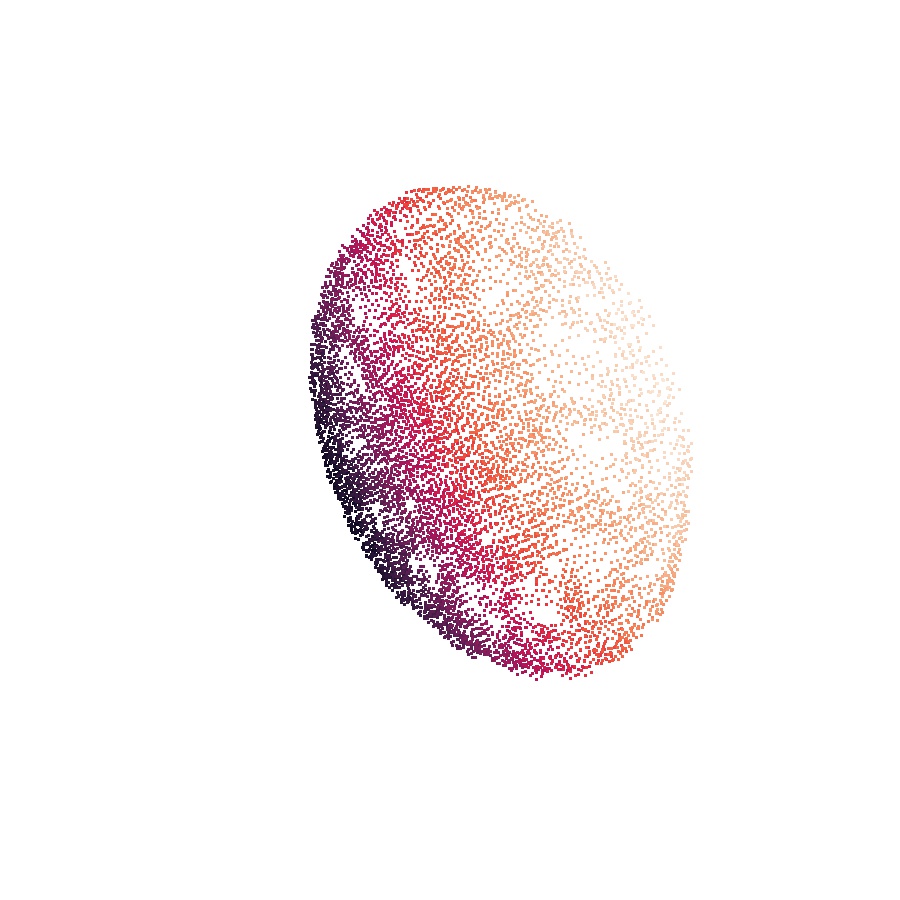}\end{subfigure} & \begin{subfigure}{0.074\textwidth}\centering\includegraphics[trim=220 170 230 110,clip,width=\textwidth]{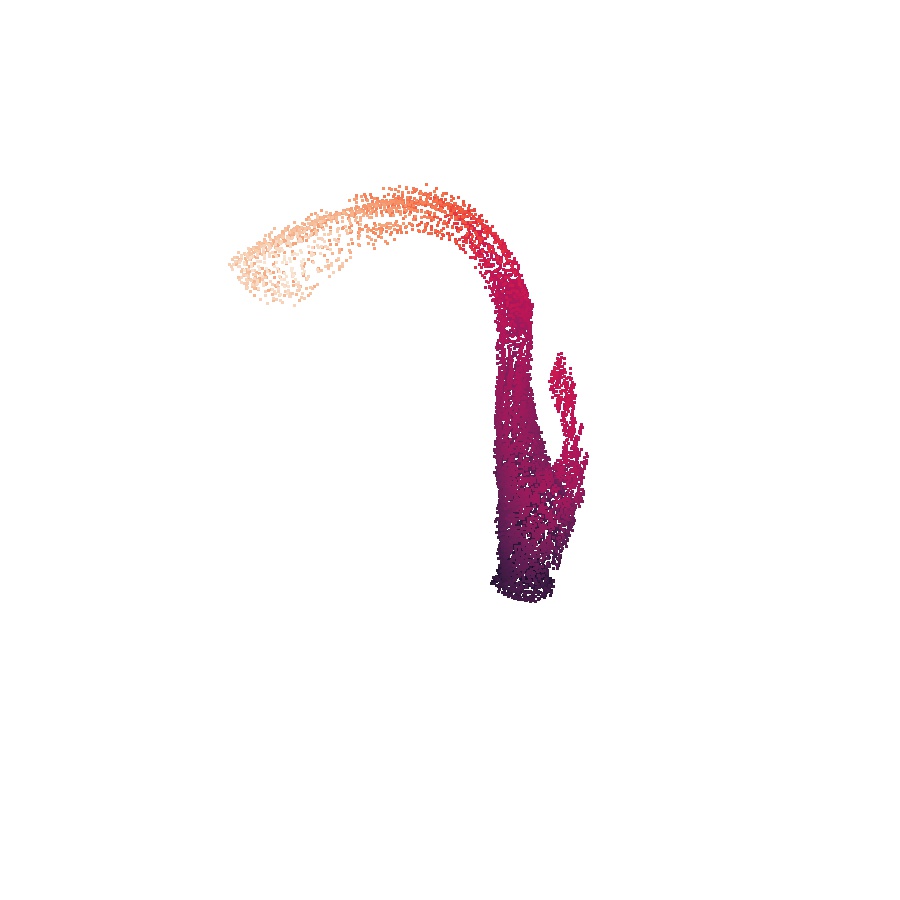}\end{subfigure} & \begin{subfigure}{0.12\textwidth}\centering\includegraphics[trim=270 280 200 370,clip,width=\textwidth]{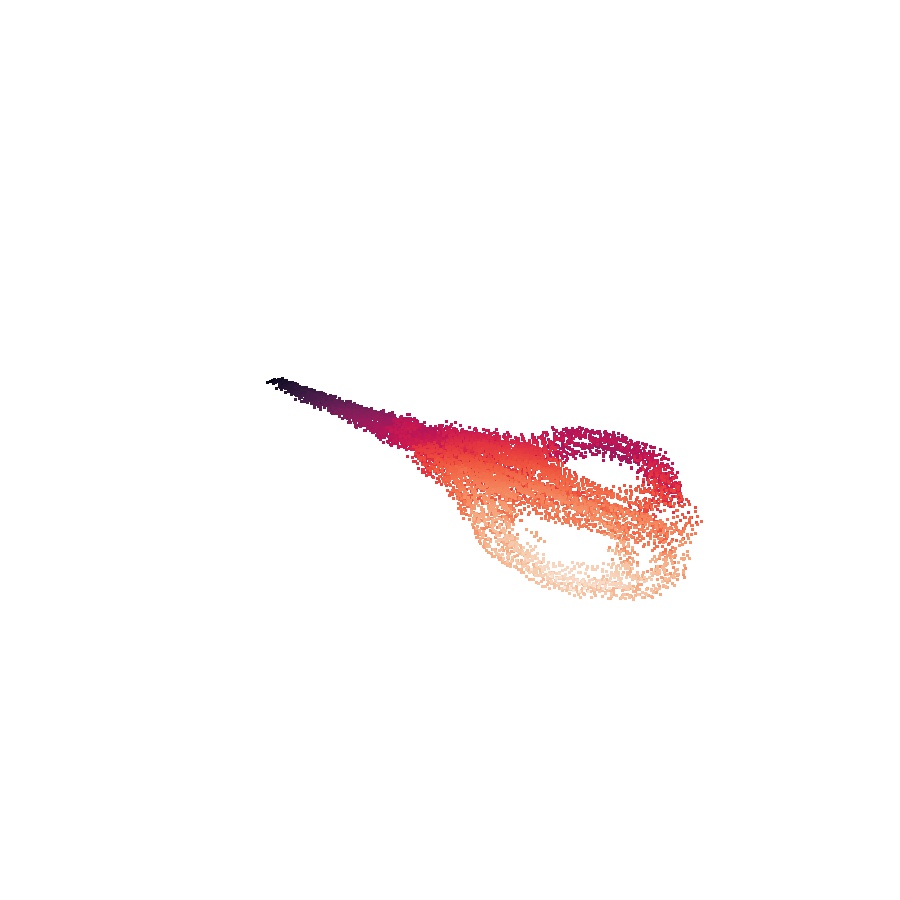}\end{subfigure} & \begin{subfigure}{0.082\textwidth}\centering\includegraphics[trim=150 200 250 250,clip,width=\textwidth]{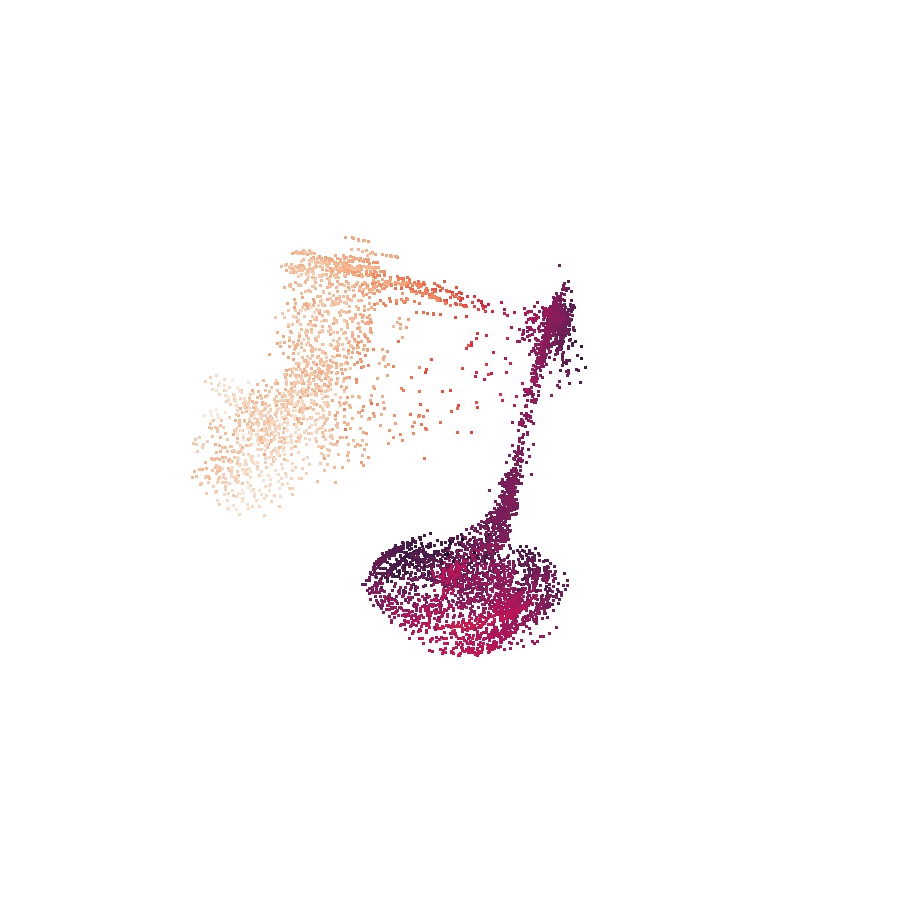}\end{subfigure} & \begin{subfigure}{0.05\textwidth}\centering\includegraphics[trim=300 200 300 200,clip,width=\textwidth]{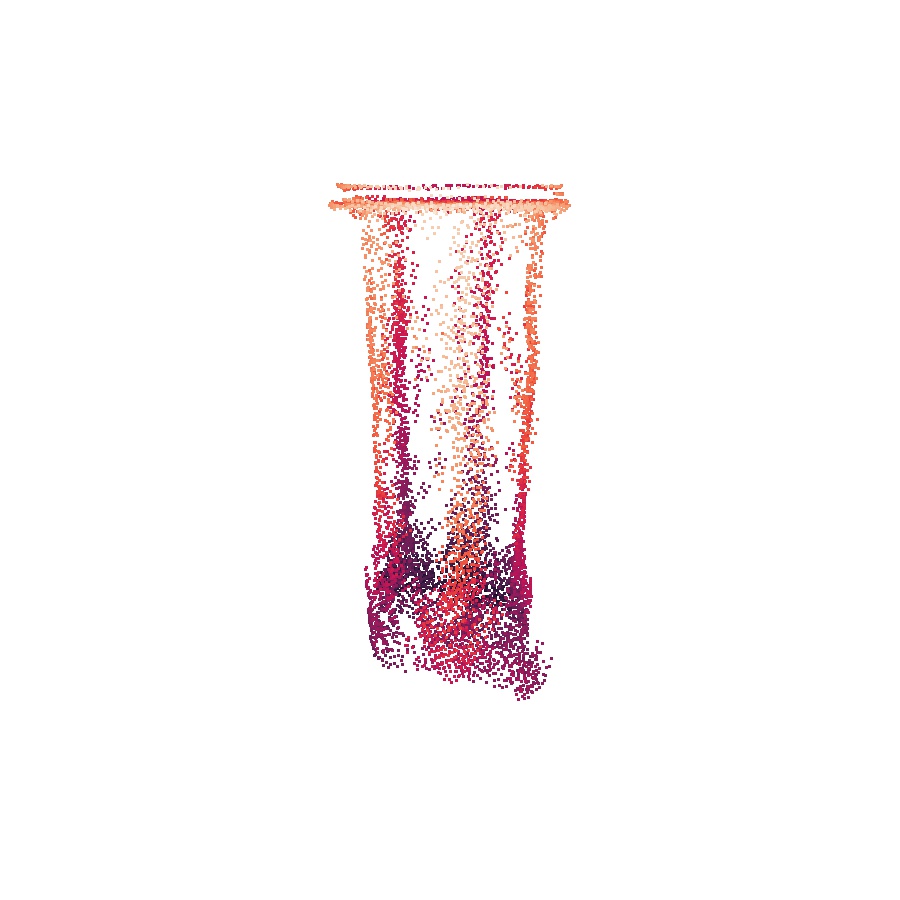}\end{subfigure} & \begin{subfigure}{0.095\textwidth}\centering\includegraphics[trim=200 250 200 250,clip,width=\textwidth]{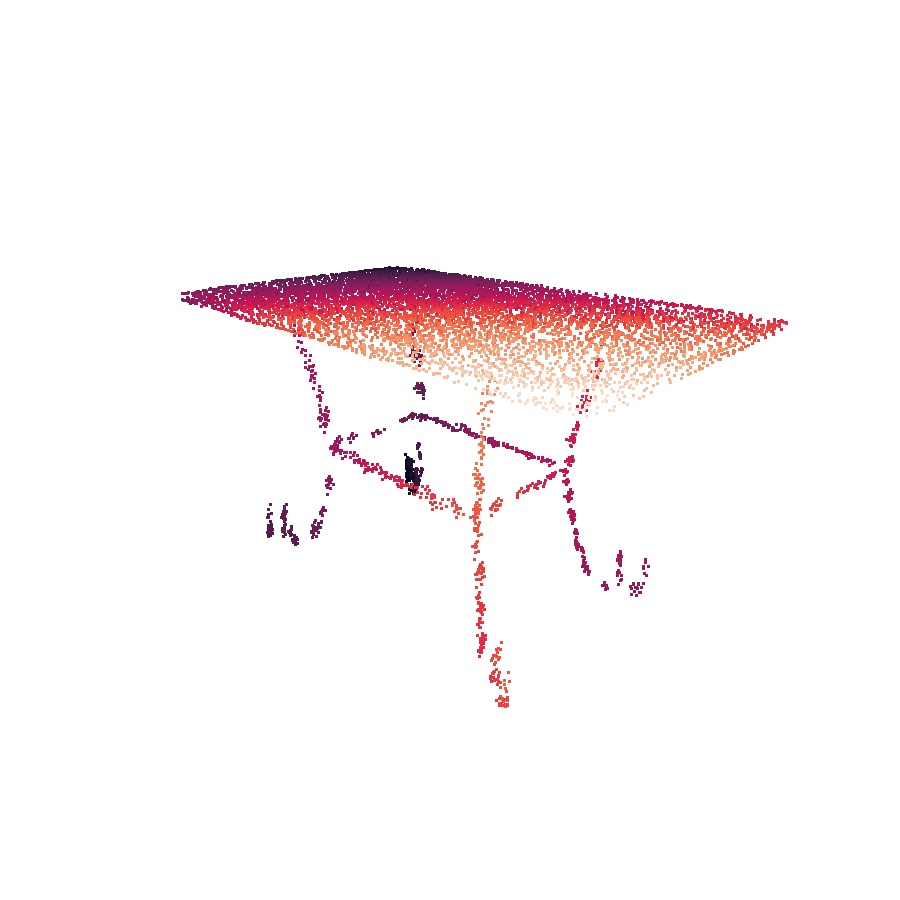}\end{subfigure} & \begin{subfigure}{0.086\textwidth}\centering\includegraphics[trim=100 150 200 100,clip,width=\textwidth]{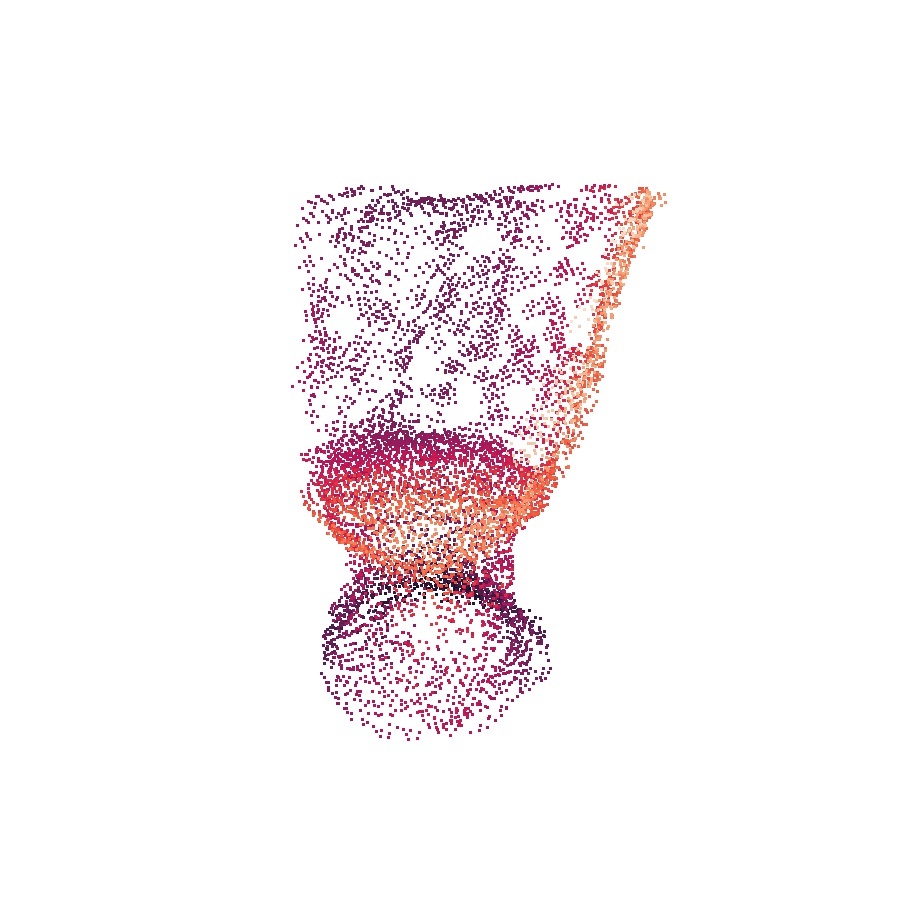}\end{subfigure} & \begin{subfigure}{0.11\textwidth}\centering\includegraphics[trim=200 300 100 250,clip,width=\textwidth]{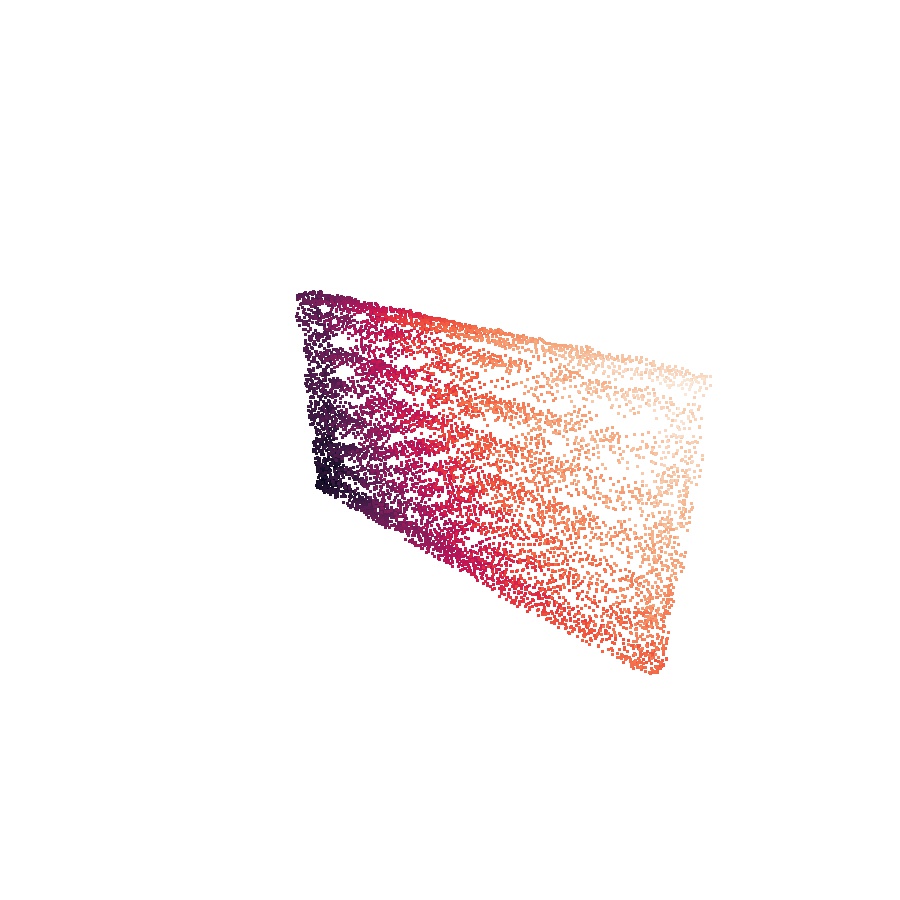}\end{subfigure} \\
GRNet & \begin{subfigure}{0.07\textwidth}\centering\includegraphics[trim=300 150 200 150,clip,width=\textwidth]{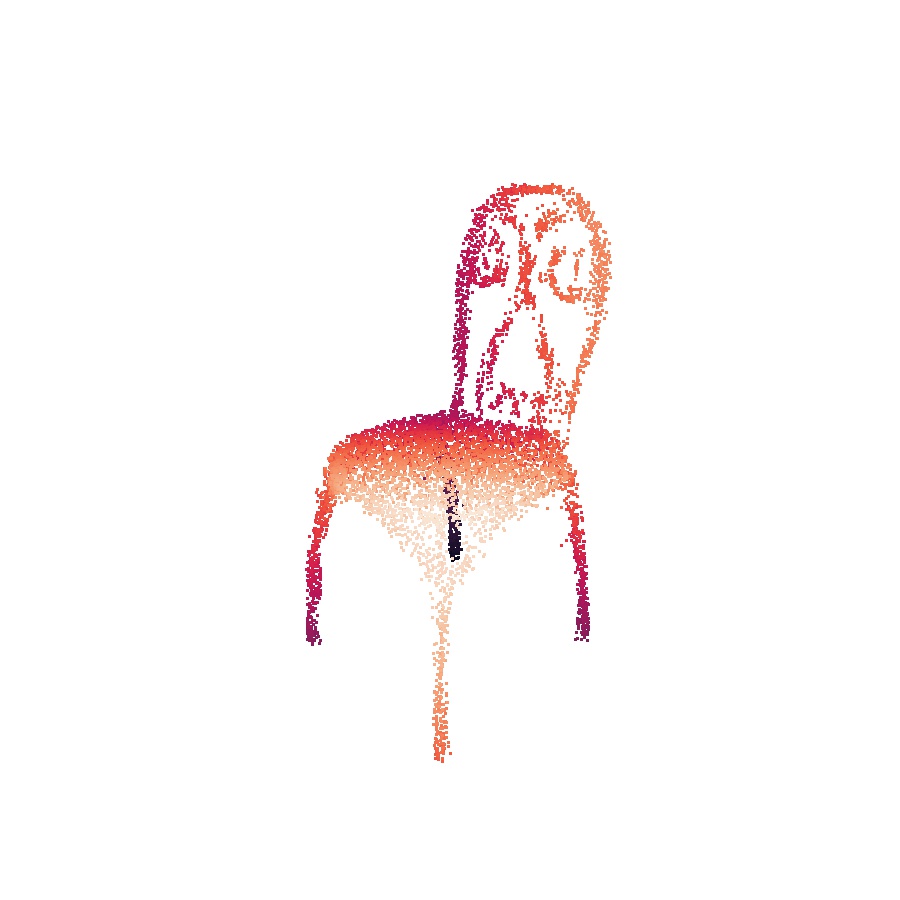}\end{subfigure} & \begin{subfigure}{0.077\textwidth}\centering\includegraphics[trim=250 150 200 150,clip,width=\textwidth]{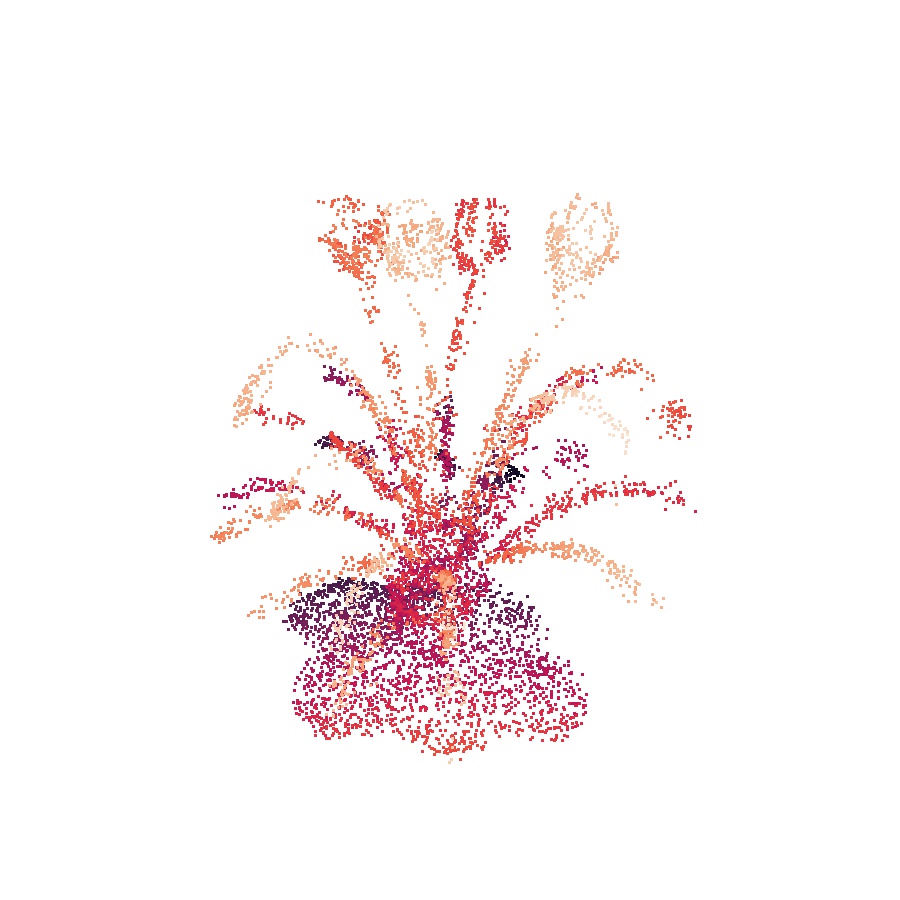}\end{subfigure} & \begin{subfigure}{0.071\textwidth}\centering\includegraphics[trim=300 150 190 140,clip,width=\textwidth]{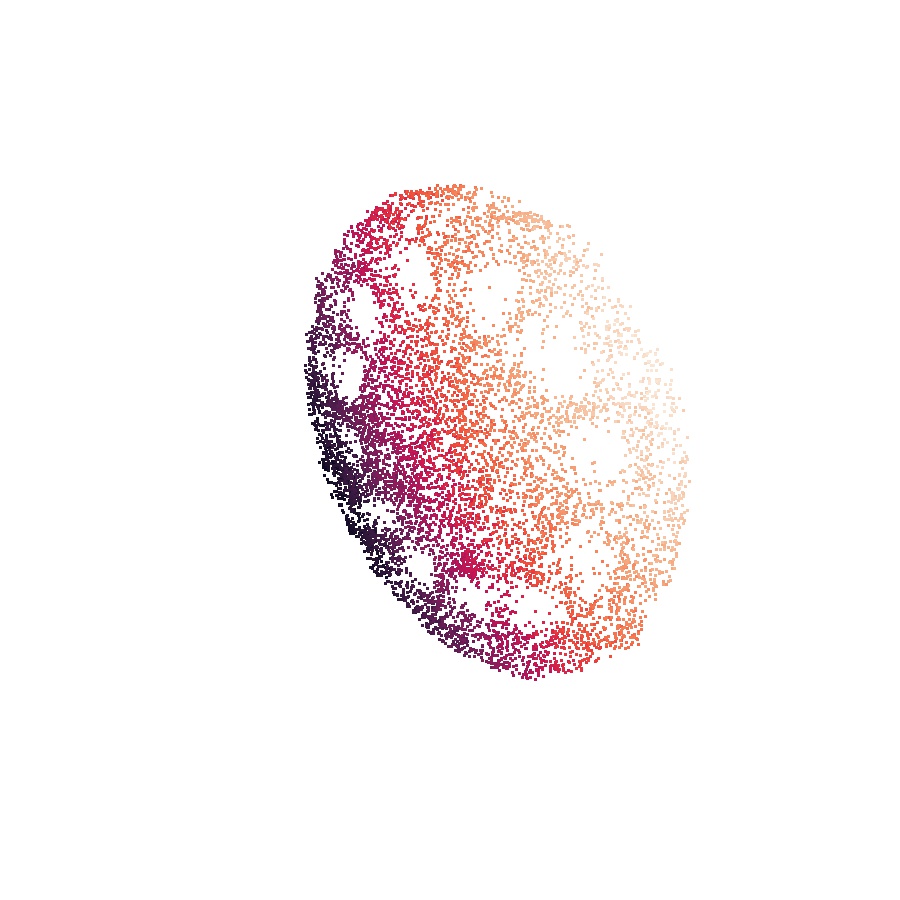}\end{subfigure} & \begin{subfigure}{0.074\textwidth}\centering\includegraphics[trim=220 170 230 110,clip,width=\textwidth]{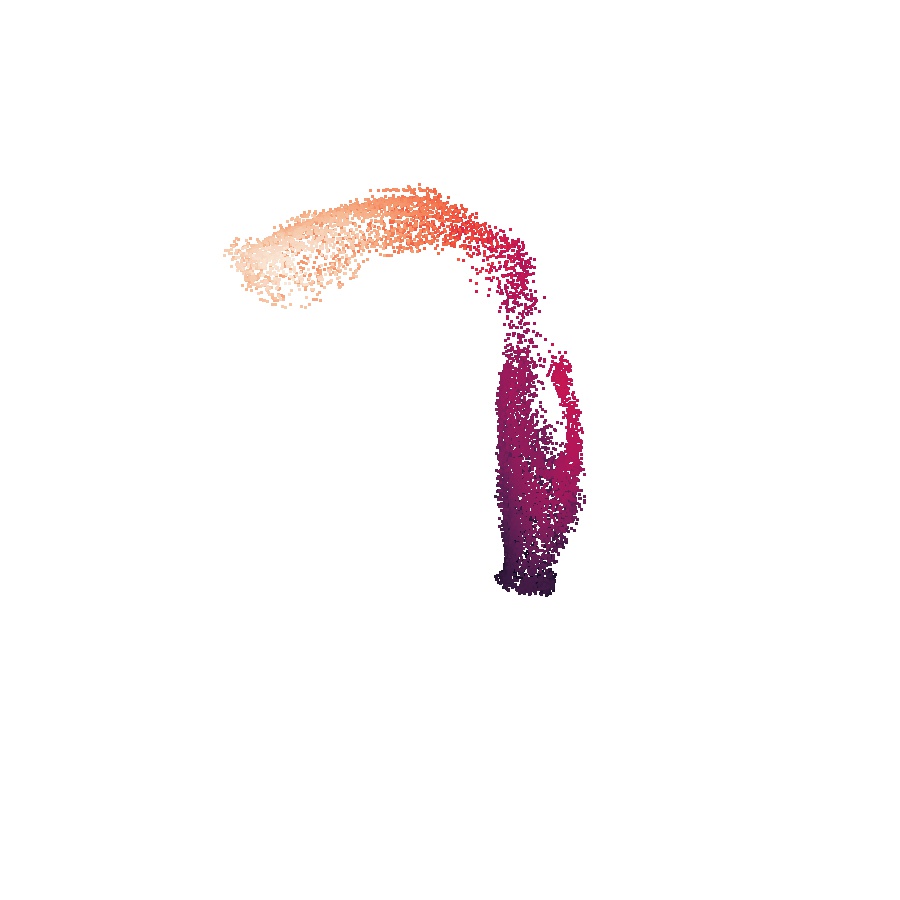}\end{subfigure} & \begin{subfigure}{0.12\textwidth}\centering\includegraphics[trim=270 280 200 370,clip,width=\textwidth]{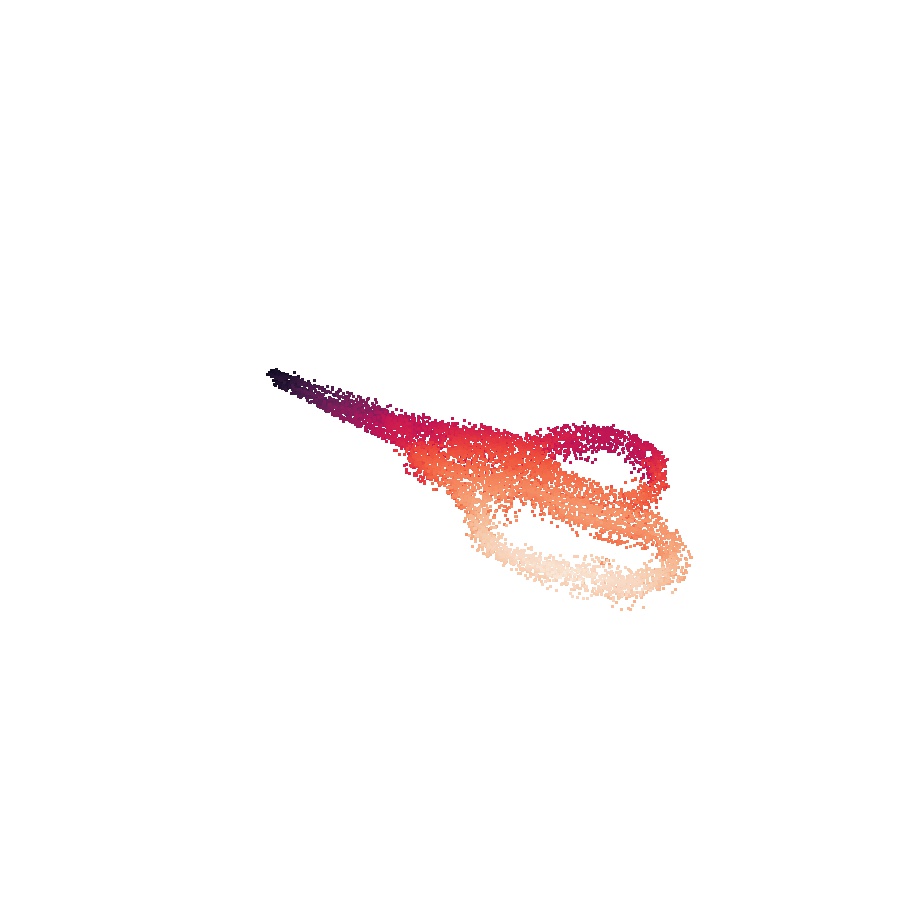}\end{subfigure} & \begin{subfigure}{0.082\textwidth}\centering\includegraphics[trim=150 200 250 250,clip,width=\textwidth]{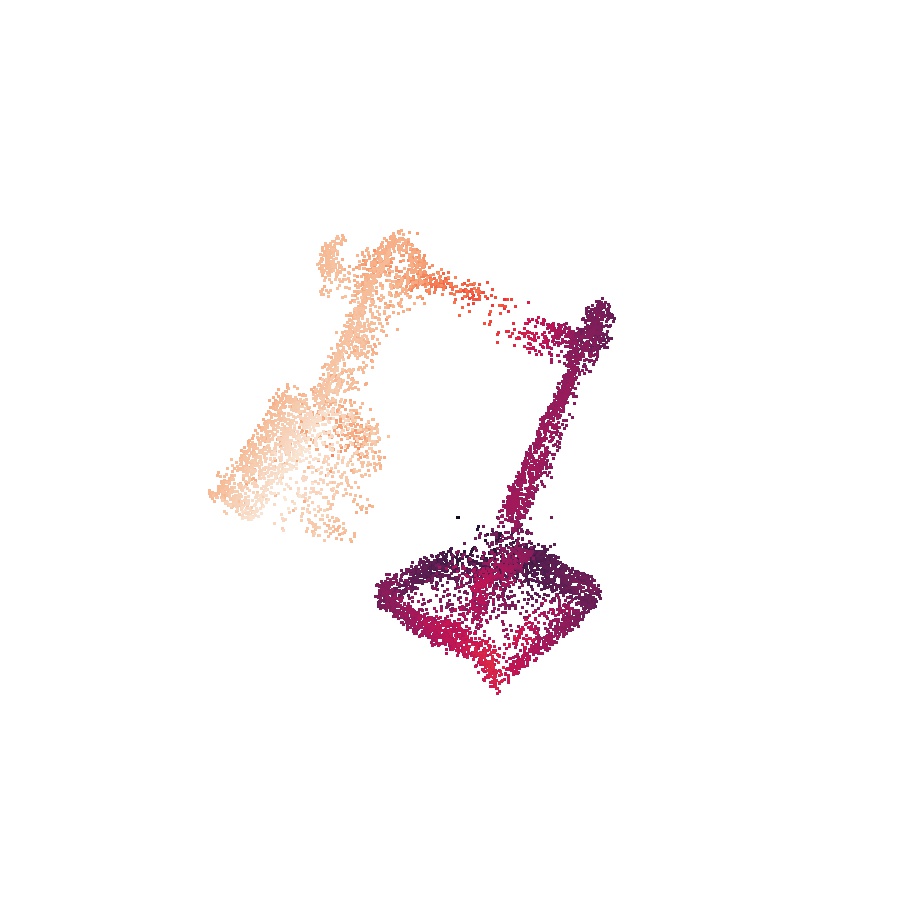}\end{subfigure} & \begin{subfigure}{0.05\textwidth}\centering\includegraphics[trim=300 200 300 200,clip,width=\textwidth]{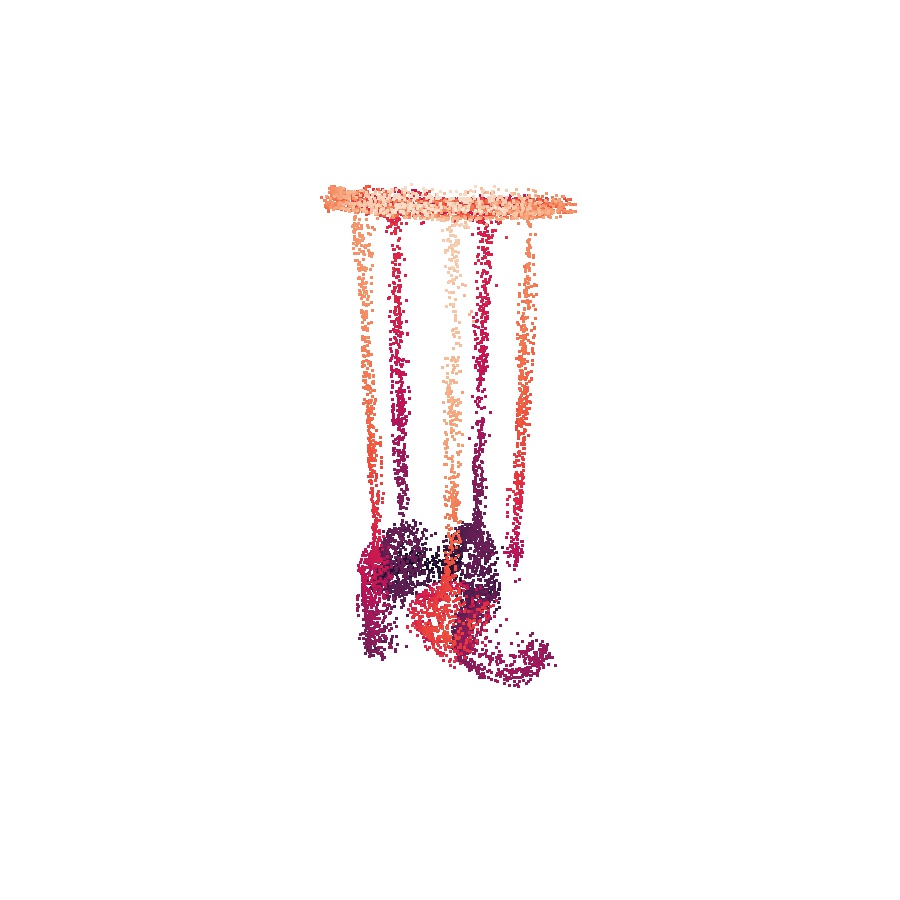}\end{subfigure} & \begin{subfigure}{0.095\textwidth}\centering\includegraphics[trim=200 250 200 250,clip,width=\textwidth]{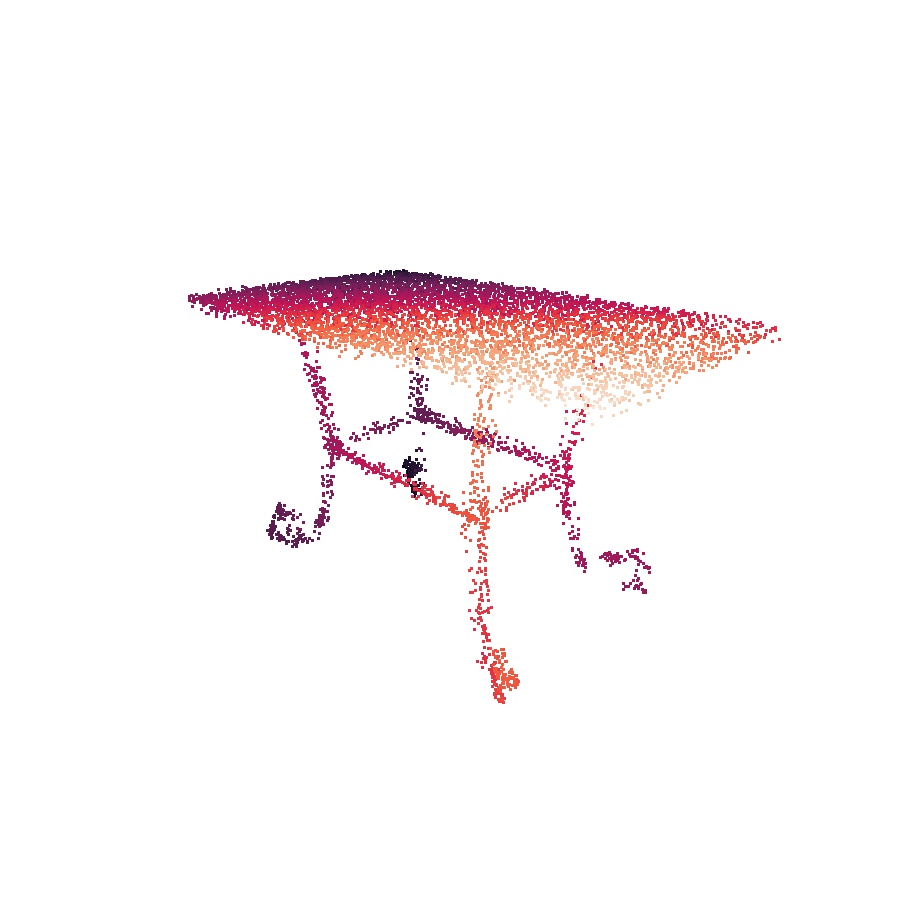}\end{subfigure} & \begin{subfigure}{0.086\textwidth}\centering\includegraphics[trim=100 150 200 100,clip,width=\textwidth]{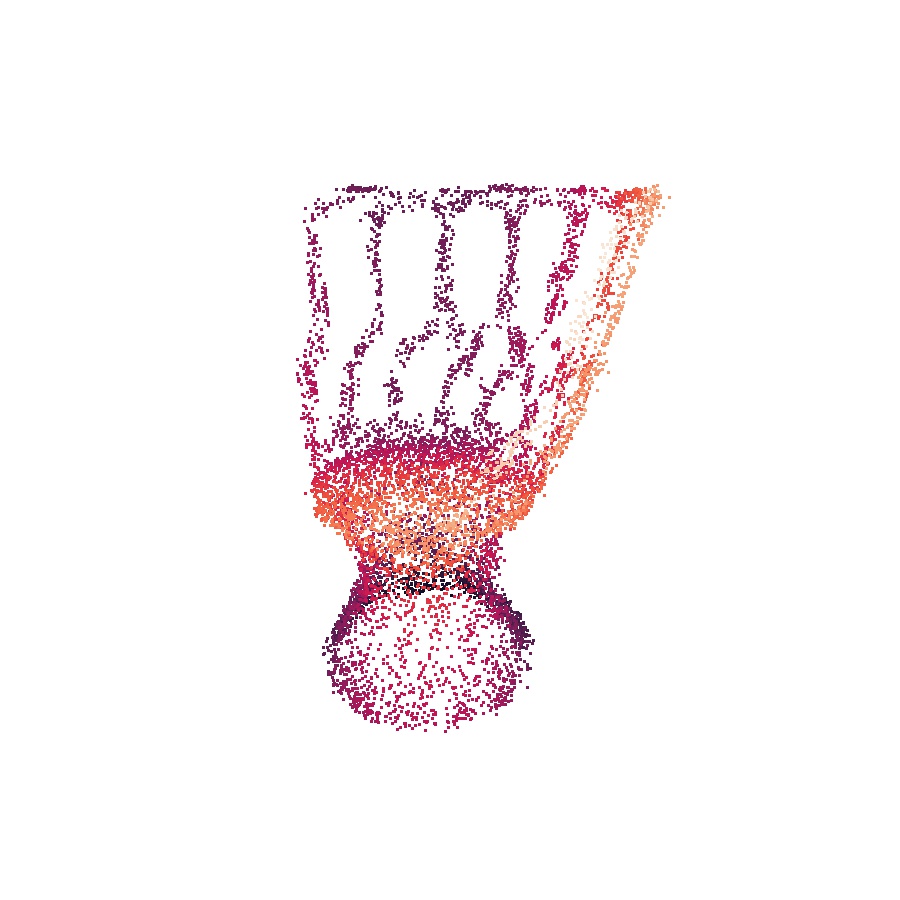}\end{subfigure} & \begin{subfigure}{0.11\textwidth}\centering\includegraphics[trim=200 300 100 250,clip,width=\textwidth]{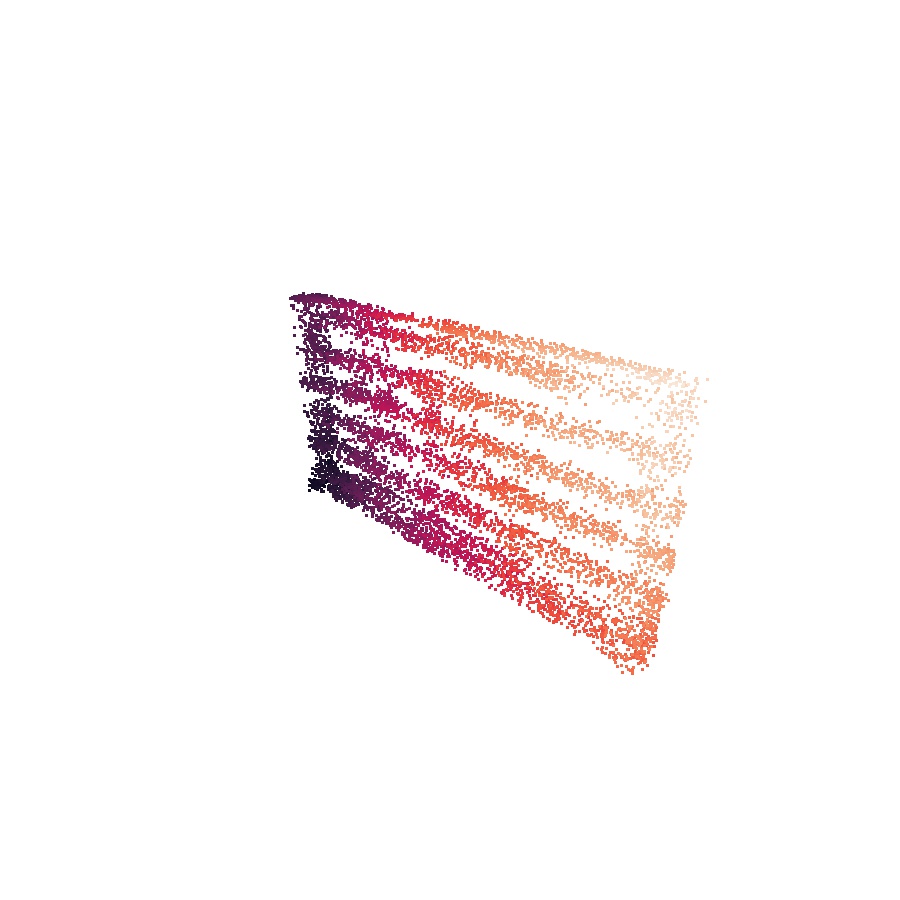}\end{subfigure} \\
Ours & \begin{subfigure}{0.07\textwidth}\centering\includegraphics[trim=300 150 200 150,clip,width=\textwidth]{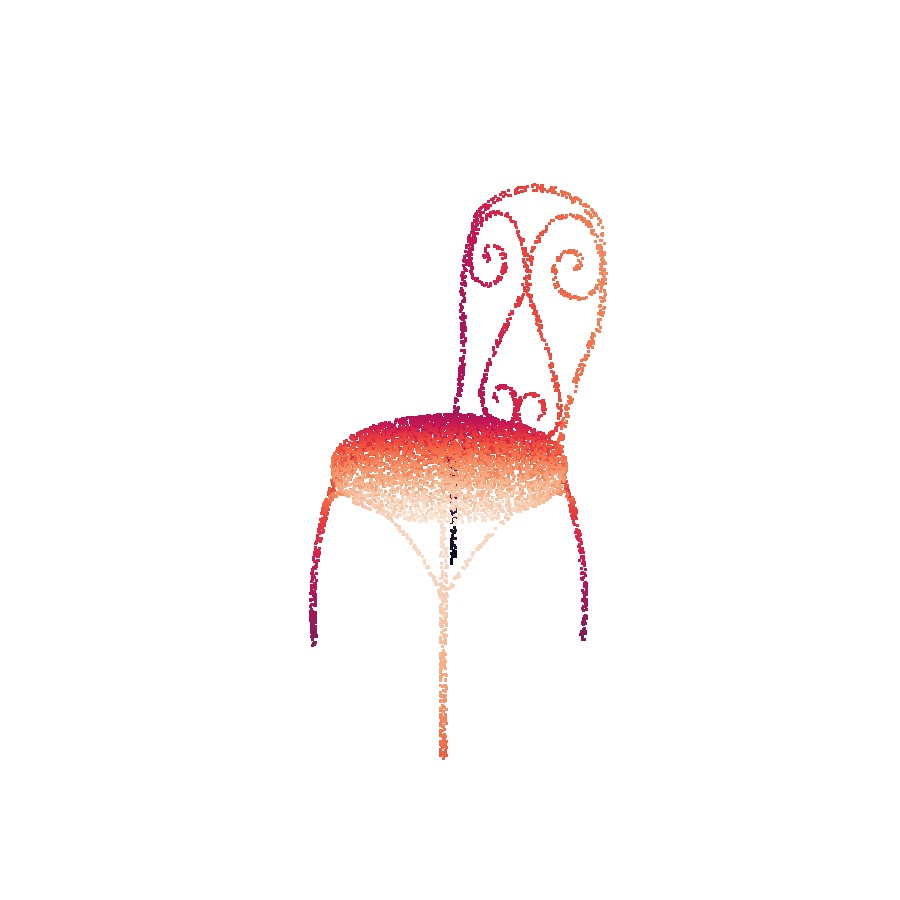}\end{subfigure} & \begin{subfigure}{0.077\textwidth}\centering\includegraphics[trim=250 150 200 150,clip,width=\textwidth]{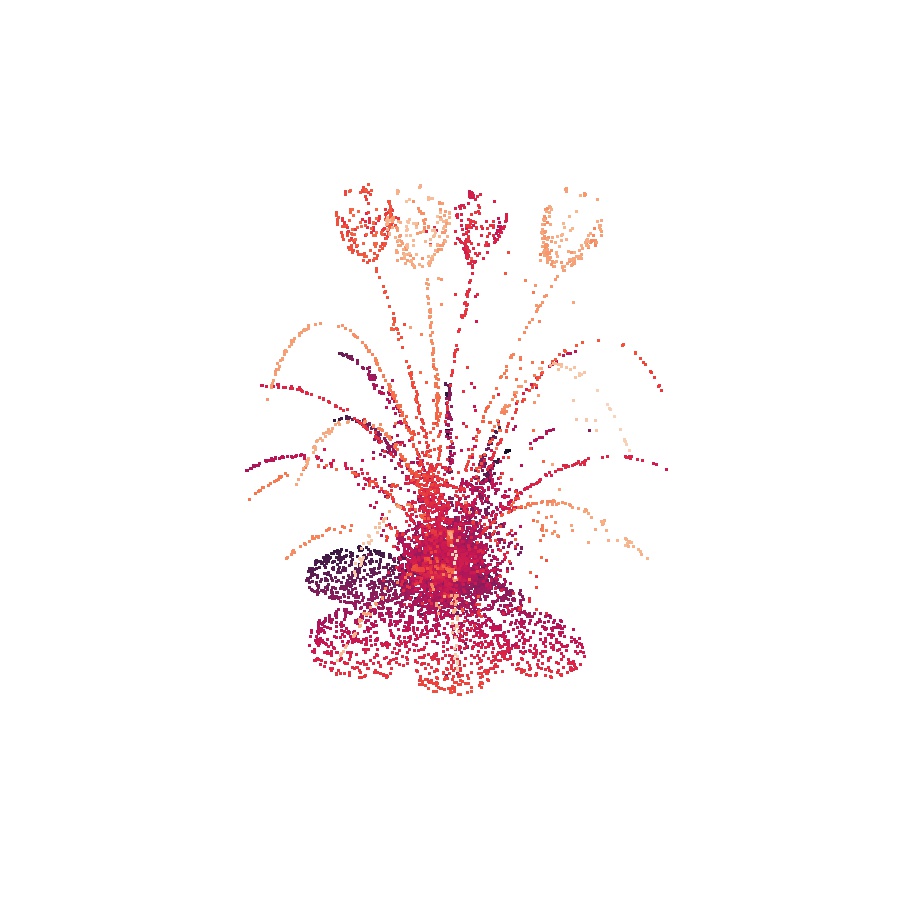}\end{subfigure} & \begin{subfigure}{0.071\textwidth}\centering\includegraphics[trim=300 150 190 140,clip,width=\textwidth]{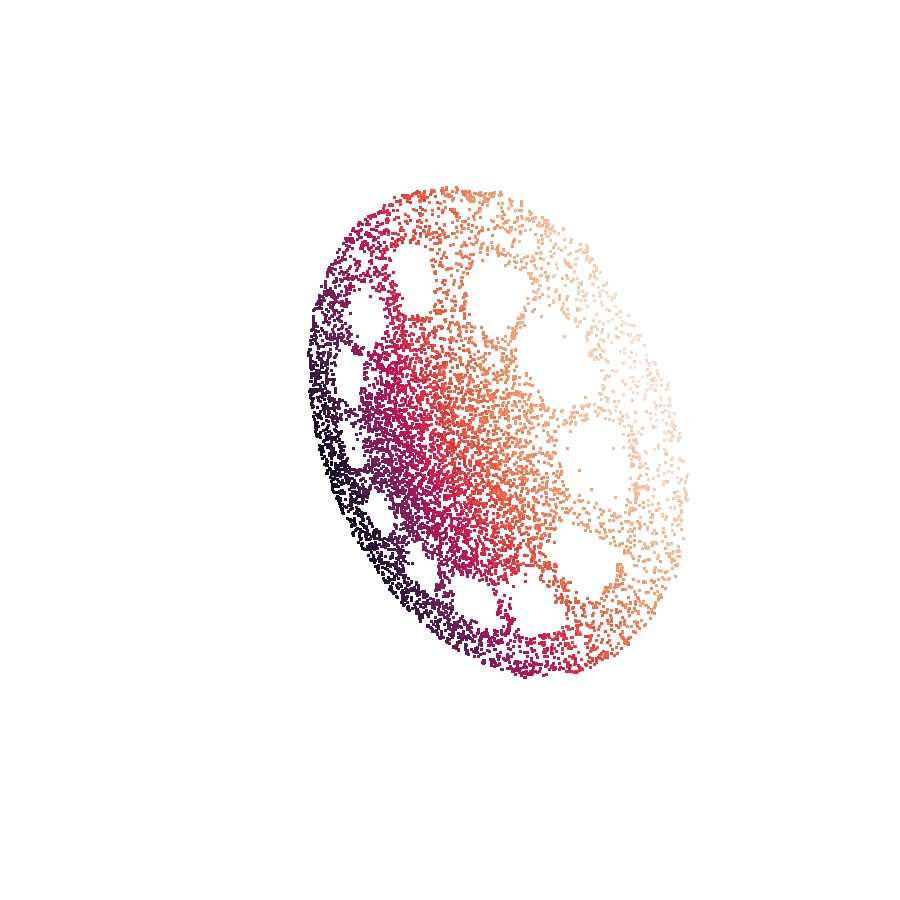}\end{subfigure} & \begin{subfigure}{0.074\textwidth}\centering\includegraphics[trim=220 170 230 110,clip,width=\textwidth]{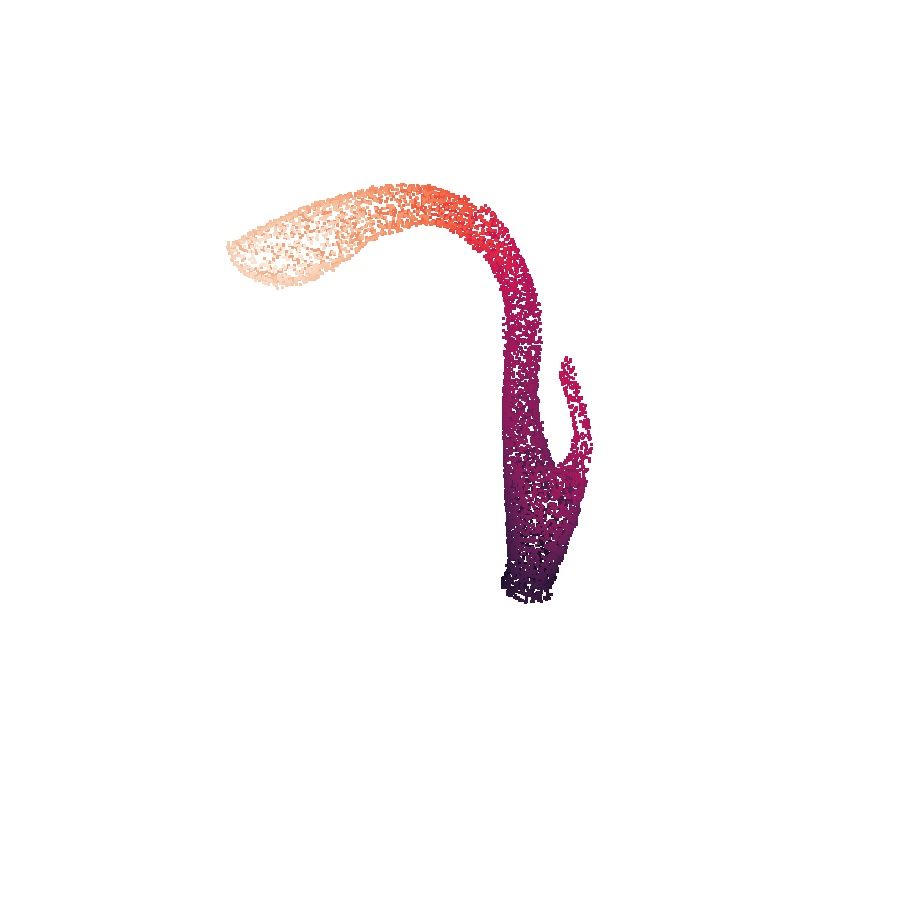}\end{subfigure} & \begin{subfigure}{0.12\textwidth}\centering\includegraphics[trim=270 280 200 370,clip,width=\textwidth]{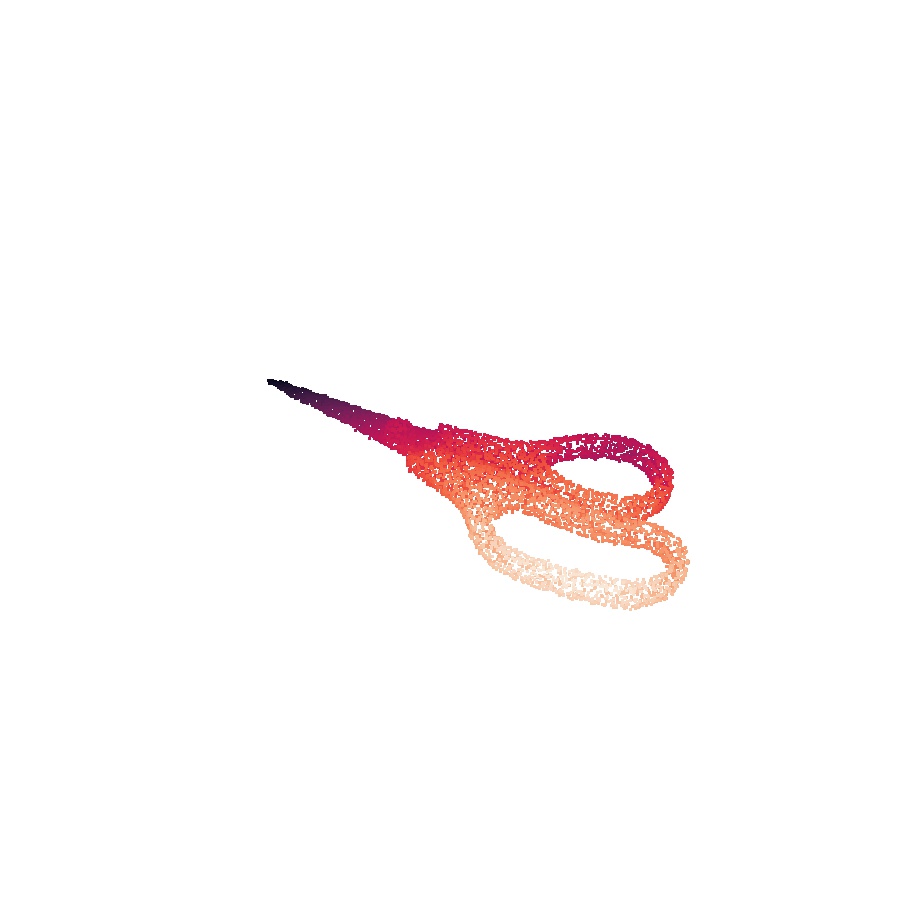}\end{subfigure} & \begin{subfigure}{0.082\textwidth}\centering\includegraphics[trim=150 200 250 250,clip,width=\textwidth]{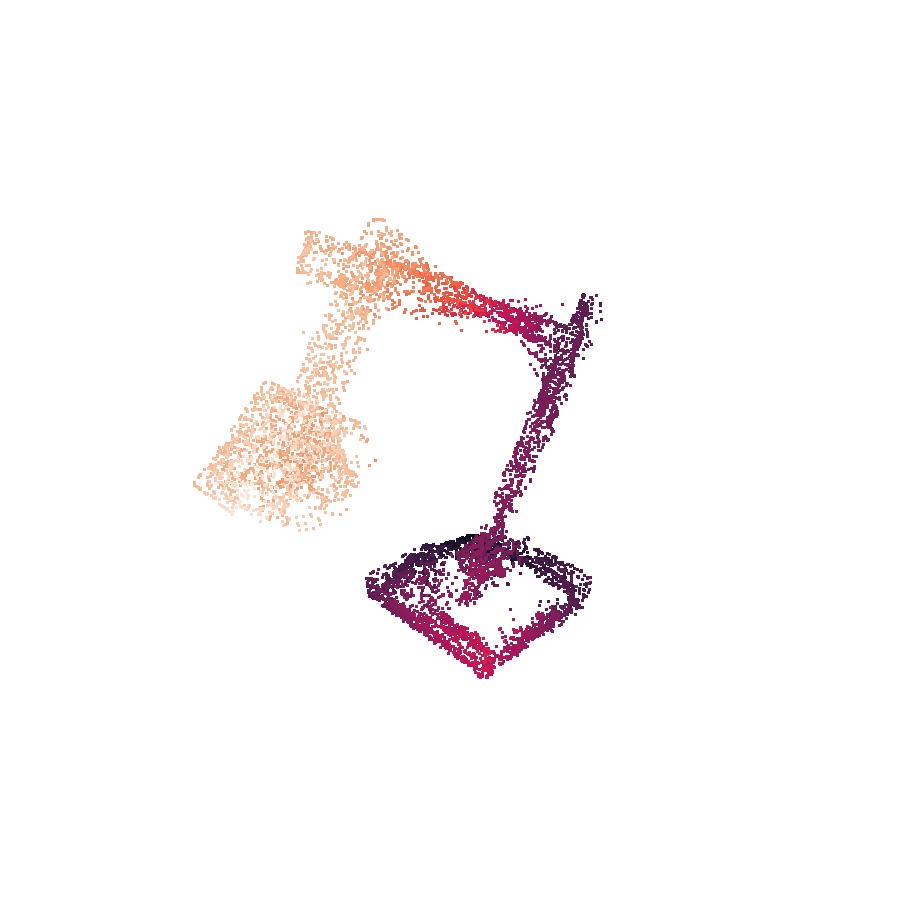}\end{subfigure} & \begin{subfigure}{0.05\textwidth}\centering\includegraphics[trim=300 200 300 200,clip,width=\textwidth]{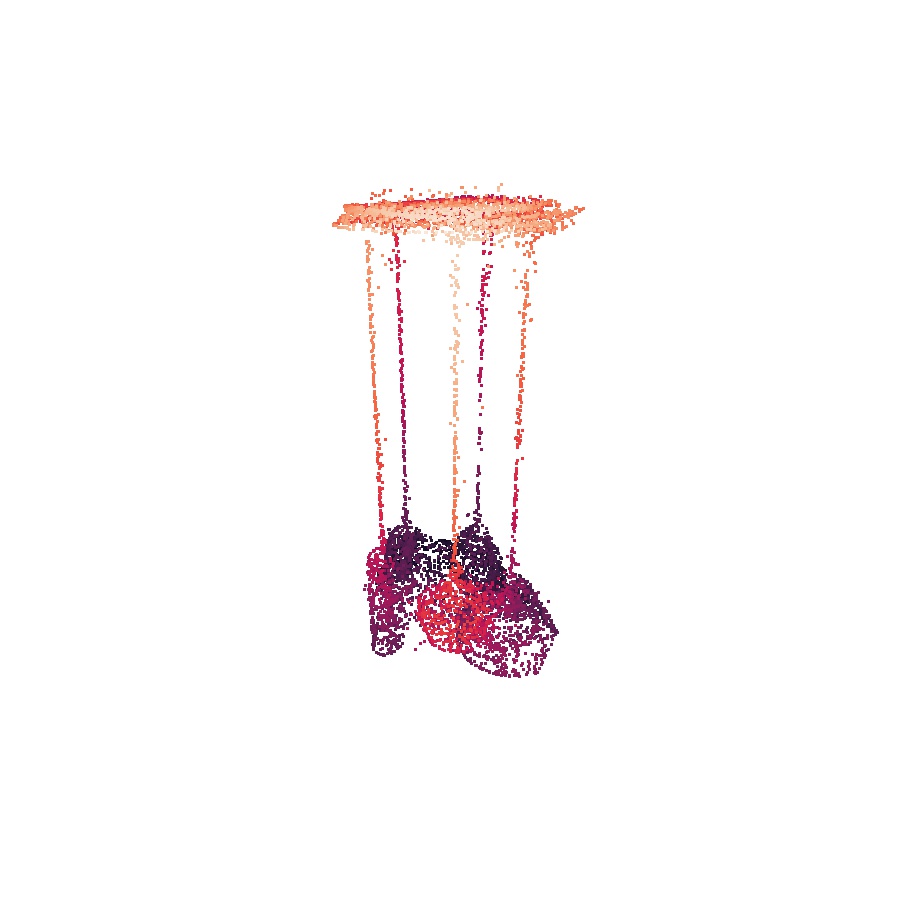}\end{subfigure} & \begin{subfigure}{0.095\textwidth}\centering\includegraphics[trim=200 250 200 250,clip,width=\textwidth]{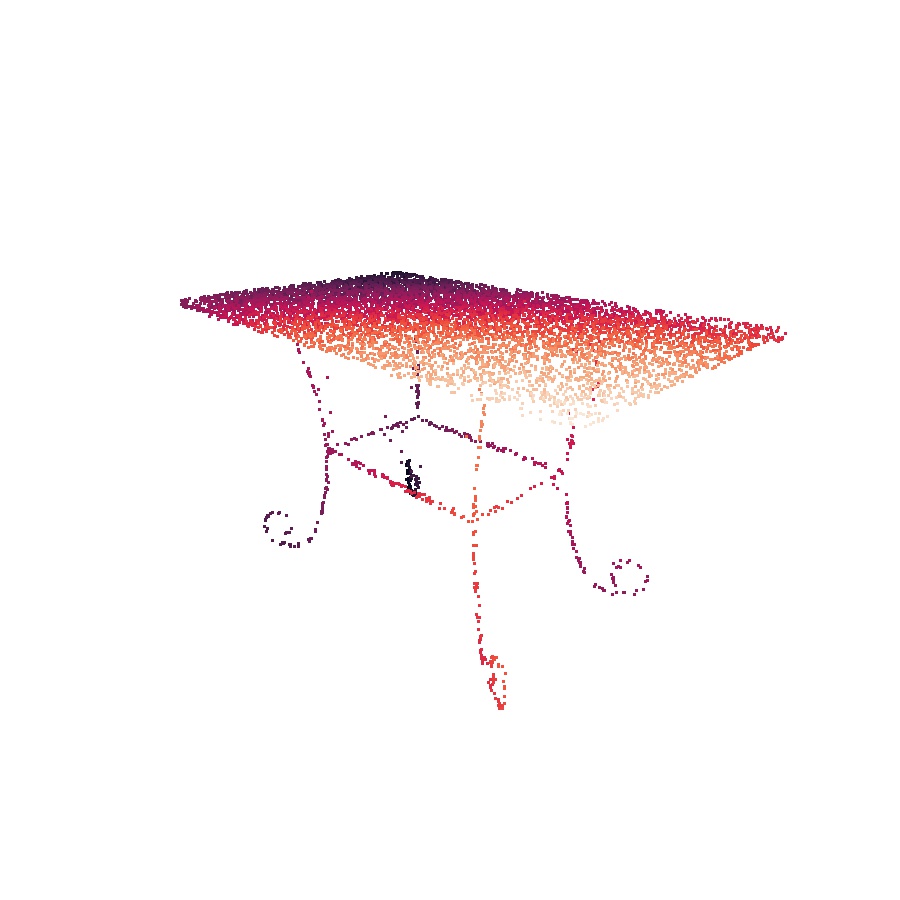}\end{subfigure} & \begin{subfigure}{0.086\textwidth}\centering\includegraphics[trim=100 150 200 100,clip,width=\textwidth]{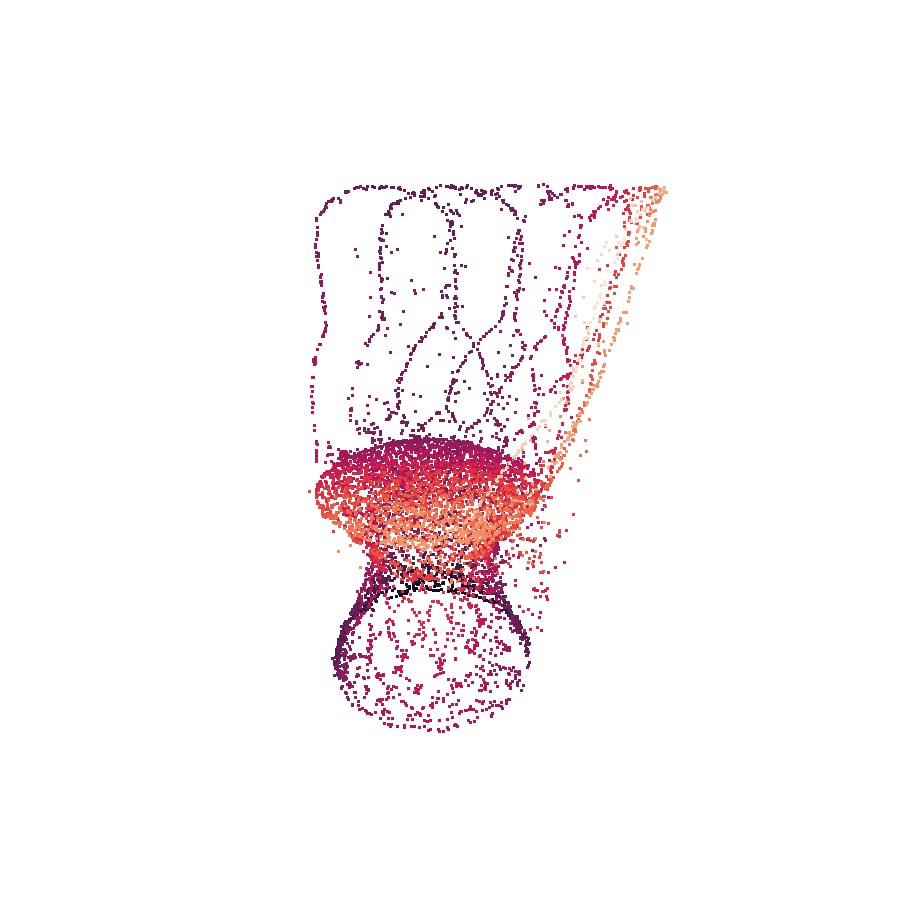}\end{subfigure} & \begin{subfigure}{0.11\textwidth}\centering\includegraphics[trim=200 300 100 250,clip,width=\textwidth]{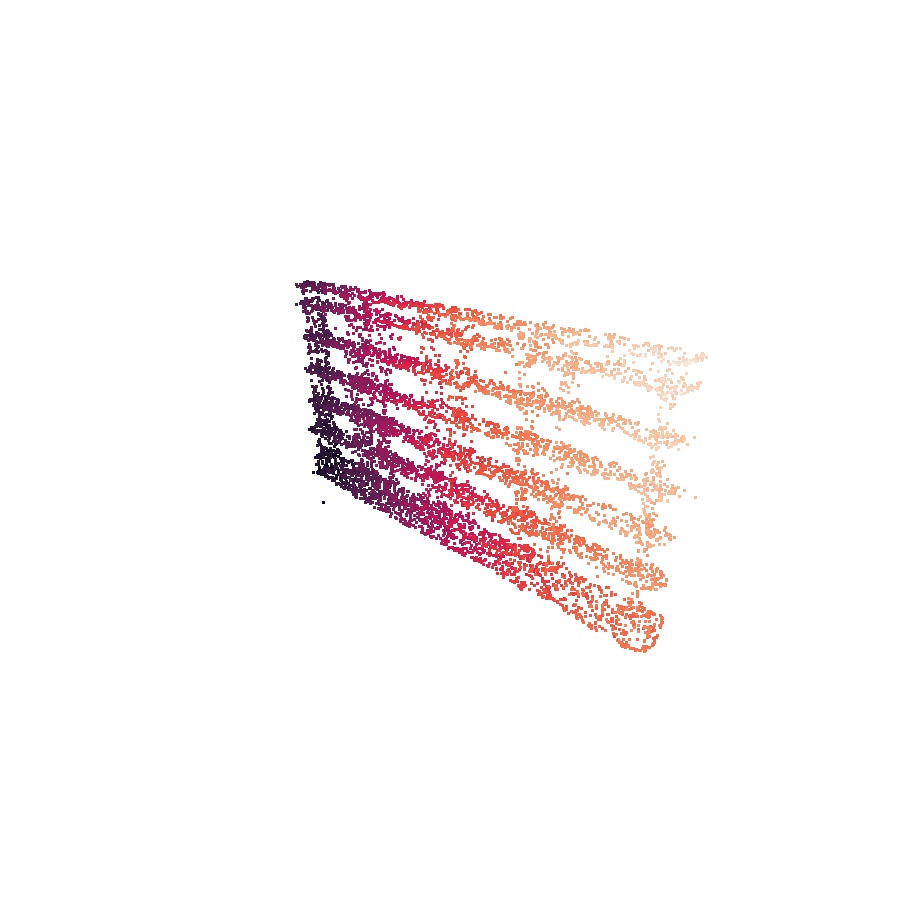}\end{subfigure} \\
GT & \begin{subfigure}{0.07\textwidth}\centering\includegraphics[trim=300 150 200 150,clip,width=\textwidth]{fig/2254_gt}\end{subfigure} & \begin{subfigure}{0.077\textwidth}\centering\includegraphics[trim=250 150 200 150,clip,width=\textwidth]{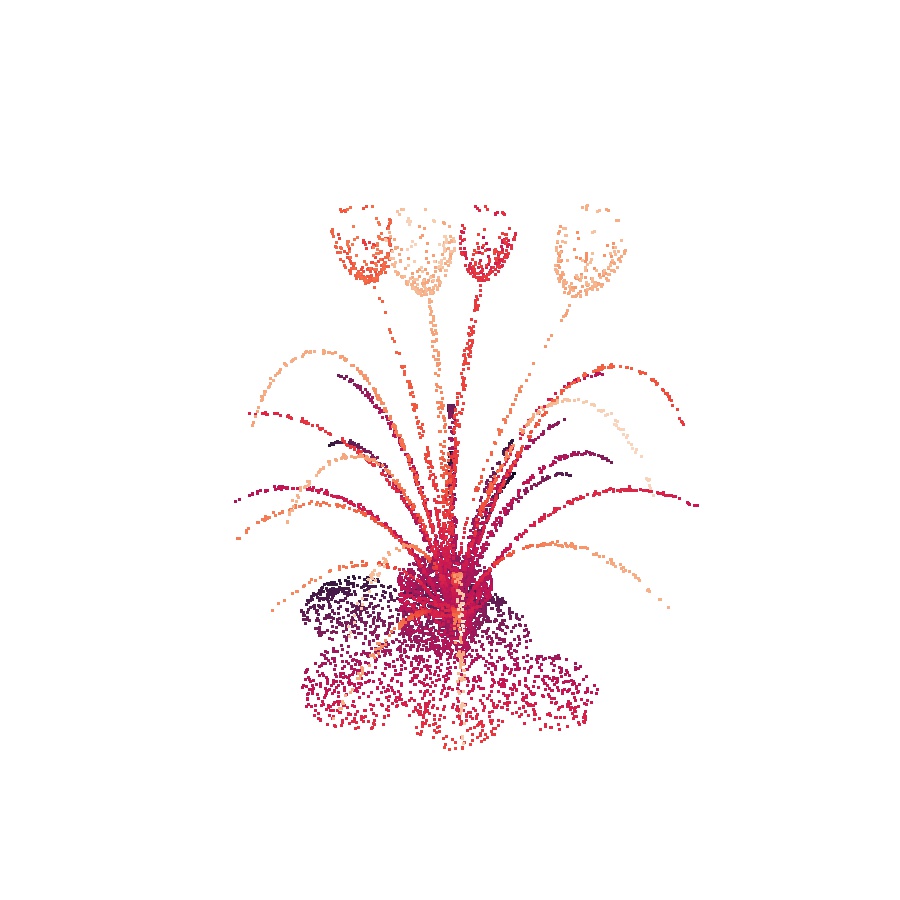}\end{subfigure} & \begin{subfigure}{0.071\textwidth}\centering\includegraphics[trim=300 150 190 140,clip,width=\textwidth]{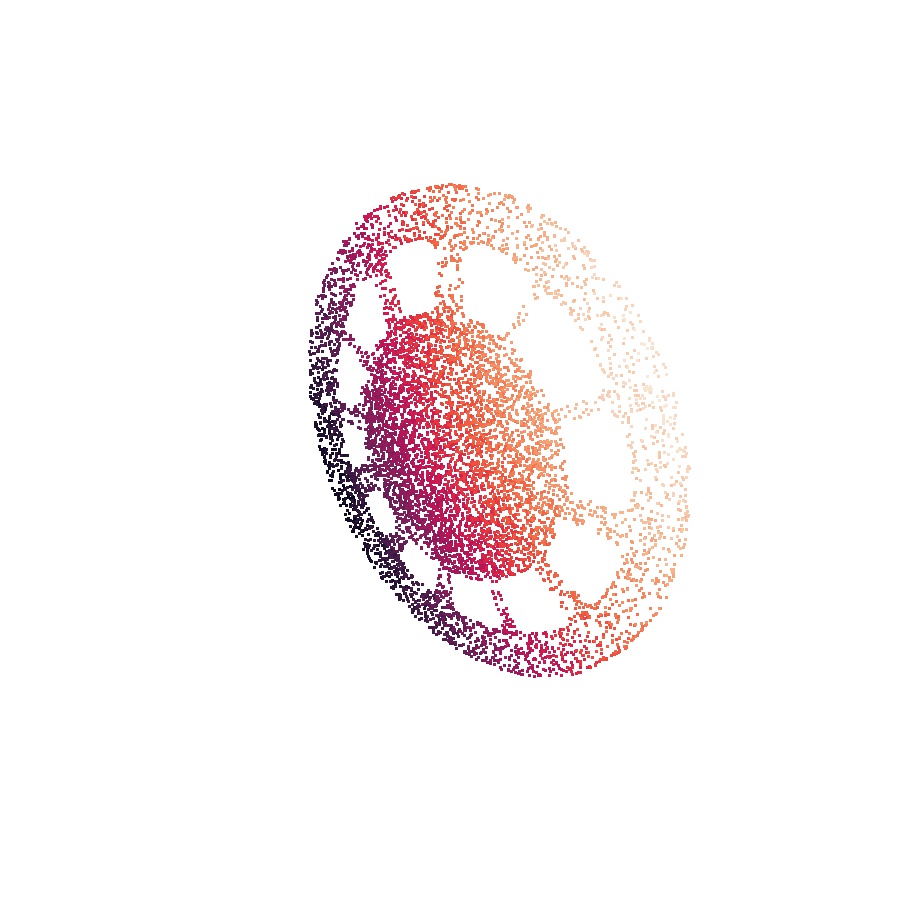}\end{subfigure} & \begin{subfigure}{0.074\textwidth}\centering\includegraphics[trim=220 170 230 110,clip,width=\textwidth]{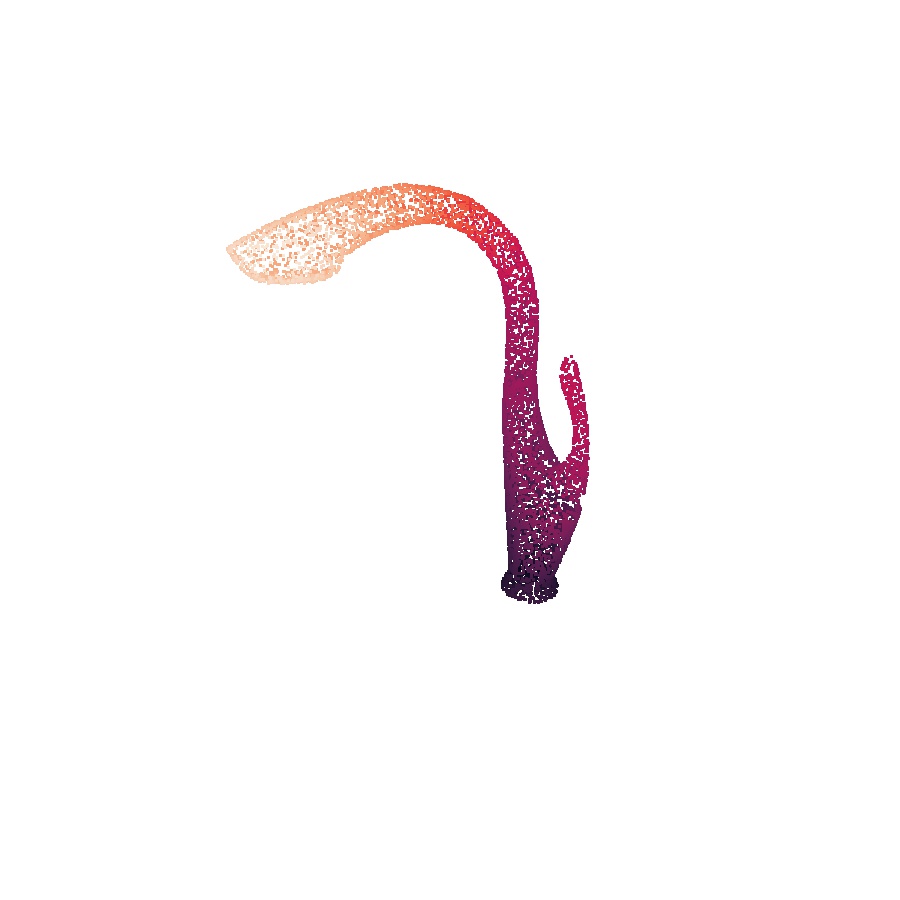}\end{subfigure} & \begin{subfigure}{0.12\textwidth}\centering\includegraphics[trim=270 280 200 370,clip,width=\textwidth]{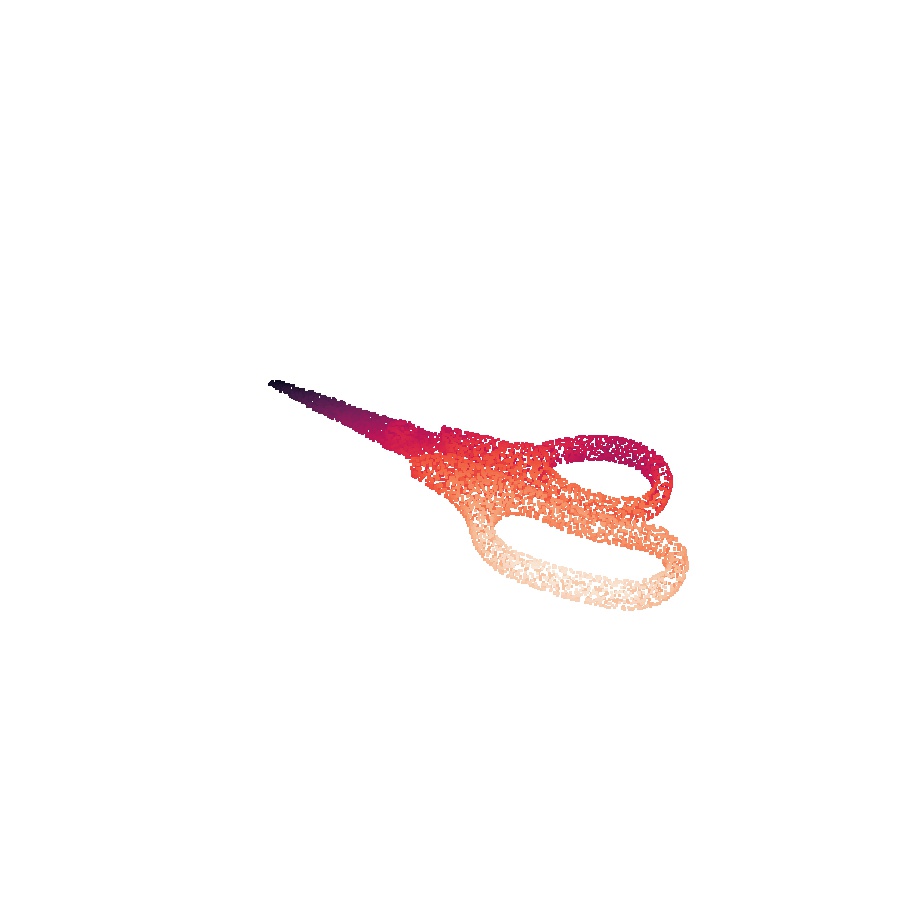}\end{subfigure} & \begin{subfigure}{0.082\textwidth}\centering\includegraphics[trim=150 200 250 250,clip,width=\textwidth]{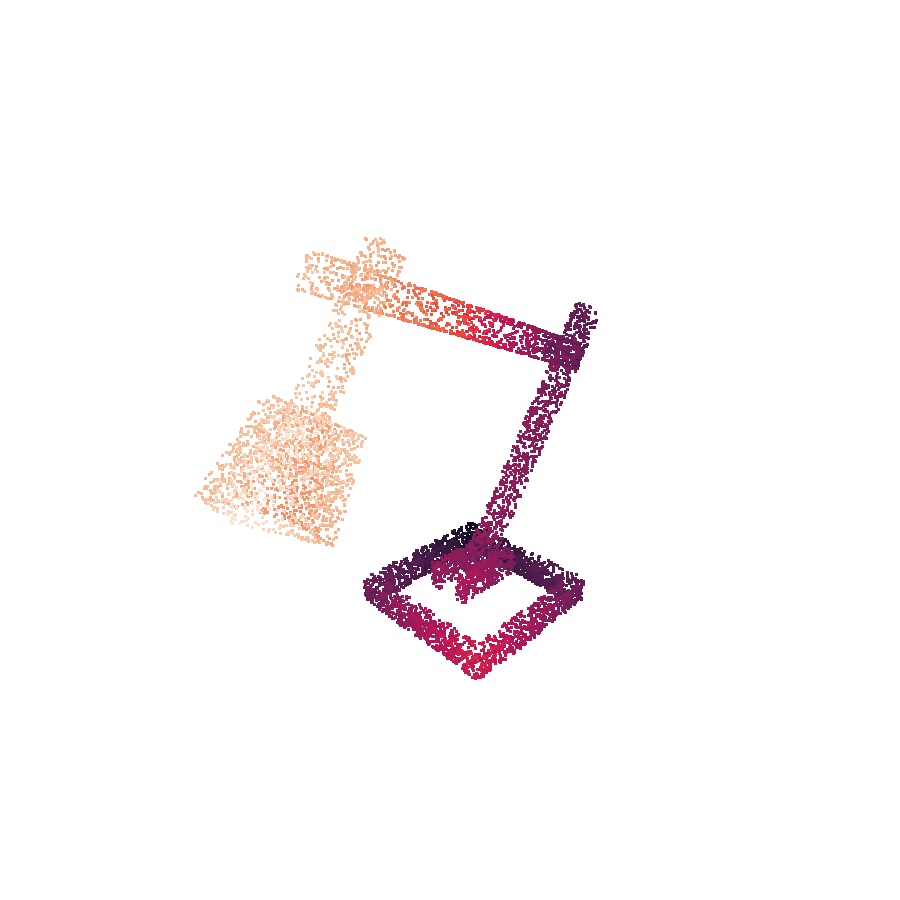}\end{subfigure} & \begin{subfigure}{0.05\textwidth}\centering\includegraphics[trim=300 200 300 200,clip,width=\textwidth]{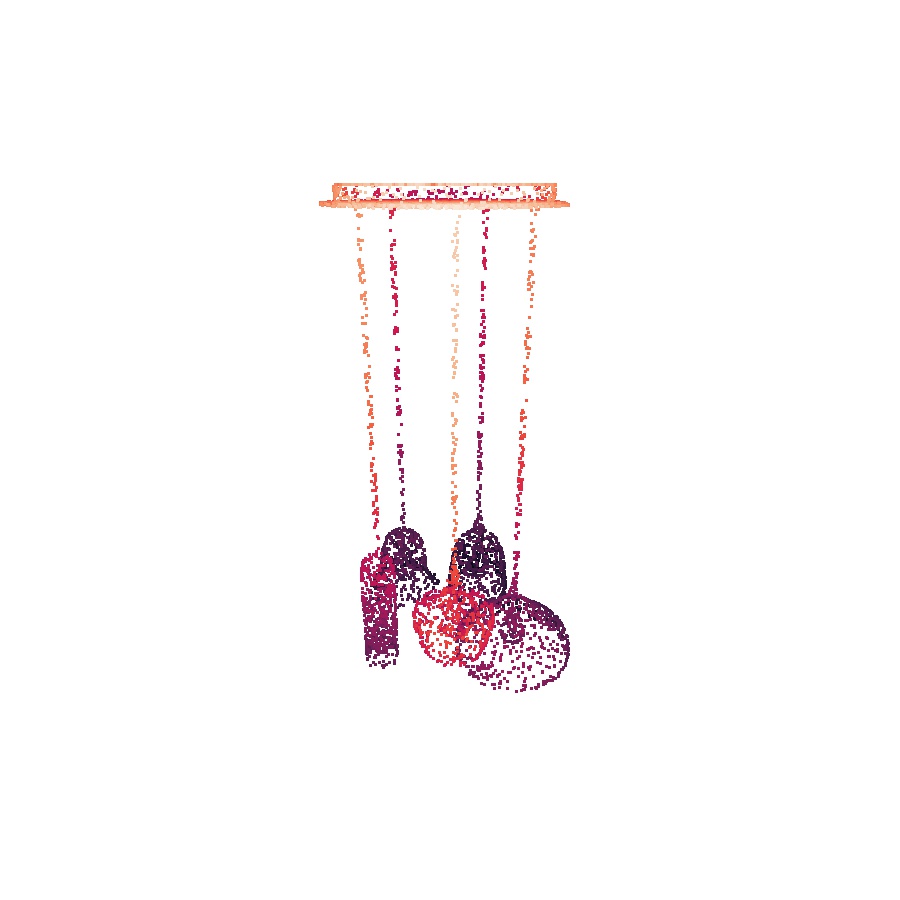}\end{subfigure} & \begin{subfigure}{0.095\textwidth}\centering\includegraphics[trim=200 250 200 250,clip,width=\textwidth]{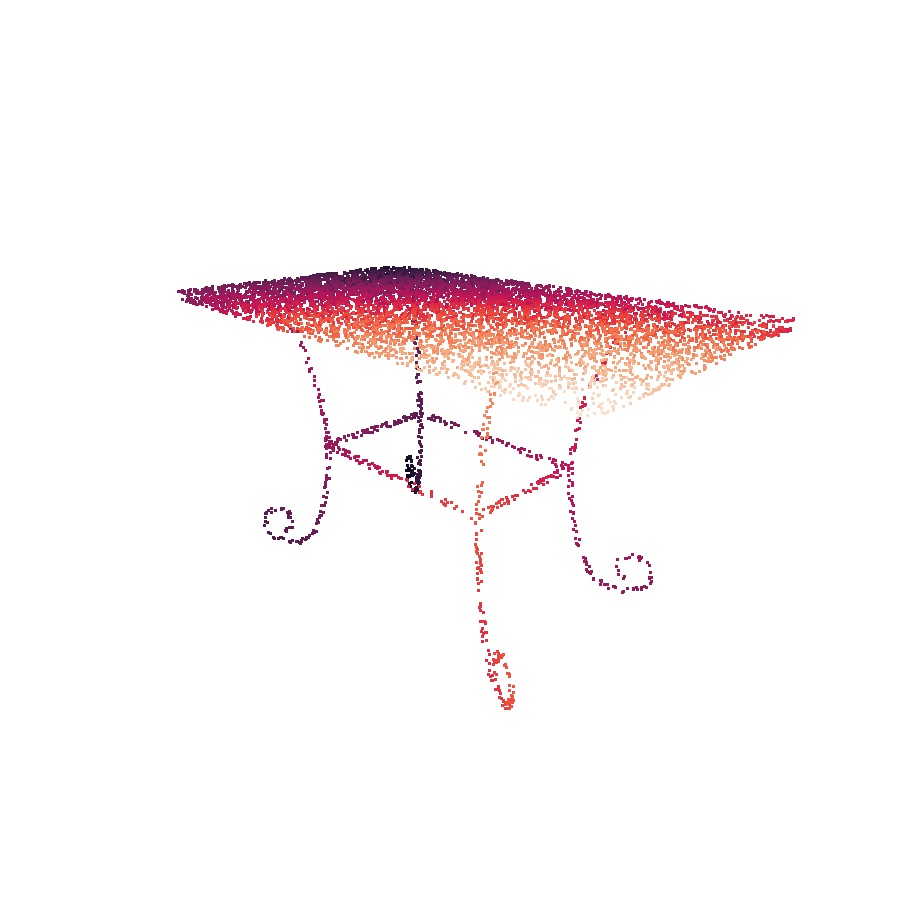}\end{subfigure} & \begin{subfigure}{0.086\textwidth}\centering\includegraphics[trim=100 150 200 100,clip,width=\textwidth]{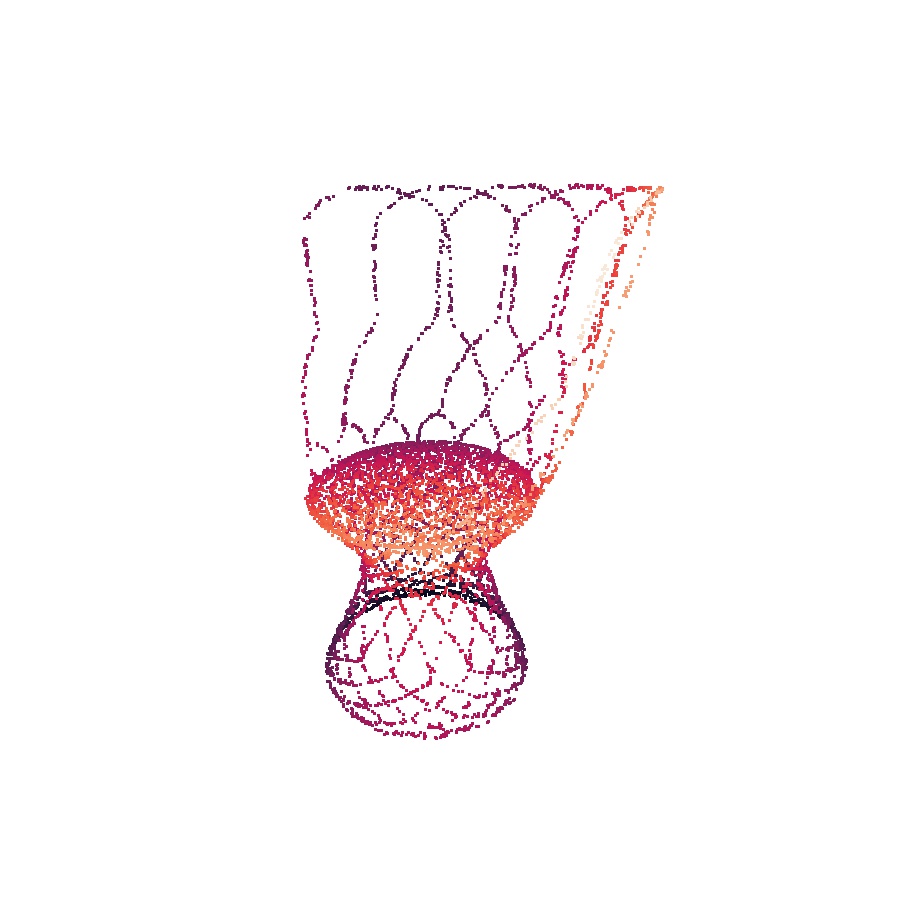}\end{subfigure} & \begin{subfigure}{0.11\textwidth}\centering\includegraphics[trim=200 300 100 250,clip,width=\textwidth]{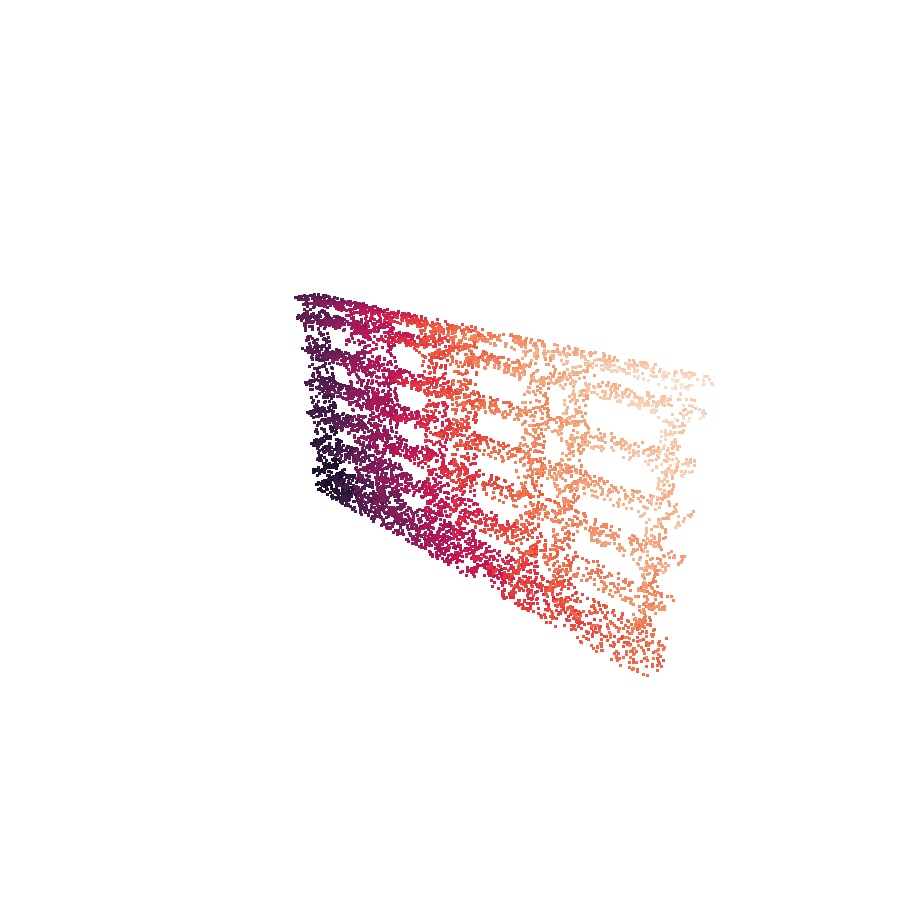}\end{subfigure} 
\end{tabular}

\vspace{-2pt}
\caption{Comparisons of different methods on point cloud completion. Note that 8,192 points are exported from each method for comparison, except SoftPoolNet (4,096 points) due to its network specification.}
\label{fig:sota_fig_big}
\vspace{-6pt}
\end{figure*}

\subsection{Comparison with Existing Methods}
We list the qualitative results in Figure~\ref{fig:sota_fig_big}. For quantitative evaluation, since some methods (PCN~\cite{yuan2018pcn}, CRN~\cite{wang2020cascaded}, GR-Net~\cite{xie2020grnet}, and MSN~\cite{liu2020morphing}) support upsampling to recover higher resolution of outputs, we compare our methods with them under the resolution of 8,192. In these methods, \textbf{PCN} completes the partial point cloud by using a stacked version of PointNet layers \cite{qi2017pointnet} to construct an auto-encoder. \textbf{CRN} combines local details of partial input with the global shape feature to synthesize detailed object shapes with a coarse-to-fine strategy. Similarly, \textbf{MSN} also recovers shapes from coarse-to-fine and involves a joint loss function to ensure even point distribution. \textbf{GRNet} is the recent approach that applies 3D convolutions to process shapes on volumetric grids. We report the comparisons using EMD \cite{liu2020morphing} and CD scores \cite{fan2017point} in Table~\ref{tab:sota-table-emd-8k} and Table~\ref{tab:sota-table-cd-8k} respectively.

For the other methods that do not support upsampling (including PF-Net~\cite{huang2020pf}, P2P-Net\cite{yin2018p2p}, SoftPoolNet\cite{wang2020softpoolnet}), we downsample the output of all the methods to the resolution of 2,048 to enable a fair comparison.
\textbf{SoftPoolNet} employs a similar encoder as PCN~\cite{yuan2018pcn} but aggregates soft pooling layers as the activation function instead of max-pooling.
\textbf{PF-Net} designs a multi-resolution pyramid decoder to recover the missing geometries on different scales. \textbf{P2P-Net} generates compact completion results by learning the bidirectional deformation between the input partial point cloud and the complete point cloud, but it struggles to recover the detailed structure, especially on invisible areas. The quantitative scores on EMD and CD are respectively listed in Table~\ref{tab:sota-table-emd-2k} and Table~\ref{tab:sota-table-cd-2k}.

\begin{table}[t]
	\centering
	\small
	\setlength{\tabcolsep}{2pt}
	\begin{tabular}{c|cccccc|c}
		\Xhline{2\arrayrulewidth}
		methods     & faucet & cabinet & table & chair & vase  & lamp  & average \\ \hline
		PCN         & 16.49  & 9.34    & 11.34 & 10.94 & 12.43 & 16.10 & 12.77    \\
		PCN+Ray     & 13.63  & 8.25    & 10.79 &  9.74 & 11.10 & 14.30 & 11.30    \\
		CRN    & 13.43  & 9.85    &  7.93 &  8.67 & 12.49 & 11.38 & 10.63    \\
		GRNet       & 10.36  & 7.75    &  7.50 &  7.74 & 11.21 & 10.74 &  9.22    \\
		MSN         & 7.71   & 6.70    &  6.52 &  6.57 &  6.89 &  7.55 &  6.99    \\
		\hline
		Ours  &  \textbf{6.31}  & \textbf{6.14}    &  \textbf{5.33} &  \textbf{5.12} &  \textbf{5.93} &  \textbf{6.76} & \textbf{5.93}
		\\ \Xhline{2\arrayrulewidth}
	\end{tabular}
	\vspace{-1pt}
    \caption{Evaluation on EMD ($\times 10^2$) with Res.=8,192 }
	\label{tab:sota-table-emd-8k}
	\vspace{-3pt}
\end{table}

\begin{table}[t]
	\centering
	\small
	\setlength{\tabcolsep}{2pt}
	\begin{tabular}{c|cccccc|c}
		\Xhline{2\arrayrulewidth}
		methods     & faucet & cabinet & table & chair & vase  & lamp  & average \\ \hline
		PCN         & 4.17   & 4.67    &  3.82 & 4.01  &  6.31 &  3.73 & 4.45    \\
		PCN+Ray     & 2.80   & 4.55    &  3.57 & 3.81  &  5.80 &  3.12 & 3.94    \\
		CRN    & 3.67   & \textbf{4.49}    &  \textbf{3.44} & 3.81  &  5.49 &  3.19 & 4.01    \\
		GRNet       & 3.28   & 4.66    &  3.73 & 3.94  &  5.53 &  3.52 & 4.11    \\
		MSN         & 4.02   & 5.75    &  4.61 & 4.81  &  5.71 &  4.34 & 4.87    \\
		\hline
		Ours  & \textbf{2.62}   & 4.72    &  3.76 & \textbf{3.62}  &  \textbf{4.54} &  \textbf{3.02} & \textbf{3.71}
		\\ \Xhline{2\arrayrulewidth}
	\end{tabular}
    \vspace{-1pt}
	\caption{Evaluation on CD ($\times 10^2$) with Res.=8,192}
	\label{tab:sota-table-cd-8k}
	\vspace{-3pt}
\end{table}

\begin{table}[t]
	\centering
	\small
	\setlength{\tabcolsep}{2pt}
	\begin{tabular}{c|cccccc|c}
		\Xhline{2\arrayrulewidth}
		methods     & faucet & cabinet & table & chair & vase  & lamp  & average  \\ \hline
		PCN         & 16.81  & 10.47   & 12.22 & 11.81 & 13.25 & 16.67 & 13.54    \\
		PCN+Ray     & 16.13  & 10.18   & 11.68 & 10.61 & 11.13 & 14.90 & 12.44    \\
		PF-Net      & 16.11  & 10.04   &  9.97 & 10.61 & 11.50 & 14.07 & 12.05    \\
		P2P-Net     & 16.09  & 11.64   & 10.73 & 12.29 & 16.36 & 13.52 & 13.44    \\
		SoftPoolNet & 15.03  & 14.30   & 11.28 & 14.05 & 17.63 & 15.89 & 14.70    \\
		CRN    & 14.00  & 11.00   &  9.09 &  9.70 & 13.32 & 12.09 & 11.53    \\
		GRNet       & 11.30  &  9.16   &  8.61 &  8.82 & 12.27 & 11.28 & 10.24    \\
		MSN         &  8.52  &  8.19   &  7.82 &  7.82 &  8.36 &  8.51 &  8.20    \\
		\hline
		Ours  &  \textbf{6.89}  &  \textbf{7.48}   &  \textbf{6.63} &  \textbf{6.63} &  \textbf{7.16} &  \textbf{7.48} & \textbf{7.05}
		\\ \Xhline{2\arrayrulewidth}
	\end{tabular}
	\vspace{-1pt}
	\caption{Evaluation on EMD ($\times 10^2$) with Res.=2,048}
	\label{tab:sota-table-emd-2k}
	\vspace{-3pt}
\end{table}

\begin{table}[t]
    \centering
	\small
	\setlength{\tabcolsep}{2pt}
	\begin{tabular}{c|cccccc|c}
		\Xhline{2\arrayrulewidth}
		methods     & faucet & cabinet & table & chair & vase  & lamp  & average \\ \hline
		PCN         & 5.62   & 7.28    & 5.95  & 6.14  &  8.71 &  5.15 & 6.48    \\
		PCN+Ray     & 4.35   & 7.14    & 5.19  & 5.98  &  7.19 &  4.61 & 5.74    \\
		PF-Net      & 8.96   & 8.15    & 6.94  & 7.48  & 10.10 &  7.56 & 8.20    \\
		P2P-Net     & 4.47   & 7.21    & \textbf{5.49}  & 5.92  &  7.62 &  4.41 & 5.85    \\
		SoftPoolNet & 5.54   & 7.85    & 6.41  & 6.59  &  8.27 &  5.56 & 6.70    \\
		CRN    & 5.14   & 7.13    & 5.59  & 5.94  &  7.96 &  4.63 & 6.06    \\
		GRNet       & 4.72   & 7.21    & 5.77  & 6.00  &  7.90 &  4.92 & 6.08    \\
		MSN         & 5.25   & 8.06    & 6.50  & 6.70  &  7.92 &  5.66 & 6.68    \\
		\hline
		Ours     & \textbf{3.90}   & \textbf{7.01}    & 5.65  & \textbf{5.61}  &  \textbf{6.68} &  \textbf{4.26} & \textbf{5.51}
		\\ \Xhline{2\arrayrulewidth}
	\end{tabular}
	\vspace{-1pt}
	\caption{Evaluation on CD ($\times 10^2$) with Res.=2,048}
	\label{tab:sota-table-cd-2k}
	\vspace{-6pt}
\end{table}

As shown in Table~\ref{tab:sota-table-emd-8k}-\ref{tab:sota-table-cd-2k}, our method outperforms existing methods on both EMD and CD scores. Our method uses EMD as a loss function and archives the lowest EMD score in all object categories. Besides, we can observe some categories present large scores among all methods, which indicates their inherent structure complexity such as \textit{faucet}, \textit{vase}, \textit{chair}, and \textit{lamp}. While our method shows superiority especially in those categories on both EMD and CD scores. The qualitative results in Figure~\ref{fig:sota_fig_big} demonstrate that PointNet-based methods like PCN, SoftPool, and MSN fail to reconstruct subtle structures and tend to generate blurred details. This is due to the limitation of PointNet architecture. GRNet uses volumetric convolution and therefore can encode the emptiness semantic implicitly, result in better details. However, GRNet is still limited by high computational cost. With the help of ray features encoded in points, our method can encode and reconstruct complex structures precisely and efficiently.

\textbf{SK-PCN}~\cite{nie2020skeleton} predicts detailed structure by learning the skeletal points first. It decouples the shape completion into structure estimation and surface reconstruction. Limited by the availability of skeleton points for training, we only evaluate SK-PCN and ours on two categories: chair and table, where each of them has 818 and 991 models, respectively. The results are listed in Table~\ref{tab:skeleton-table}.

\begin{table}[h]
\centering
		\begin{tabular}{c|cc}
			\Xhline{2\arrayrulewidth}
			Category & SK-PCN           & Ours            \\ \hline
			Chair    & 8.46 / 4.75      & 4.01 / 3.09     \\
			Table    & 8.98 / 4.37      & 5.53 / 3.94     \\ \hline
			Average  & 8.72 / 4.56      & 4.77 / 3.51     \\ \Xhline{2\arrayrulewidth}
		\end{tabular}
	\vspace{-2pt}
	\caption{Comparison with SK-PCN (EMD / CD $\times 10^2$)}
	\label{tab:skeleton-table}
	\vspace{-6pt}
\end{table}


\subsection{Effectiveness of Emptiness Encoding}
\begin{figure}[h]
	\centering
	\begin{subfigure}[t]{0.115\textwidth}
		\includegraphics[width=\textwidth, trim=200 200 150 200, clip]
		{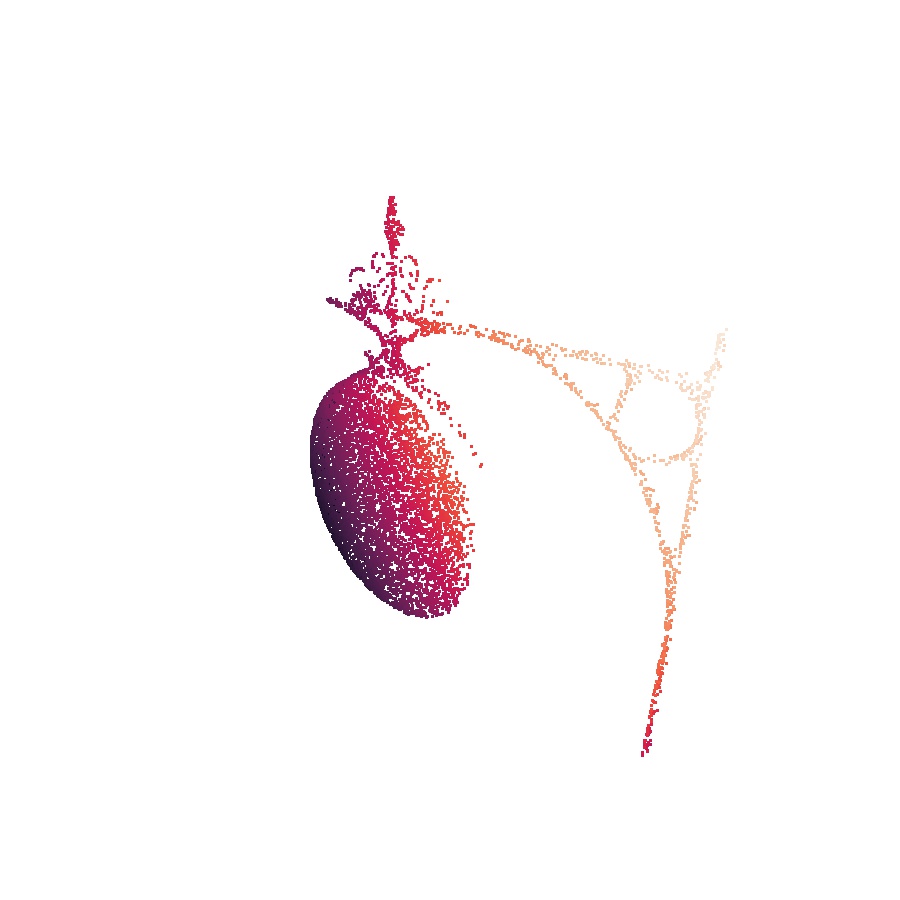}
		\caption{Input}
	\end{subfigure}
	\begin{subfigure}[t]{0.115\textwidth}
		\includegraphics[width=\textwidth, trim=200 200 150 200, clip]
		{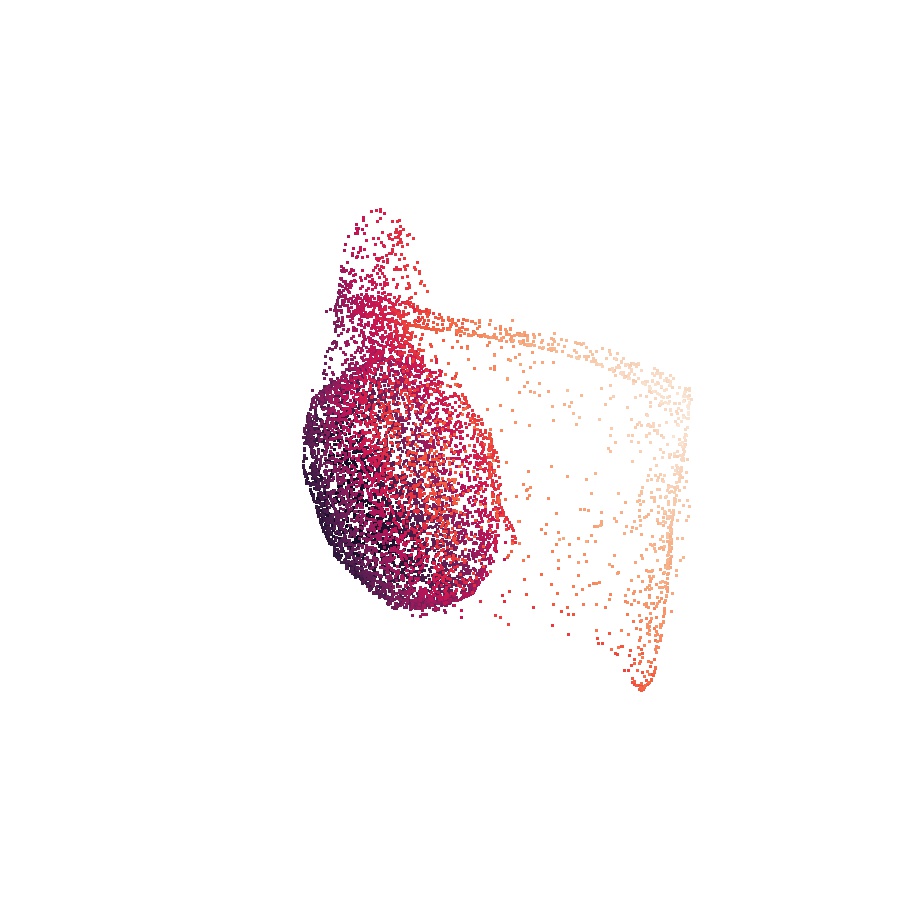}
		\caption{v. MSN+ray}
	\end{subfigure}
	\begin{subfigure}[t]{0.115\textwidth}
		\includegraphics[width=\textwidth, trim=200 200 150 200, clip]
		{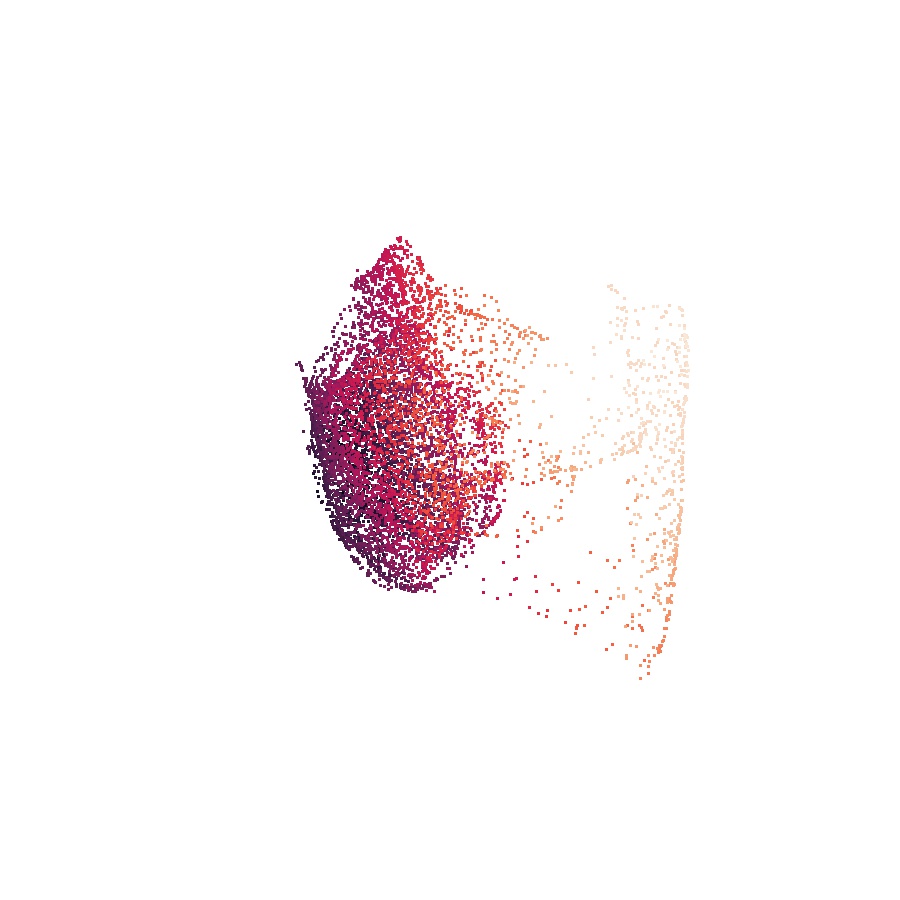}
		\caption{v. MSN}
	\end{subfigure}
	\begin{subfigure}[t]{0.115\textwidth}
		\includegraphics[width=\textwidth, trim=200 200 150 200, clip]
		{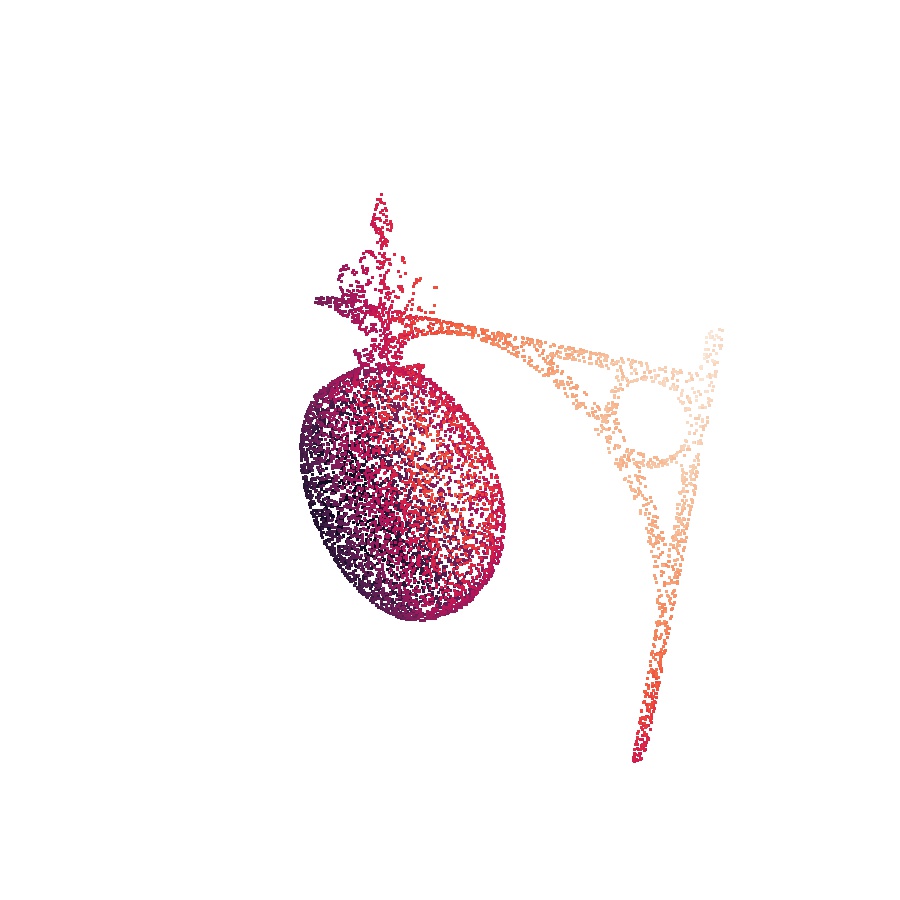}
		\caption{GT}
	\end{subfigure}
	\vspace{-2pt}
	\caption{Coarse results comparison with vanilla MSN+ray.}
	\label{fig:gfv-coarse-compare}
	\vspace{-6pt}
\end{figure}

To verify the effectiveness of embedding emptiness information in feature encoding, we compare the coarse output from MSN~\cite{liu2020morphing} (vanilla MSN) with the one embedded with ray features as in Section~\ref{sec:globalfeat} (vanilla MSN+ray). The output resolution of the two methods is set to 8,192 points. We report the quantitative results in Table \ref{tab:coarse-table} on both EMD and CD scores. Besides, we also augment PCN~\cite{yuan2018pcn} into PCN+ray. The results are listed in Table~\ref{tab:sota-table-emd-8k}-\ref{tab:sota-table-cd-2k}. Figure~\ref{fig:gfv-coarse-compare} shows a qualitative comparison. Both the qualitative and quantitative results suggest that, encoding emptiness feature improves the representation ability of the encoder, which further boosts the performance in shape decoding even using a coarse decoder (from a global feature). The reason could be that the emptiness feature indicates a clear shape contour. There could be fewer output points within empty areas from MSN+ray compared with using vanilla MSN.


\begin{table}[h]
\centering
		\begin{tabular}{c|cc}
			\Xhline{2\arrayrulewidth}
			Category & vanilla MSN      & vanilla MSN+Ray \\ \hline
			Faucet   & 8.61 / 5.03      & 6.59 / 3.63     \\
			Cabinet  & 7.17 / 6.07      & 6.13 / 5.24     \\
			Table    & 8.59 / 5.75      & 6.10 / 4.66     \\
			Chair    & 7.51 / 5.59      & 5.84 / 4.50     \\
			Vase     & 8.67 / 6.75      & 6.63 / 5.62     \\
			Lamp     & 8.59 / 5.42      & 7.84 / 4.38     \\ \hline
			Average  & 8.19 / 5.77      & 6.52 / 4.67     \\ \Xhline{2\arrayrulewidth}
		\end{tabular}
	\vspace{-2pt}
	\caption{MSN + emptiness encoding (EMD / CD $\times 10^2$)}
	\label{tab:coarse-table}
	\vspace{-6pt}
\end{table}


\subsection{Encoding Emptiness with 2D Convolution}
Apart from using 3D rays to encode emptiness, we also adopt 2D CNNs to perceive the empty regions on the depth map. In our experiments, the depth map in Figure~\ref{fig:architecture} is concatenated with the emptiness $mask$ in Equation~\ref{eq:01} to construct a 4-channel image $I$. Then we build a ME-PCN-2DConv network by replacing MLPs in the global encoder with three 2D convolution layers + Max-pooling, which outputs the global feature with the same dimension (2048-D) for the following shape decoding.


\begin{figure}[h]
	\centering
	\begin{subfigure}[t]{0.110\textwidth}
		\includegraphics[width=\textwidth, trim=300 180 300 180, clip]
		{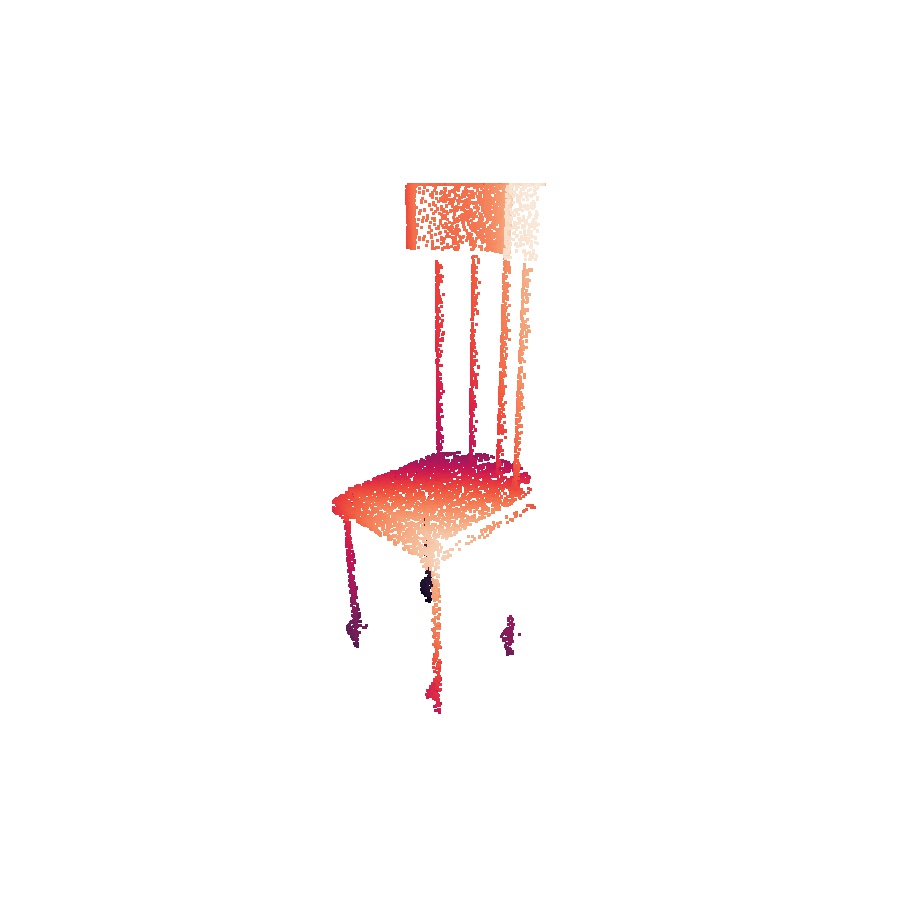}
		\caption{Input}
	\end{subfigure}
	\begin{subfigure}[t]{0.110\textwidth}
		\includegraphics[width=\textwidth, trim=300 180 300 180, clip]
		{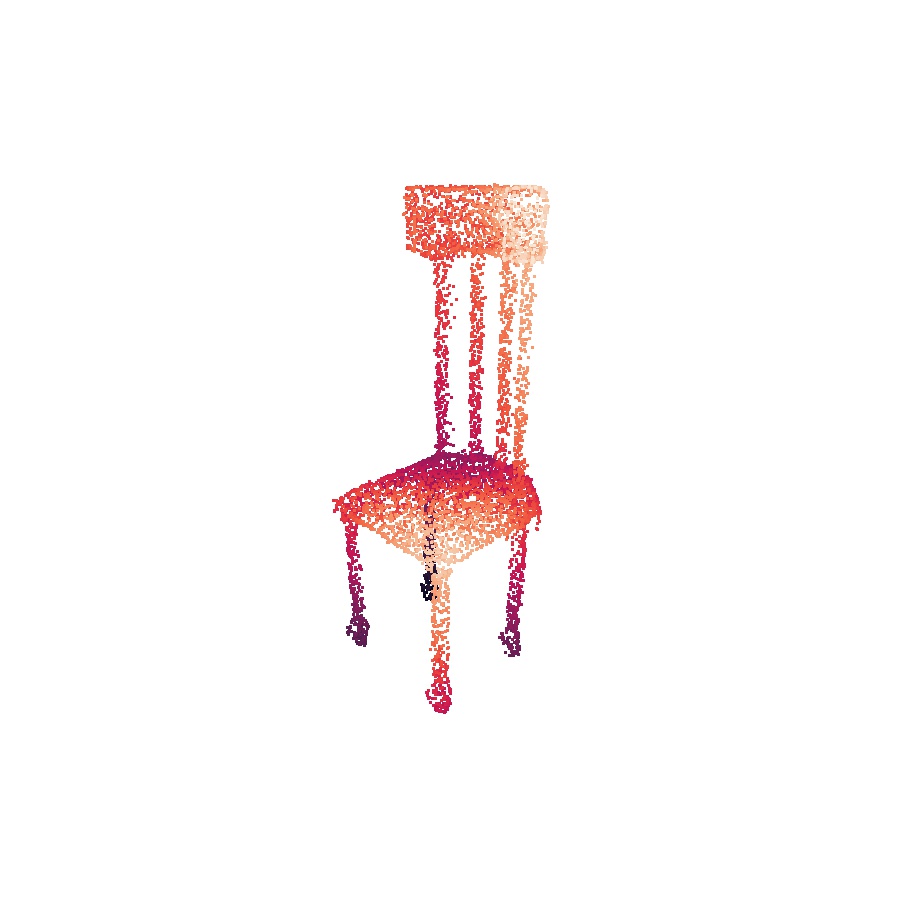}
		\caption{Ours}
	\end{subfigure}
	\begin{subfigure}[t]{0.110\textwidth}
		\includegraphics[width=\textwidth, trim=300 180 300 180, clip]
		{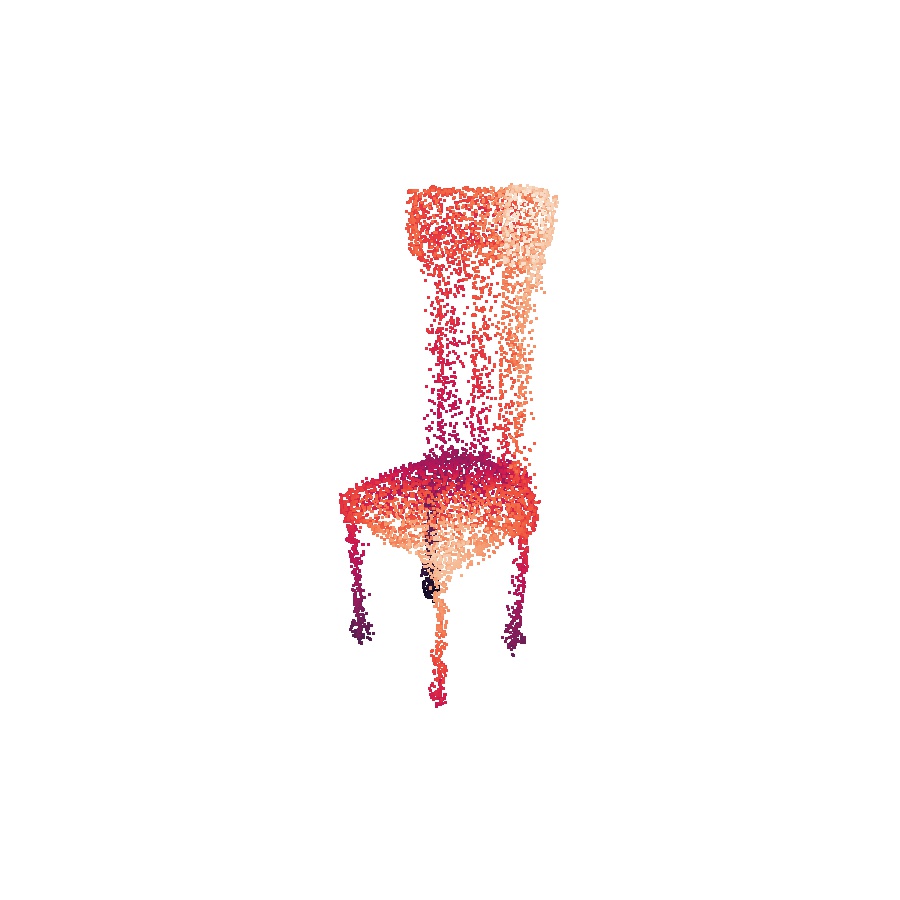}
		\caption{w. 2DConv}
	\end{subfigure}
	\begin{subfigure}[t]{0.110\textwidth}
		\includegraphics[width=\textwidth, trim=300 180 300 180, clip]
		{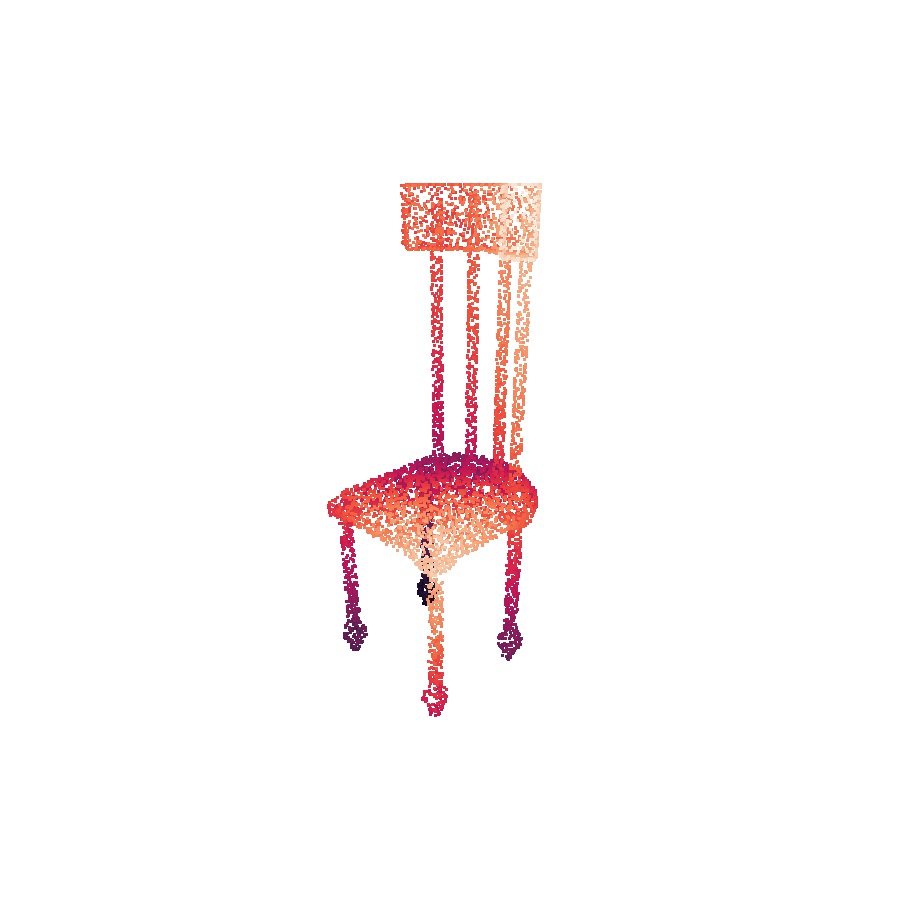}
		\caption{GT}
	\end{subfigure}
\vspace{-2pt}
	\caption{Using 2D convolutions for emptiness encoding.}
	\label{fig:encoding-2d-conv}
	\vspace{-6pt}
\end{figure}

For coarse-to-fine decoding, we project predicted coarse points back to a 2D image plane. For each coarse point, we obtain its point feature by querying the 2D feature map learned by another 2D CNN from image $I$. The 2D CNN is constructed in a similar way as above, and obtains a 64-channel feature map from $I$. We compare ME-PCN-2DConv with our ME-PCN. The qualitative and quantitative results are listed in Figure~\ref{fig:encoding-2d-conv} and Table~\ref{tab:conv2d-table}. They show that ME-PCN presents better results. Besides, MEM-PCN-2DConv consumes more net parameters and costs 2-5 times of GPU memory led by the 2D convolutions, which demonstrates the efficiency of representing emptiness using rays.

\begin{table}[t]
\centering
\begin{tabular}{c|cc}
			\Xhline{2\arrayrulewidth}
			Category & ME-PCN-2DConv       & Ours         \\ \hline
			Faucet   & 7.46 / 3.31      & 6.31 / 2.62     \\
			Cabinet  & 6.26 / 5.00      & 6.14 / 4.72     \\
			Table    & 5.70 / 4.14      & 5.33 / 3.76     \\
			Chair    & 5.91 / 4.13      & 5.12 / 3.62     \\
			Vase     & 6.97 / 5.35      & 5.93 / 4.54     \\
			Lamp     & 6.74 / 3.54      & 6.76 / 3.02     \\ \hline
			Average  & 6.51 / 4.25      & 5.93 / 3.71     \\ \Xhline{2\arrayrulewidth}
\end{tabular}
\vspace{-2pt}
\caption{2D emptiness encoding (EMD / CD $\times 10^2$)}
\label{tab:conv2d-table}
\vspace{-6pt}
\end{table}

\subsection{Robustness of Emptiness Encoding}
\begin{figure}[h]
	\centering
	\begin{subfigure}[t]{0.21\textwidth}
		\includegraphics[width=\textwidth, trim=515 300 570 350, clip]
		{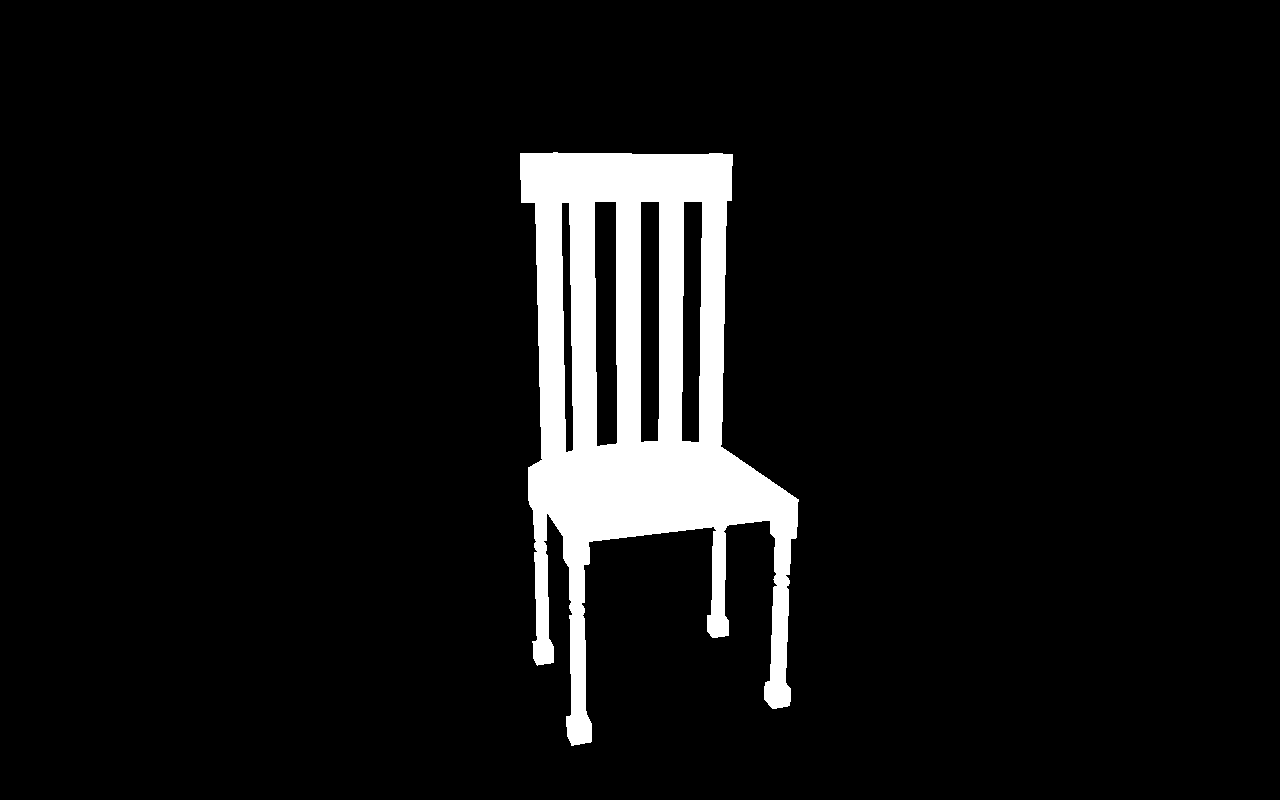}
		\caption{Before}
	\end{subfigure}
	\begin{subfigure}[t]{0.21\textwidth}
		\includegraphics[width=\textwidth, trim=515 300 570 350, clip]
		{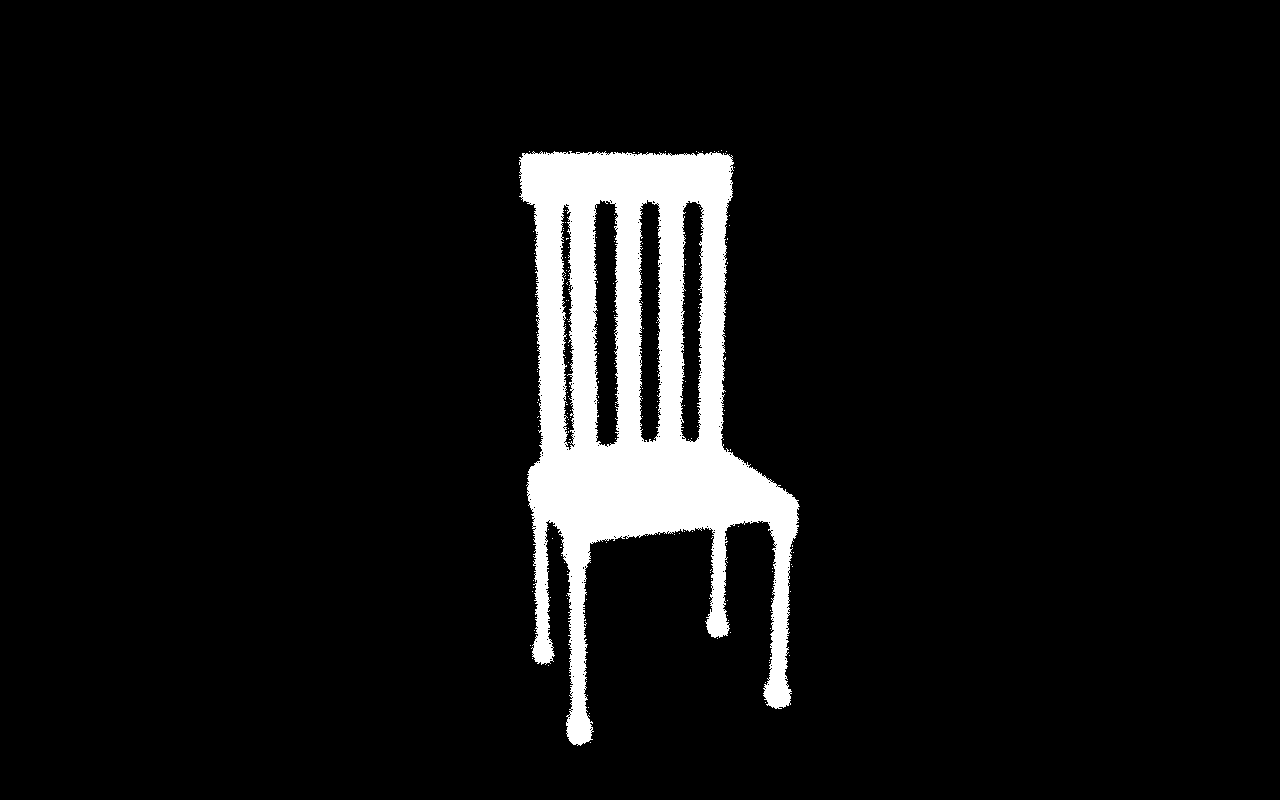}
		\caption{After}
	\end{subfigure}

	\caption{Adding noise to simulate the mask from real world data: a) mask of chair back without noise; b) mask with strong noise on  boundaries;}
	\label{fig:robustness-mask-compare}

\end{figure}

The emptiness masks in our method are generated from depth maps. However, in the real world, such a mask can be inaccurate considering the noises in depth scans. To verify our robustness to noisy masks, we simulate the masks from real-world depth/RGB data, and add strong Gaussian noise (see Figure~\ref{fig:robustness-mask-compare}) to the boundaries of masks. We fine-tune our model on chair and table categories using the noisy mask for 2000 iterations.
Test results show that the EMD score on chair / table increase from 5.12 / 5.33 to 5.26 / 5.56 respectively. CD score on chair/table increase from 3.62 / 3.76 to 3.67 / 3.92 respectively. The performance-degradation is less than $3\%$, which verifies the robustness of our method.

\section{Conclusion}
We provide a novel feature encoding modality for point completion, namely ME-PCN. It leverages the 3D emptiness in shape space to make neural networks sensitive to shape boundaries. In our method, we complete surface points by learning from both the shape occupancy and emptiness. A ray-based emptiness encoding strategy is proposed to perceive the emptiness clues on shape boundaries. It enables our method to recover enriched surface details while keeping consistent local topology. Verified by the ablation studies, our emptiness encoding is effective, moreover, robust and efficient. Extensive experiments demonstrate that our method achieves much better shape completion quality and largely outperforms the state-of-the-art on EMD and CD metrics. Though there is still some gap between our predictions and the GT, e.g., it may fail to capture small and complex topology, like decorations on the table stand, our method works well in general cases. We hope our work can serve as a universal improvement strategy for point completion and draw attention to the information in the `emptiness', even in the larger community.

\paragraph*{Acknowledgments} This work was sponsored by Hong Kong Research Grants Council under General Research Funds (HKU17206218), and partially supported by NSFC-61902334 and Shenzhen General project JCYJ20190814112007258. It was also supported by CCF-Tencent AI Lab RhinoBird Fund.

\clearpage

\twocolumn[
\vspace{3em}
\part*{\centering Supplementary Material}
\vspace{3em}
]

\setcounter{section}{0}

\renewcommand*{\thesection}{A\arabic{section}}
\renewcommand*{\thesubsection}{A\arabic{section}.\arabic{subsection}}

\renewcommand{\theHsection}{A\arabic{section}}
\renewcommand*{\theHsubsection}{A\arabic{section}.\arabic{subsection}}

\section{Network Architecture and Parameters}
\label{sec:network}

We provide the details of parameters and layer information of our network in this section. It discusses the encoder for $\mathcal{R}^{*}_{ray}$, $\mathcal{R}^{d}_{ray1}$ and $\mathcal{R}^{d}_{ray1}$.  The coarse decoder is adopted from \cite{liu2020morphing} and will not be covered here.

\paragraph{Emptiness $\mathcal{R}^{*}_{ray}$ Encoder} Our approach takes a partial point
cloud $\mathbf{Q}$ and sampled rays $\mathcal{R}^{*}_{ray}$ as inputs and encodes them into a global feature vector (GFV) with emptiness semantics, which will be used to predict complete point cloud with a coarse-to-fine strategy.
The parameters for each layer in our encoder is shown in Table \ref{tab:ray-encoder-msn}. For vanilla MSN+ray, we use the same encoder architecture as in Table~\ref{tab:ray-encoder-msn}.

\begin{table}[h]
	\centering
	\begin{tabular}{c|cc}
		\Xhline{2\arrayrulewidth}
		Input       & \multicolumn{1}{c|}{Partial Points $Q$} & Ray Samples $\mathcal{R}^{*}_{ray}$   \\ \hline
		3 MLPs      & \multicolumn{1}{c|}{64, 128, 1024} & 64, 128, 1024  \\ \hline
		Max-pool    & \multicolumn{1}{c|}{1024}          & \multicolumn{1}{c}{1024}           \\ \hline
		Concatenate & \multicolumn{2}{c}{2048}      \\ \hline
		Output      & \multicolumn{2}{c}{Global Feature Vector $g'$}\\ \Xhline{2\arrayrulewidth}
	\end{tabular}
	\caption{Encoder of ME-PCN}
	\label{tab:ray-encoder-msn}
	\vspace{-5pt}
\end{table}

\begin{table}[h]
	\centering
	\begin{tabular}{c|cccc}
		\Xhline{2\arrayrulewidth}
		Input       & \multicolumn{2}{c|}{Partial Points $\mathbf{Q}$} 
		& \multicolumn{2}{c}{Ray Samples $\mathcal{R}^{*}_{ray}$}  \\ \hline
		2 MLPs      & \multicolumn{2}{c|}{128, 256} & \multicolumn{2}{c}{128, 256} \\ \hline
		Max-pool    & \multirow{2}{*}{$f_{i}:256$} & \multicolumn{1}{|c|}{256} & \multicolumn{1}{c|}{256} & \multirow{2}{*}{$g_i:256$}    \\ \cline{1-1} \cline{3-4}
		Concatenate &  & \multicolumn{2}{|c|}{GFV $g'$: 512} &      \\ \hline
		Tile \& Cat & \multicolumn{2}{c|}{$\widetilde{F}: \left|Q\right|\times768$} &   \multicolumn{2}{c}{$\widetilde{G}: \left|\mathcal{R}^{*}_{ray}\right|\times768$}    \\ \hline
		2 MLPs      & \multicolumn{2}{c|}{512, 1024} & \multicolumn{2}{c}{512, 1024} \\ \hline
		Max-pool    & \multicolumn{2}{c|}{1024}          & \multicolumn{2}{c}{1024}           \\ \hline
		Concatenate & \multicolumn{4}{c}{2048}      \\ \hline
		Output      & \multicolumn{4}{c}{Global Feature Vector $g^{*}$} \\ \Xhline{2\arrayrulewidth}
	\end{tabular}
	\caption{PCN+ray $\mathcal{R}^{*}_{ray}$ Encoder}
	\label{tab:ray-encoder-pcn}
	\vspace{-5pt}
\end{table}

The core part of Feature Encoding (FE) layers is 3 separate Multilayer Perceptron Layers (MLP), which transform input points $\{q_{i}|q_{i}\in\mathbf{Q}\}$ or ray samples $\{\mathbf{r}|\mathbf{r}\in\mathcal{R}^{*}_{ray}\}$ into point features. A point-wise
max-pooling is respectively performed on point features to obtain global features. Lastly, global features are concatenated together to form a single global feature vector.

PCN~\cite{yuan2018pcn} has a deeper encoder by appending extra FE layers to learn a more abstract and latent global feature. To keep consistent with its original network, we construct an encoder for PCN+ray as shown in Table \ref{tab:ray-encoder-pcn}.

It first concatenates the global feature $g'$ to each $f_{i}$ and $g_i$ to obtain augmented point feature matrices $\widetilde{F}$ and $\widetilde{G}$ whose rows are the concatenated feature vectors $[f_i, g']$ and $[g_i, g']$. Then, $\widetilde{F}$ and $\widetilde{G}$ are respectively passed through another two-layer MLP followed by point-wise max-pooling (as the first FE layer). The updated global feature vector $g^{*}$ is
outputted by concatenating the pooling results.

\paragraph{Emptiness $\mathcal{R}^{d}_{ray1}$ and $\mathcal{R}^{d}_{ray2}$ Encoder} Rays $\mathcal{R}^{d}_{ray1}$ and $\mathcal{R}^{d}_{ray2}$ are sampled using the coarse points $\mathbf{P}^{c}$. The sampled empty rays $\mathcal{R}^{d}_{ray1}\in\mathbf{R}^{N_{c}\times K\times 9}$ tell whether a coarse point in $\mathbf{P}^{c}$ is in an `empty' region while $\mathcal{R}^{d}_{ray2}\in\mathbf{R}^{N_{c}\times K\times 15}$ represents the boundaries of real shapes. For the sampling method, we refer readers to our paper. In our experiments, $N_c = 8192$ for our method and vanilla MSN+ray, and $N_c = 1024$ for PCN+ray. Both networks use the same emptiness encoder for local feature extraction, as shown in Table \ref{tab:local-ray-encoder}.

\begin{table}[h]
	\centering
	\begin{tabular}{c|cc}
		\Xhline{2\arrayrulewidth}
		Input       & \multicolumn{1}{c|}{Ray Samples $\mathcal{R}^{d}_{ray1}$} & Ray Samples $\mathcal{R}^{d}_{ray2}$   \\ \hline
		3 MLPs      & \multicolumn{1}{c|}{16, 32, 32} & 16, 32, 32  \\ \hline
		Grouping    & \multicolumn{1}{c|}{$N_c\times32$}          & \multicolumn{1}{c}{$N_c\times32$}           \\ \hline
		Concatenate & \multicolumn{2}{c}{$N_c\times64$}      \\ \hline
		Output      & \multicolumn{2}{c}{Local Feature Vector}\\ \Xhline{2\arrayrulewidth}
	\end{tabular}
	\caption{Local Emptiness Encoder for ray $\mathcal{R}^{d}_{ray1}$ and $\mathcal{R}^{d}_{ray2}$}
	\label{tab:local-ray-encoder}
	\vspace{-5pt}
\end{table}

Rays in $\mathcal{R}^{d}_{ray1}$ represent empty space neighboring to coarse points, while $\mathcal{R}^{d}_{ray2}$ informs the coarse points with the real shape boundary. Two FE layers are respectively used to encode $\mathcal{R}^{d}_{ray1}$ and $\mathcal{R}^{d}_{ray2}$. Grouping operation is a PointConv-style aggregation defined in \cite{wu2019pointconv}. We train the entire network end-to-end with the loss function in Equation~(10) of the paper with $\lambda_{1}=1.0$ and $\lambda_{2}=0.1$.

\begin{figure*}[t]
	\centering
	\begin{subfigure}[t]{0.075\textwidth}
		\centering
		\stackon[0pt]
		{\reflectbox{\includegraphics[height=40pt, trim=120 0 60 0, clip]{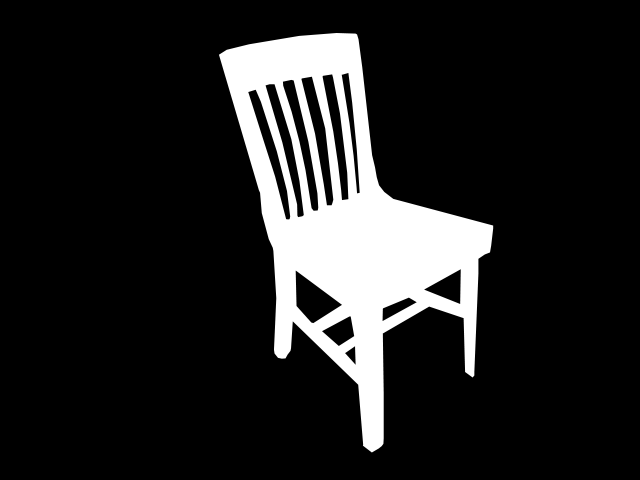}}}
		{\reflectbox{\includegraphics[height=40pt, trim=120 0 60 0, clip]{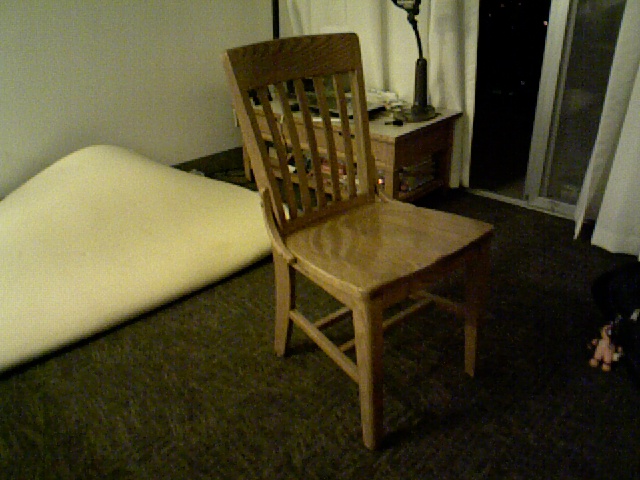}}}
		
		\stackon[0pt]
		{\reflectbox{\includegraphics[height=40pt, trim=120 0 60 0, clip]{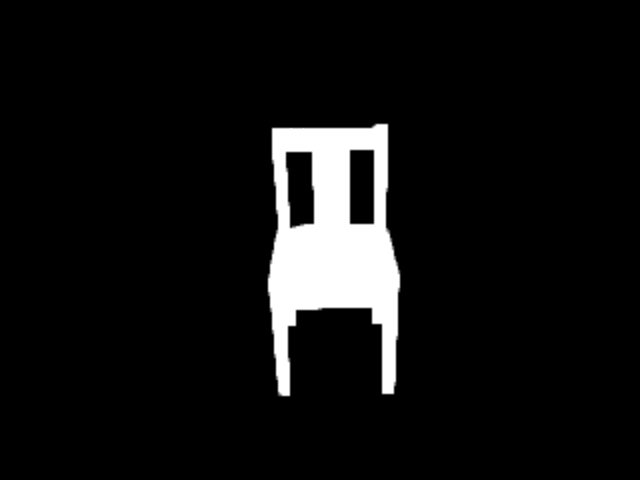}}}
		{\reflectbox{\includegraphics[height=40pt, trim=120 0 60 0, clip]{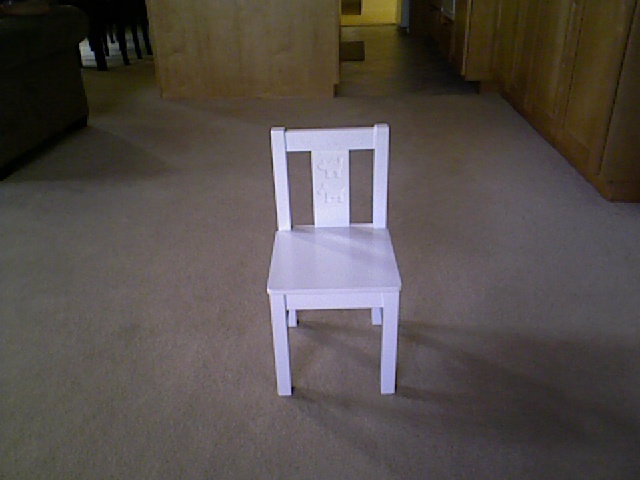}}}
		
		\caption{Scans}
	\end{subfigure}
	\begin{subfigure}[t]{0.1\textwidth}
		\centering
		\includegraphics[height=80pt, trim=480 200 450 320, clip]  
		{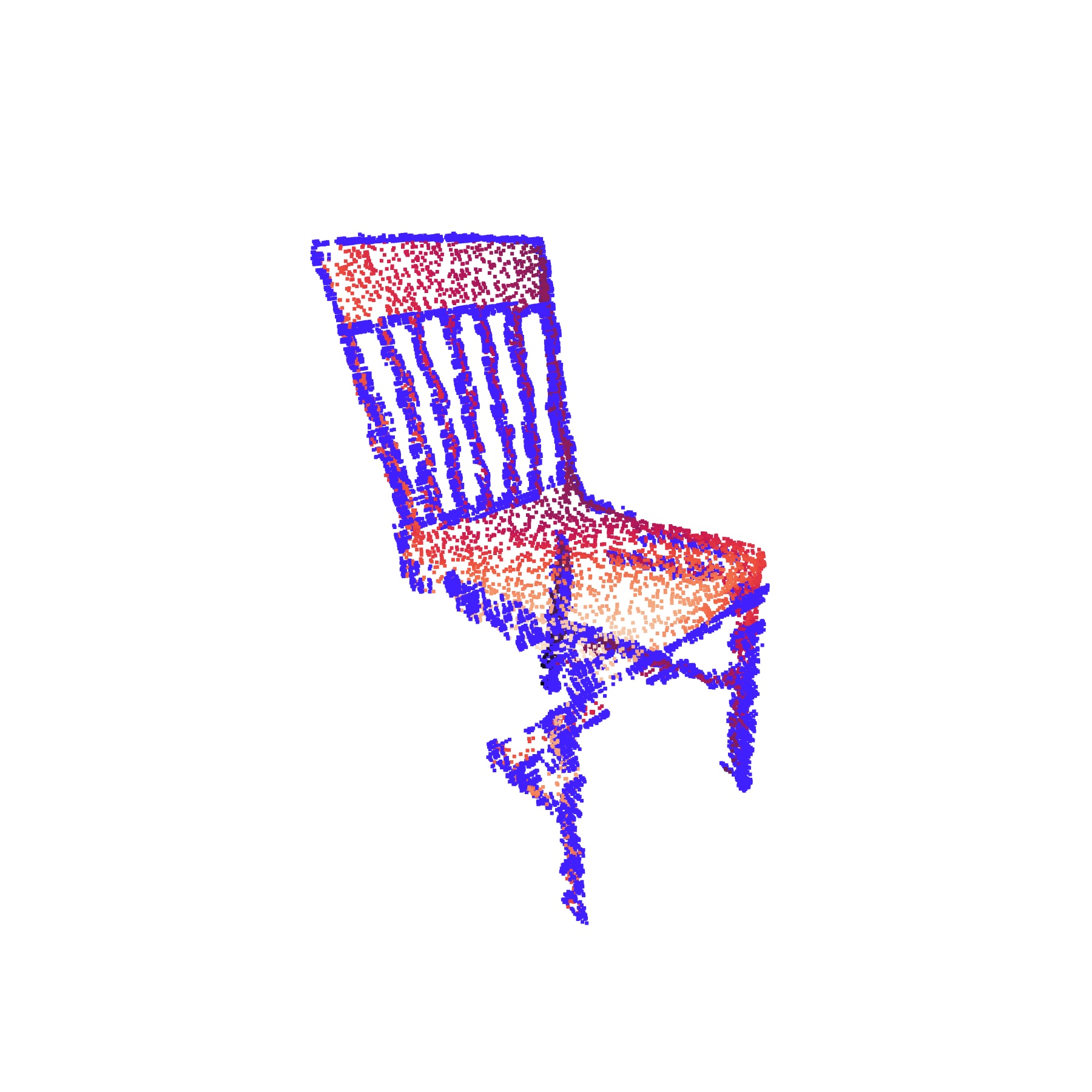}
		
		\includegraphics[height=80pt, trim=450 170 420 310, clip]  
		{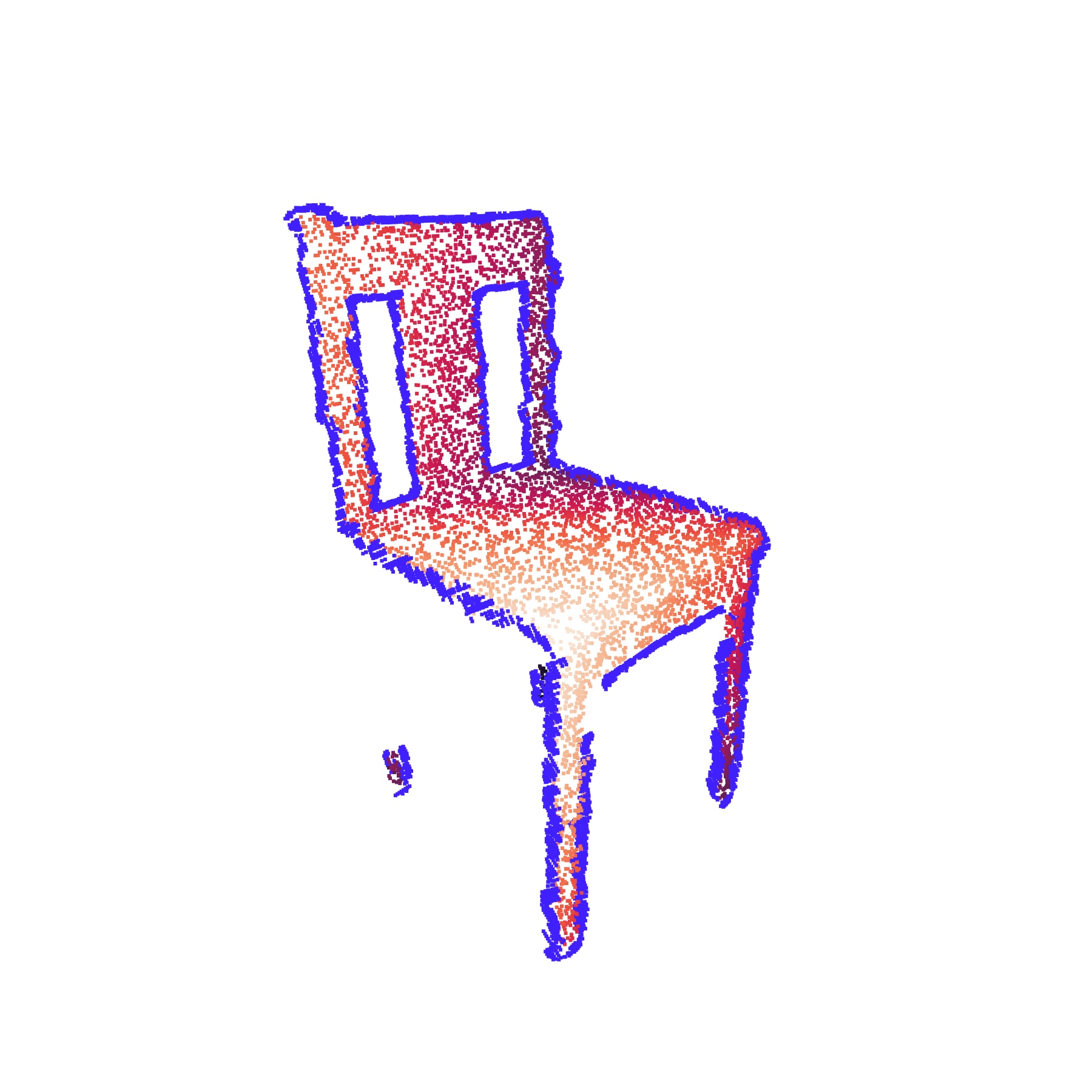}
		
		\caption{Input}
	\end{subfigure}
	\begin{subfigure}[t]{0.1\textwidth}
		\centering
		\includegraphics[height=80pt, trim=480 200 450 320, clip]
		{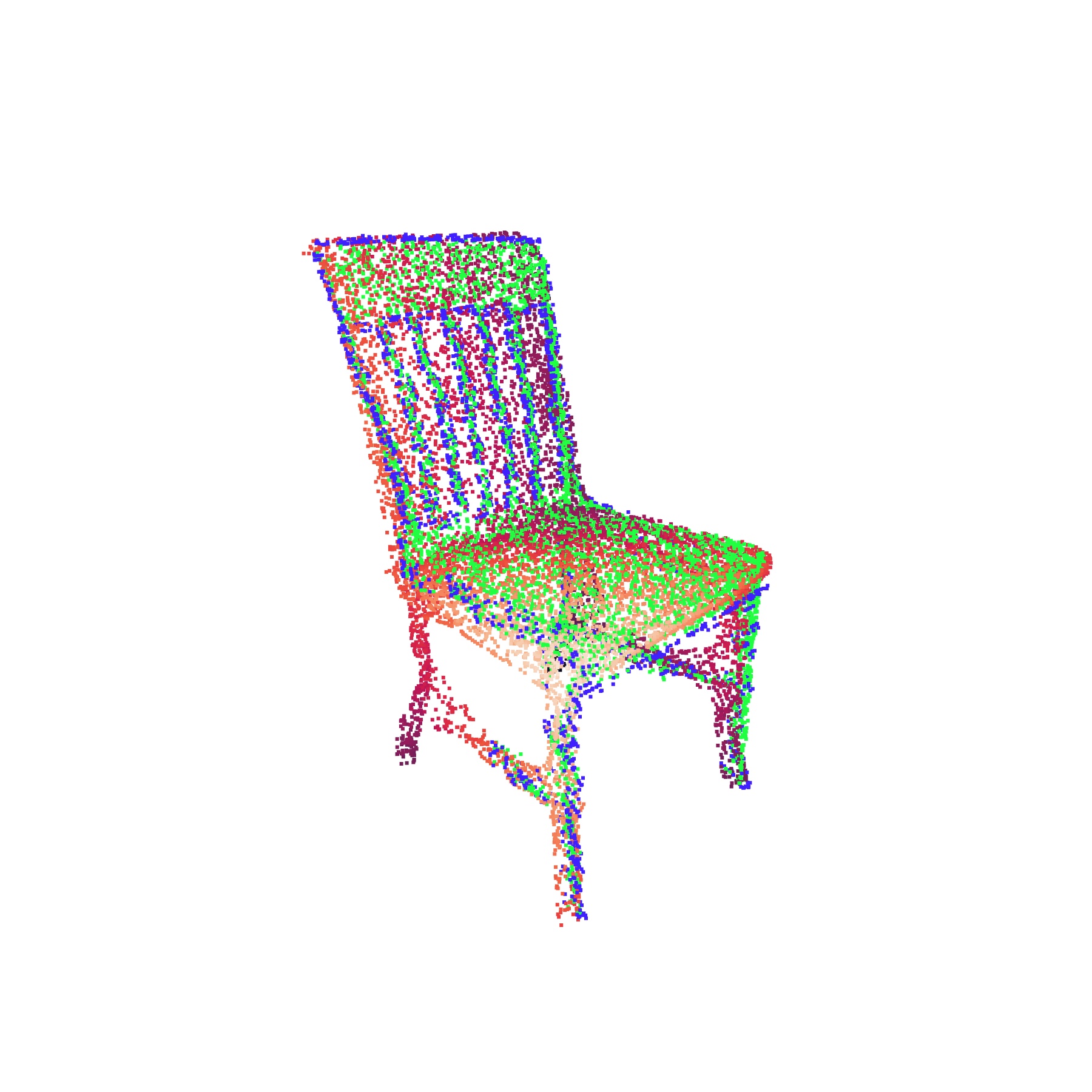}
		
		\includegraphics[height=80pt, trim=450 170 420 310, clip]
		{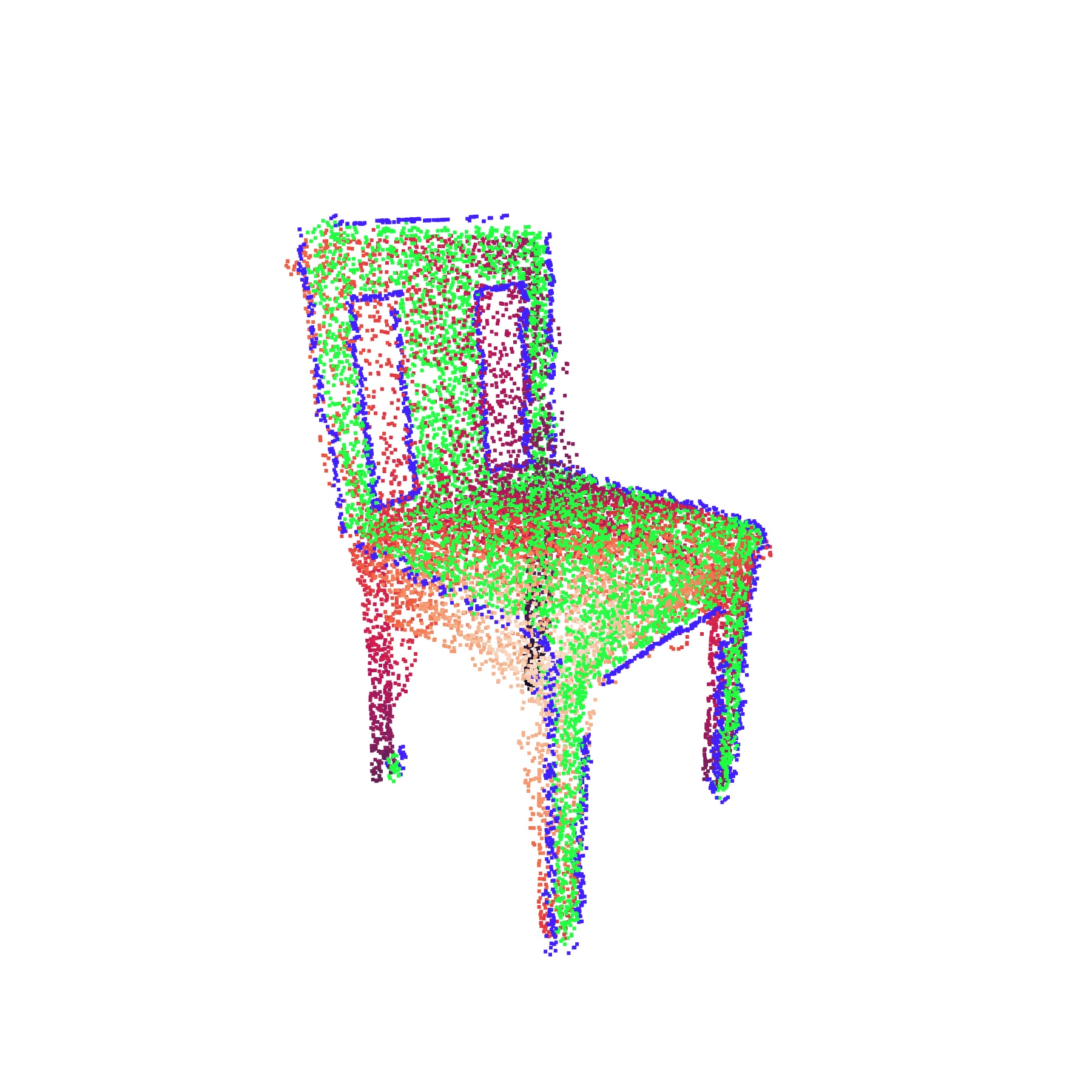}
		
		\caption{Coarse}
	\end{subfigure}
	\begin{subfigure}[t]{0.1\textwidth}
		\centering
		\includegraphics[height=80pt, trim=480 200 450 320, clip]
		{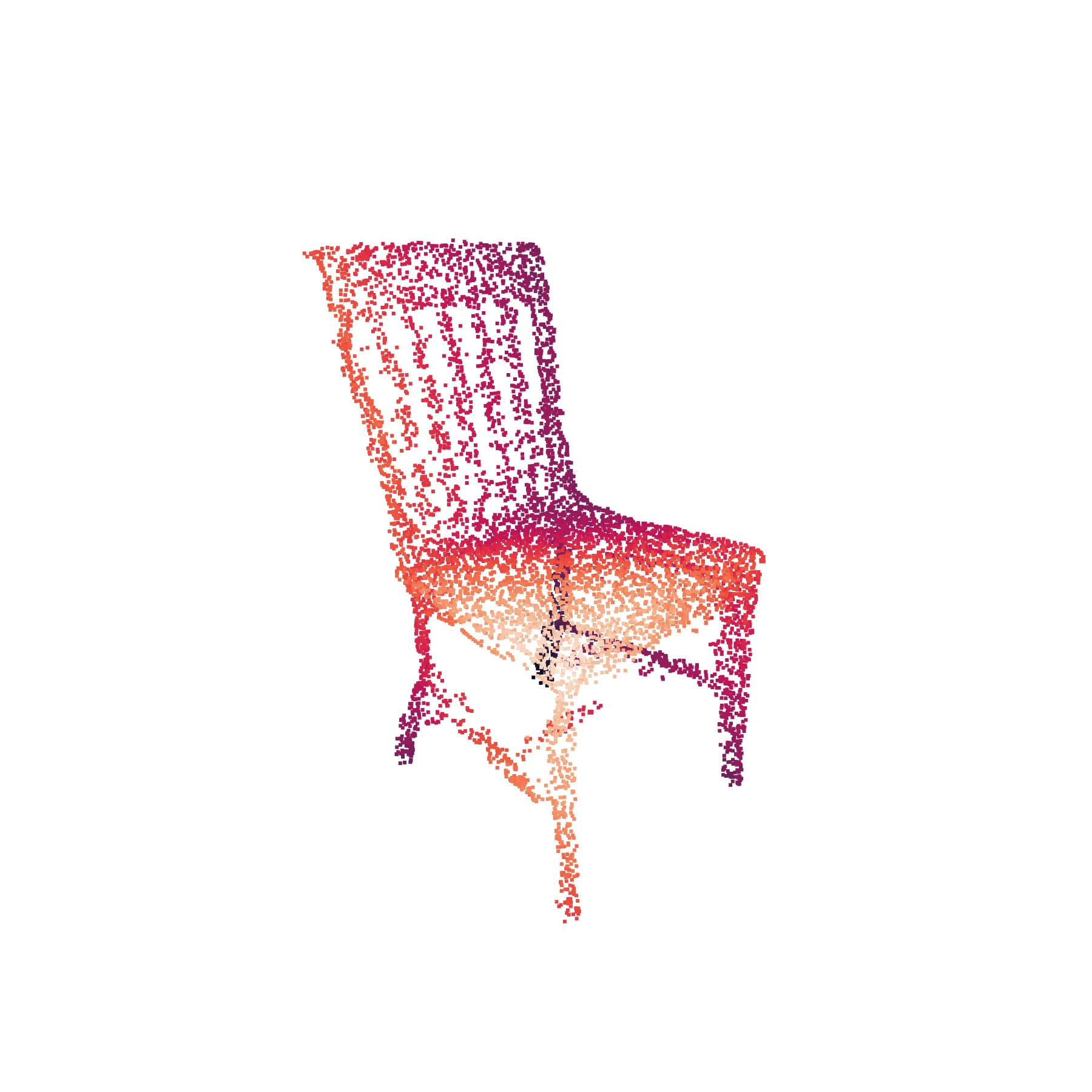}
		
		\includegraphics[height=80pt, trim=450 170 420 310, clip]
		{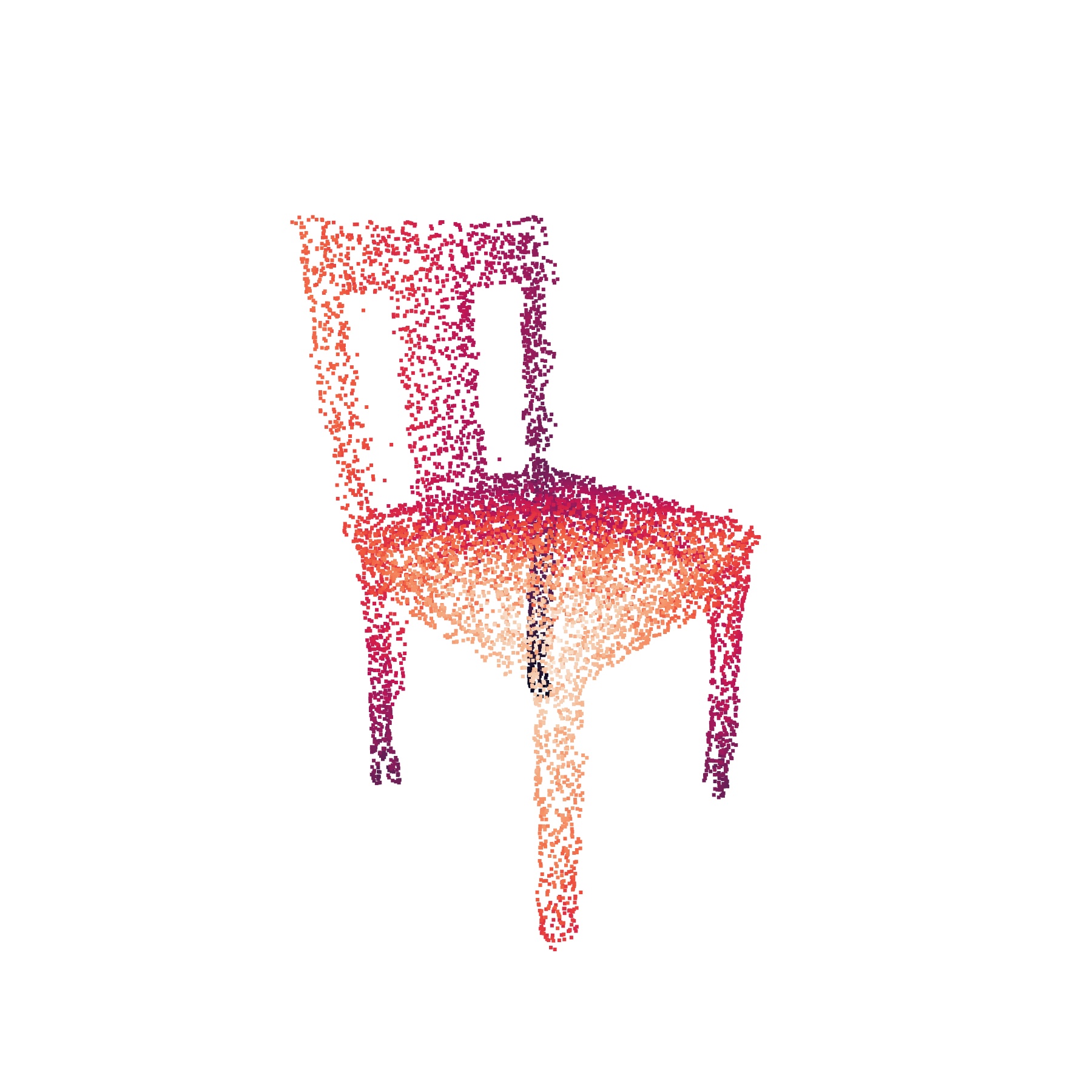}
		
		\caption{Output}
	\end{subfigure}
	\begin{subfigure}[t]{0.1\textwidth}
		\centering
		\includegraphics[height=80pt, trim=500 200 450 320, clip]
		{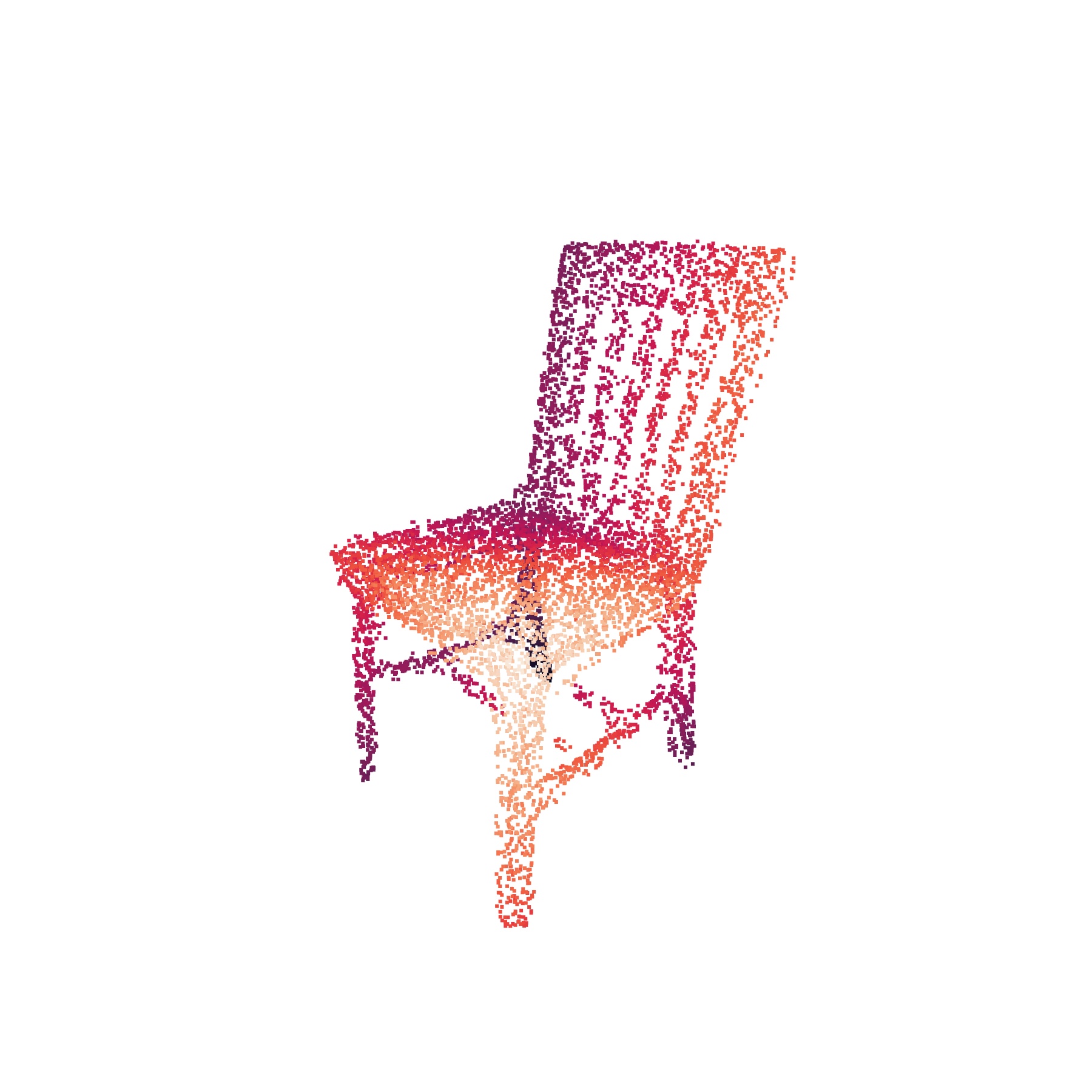}
		
		\includegraphics[height=80pt, trim=480 170 420 310, clip]
		{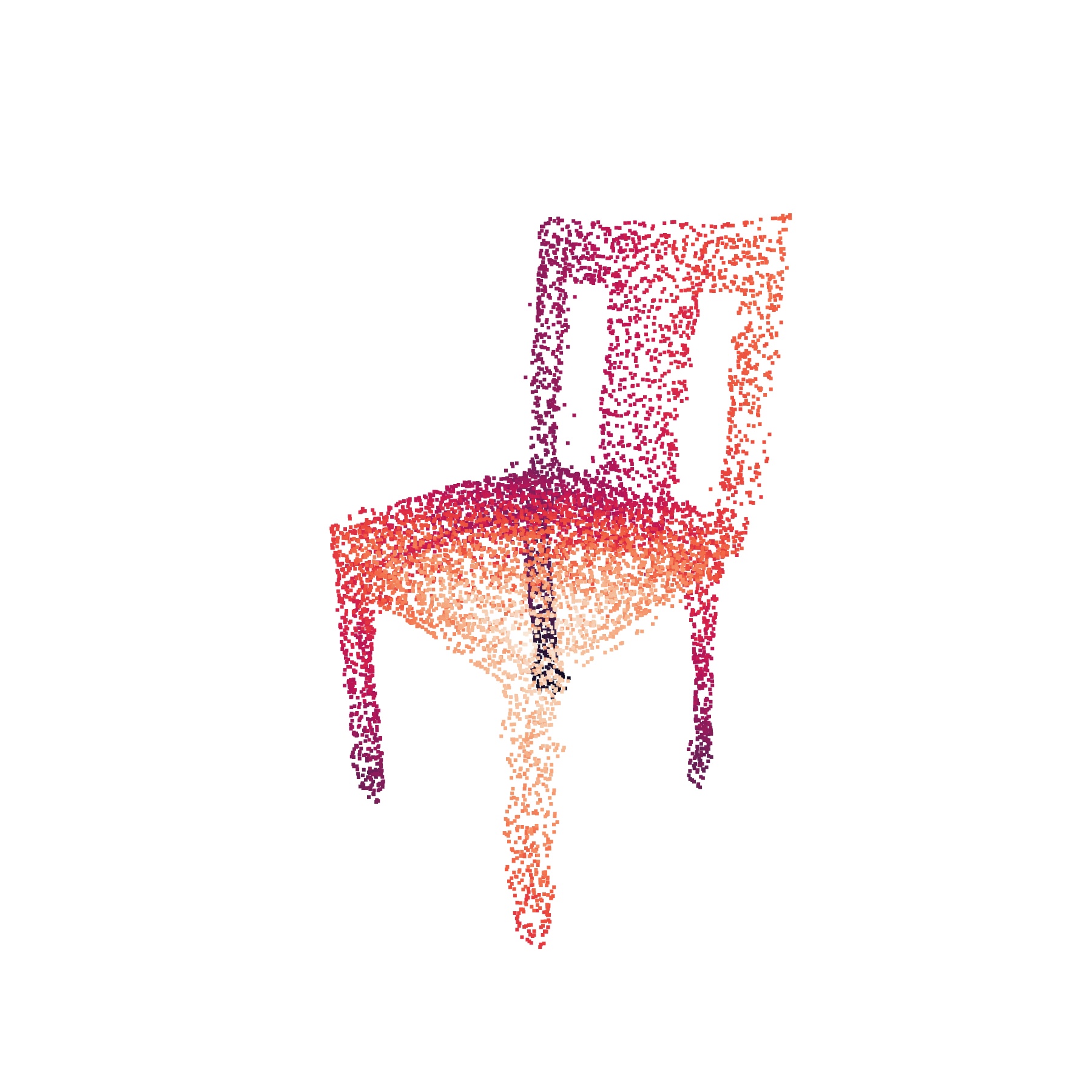}
		
		\caption{Output}
	\end{subfigure}
	\rulesep
	\begin{subfigure}[t]{0.075\textwidth}
		\centering
		\stackon[0pt]
		{\reflectbox{\includegraphics[height=40pt, trim=90 0 90 0, clip]{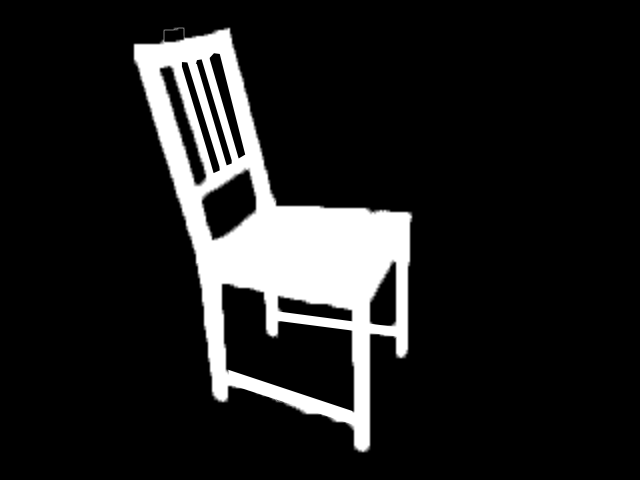}}}
		{\reflectbox{\includegraphics[height=40pt, trim=90 0 90 0, clip]{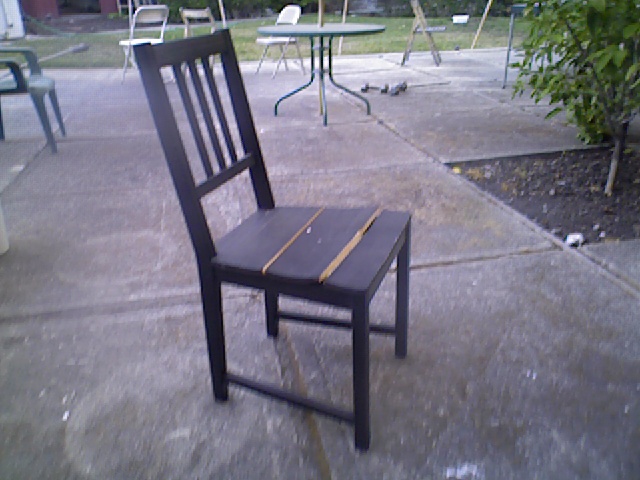}}}
		
		\stackon[0pt]
		{\reflectbox{\includegraphics[height=40pt, trim=90 0 90 0, clip]{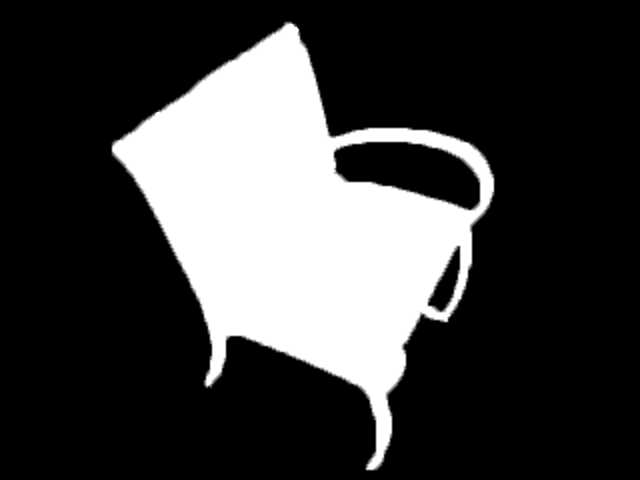}}}
		{\reflectbox{\includegraphics[height=40pt, trim=90 0 90 0, clip]{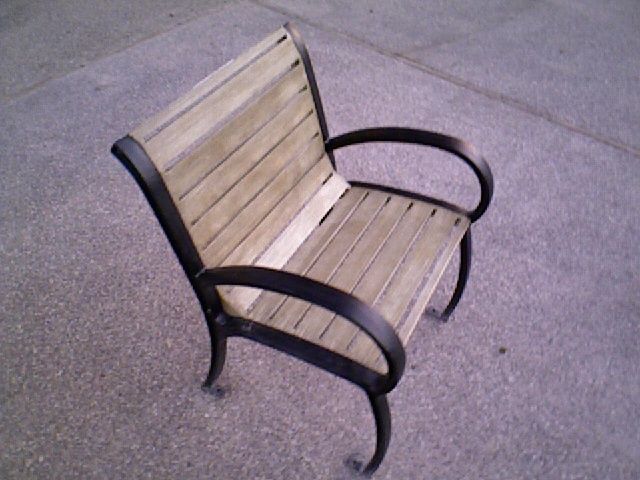}}}
		
		\caption{Scans}
	\end{subfigure}
	\begin{subfigure}[t]{0.1\textwidth}
		\centering
		\includegraphics[height=80pt, trim=515 200 450 320, clip]  
		{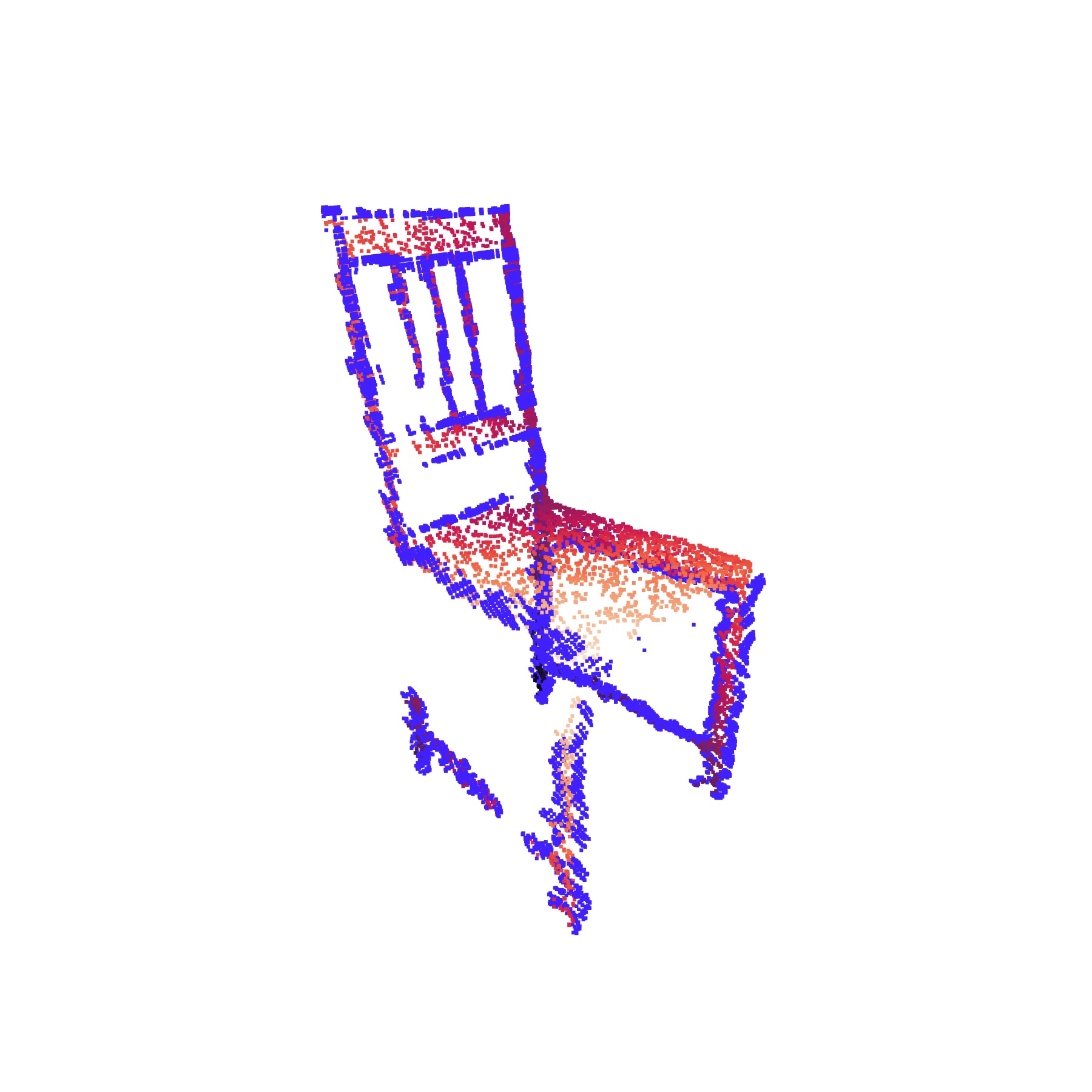}
		
		\includegraphics[height=80pt, trim=300 0 300 0, clip]  
		{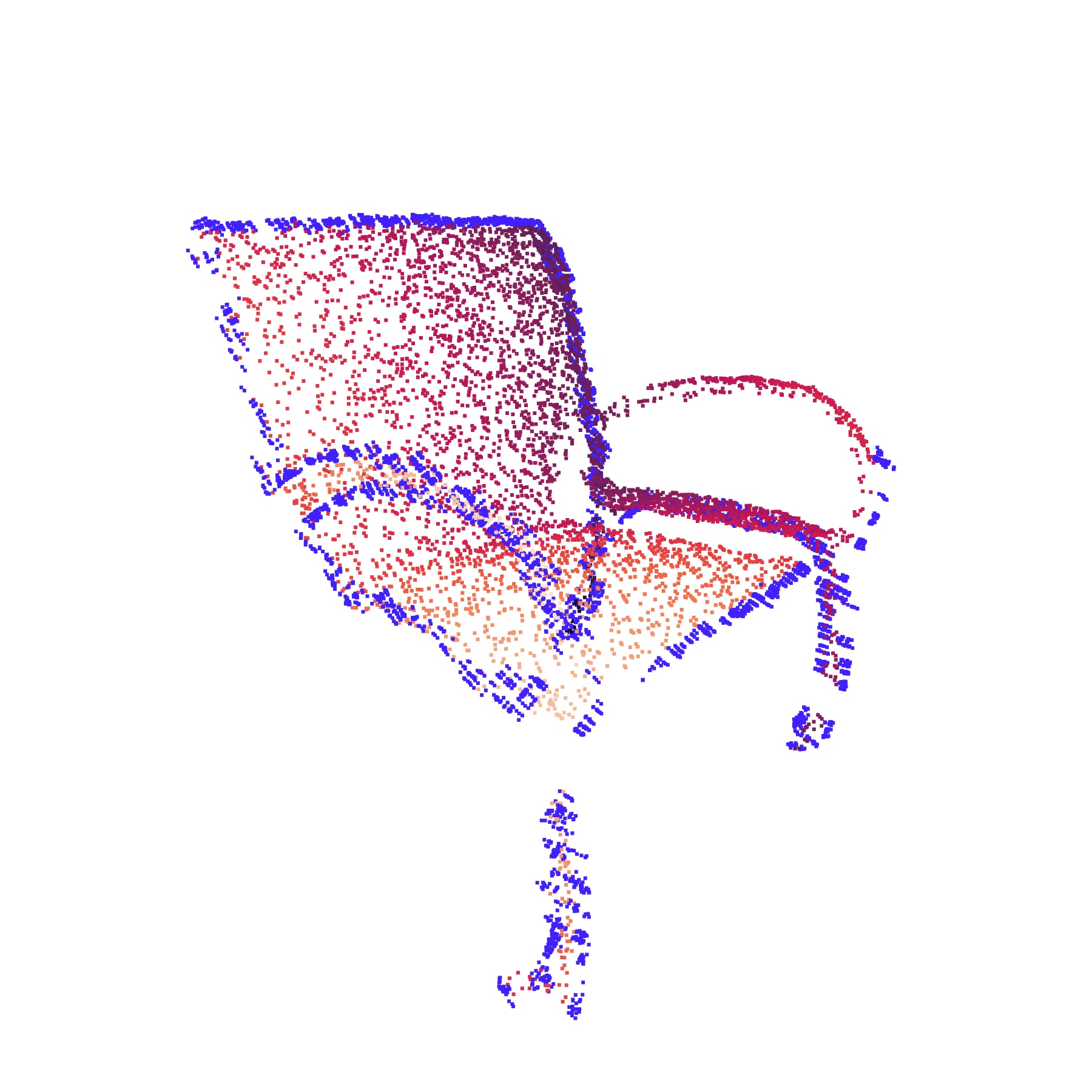}
		
		\caption{Input}
	\end{subfigure}
	\begin{subfigure}[t]{0.1\textwidth}
		\centering
		\includegraphics[height=80pt, trim=515 200 450 320, clip]
		{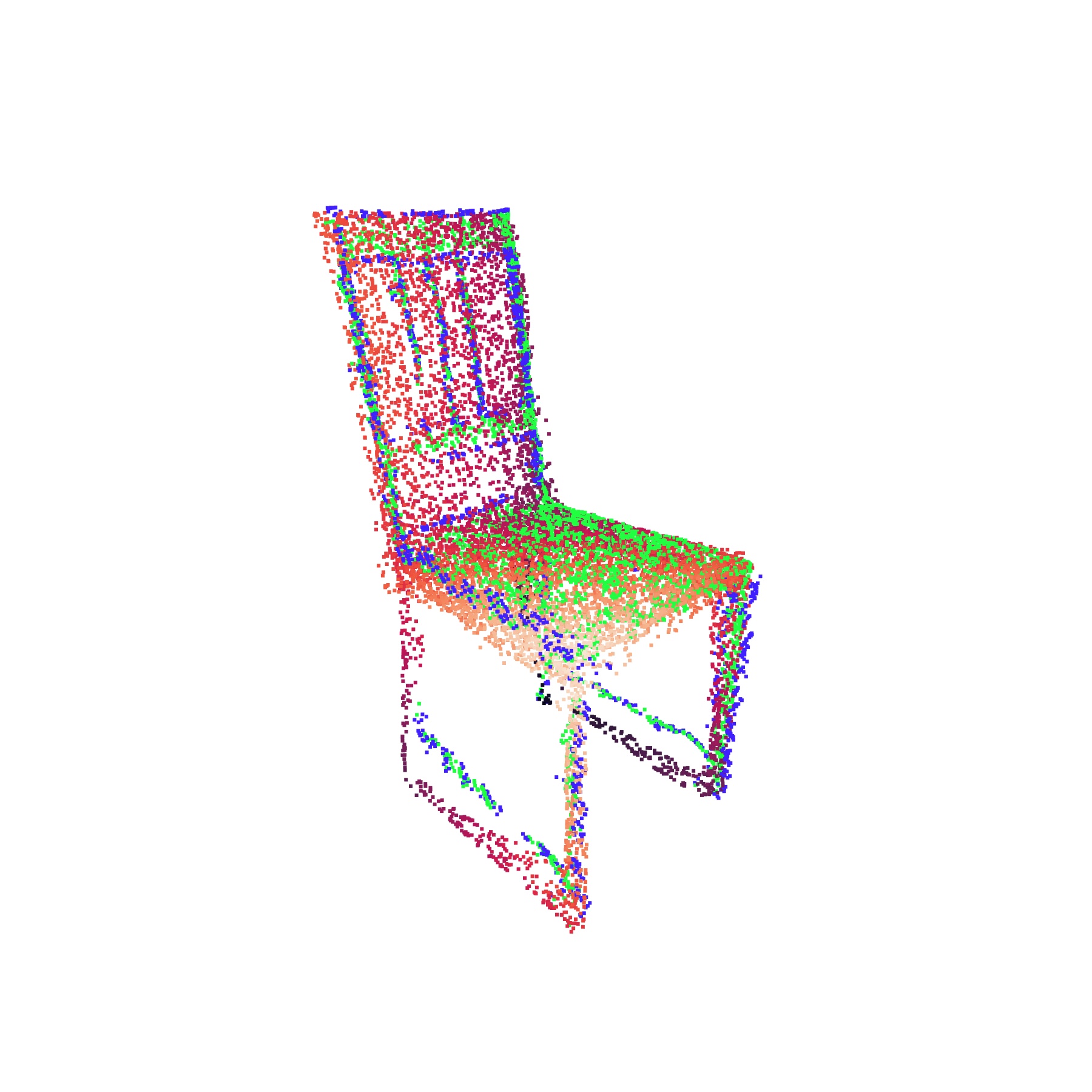}
		
		\includegraphics[height=80pt, trim=300 0 300 0, clip]
		{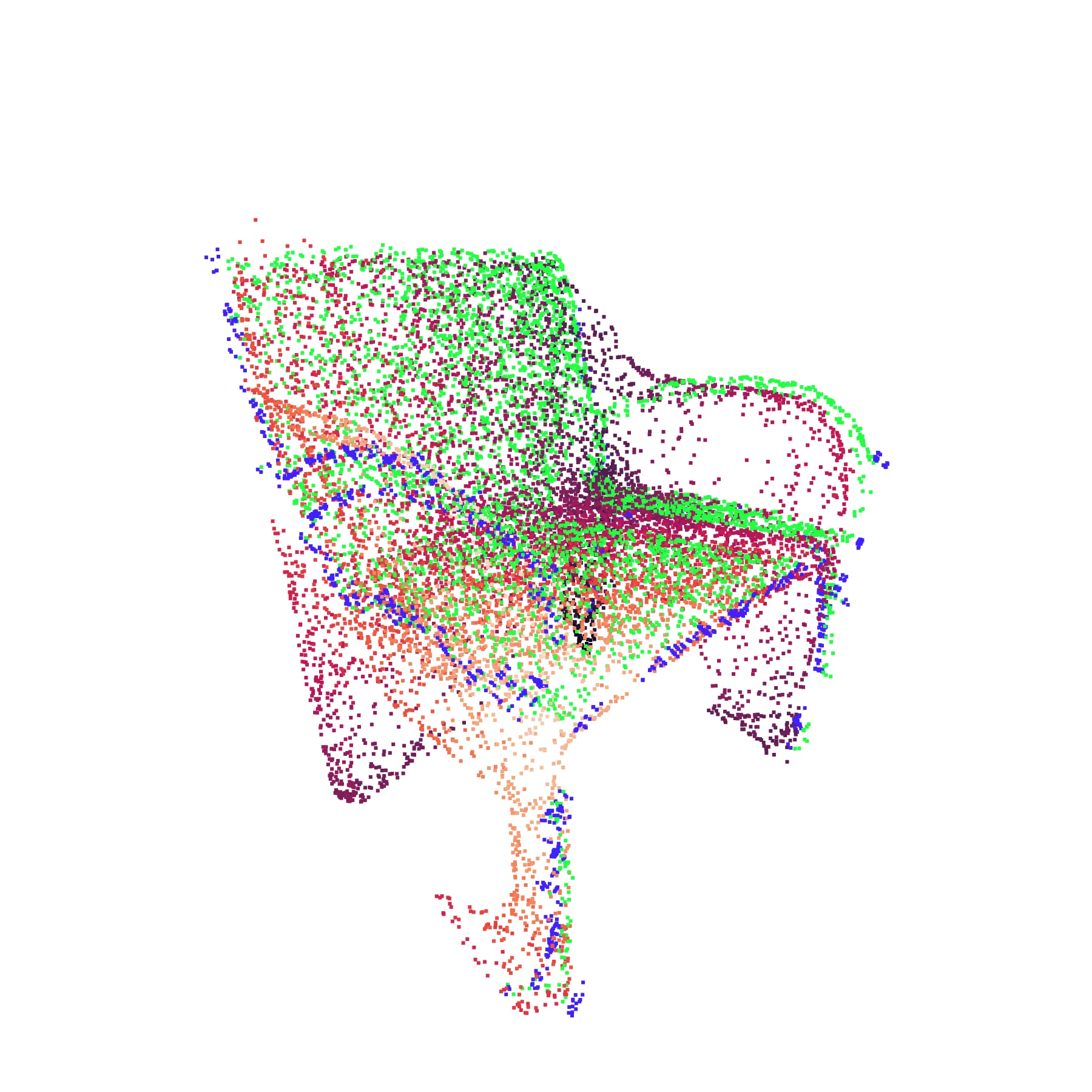}
		
		\caption{Coarse}
	\end{subfigure}
	\begin{subfigure}[t]{0.1\textwidth}
		\centering
		\includegraphics[height=80pt, trim=515 200 450 320, clip]
		{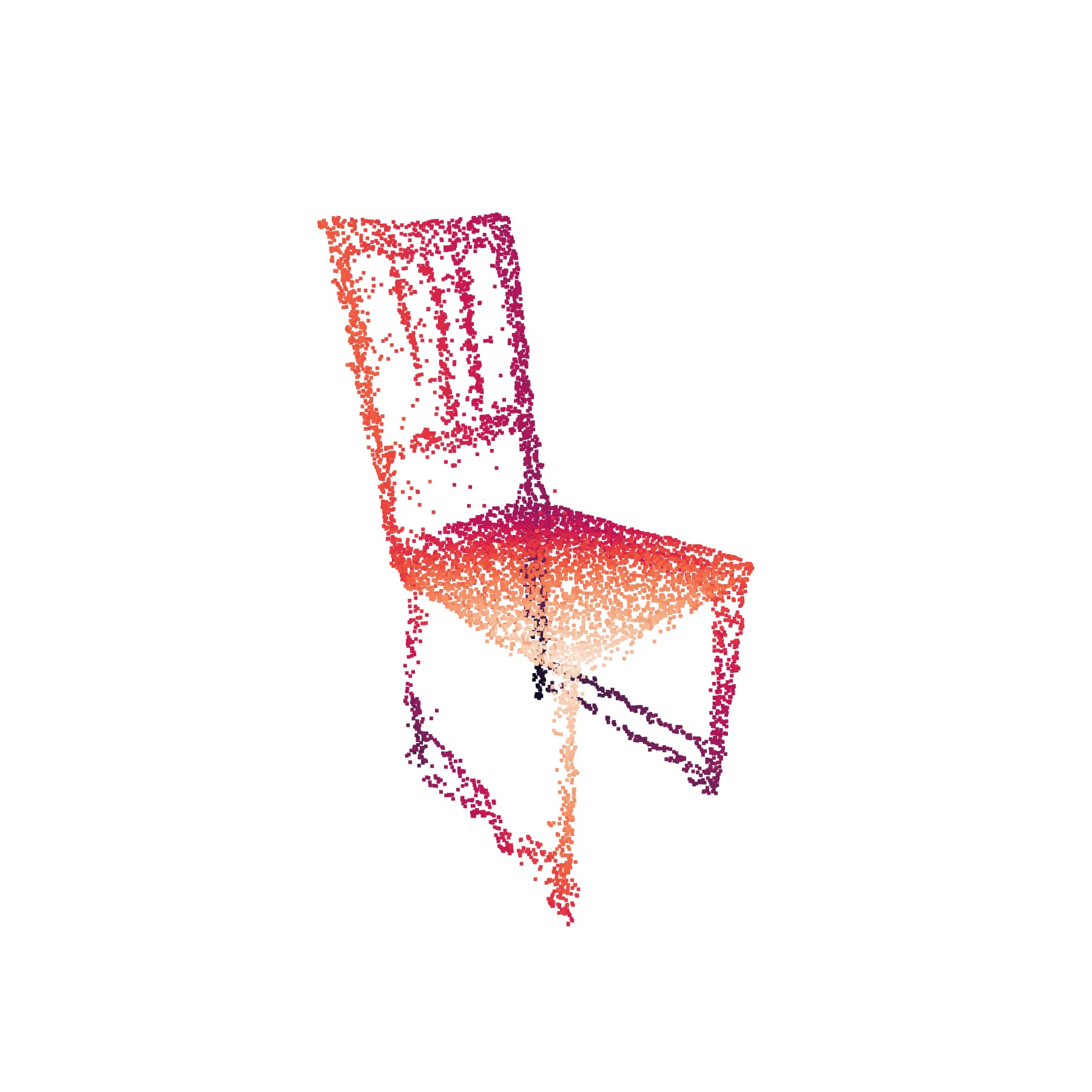}
		
		\includegraphics[height=80pt, trim=300 0 300 0, clip]
		{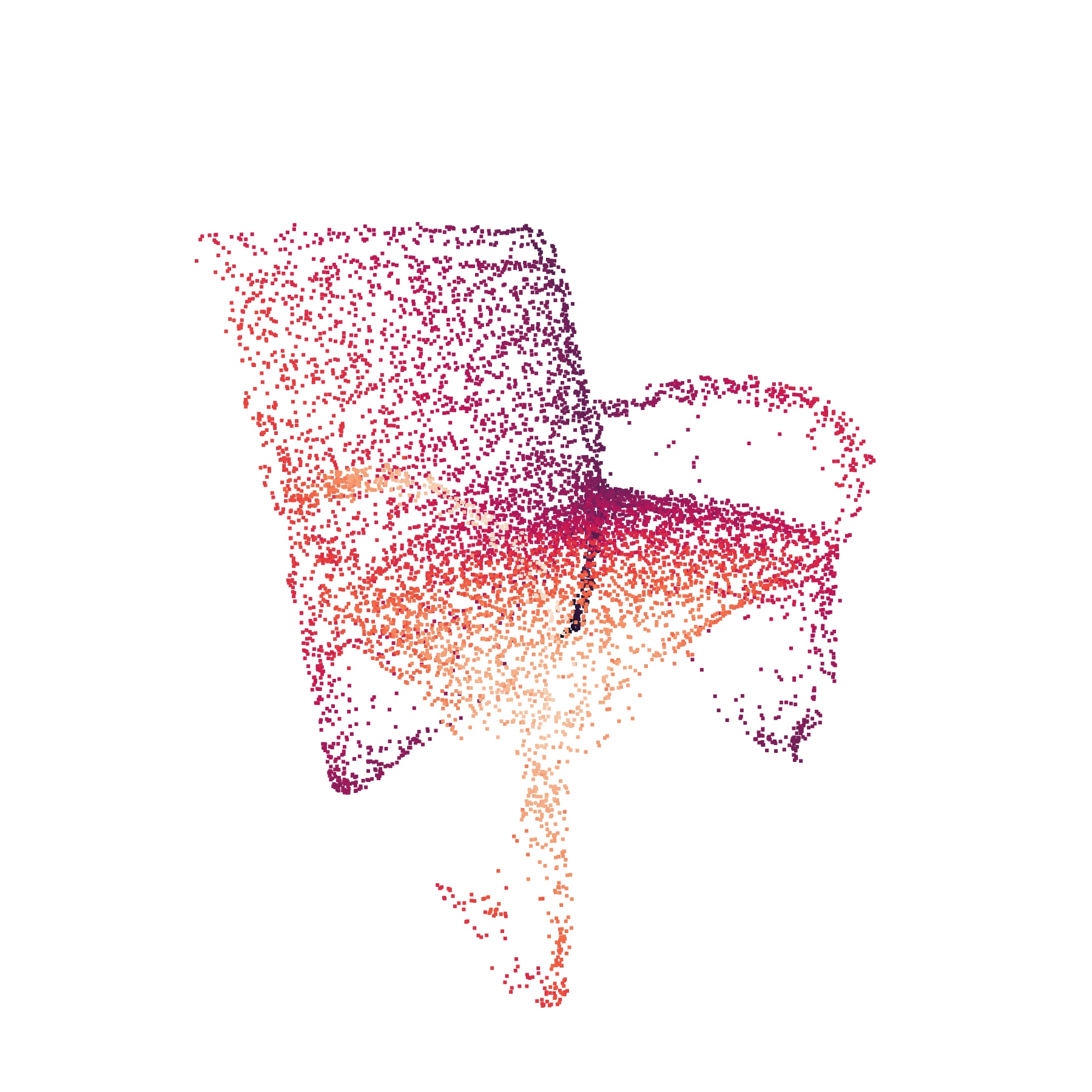}
		
		\caption{Output}
	\end{subfigure}
	\begin{subfigure}[t]{0.1\textwidth}
		\centering
		\includegraphics[height=80pt, trim=535 200 450 320, clip]
		{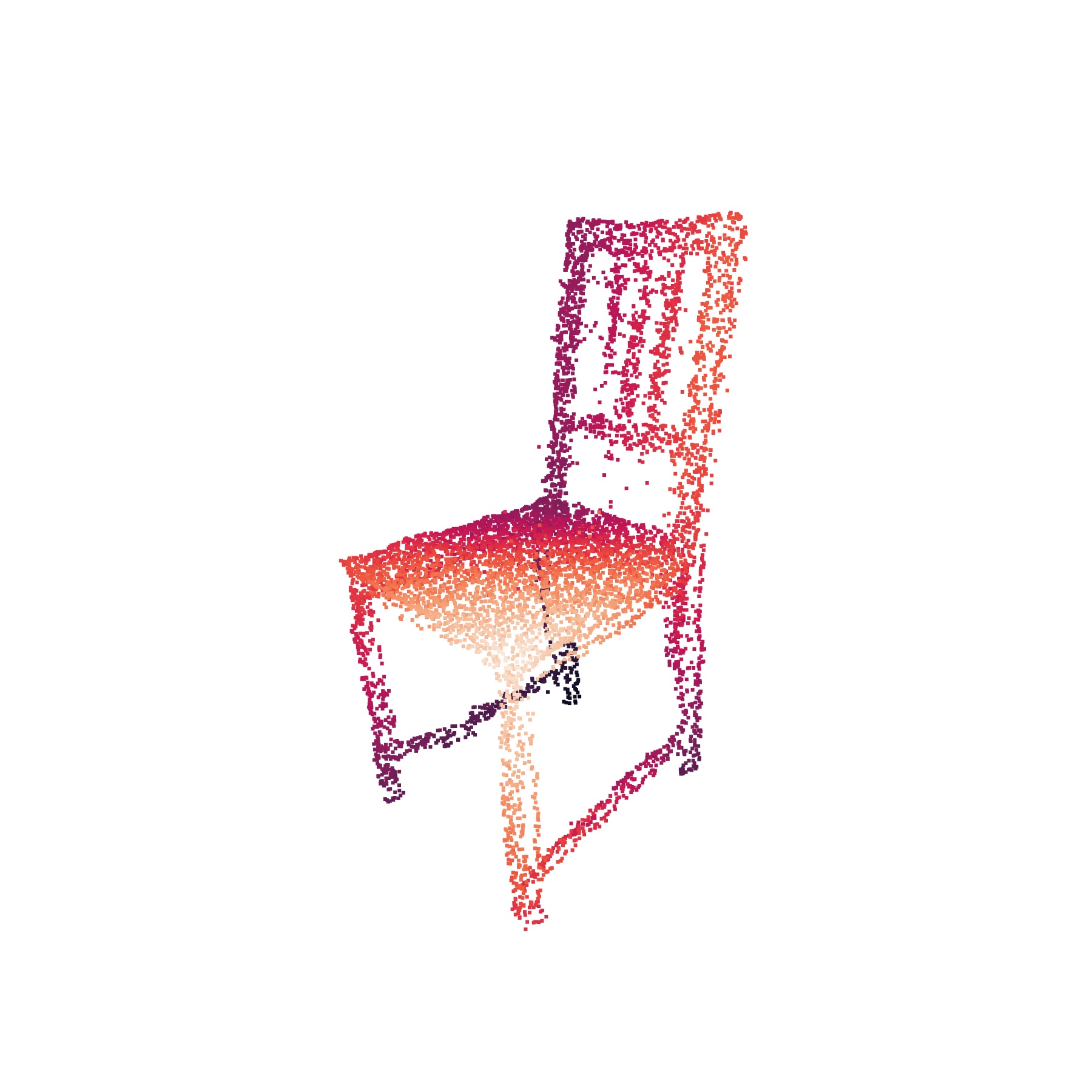}
		
		\includegraphics[height=80pt, trim=300 0 300 0, clip]
		{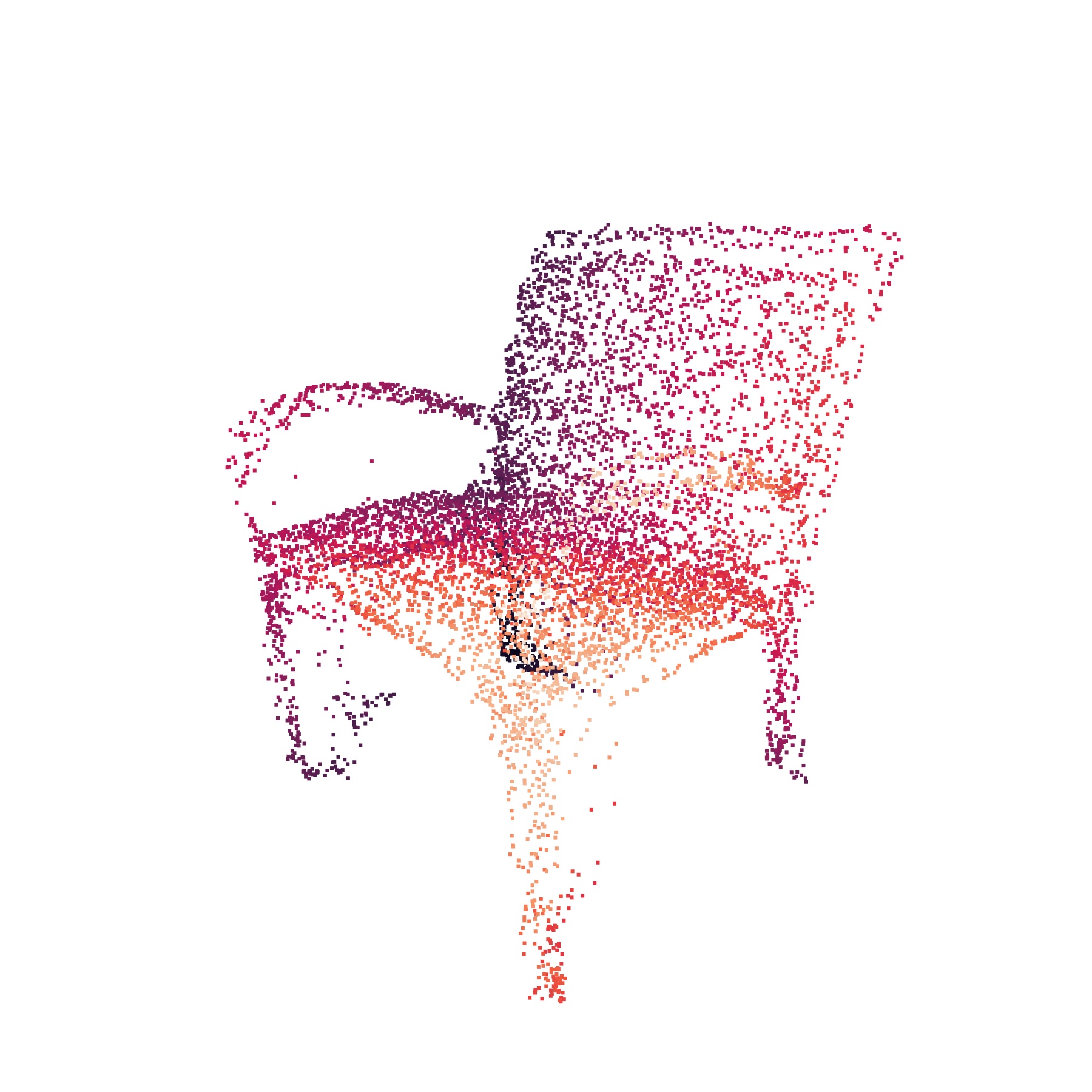}
		
		\caption{Output}
	\end{subfigure}
	
	\caption{Tests on real scans. From left to right: image and mask of the target object; input partial scan; predicted coarse point cloud; predicted final point cloud from two different viewpoints. Note that blue points in (b),(f)/(c),(g) are sampled from the empty rays $\mathcal{R}_{ray}$/$\mathcal{R}^{*}_{ray}$. Green points in (c), (g) are sampled from visible points.}
	\label{fig:realdata-scan}
\end{figure*}

\section{Tests on Real Scans}
\label{sec:realcase}

We train our network with ShapeNet and test it on real scans to investigate its generalization ability. Four real scans are taken from \cite{nie2020skeleton} for the testing, where the 3D points are back-projected from a single depth map and aligned to a canonical system. The corresponding 2D masks are extracted from RGB images.

From the results in Figure \ref{fig:realdata-scan}, we can see that our method can achieve plausible results `in the wild'. The ray feature works quite well even though the depth scans are captured by users without training. 

\section{Robustness of Encoding Emptiness}
\begin{figure}[h]
	\centering
	\begin{subfigure}[t]{0.23\textwidth}
		\includegraphics[width=\textwidth, trim=410 40 440 180, clip]  
		{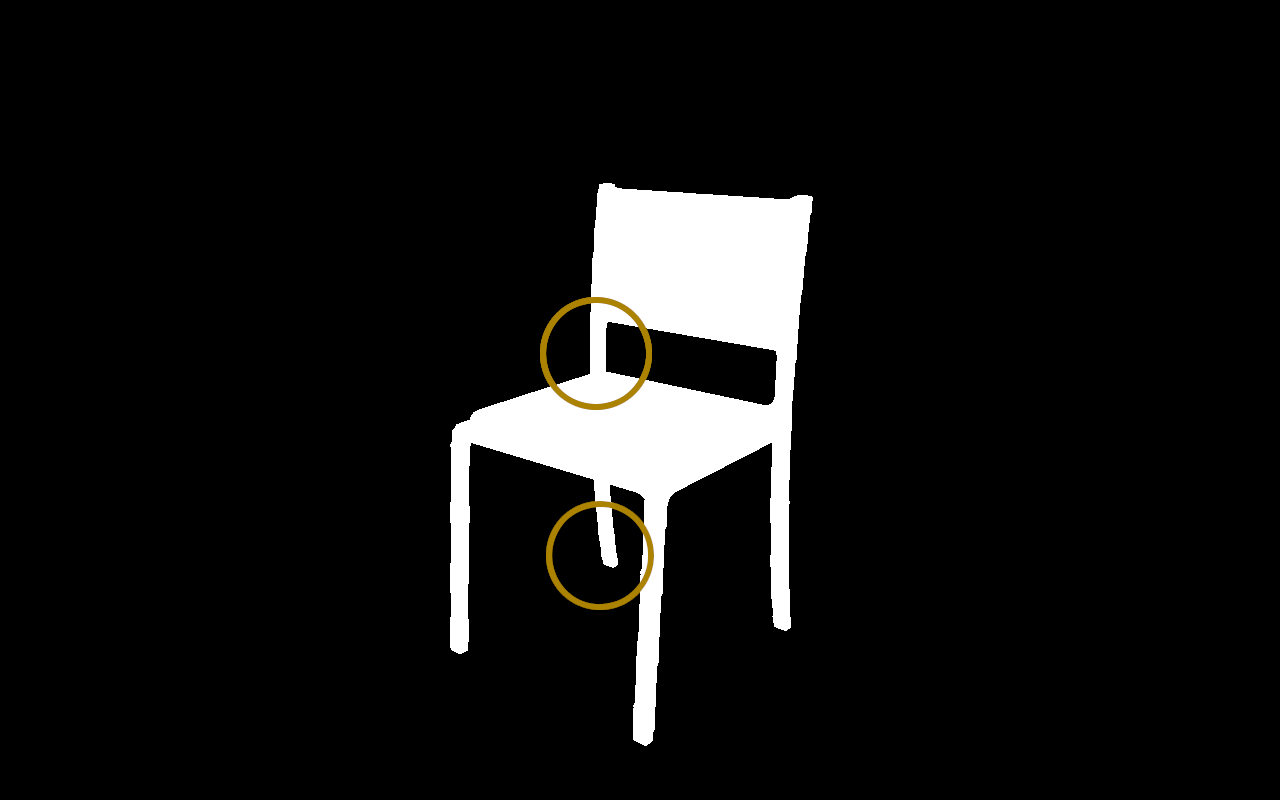}
		\caption{Before}
	\end{subfigure}
	\begin{subfigure}[t]{0.23\textwidth}
		\includegraphics[width=\textwidth, trim=410 40 440 180, clip]
		{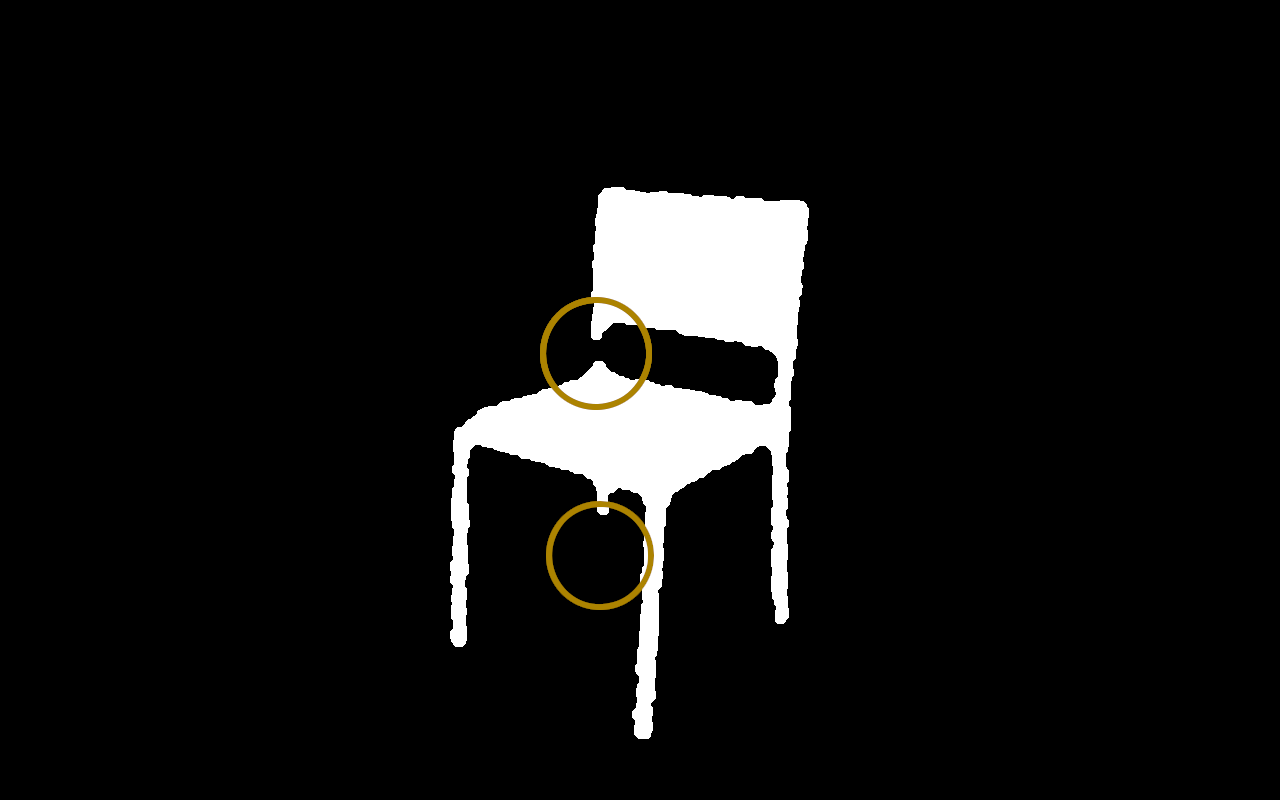}  
		\caption{After}
	\end{subfigure}
	\caption{Adding noise to simulate the mask from real world data: a) mask of chair back without noise; b) mask with noises on  boundaries.}
	\label{fig:robustness-mask-compare-supp}
\end{figure}

In real-world applications, the input point clouds are usually involved with background points. For such a scenario, former works usually adopt a binary mask on the image plane to filter background points \cite{yin2018p2p,nie2020skeleton}. This kind of masks can be directly used in our cases to learn emptiness. Since the mask should be given for most methods in real applications, in this section, we would like to explore the robustness of our method to imperfect masks. We simulate the masks extracted from real-world depth/RGB data, and add random segmentation errors (see Figure~\ref{fig:robustness-mask-compare-supp}) to the boundaries of masks. We fine-tune our model on all categories using the noisy mask for 2 epochs.
Test results are shown in Table \ref{tab:ablation-table}. From the results we can see that the performance measured in both EMD and CD is only degraded slightly, which verifies the robustness of our method. It further demonstrates that, for synthesized depth scans, we can obtain masks by thresholding the depth maps; for real scans, we can use segmentation methods to extract masks from RGB/RGBD images. For both cases, our method can deliver faithful results.

\begin{table*}[t]
	\centering
	\begin{subtable}{\textwidth}
	\centering
	\setlength{\tabcolsep}{2pt}
	\begin{tabular}{c|cccccccccccccc|c}
		\Xhline{2\arrayrulewidth}
		methods     & faucet & cabinet & table & chair & vase  & lamp  & bottle & clock 
     & display & knife & mug  & fridge  & scissors & trashcan &  average \\ \hline
     
		PCN         & 16.49  & 9.34    & 11.34 & 10.94 & 12.43 & 16.10 & 8.07 & 8.48 
	 & 10.24   & 8.26  & 9.49 & 9.21          &  10.98   & 10.50    &  10.85  \\
	 
		PCN+Ray     & 13.63  & 8.25    & 10.79 &  9.74 & 11.10 & 14.30 & 5.91 & 6.51
	 &  8.21   & 6.88  & 7.02 & 6.97   &  9.95   & 7.89 & 9.08   \\
		
		CRN         & 13.43  & 9.85    &  7.93 &  8.67 & 12.49 & 11.38 & 10.23 & 7.76
	 & 8.47    & 5.47  & 12.16 & 11.25 & 7.22      & 12.61    &  9.92    \\
		
		GRNet       & 10.36  & 7.75    &  7.50 &  7.74 & 11.21 & 10.74 &  9.11 & 7.52
	 & 7.18    & 8.53  & 9.46  & 8.62   &  7.64  &  9.76 & 8.79  \\
		
		MSN         & 7.71   & 6.70    &  6.52 &  6.57 &  6.89 &  7.55 &  5.17 & 5.77
	 & 6.06    & 4.51  & 5.63  & 6.68   & 4.29  &  6.26  & 6.17   \\
		\hline
		
		Ours  &  \textbf{6.31}  & \textbf{6.14}    &  \textbf{5.33} &  \textbf{5.12} &  \textbf{5.93} &  \textbf{6.76} & \textbf{4.06} & \textbf{4.45}
	 &  \textbf{4.46}   &  \textbf{3.65} & \textbf{4.29}   &  \textbf{4.97} 
	 &  \textbf{3.69}  &  \textbf{4.92}  &  \textbf{5.01}
		\\ \Xhline{2\arrayrulewidth}
	\end{tabular}
    \caption{Evaluation on EMD ($\times 10^2$) with Res.=8,192 }
    \vspace{1em}
	\label{tab:sota-table-emd-8k-supp}
	\end{subtable}

	\begin{subtable}{\textwidth}
	\centering
	\setlength{\tabcolsep}{2pt}
	\begin{tabular}{c|cccccccccccccc|c}
		\Xhline{2\arrayrulewidth}
		methods     & faucet & cabinet & table & chair & vase  & lamp  & bottle & clock 
		& display & knife & mug  & fridge  & scissors & trashcan &  average \\ \hline
		
		PCN         & 4.17   & 4.67    &  3.82 & 4.01  &  6.31 &  3.73 & 3.75 & 4.67 
		& 4.15    & 1.82  & 5.93 & 4.65    & 2.33     & 5.33     & 4.24           \\
		
		PCN+Ray     & 2.80   & 4.55    &  3.57 & 3.81  &  5.80 &  3.12 & 3.24 & 3.67
		& \textbf{2.97} & 1.57 & 4.34 & \textbf{3.51} &  \textbf{1.26}  &  4.50  & 3.48    \\
		
		CRN    & 3.67   & \textbf{4.49}    &  \textbf{3.44} & 3.81  &  5.49 &  3.19 & 3.35 & 4.32 
		& 4.00 & 1.64   & 5.58 &  4.52  &   2.06  &  5.08  &  3.90    \\
		
		GRNet       & 3.28   & 4.66    &  3.73 & 3.94  &  5.53 &  3.52 & 4.51 & 4.77
		& 4.08 & 2.03    & 6.17  & 4.99   & 2.15   &  5.37 & 4.20   \\
		
		MSN         & 4.02   & 5.75    &  4.61 & 4.81  &  5.71 &  4.34 &  4.55 & 4.86
		& 4.45 & 1.89   &  5.42  & 5.25 & 2.04  & 5.49 & 4.51  \\
		\hline
		Ours  & \textbf{2.62}   & 4.72    &  3.76 & \textbf{3.62}  &  \textbf{4.54} &  \textbf{3.02} & \textbf{3.11}   & \textbf{3.59} 
		& 3.52    &  \textbf{1.46}  & \textbf{4.28}  & 4.17  &  1.51  &  \textbf{4.48} & \textbf{3.46}  
		\\ \Xhline{2\arrayrulewidth}
	\end{tabular}
	\caption{Evaluation on CD ($\times 10^2$) with Res.=8,192}
	\label{tab:sota-table-cd-8k-supp}
	\end{subtable}
	\caption{Comparison with Existing Methods. Evaluation with Res.=8,192}
	\label{tab:sota-table-8k}
\end{table*}

\begin{table*}[t]
	\centering
	\begin{subtable}{\textwidth}
	\setlength{\tabcolsep}{2pt}
	\begin{tabular}{c|cccccccccccccc|c}
		\Xhline{2\arrayrulewidth}
		methods     & faucet & cabinet & table & chair & vase  & lamp  & bottle & clock 
		& display & knife & mug  & fridge  & scissors & trashcan &  average \\ \hline
		
		PCN         & 16.81  & 10.47   & 12.22 & 11.81 & 13.25 & 16.67 & 8.62 & 9.63
		& 11.07   & 8.64  & 10.83 & 10.47  & 11.58    & 11.64   & 11.69    \\
		
		PCN+Ray     & 16.13  & 10.18   & 11.68 & 10.61 & 11.13 & 14.90 & 7.02 & 8.16
		& 9.40    & 7.23  & 8.90  & 8.93   & 9.79    & 9.63    & 10.26    \\
		
		PF-Net      & 16.11  & 10.04   &  9.97 & 10.61 & 11.50 & 14.07 & 9.17 & 10.96
		& 9.55    & 10.04 & 10.21 & 10.55  & 11.02    & 9.79  & 10.97    \\
		
		P2P-Net     & 16.09  & 11.64   & 10.73 & 12.29 & 16.36 & 13.52 & 18.10 & 11.95
		& 11.00   & 13.28 & 20.55 & 15.63 & 8.78 & 16.59 & 14.04   \\
		
		SoftPoolNet & 15.03  & 14.30   & 11.28 & 14.05 & 17.63 & 15.89 & 18.35 & 13.05
		& 10.52   & 10.66 & 17.58 & 17.34 & 11.47 & 18.87 & 14.72    \\
		
		CRN         & 14.00  & 11.00   &  9.09 &  9.70 & 13.32 & 12.09 & 11.02 & 9.01
		& 9.39    & 5.84  & 13.07 & 12.41 & 7.69 & 13.18 & 10.77    \\
		
		GRNet       & 11.30  &  9.16   &  8.61 &  8.82 & 12.27 & 11.28 &  9.98 & 8.83
		& 8.24    & 9.07  &  11.06 & 9.91 & 7.70 & 11.12 & 9.81   \\
		
		MSN         &  8.52  &  8.19   &  7.82 &  7.82 &  8.36 &  8.51 &  6.43 & 7.41
		& 7.14    & 4.92  &  7.84  & 8.28 & 4.94 & 8.29 & 7.46   \\
		\hline
		Ours  &  \textbf{6.89}  &  \textbf{7.48}   &  \textbf{6.63} &  \textbf{6.63} &  \textbf{7.16} &  \textbf{7.48} & \textbf{5.53} & \textbf{6.19}
		& \textbf{6.02} & \textbf{4.44} & \textbf{6.67} & \textbf{6.87} & \textbf{4.00} & \textbf{7.04} & \textbf{6.36}   
		\\ \Xhline{2\arrayrulewidth}
	\end{tabular}
	\caption{Evaluation on EMD ($\times 10^2$) with Res.=2,048}
	\vspace{1em}
	\label{tab:sota-table-emd-2k-supp}
	\end{subtable}

	\begin{subtable}{\textwidth}
    \centering
	\setlength{\tabcolsep}{2pt}
	\begin{tabular}{c|cccccccccccccc|c}
		\Xhline{2\arrayrulewidth}
		methods     & faucet & cabinet & table & chair & vase  & lamp  & bottle & clock 
		& display & knife & mug  & fridge  & scissors & trashcan &  average \\ \hline
		
		PCN         & 5.62   & 7.28    & 5.95  & 6.14  &  8.71 &  5.15 & 5.53 & 6.97
		& 6.29   & 2.64  & 8.79 & 7.38  & 3.26    & 8.09   & 6.27    \\
		
		PCN+Ray     & 4.35   & 7.14    & 5.19  & 5.98  &  7.19 &  4.61 & 4.45 & 6.14
		& \textbf{5.23}   & 2.35  & 7.21 & \textbf{6.33} & \textbf{2.17} & 7.28 & 5.40    \\
		
		PF-Net      & 8.96   & 8.15    & 6.94  & 7.48  & 10.10 &  7.56 & 6.96 & 8.67
		& 7.16   & 4.12  & 9.80  & 8.54  & 5.24  & 9.08 & 7.77   \\
		
		P2P-Net     & 4.47   & 7.21 & \textbf{5.49} & 5.92  &  7.62 &  4.41 & 7.01 & 6.79 
		& 6.45   & 2.77 & 8.71  & 8.22  & 2.47 & 8.42 & 6.14    \\
		
		SoftPoolNet & 5.54   & 7.85    & 6.41  & 6.59  &  8.27 &  5.56 & 6.67 & 7.63 
		& 6.59  & 3.05 & 8.90 & 8.24  & 3.45  & 8.57 & 6.67    \\
		
		CRN    & 5.14   & 7.13    & 5.59  & 5.94  &  7.96 &  4.63  & 5.28 & 6.72 
		& 6.12 & 2.48   & 8.49 &  7.30  &   2.96  &  7.94  &  5.98    \\
		
		GRNet       & 4.72   & 7.21    & 5.77  & 6.00  &  7.90 &  4.92 & 6.24 & 7.06
		& 6.17 & 2.89 & 8.86 & 7.62 & 3.03 & 8.14 & 6.18   \\
		
		MSN         & 5.25   & 8.06    & 6.50  & 6.70  &  7.92 &  5.66 & 6.16 & 7.01
		& 6.36 & 2.67 & 8.16 & 7.62 & 2.88 & 8.07 & 6.36    \\
		\hline
		Ours     & \textbf{3.90}   & \textbf{7.01}    & 5.65  & \textbf{5.61}  &  \textbf{6.68} &  \textbf{4.26} & \textbf{4.92} & \textbf{5.88} 
		& 5.55 & \textbf{2.25} & \textbf{7.19} & 6.73 & 2.34 & \textbf{7.13} & \textbf{5.36}   
		\\ \Xhline{2\arrayrulewidth}
	\end{tabular}
	\caption{Evaluation on CD ($\times 10^2$) with Res.=2,048}
	\label{tab:sota-table-cd-2k-supp}
	\end{subtable}
	\caption{Comparison with existing methods. Evaluation with Res.=2,048}
	\label{tab:sota-table-2k}
\end{table*}

\begin{table*}[t]
	\centering
	\begin{tabular}{c|cccccc}
		\Xhline{2\arrayrulewidth}
		Category & only pts in rays  & remove $\mathcal{R}^{d}_{ray2}$ & remove $\mathcal{R}^{d}_{ray1}$ & noisy boundary & MSN & Ours         \\ \hline
		Faucet   & 6.54 / 2.61 & 7.88 / 4.34 & 6.88 / 2.82  & 6.94 / 2.90  & 7.71 / 4.02 & 6.31 / 2.62     \\
		Cabinet  & 6.35 / 4.72 & 6.55 / 5.52 & 6.72 / 5.02  & 6.71 / 4.99  & 6.70 / 5.75 & 6.14 / 4.72     \\
		Table    & 6.24 / 4.78 & 6.12 / 4.71 & 5.91 / 3.96  & 5.89 / 3.95  & 6.52 / 4.61 & 5.33 / 3.76     \\
		Chair    & 7.43 / 5.71 & 6.07 / 4.73 & 5.33 / 4.02  & 5.15 / 3.97  & 6.57 / 4.81 & 5.12 / 3.62     \\
		Vase     & 6.01 / 4.54 & 7.18 / 5.28 & 6.62 / 4.83  & 6.68 / 4.94  & 6.89 / 5.71 & 5.93 / 4.54     \\
		Lamp     & 7.14 / 2.95 & 8.71 / 4.19 & 7.30 / 3.23  & 7.21 / 3.27  & 7.55 / 4.34 & 6.76 / 3.02     \\ \hline
		Average  & 6.62 / 4.22 & 7.09 / 4.80 & 6.46 / 3.98  & 6.43 / 4.00  & 6.99 / 4.87 & 5.93 / 3.71     \\ \Xhline{2\arrayrulewidth}
	\end{tabular}
	\caption{Ablation study on different of configurations (EMD / CD $\times 10^2$)}
	\label{tab:ablation-table}
\end{table*}


\section{More Qualitative Comparisons}
\label{sec:qualitative}

\begin{figure*}[t]
	\centering
	\renewcommand{\arraystretch}{0}
	\renewcommand{\arrayrulewidth}{0pt}%
	\setlength{\tabcolsep}{2pt}
	
\resizebox{2.17\columnwidth}{!}{

\begin{tabular}{cccccccccc}
& Chair & Knife & Faucet & Table & Chair & Cabinet & Vase & Lamp & Table \\
Input & \begin{subfigure}{0.065\textwidth}\centering\includegraphics[trim=300 150 250 150,clip,width=\textwidth]{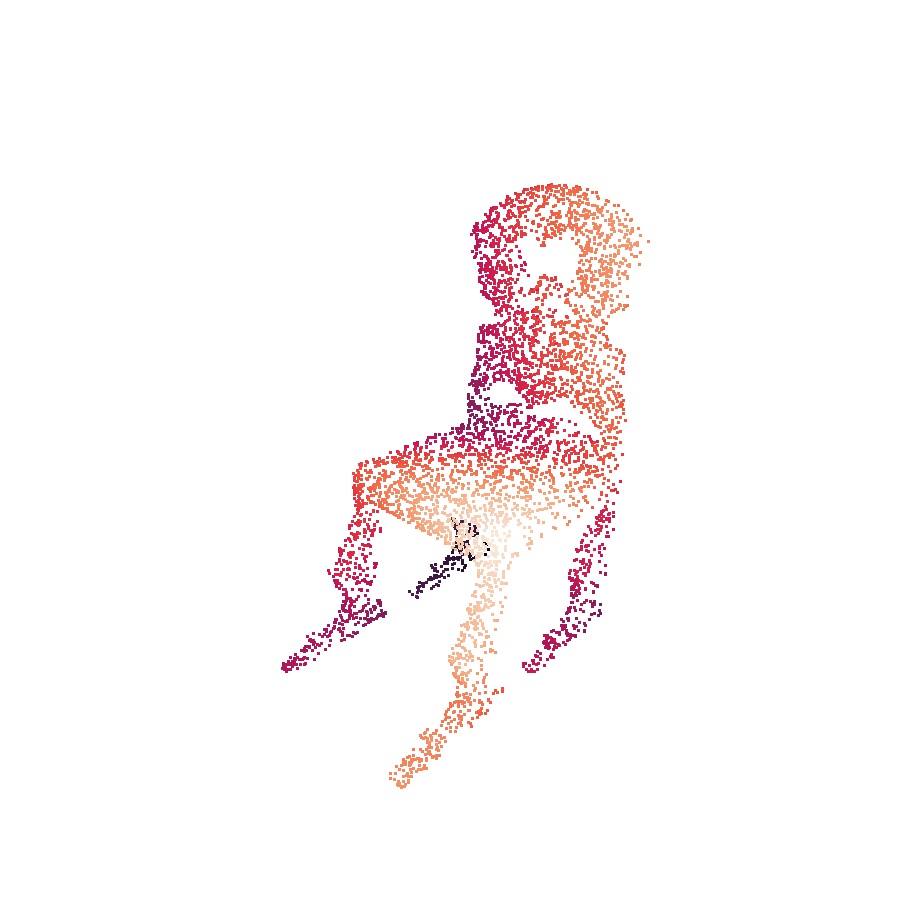}\end{subfigure} & \begin{subfigure}{0.03\textwidth}\centering\includegraphics[trim=380 250 400 190,clip,width=\textwidth]{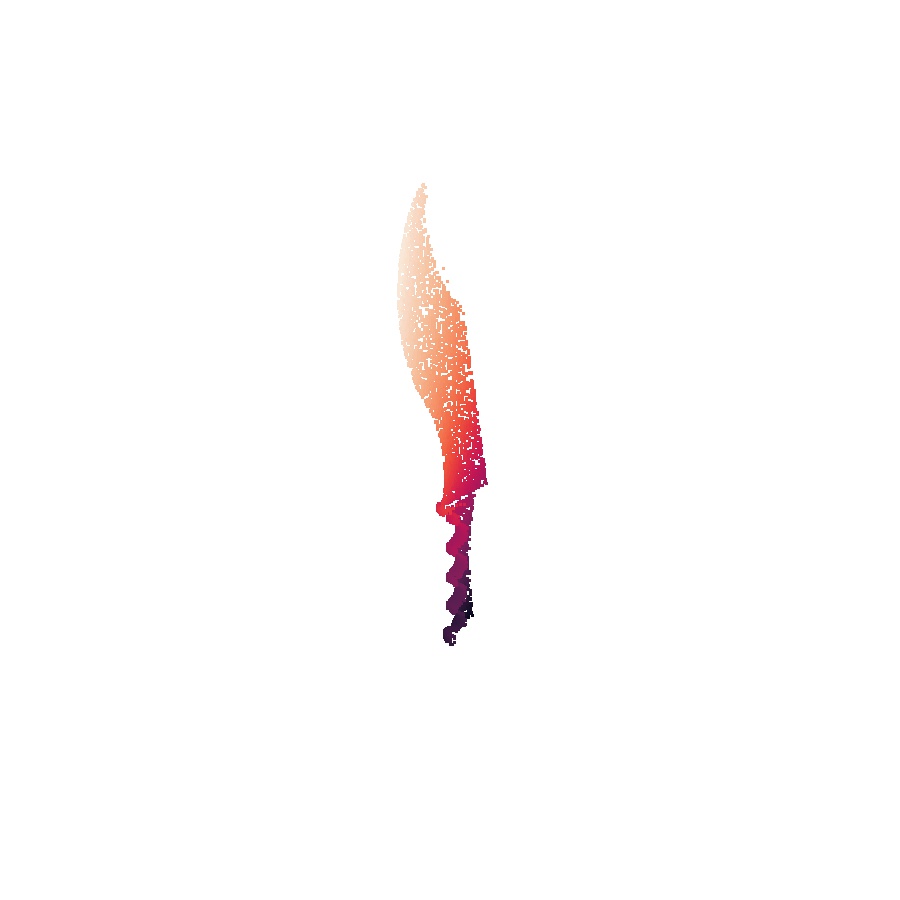}\end{subfigure} & \begin{subfigure}{0.09\textwidth}\centering\includegraphics[trim=200 170 220 100,clip,width=\textwidth]{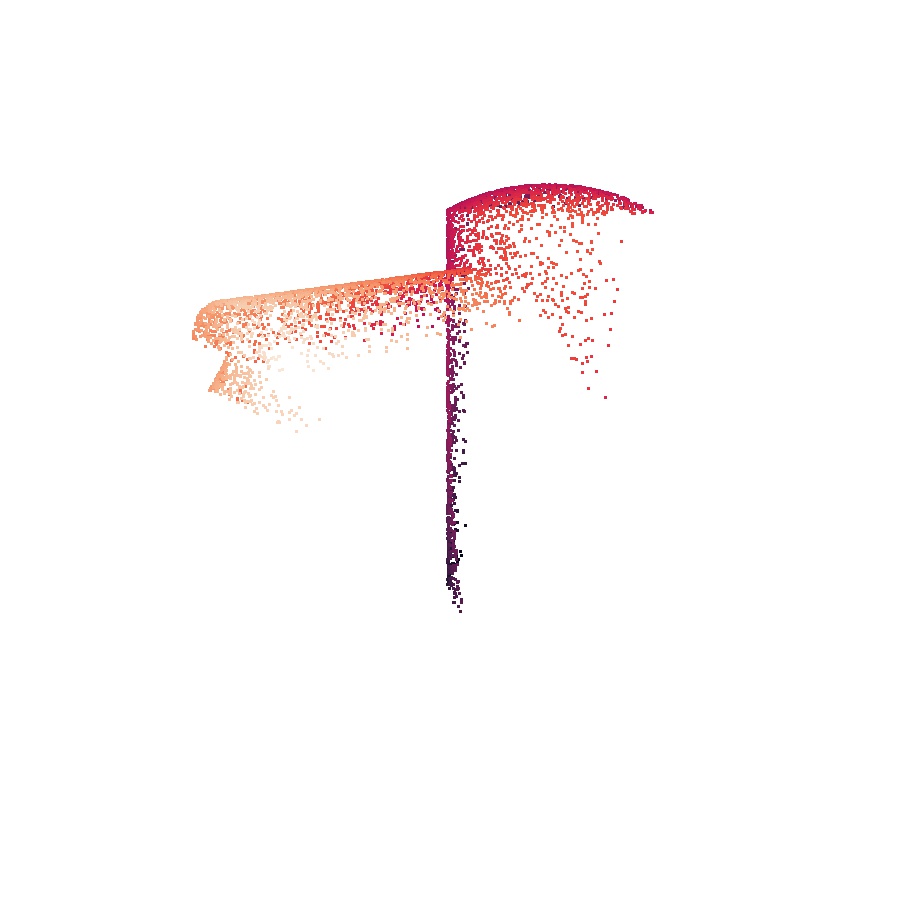}\end{subfigure} & \begin{subfigure}{0.1\textwidth}\centering\includegraphics[trim=100 170 120 110,clip,width=\textwidth]{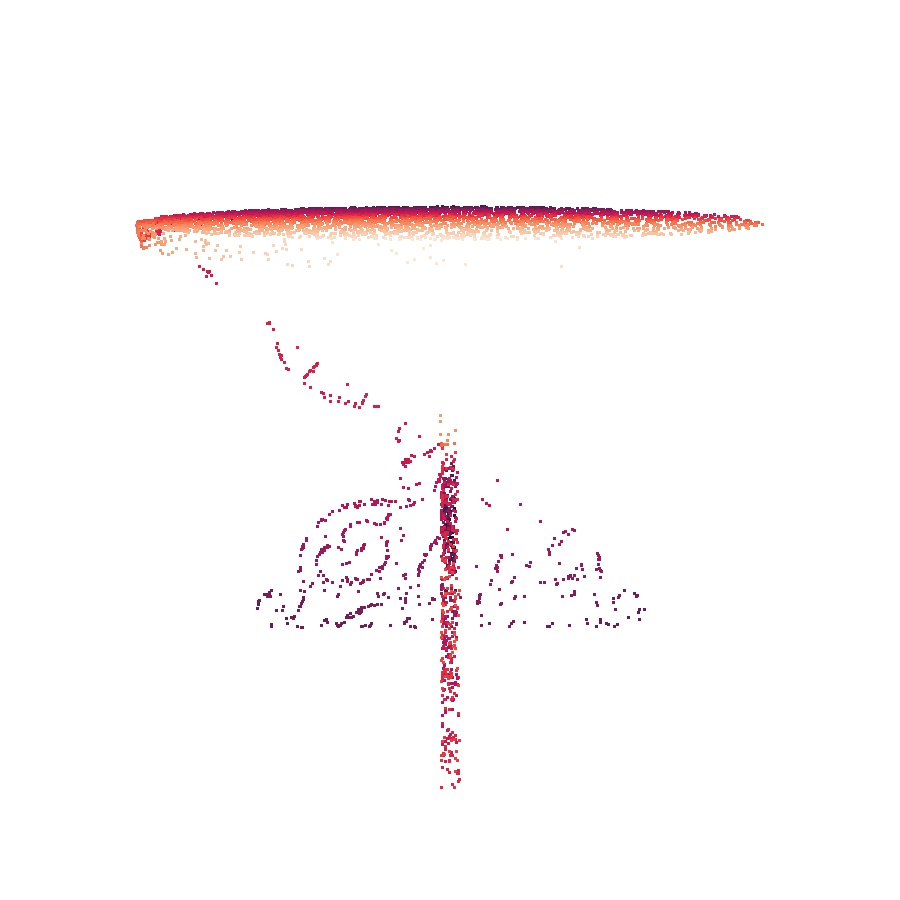}\end{subfigure} & \begin{subfigure}{0.065\textwidth}\centering\includegraphics[trim=250 120 220 100,clip,width=\textwidth]{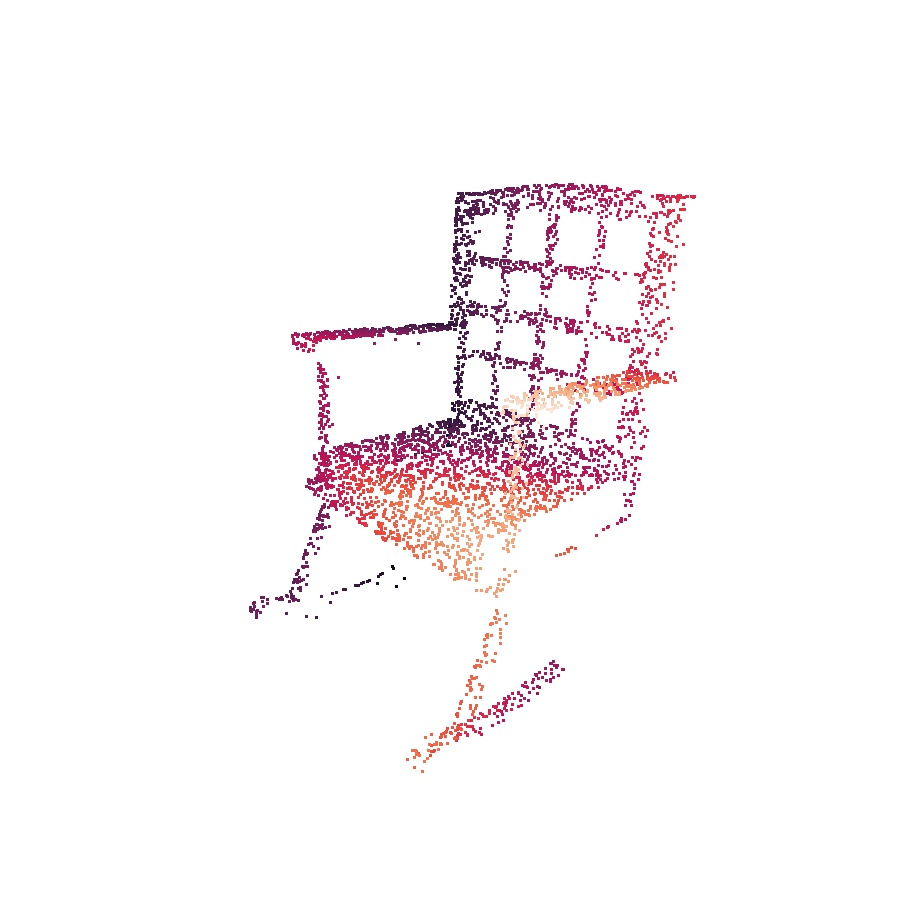}\end{subfigure} & \begin{subfigure}{0.14\textwidth}\centering\includegraphics[trim=130 250 180 190,clip,width=\textwidth]{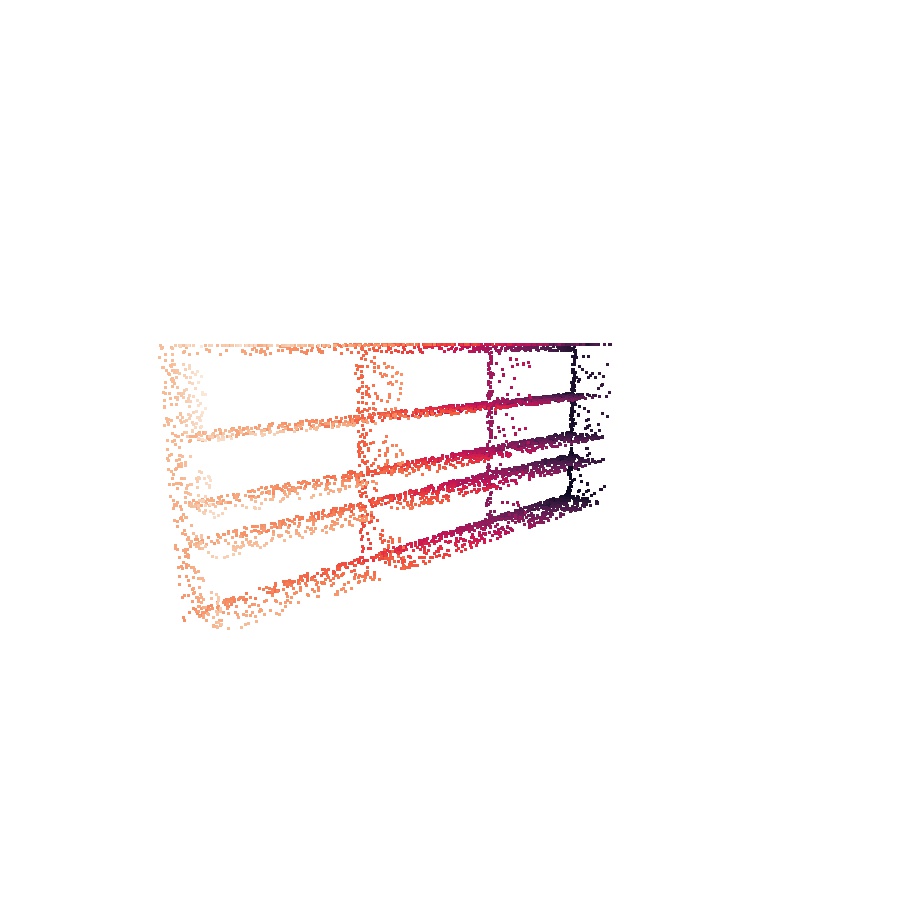}\end{subfigure} & \begin{subfigure}{0.04\textwidth}\centering\includegraphics[trim=370 220 340 190,clip,width=\textwidth]{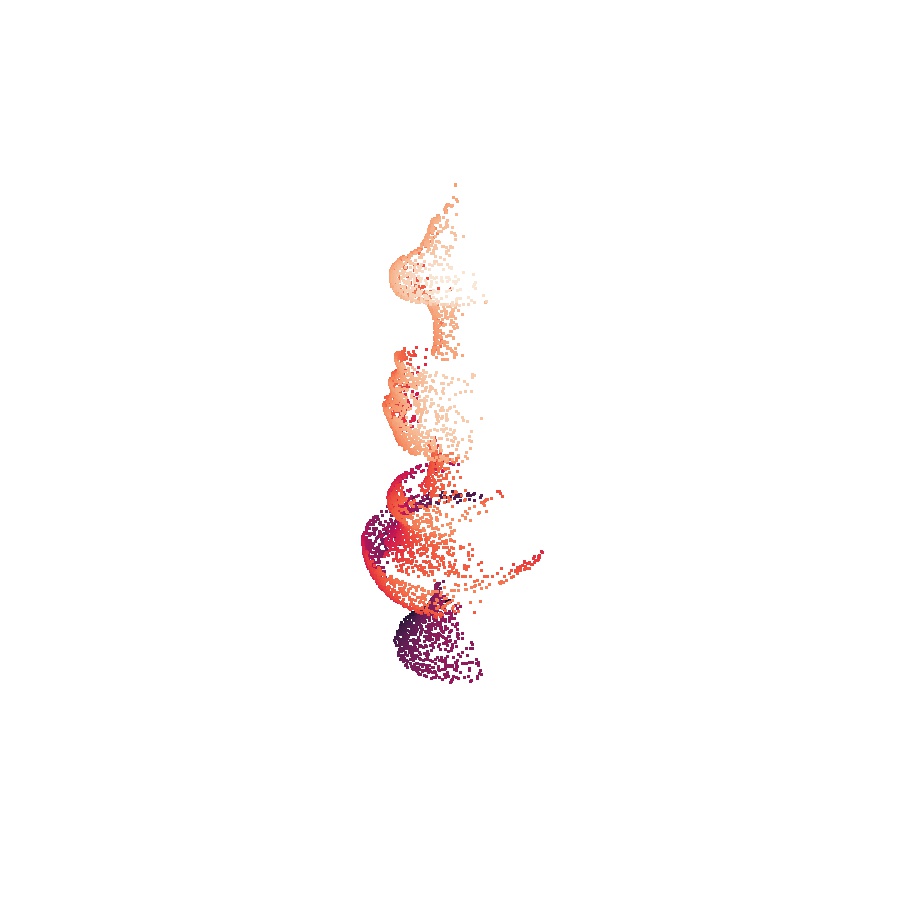}\end{subfigure} & \begin{subfigure}{0.065\textwidth}\centering\includegraphics[trim=260 260 320 160,clip,width=\textwidth]{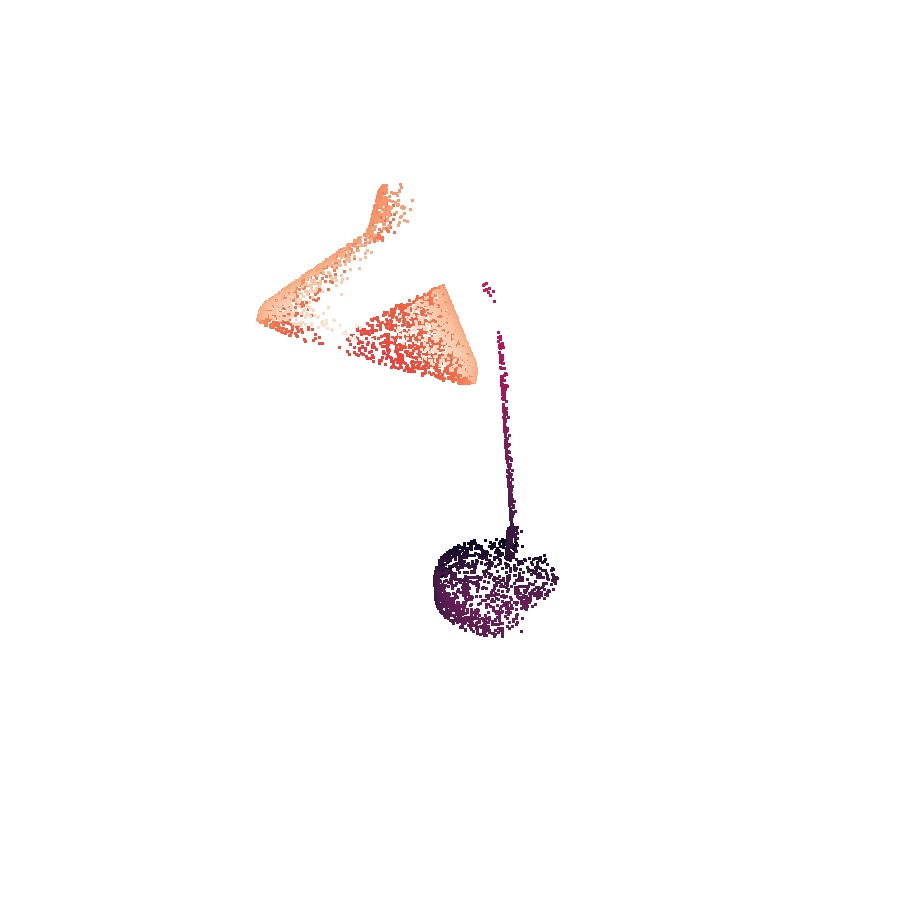}\end{subfigure} & \begin{subfigure}{0.13\textwidth}\centering\includegraphics[trim=100 200 150 190,clip,width=\textwidth]{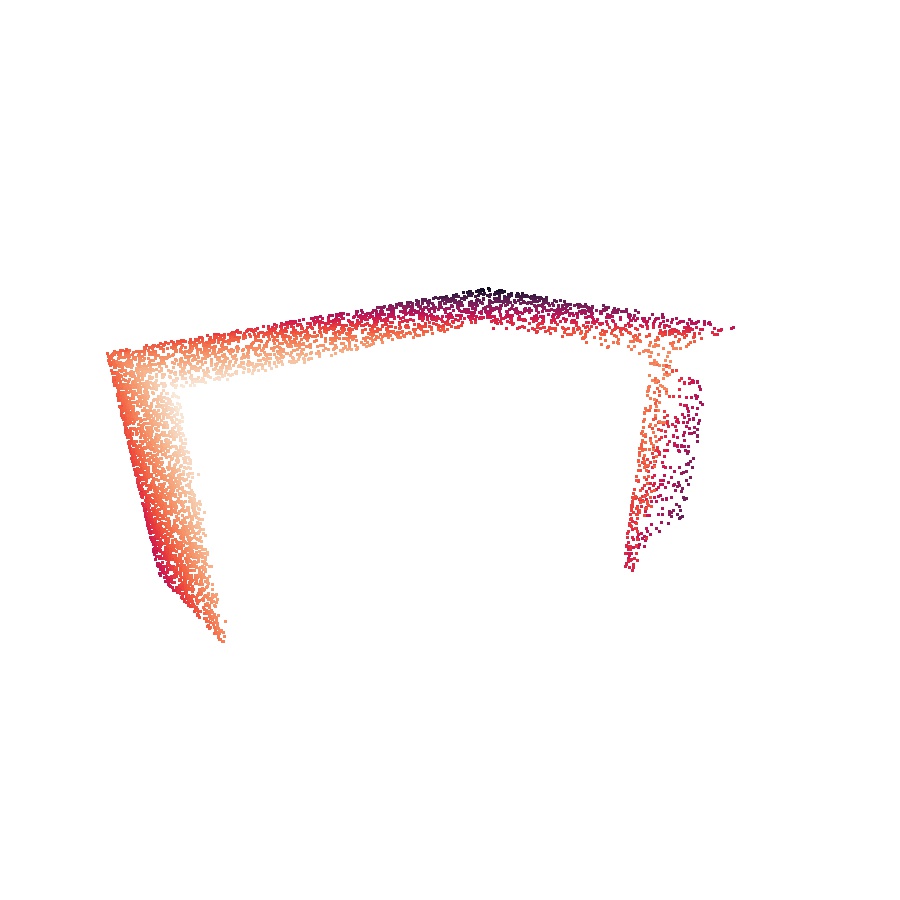}\end{subfigure} \\
PCN & \begin{subfigure}{0.065\textwidth}\centering\includegraphics[trim=300 150 250 150,clip,width=\textwidth]{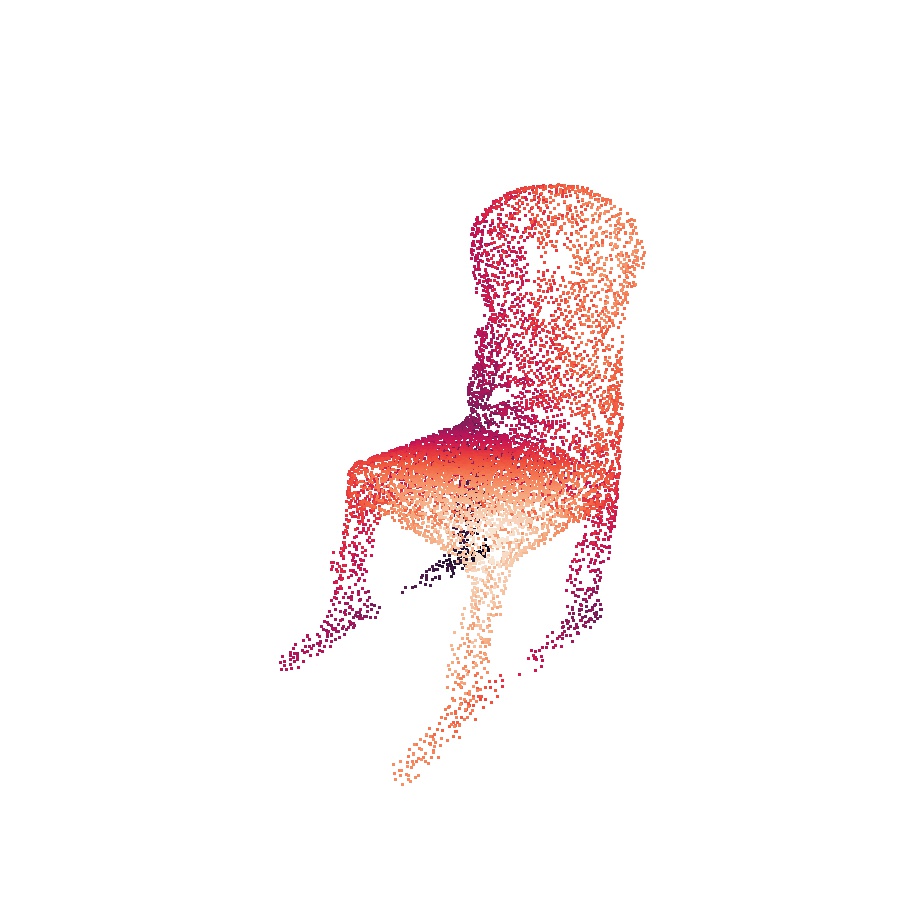}\end{subfigure} & \begin{subfigure}{0.03\textwidth}\centering\includegraphics[trim=380 230 400 190,clip,width=\textwidth]{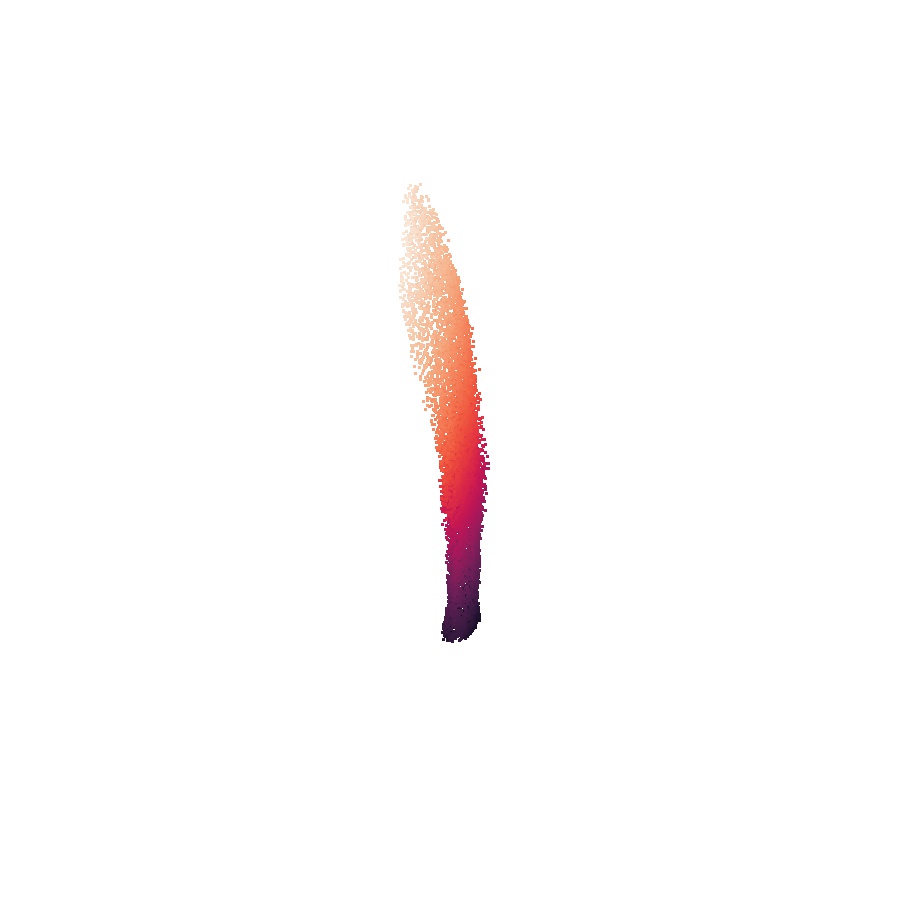}\end{subfigure} & \begin{subfigure}{0.09\textwidth}\centering\includegraphics[trim=200 170 220 100,clip,width=\textwidth]{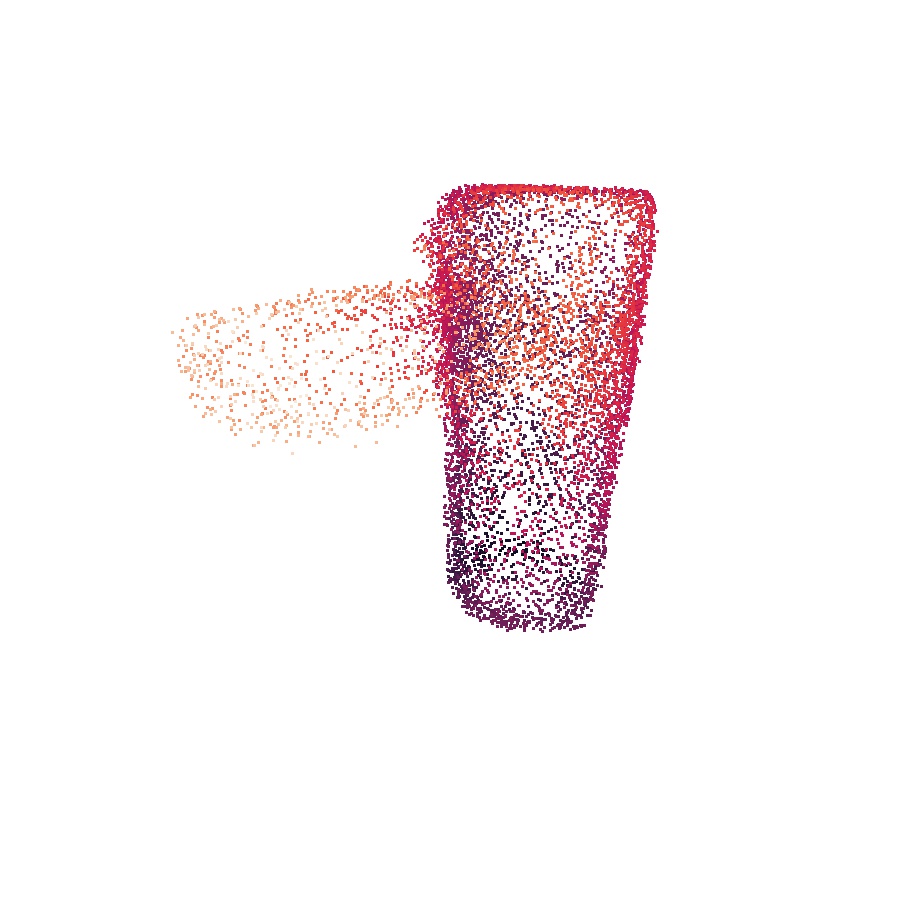}\end{subfigure} & \begin{subfigure}{0.1\textwidth}\centering\includegraphics[trim=100 170 120 110,clip,width=\textwidth]{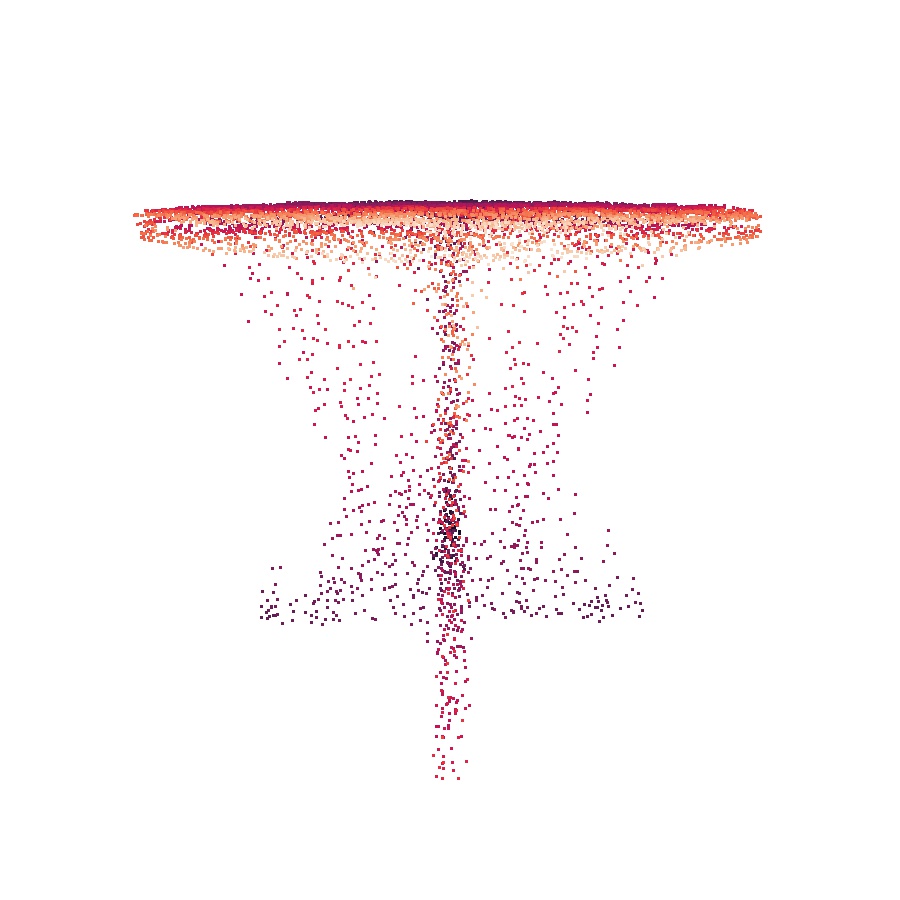}\end{subfigure} & \begin{subfigure}{0.065\textwidth}\centering\includegraphics[trim=250 120 220 100,clip,width=\textwidth]{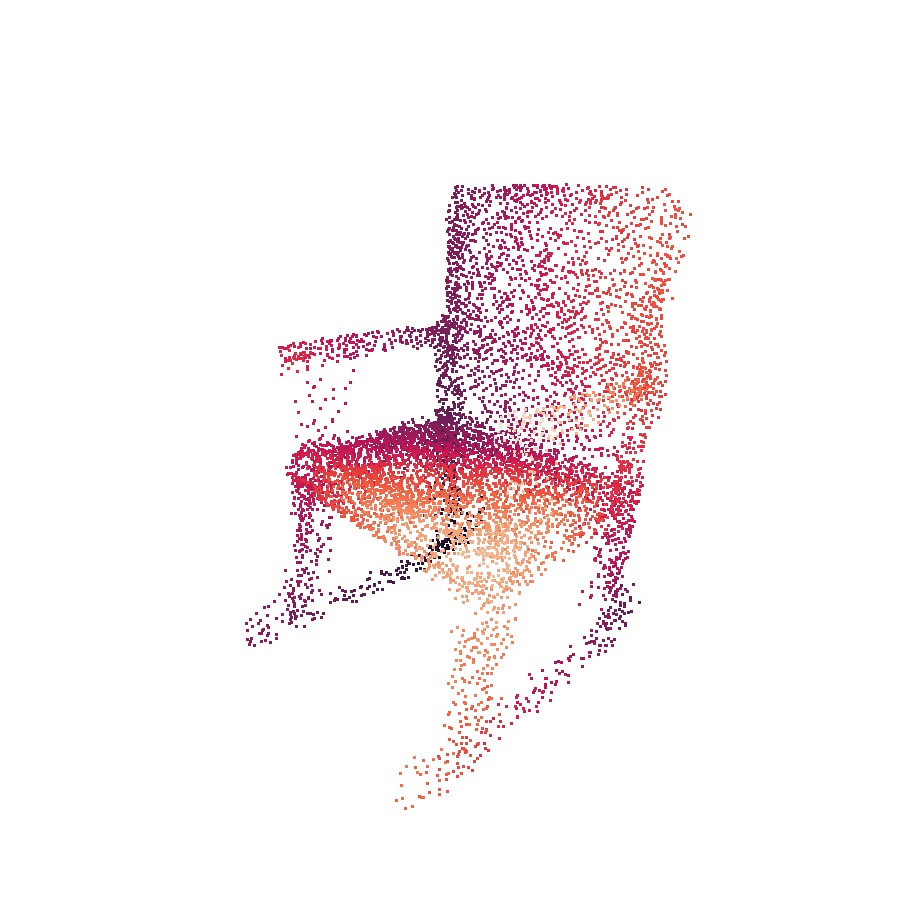}\end{subfigure} & \begin{subfigure}{0.14\textwidth}\centering\includegraphics[trim=130 250 230 190,clip,width=\textwidth]{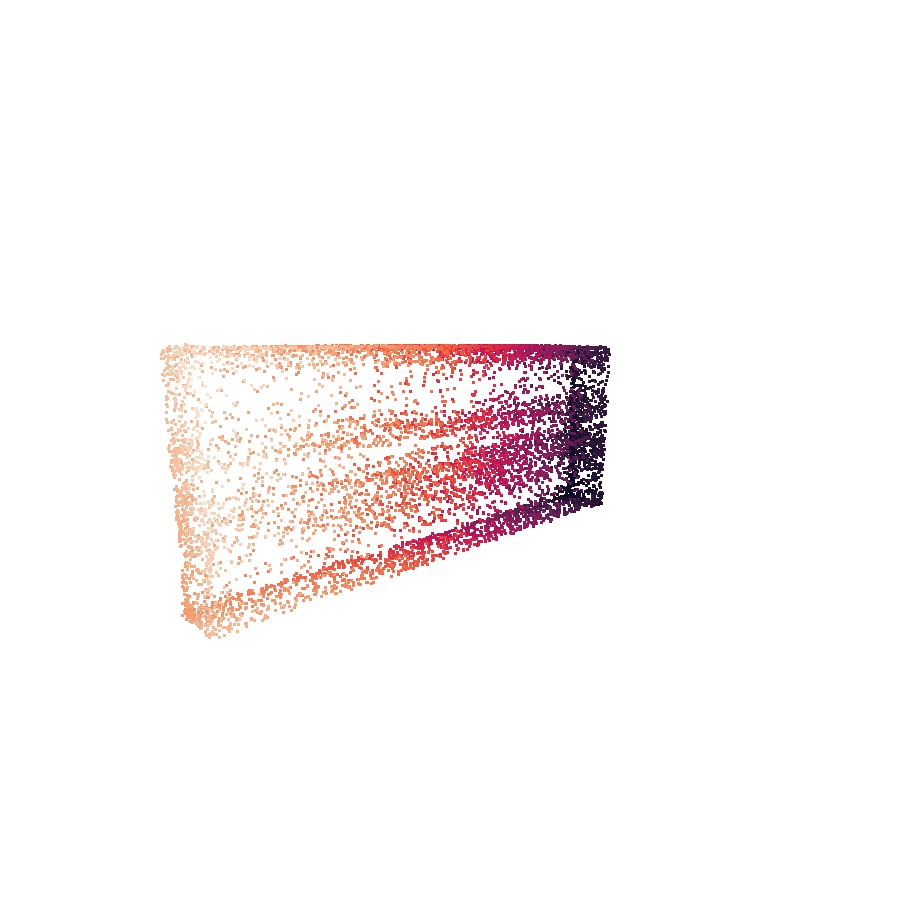}\end{subfigure} & \begin{subfigure}{0.04\textwidth}\centering\includegraphics[trim=370 220 340 190,clip,width=\textwidth]{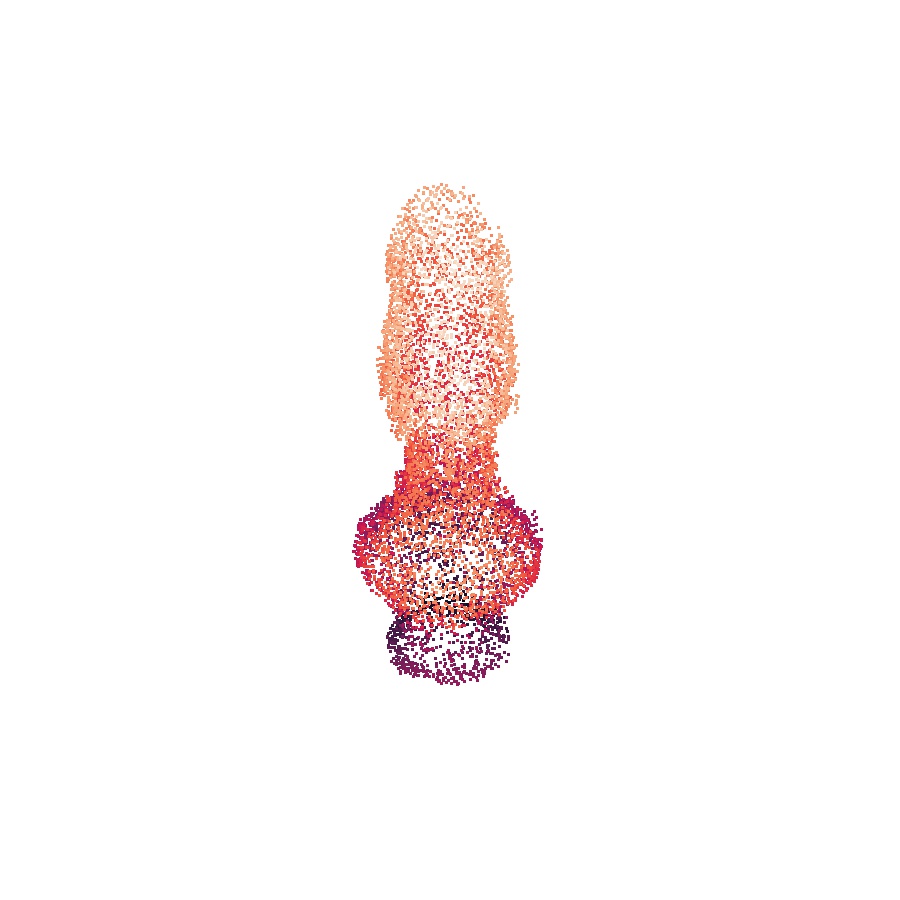}\end{subfigure} & \begin{subfigure}{0.065\textwidth}\centering\includegraphics[trim=260 260 320 160,clip,width=\textwidth]{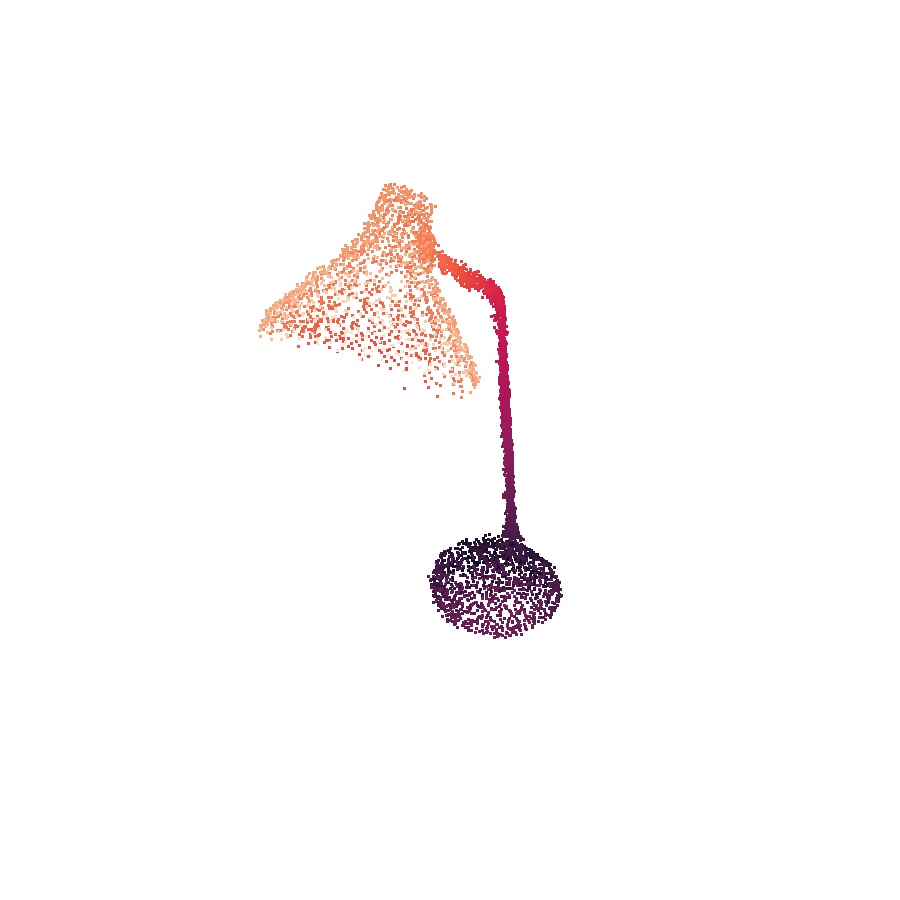}\end{subfigure} & \begin{subfigure}{0.13\textwidth}\centering\includegraphics[trim=100 200 150 190,clip,width=\textwidth]{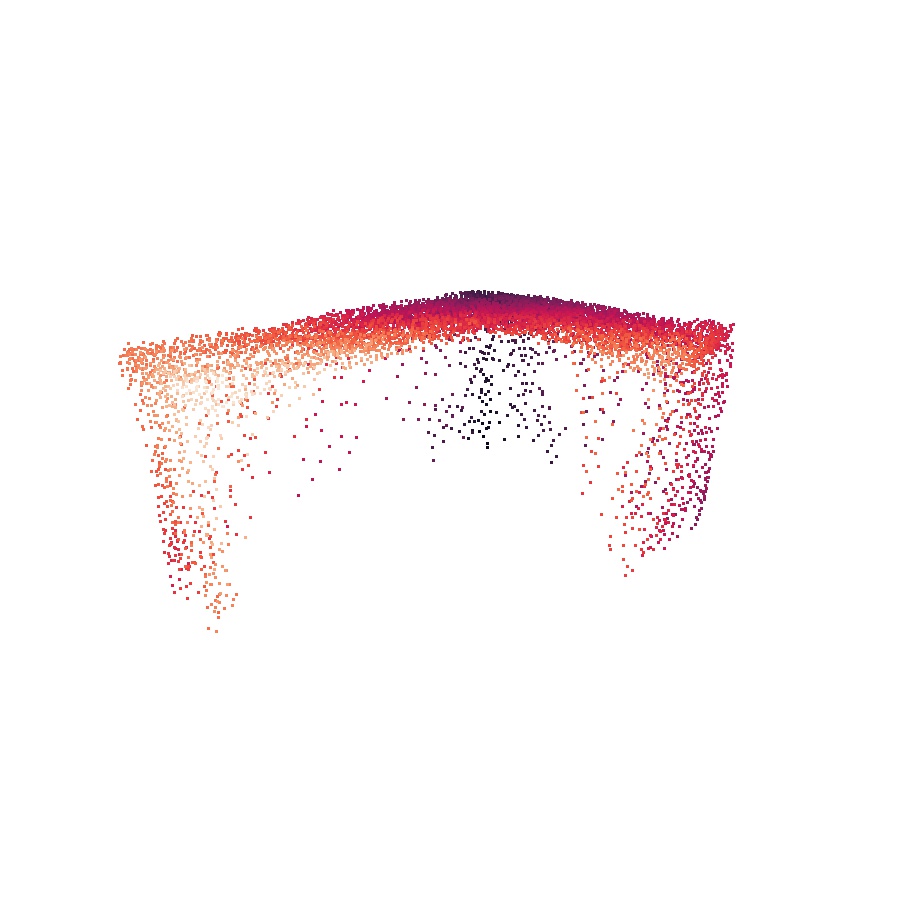}\end{subfigure} \\
SoftPool & \begin{subfigure}{0.065\textwidth}\centering\includegraphics[trim=200 150 350 150,clip,width=\textwidth]{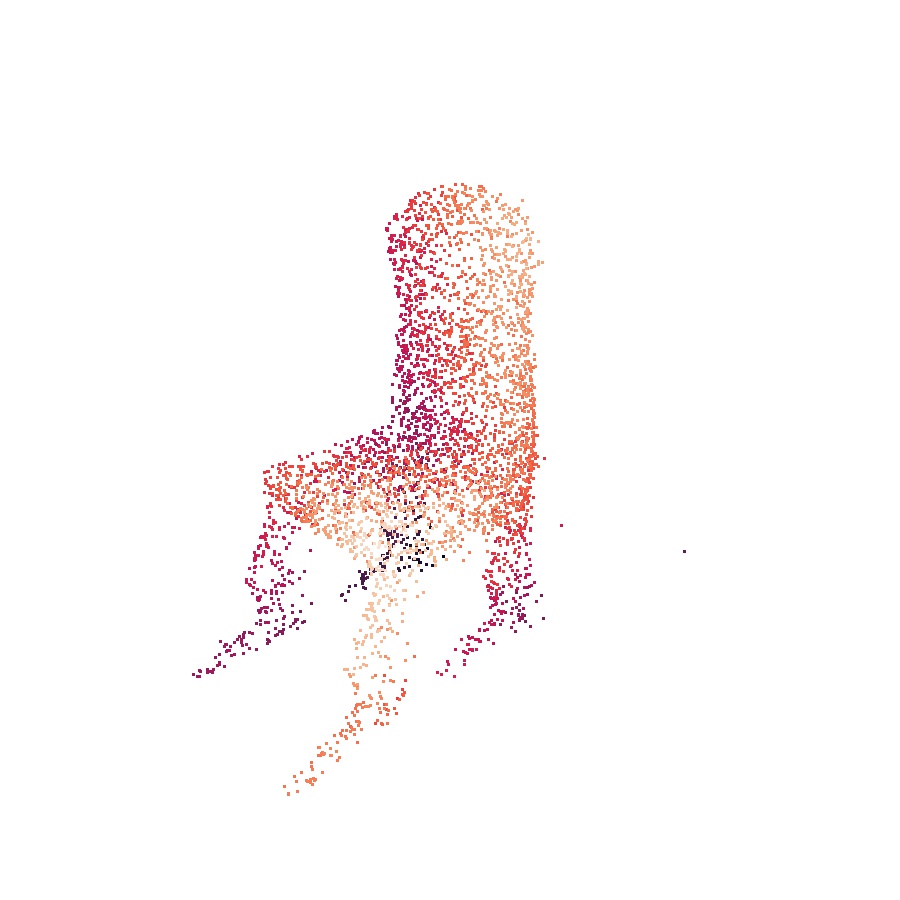}\end{subfigure} & \begin{subfigure}{0.03\textwidth}\centering\includegraphics[trim=330 200 460 190,clip,width=\textwidth]{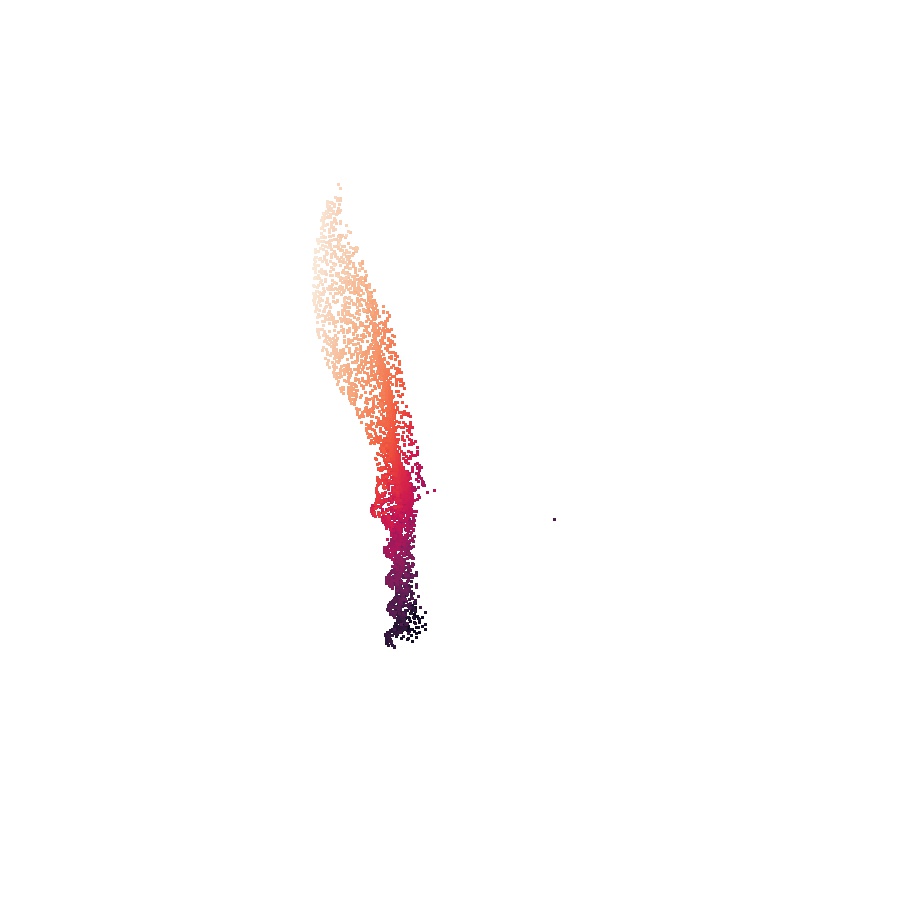}\end{subfigure} & \begin{subfigure}{0.09\textwidth}\centering\includegraphics[trim=200 180 290 100,clip,width=\textwidth]{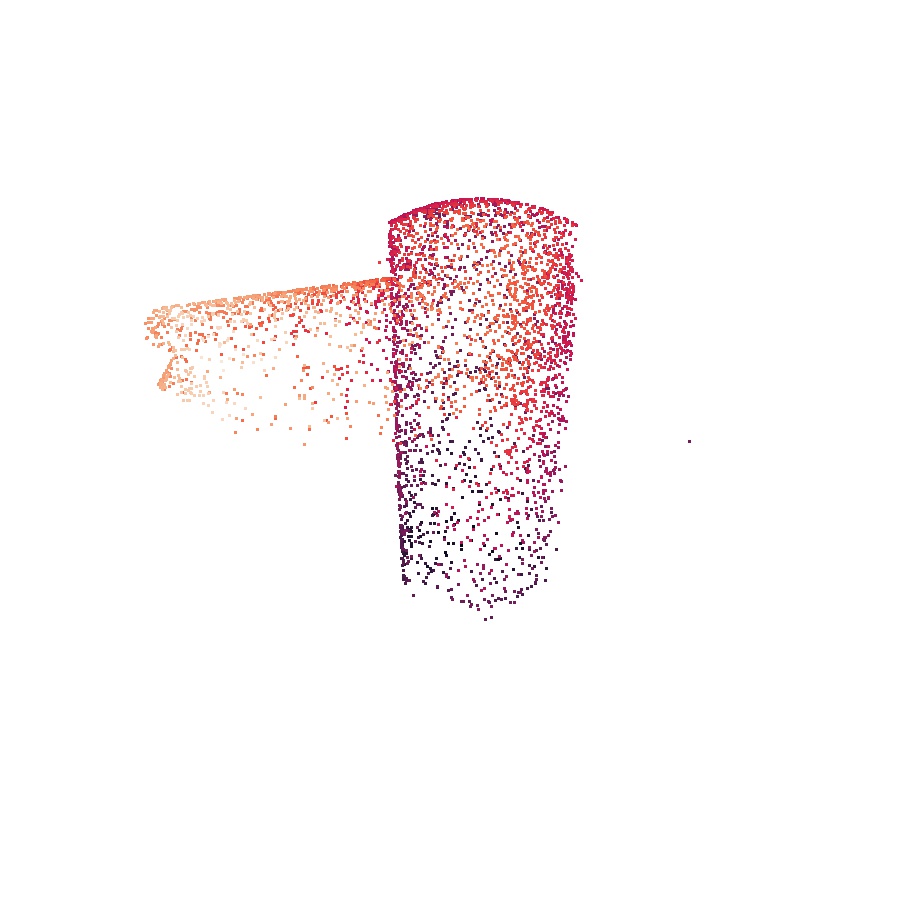}\end{subfigure} & \begin{subfigure}{0.1\textwidth}\centering\includegraphics[trim=100 170 120 110,clip,width=\textwidth]{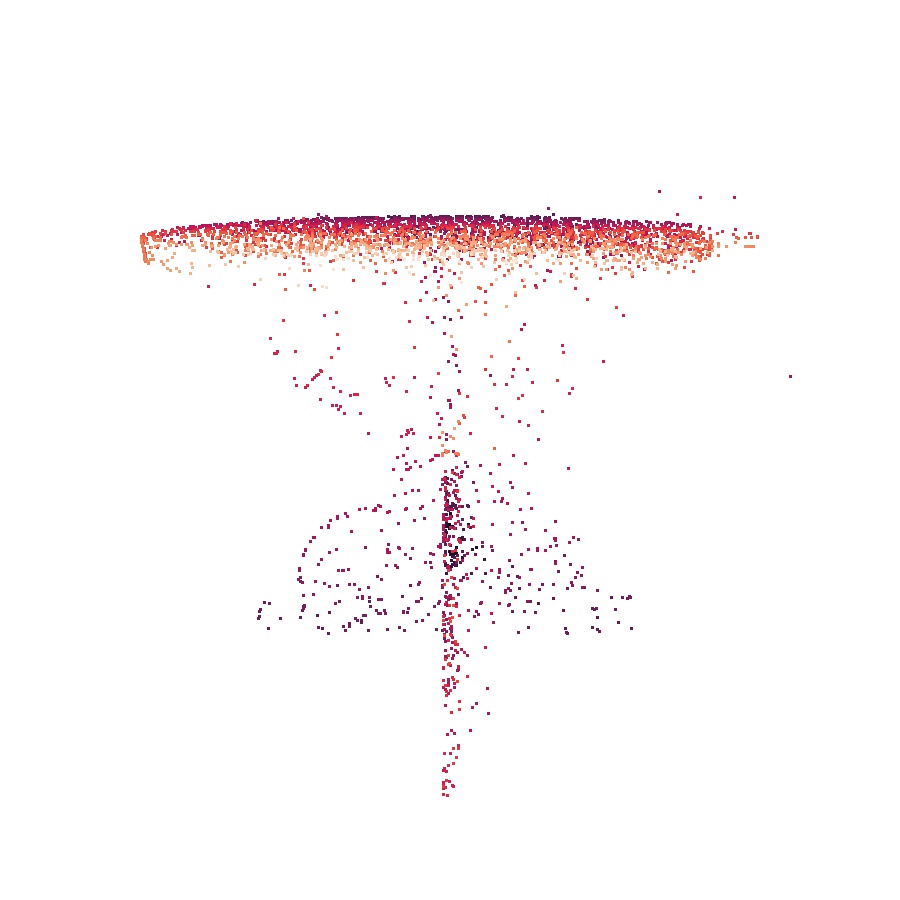}\end{subfigure} & \begin{subfigure}{0.065\textwidth}\centering\includegraphics[trim=220 120 250 100,clip,width=\textwidth]{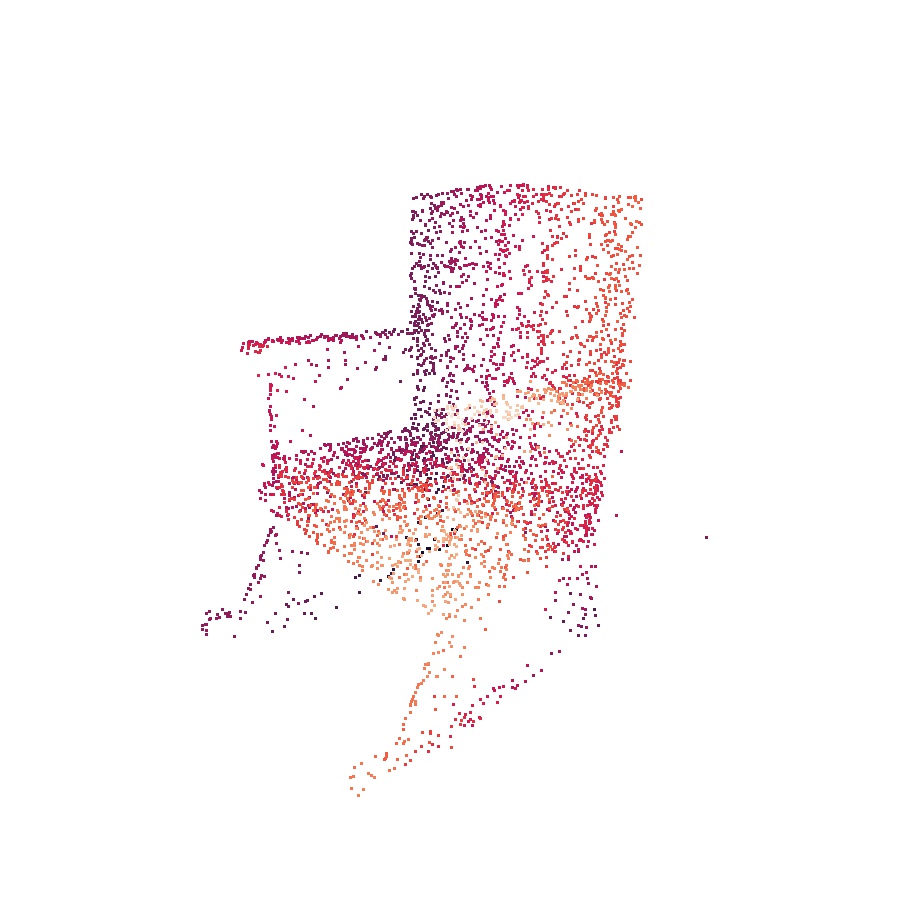}\end{subfigure} & \begin{subfigure}{0.14\textwidth}\centering\includegraphics[trim=120 250 250 190,clip,width=\textwidth]{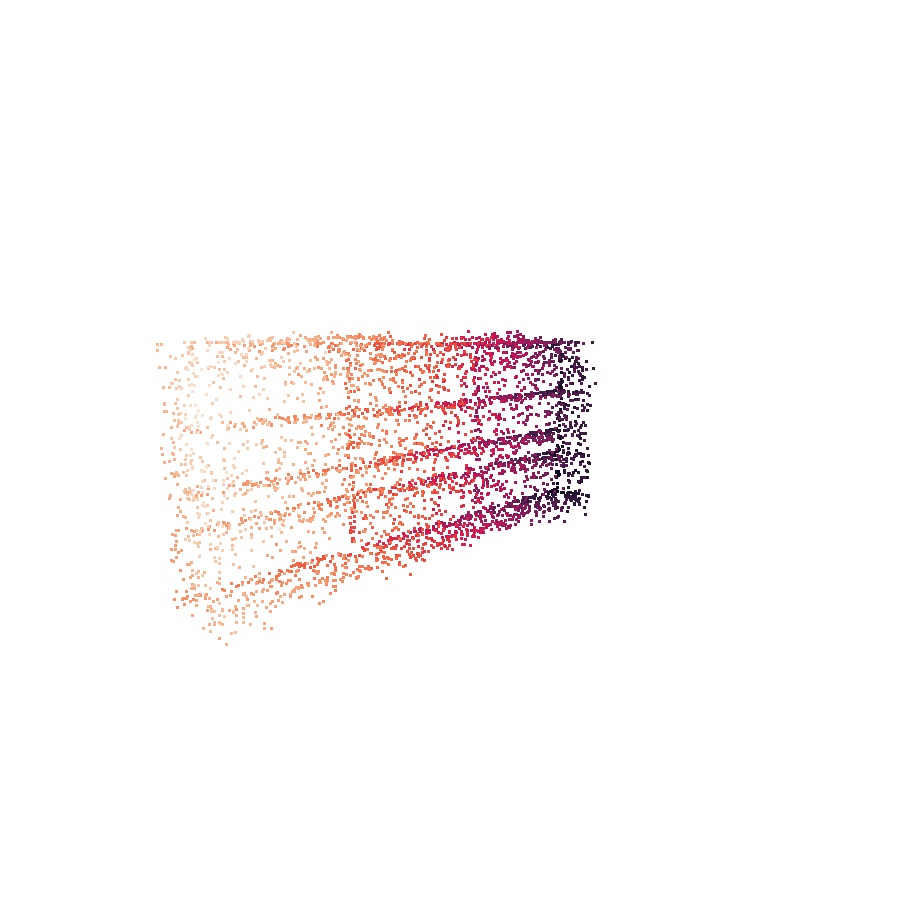}\end{subfigure} & \begin{subfigure}{0.04\textwidth}\centering\includegraphics[trim=330 220 370 190,clip,width=\textwidth]{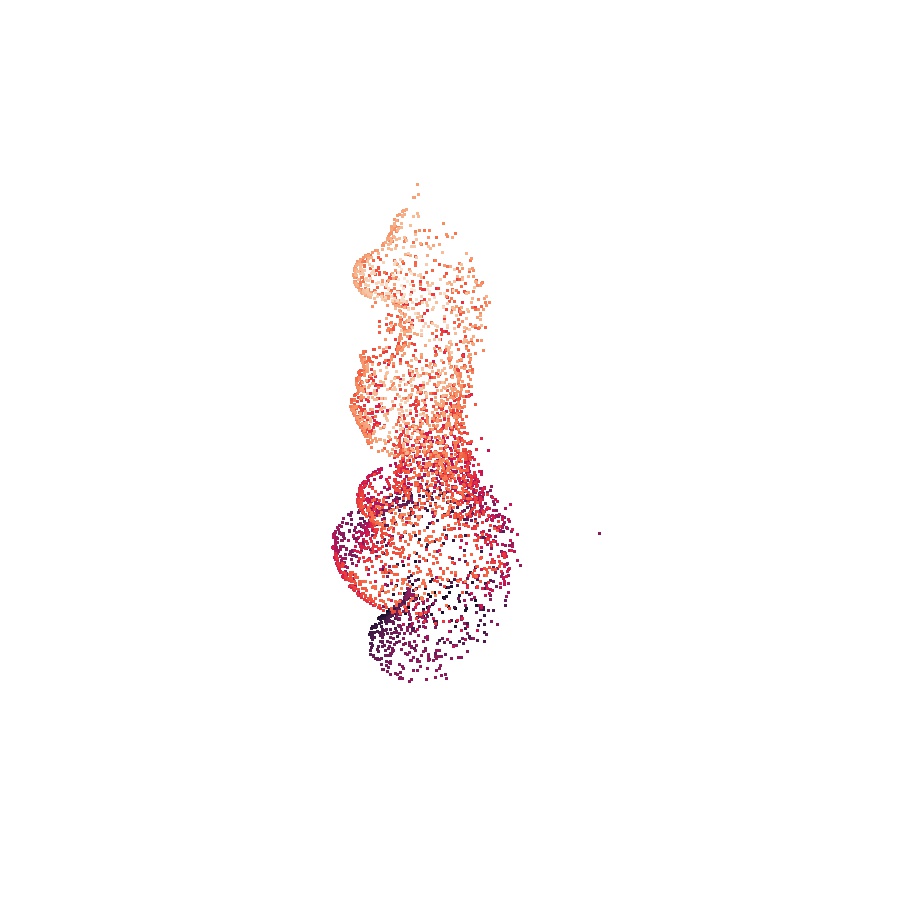}\end{subfigure} & \begin{subfigure}{0.065\textwidth}\centering\includegraphics[trim=180 260 370 160,clip,width=\textwidth]{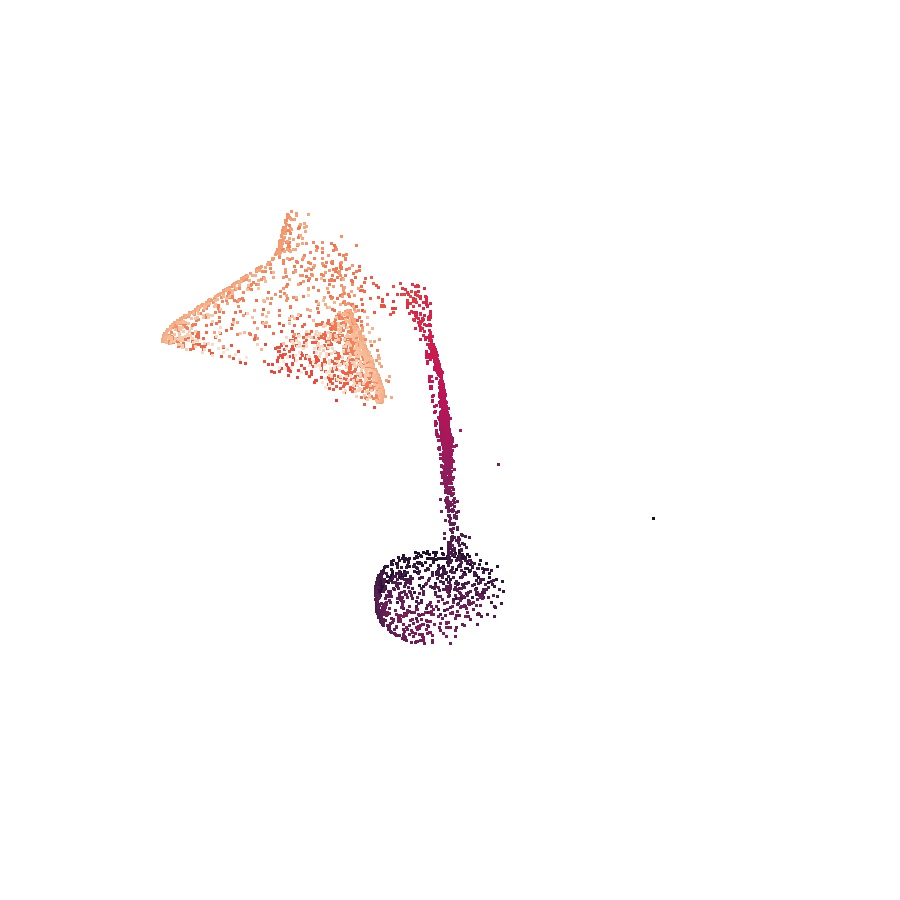}\end{subfigure} & \begin{subfigure}{0.13\textwidth}\centering\includegraphics[trim=80 200 240 180,clip,width=\textwidth]{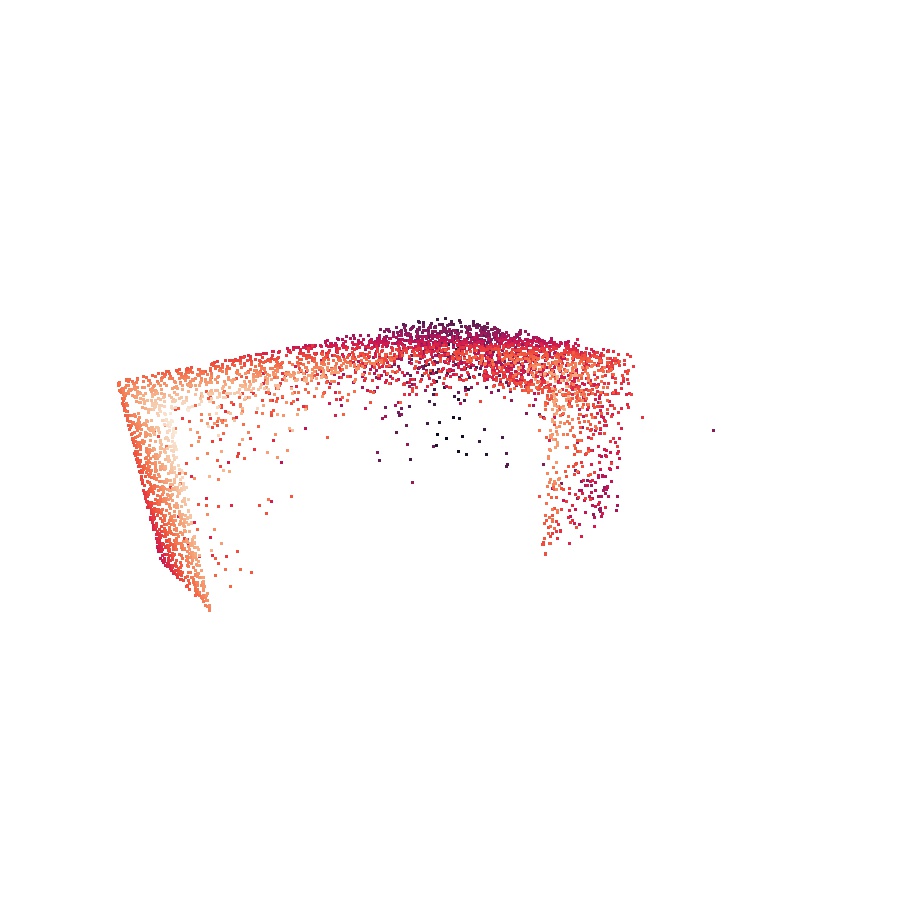}\end{subfigure} \\
MSN & \begin{subfigure}{0.065\textwidth}\centering\includegraphics[trim=300 150 250 150,clip,width=\textwidth]{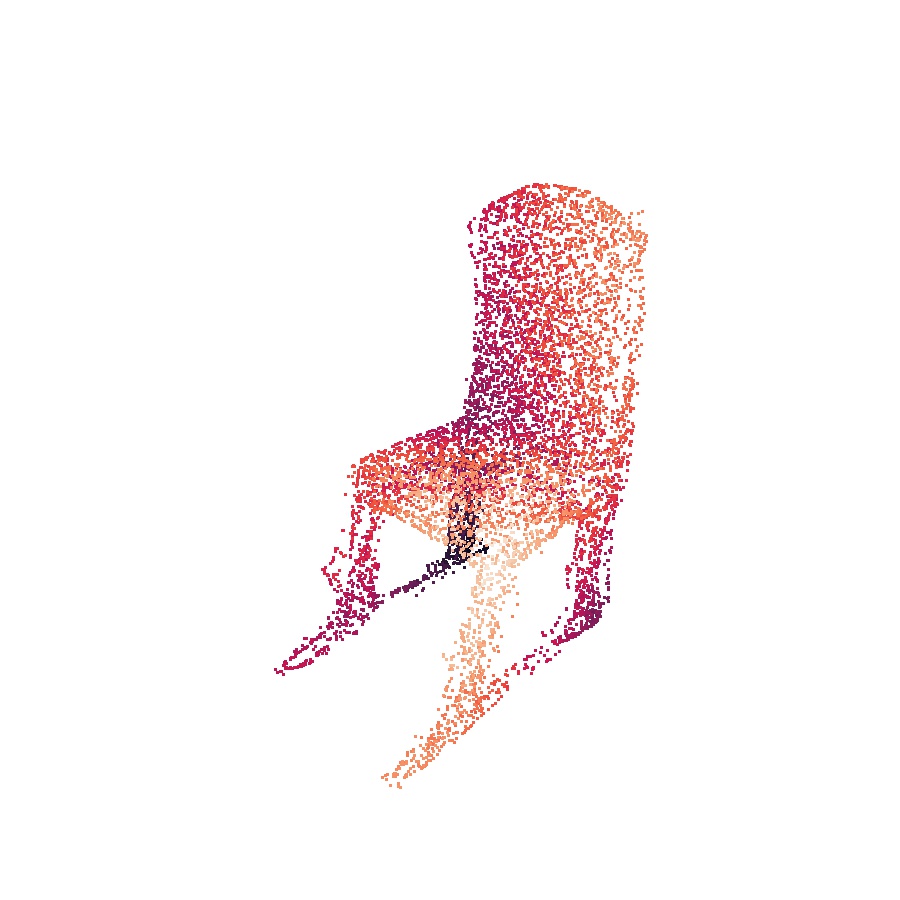}\end{subfigure} & \begin{subfigure}{0.03\textwidth}\centering\includegraphics[trim=380 230 400 190,clip,width=\textwidth]{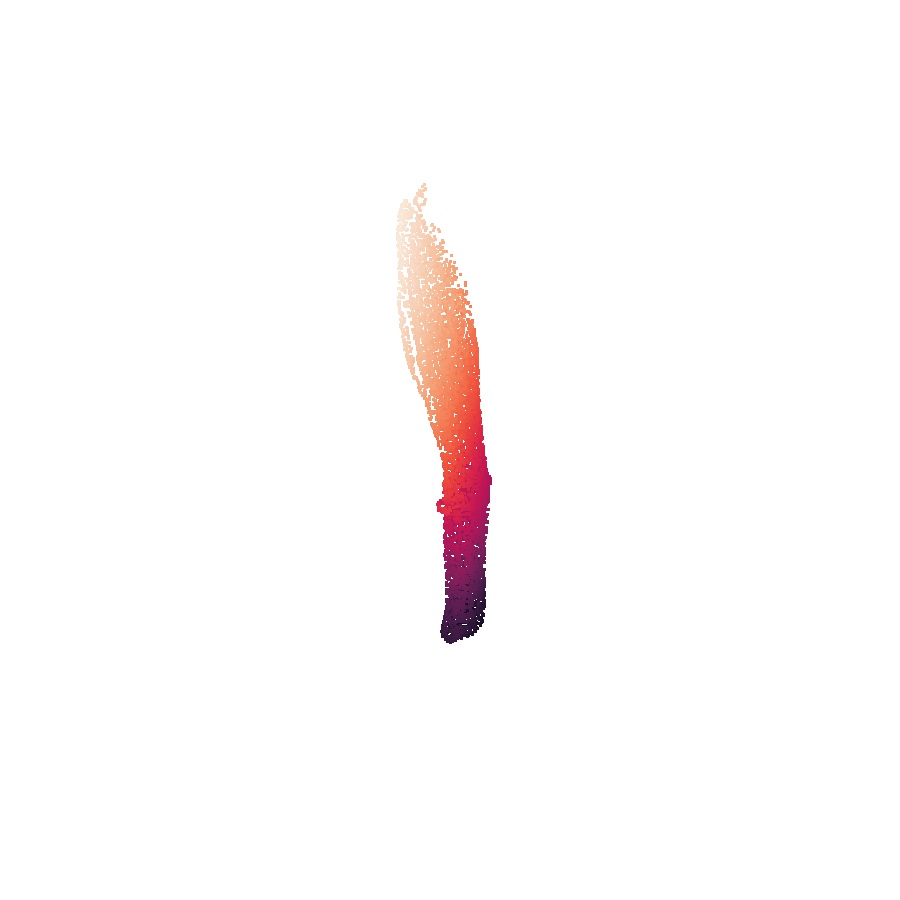}\end{subfigure} & \begin{subfigure}{0.09\textwidth}\centering\includegraphics[trim=200 170 220 100,clip,width=\textwidth]{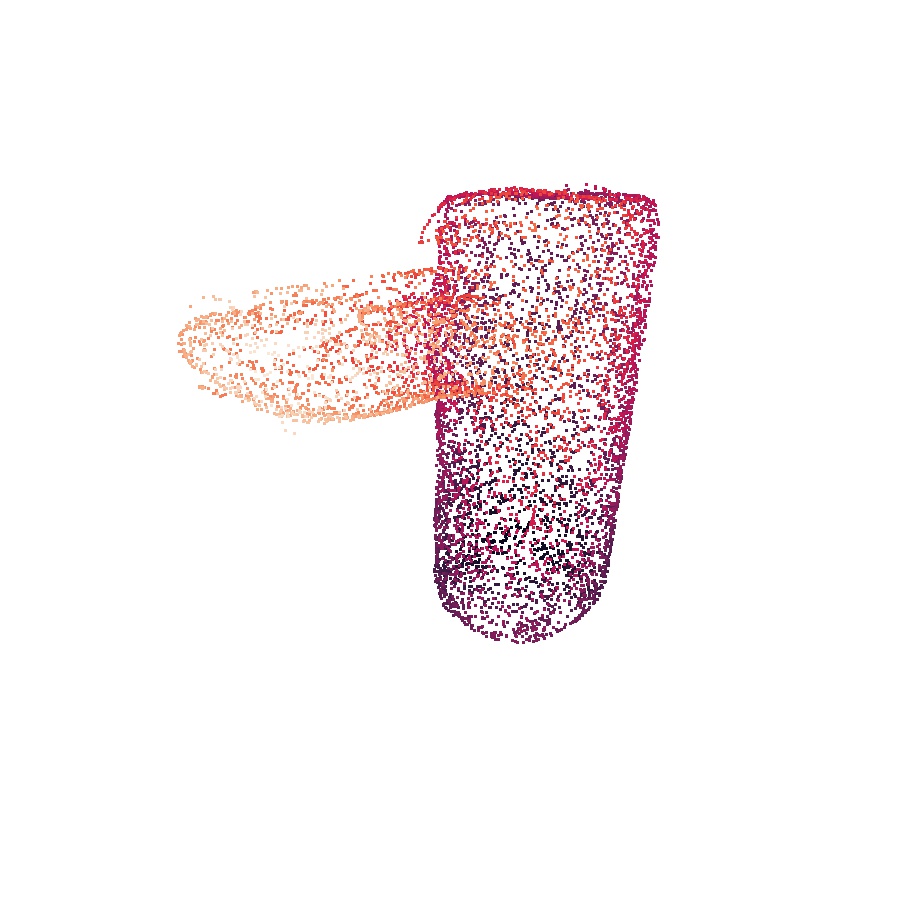}\end{subfigure} & \begin{subfigure}{0.1\textwidth}\centering\includegraphics[trim=100 170 120 110,clip,width=\textwidth]{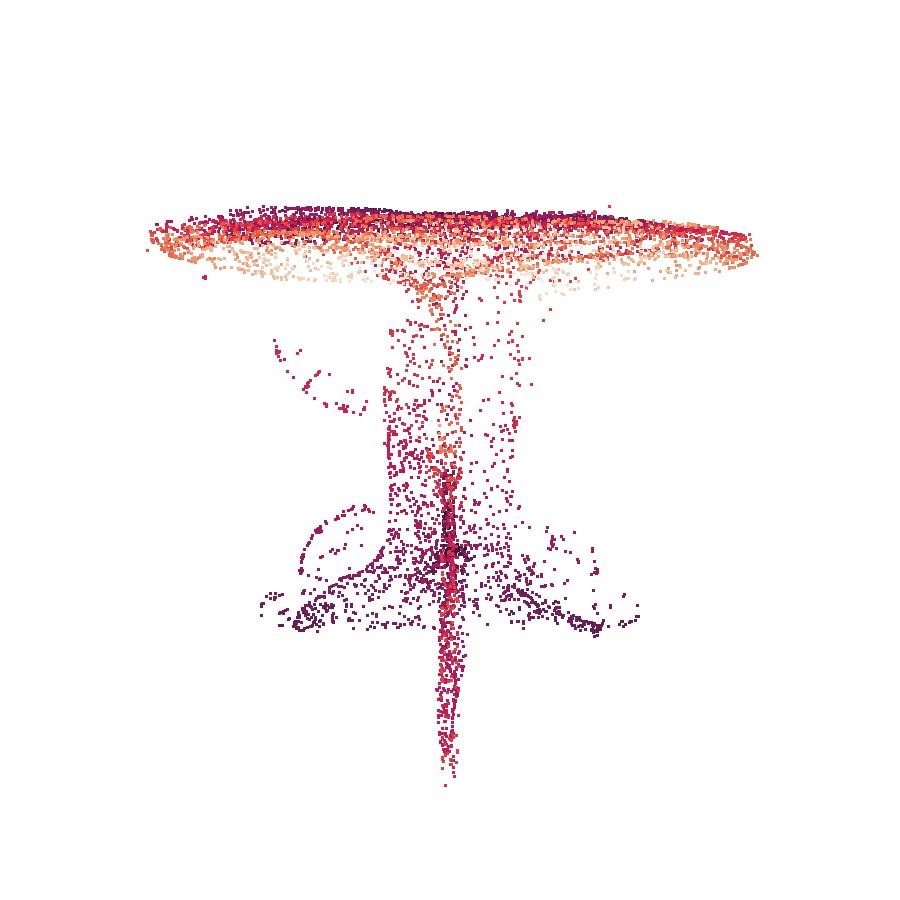}\end{subfigure} & \begin{subfigure}{0.065\textwidth}\centering\includegraphics[trim=250 120 220 100,clip,width=\textwidth]{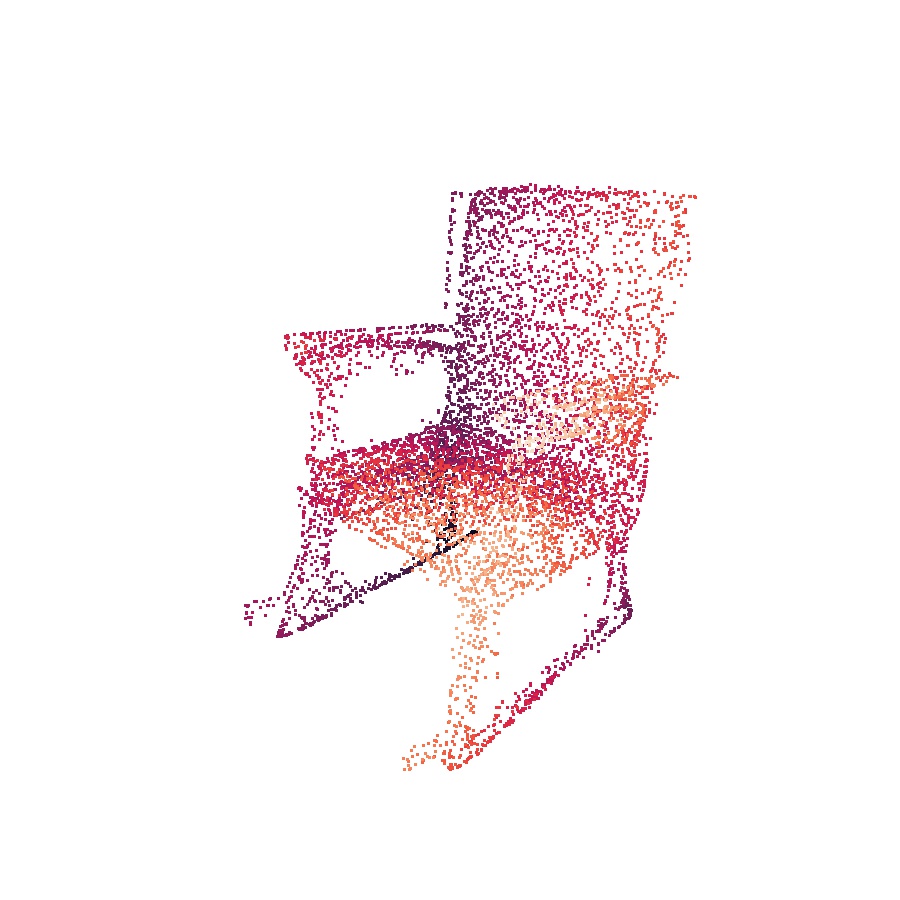}\end{subfigure} & \begin{subfigure}{0.14\textwidth}\centering\includegraphics[trim=130 250 230 190,clip,width=\textwidth]{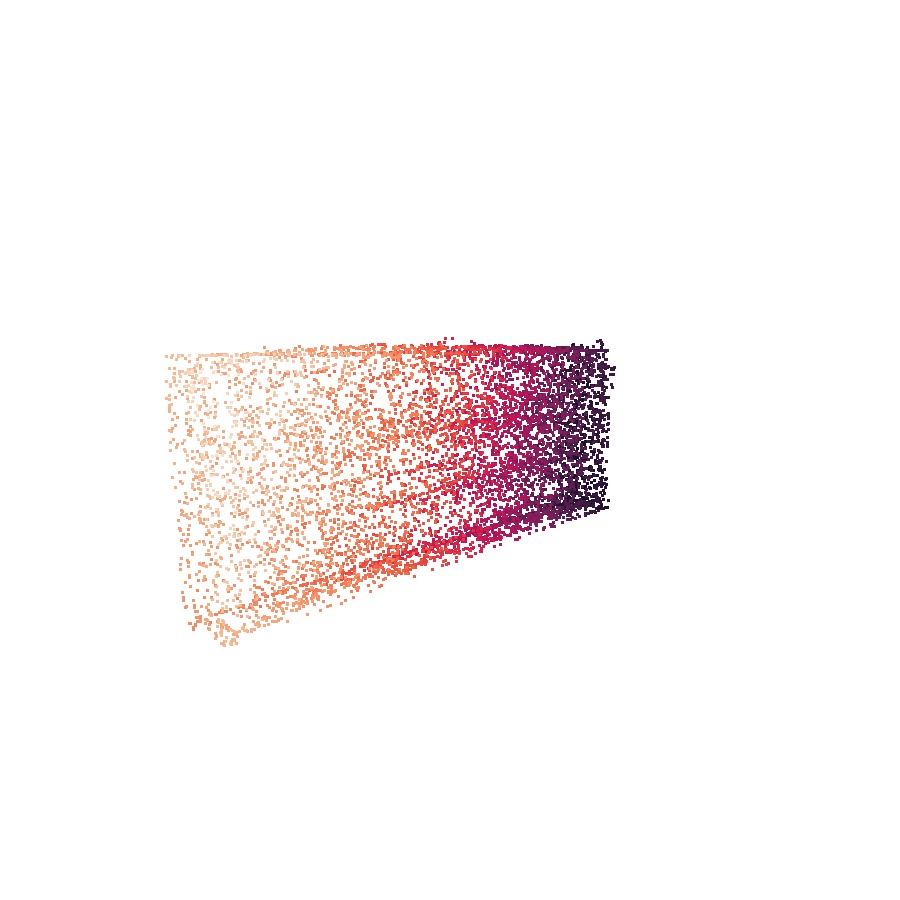}\end{subfigure} & \begin{subfigure}{0.04\textwidth}\centering\includegraphics[trim=370 220 340 190,clip,width=\textwidth]{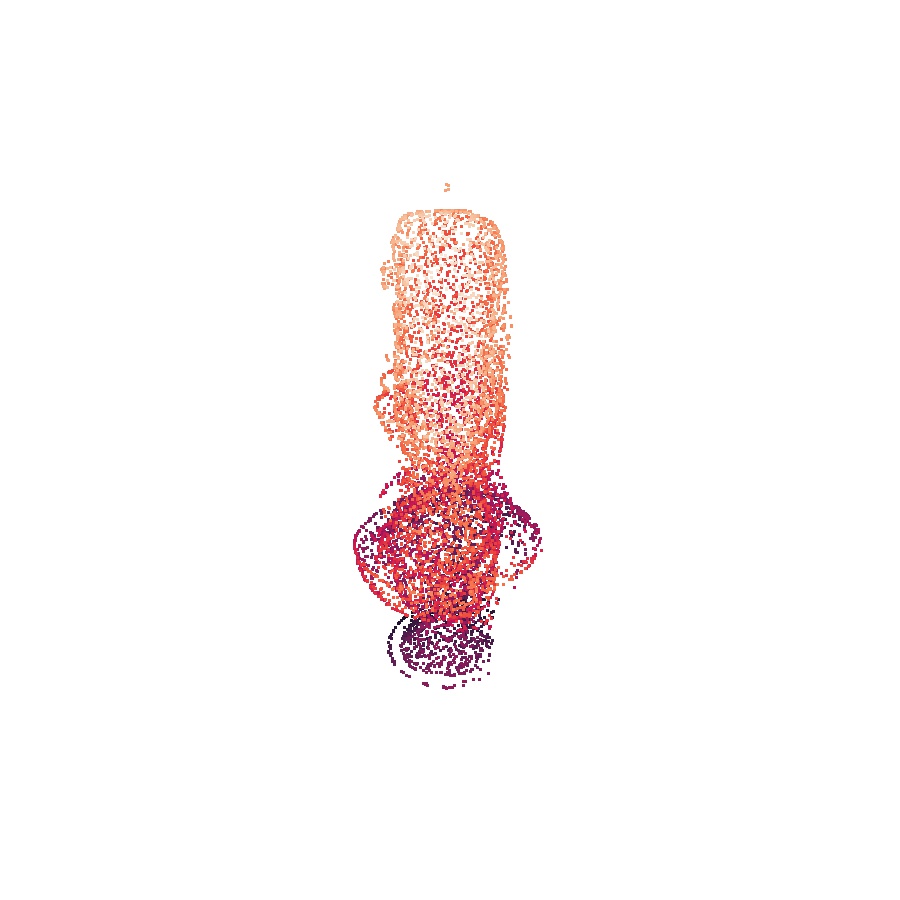}\end{subfigure} & \begin{subfigure}{0.065\textwidth}\centering\includegraphics[trim=260 260 320 130,clip,width=\textwidth]{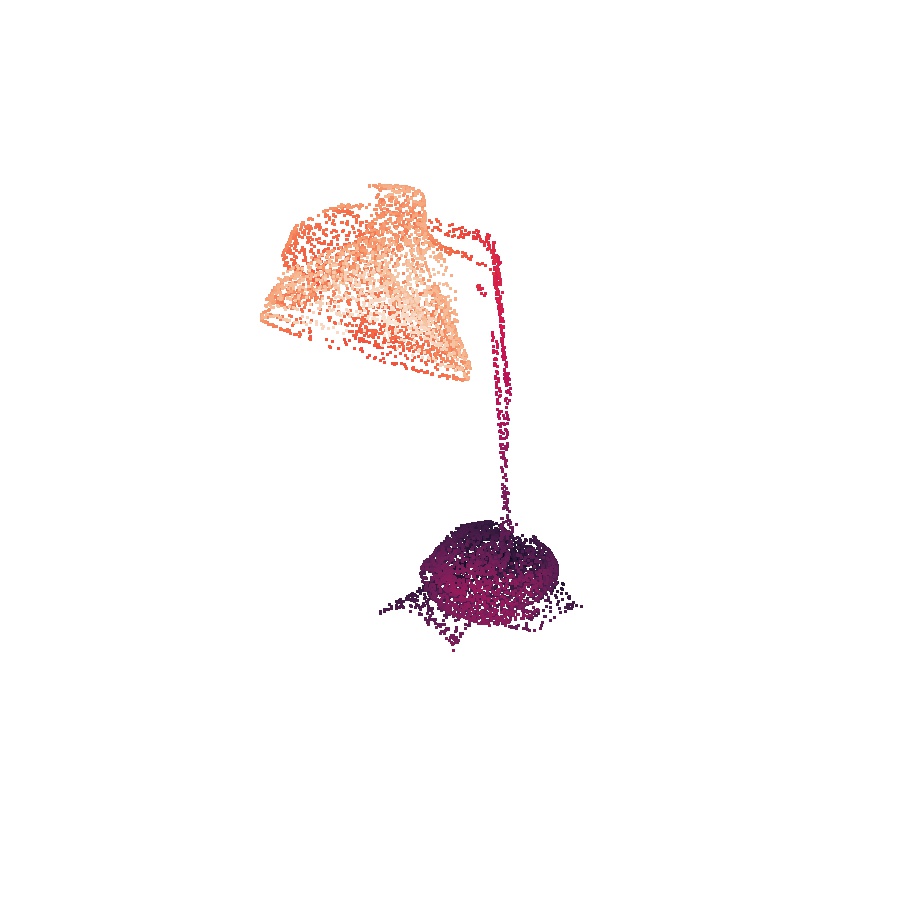}\end{subfigure} & \begin{subfigure}{0.13\textwidth}\centering\includegraphics[trim=100 200 150 180,clip,width=\textwidth]{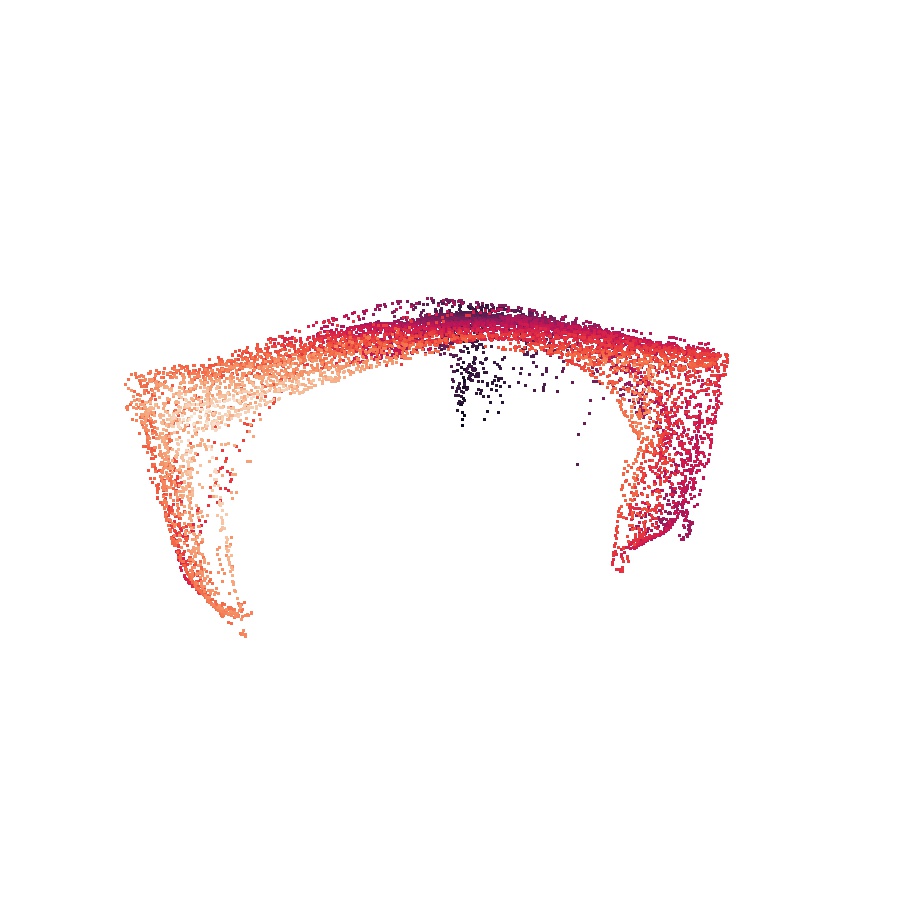}\end{subfigure} \\
CRN & \begin{subfigure}{0.065\textwidth}\centering\includegraphics[trim=300 150 250 150,clip,width=\textwidth]{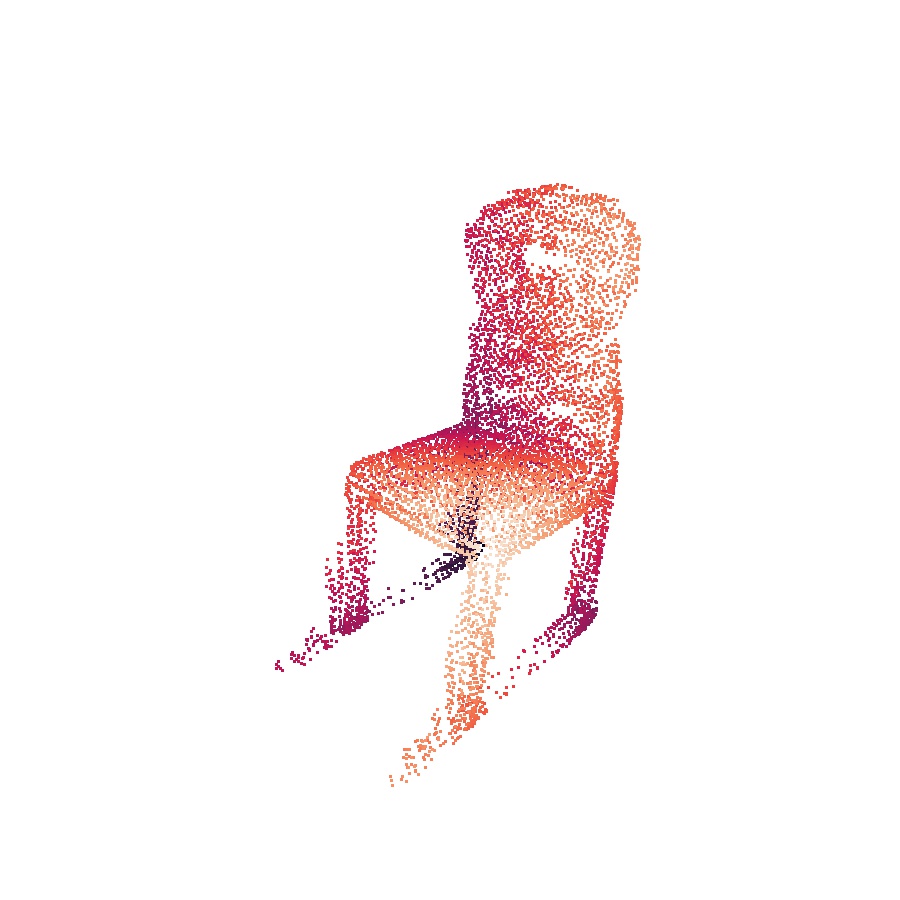}\end{subfigure} & \begin{subfigure}{0.03\textwidth}\centering\includegraphics[trim=380 230 400 190,clip,width=\textwidth]{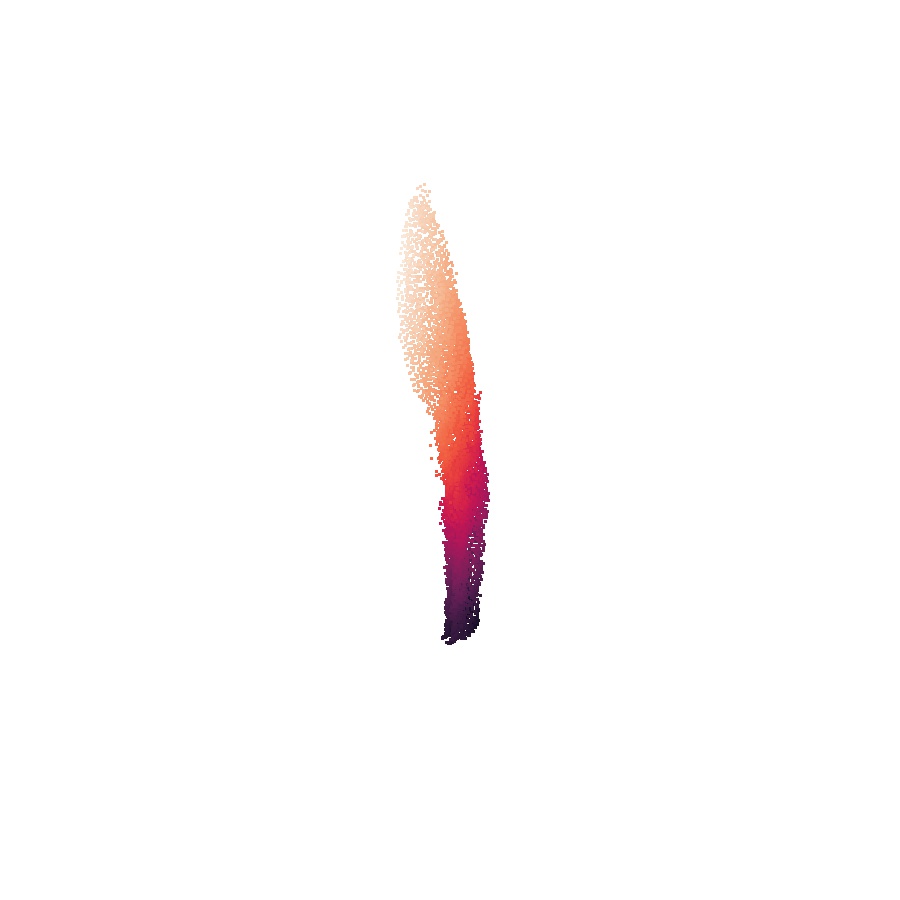}\end{subfigure} & \begin{subfigure}{0.09\textwidth}\centering\includegraphics[trim=200 170 220 100,clip,width=\textwidth]{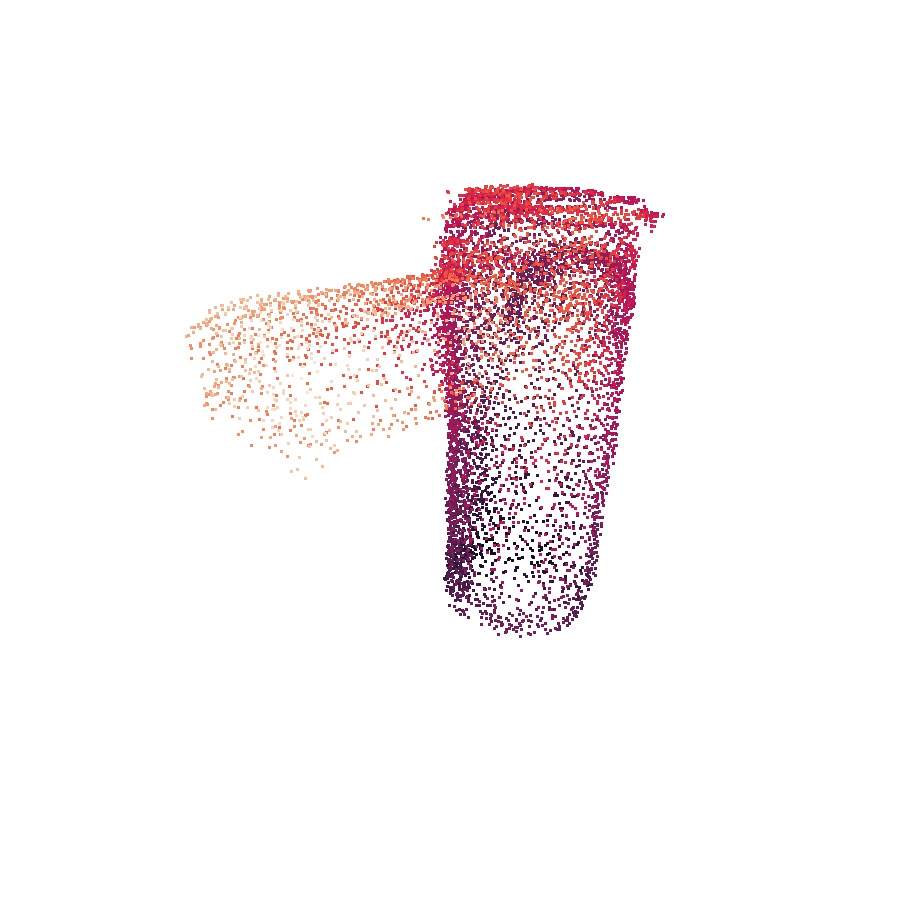}\end{subfigure} & \begin{subfigure}{0.1\textwidth}\centering\includegraphics[trim=100 170 120 110,clip,width=\textwidth]{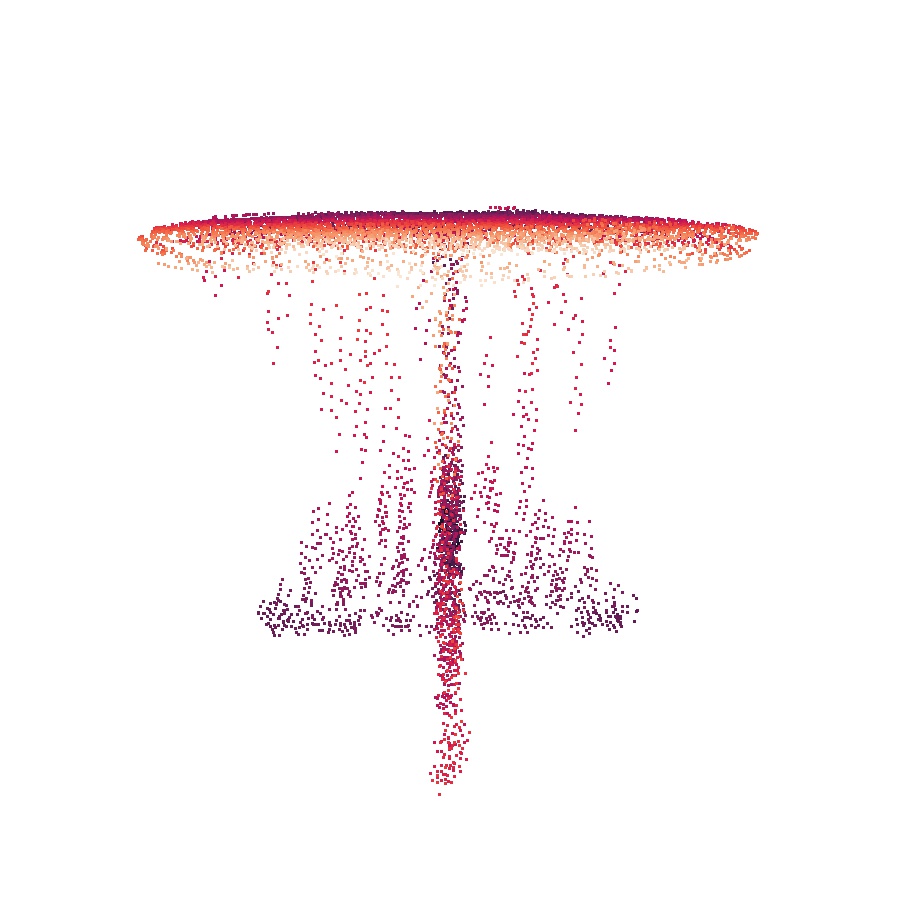}\end{subfigure} & \begin{subfigure}{0.065\textwidth}\centering\includegraphics[trim=250 120 230 100,clip,width=\textwidth]{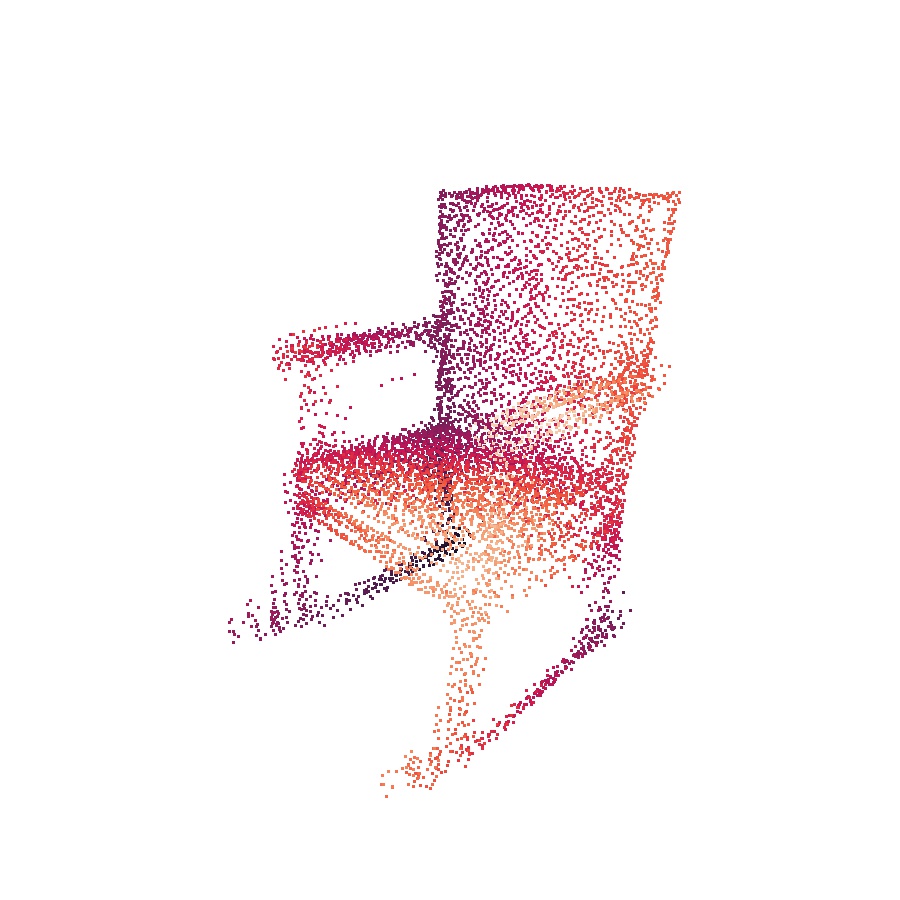}\end{subfigure} & \begin{subfigure}{0.14\textwidth}\centering\includegraphics[trim=120 250 230 190,clip,width=\textwidth]{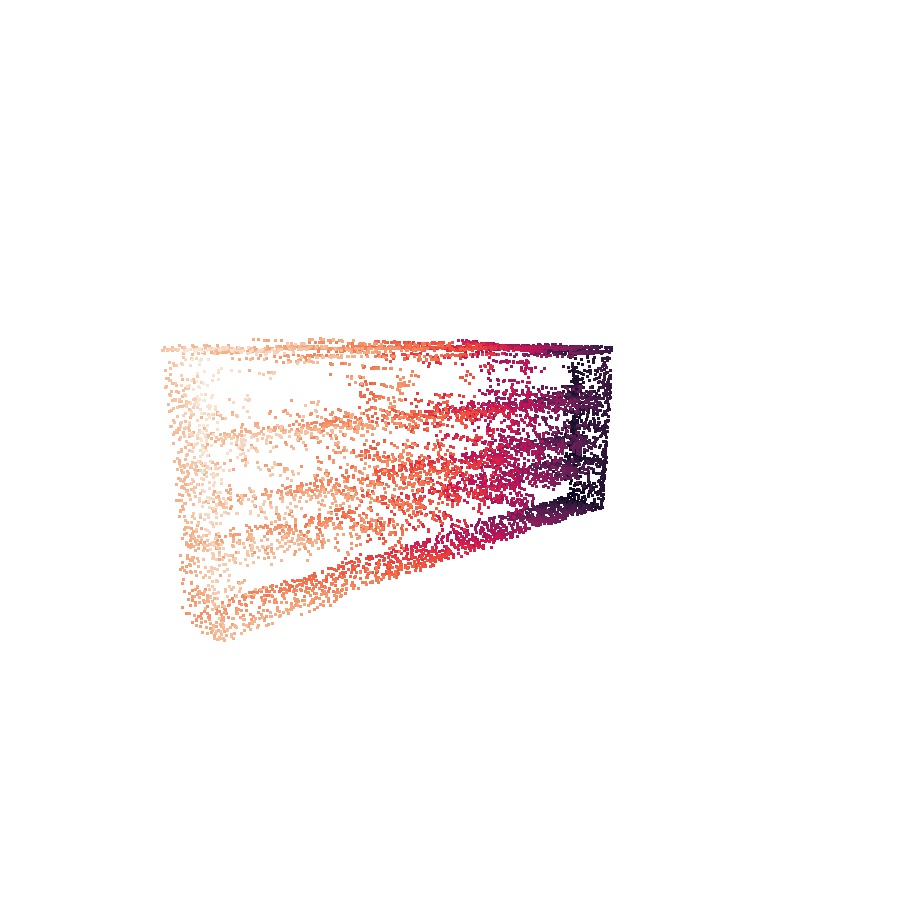}\end{subfigure} & \begin{subfigure}{0.04\textwidth}\centering\includegraphics[trim=370 220 340 190,clip,width=\textwidth]{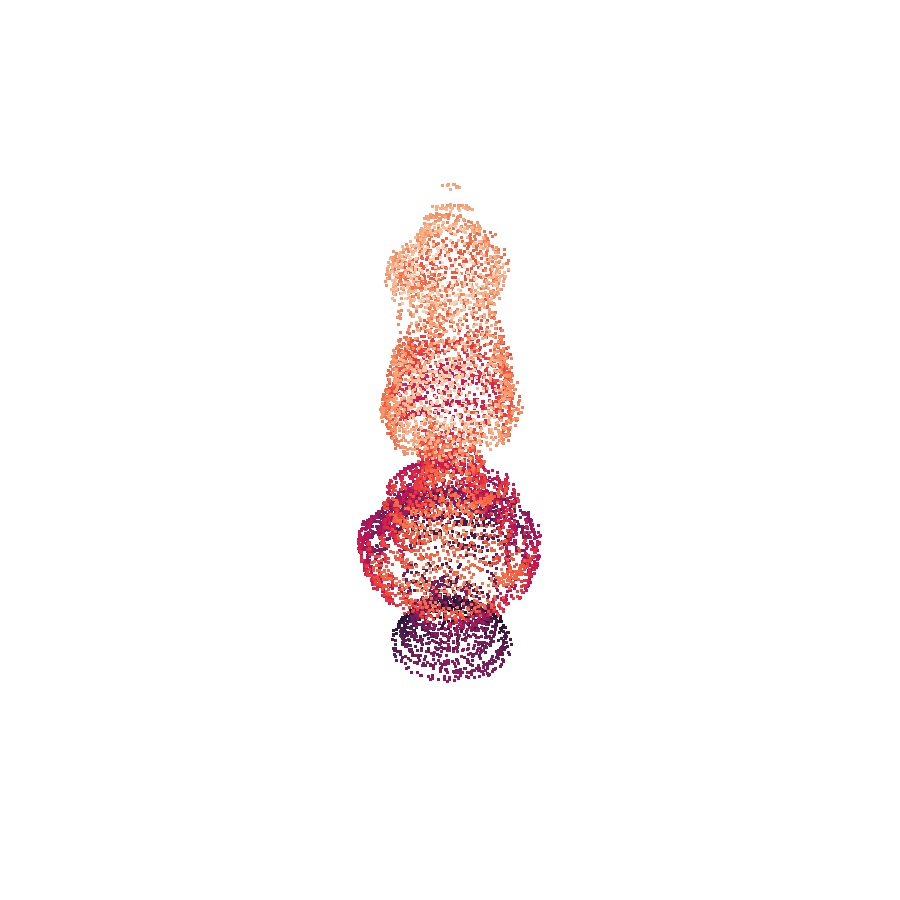}\end{subfigure} & \begin{subfigure}{0.065\textwidth}\centering\includegraphics[trim=260 260 320 160,clip,width=\textwidth]{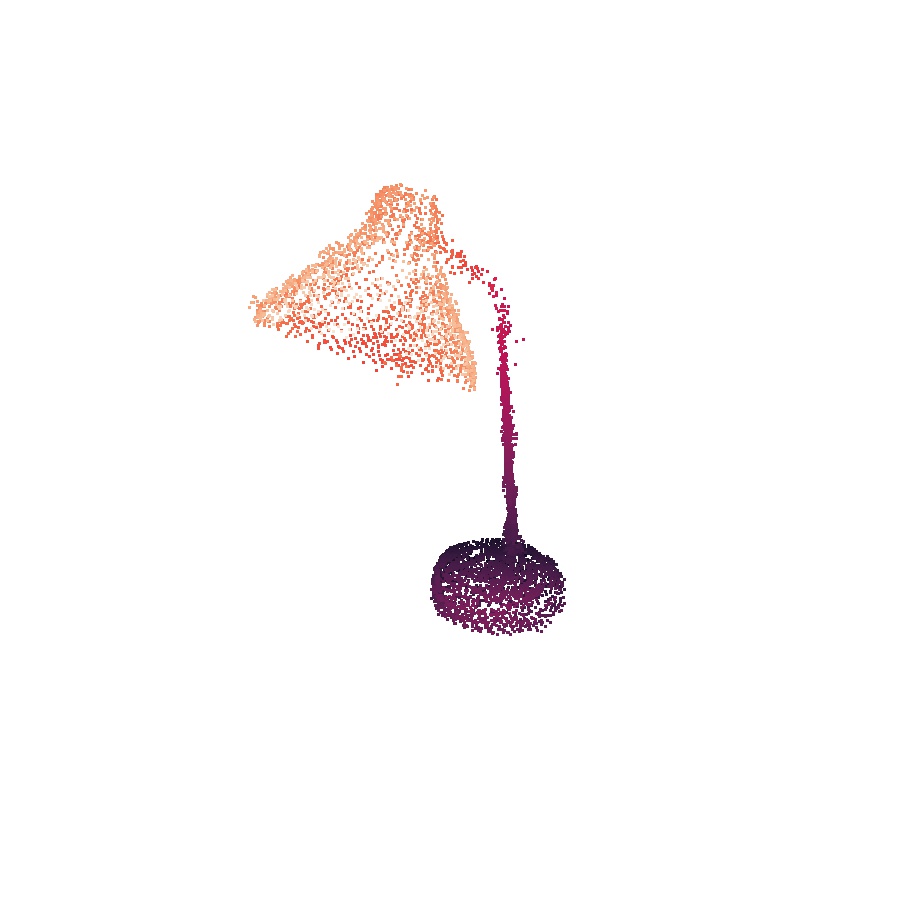}\end{subfigure} & \begin{subfigure}{0.13\textwidth}\centering\includegraphics[trim=100 200 150 180,clip,width=\textwidth]{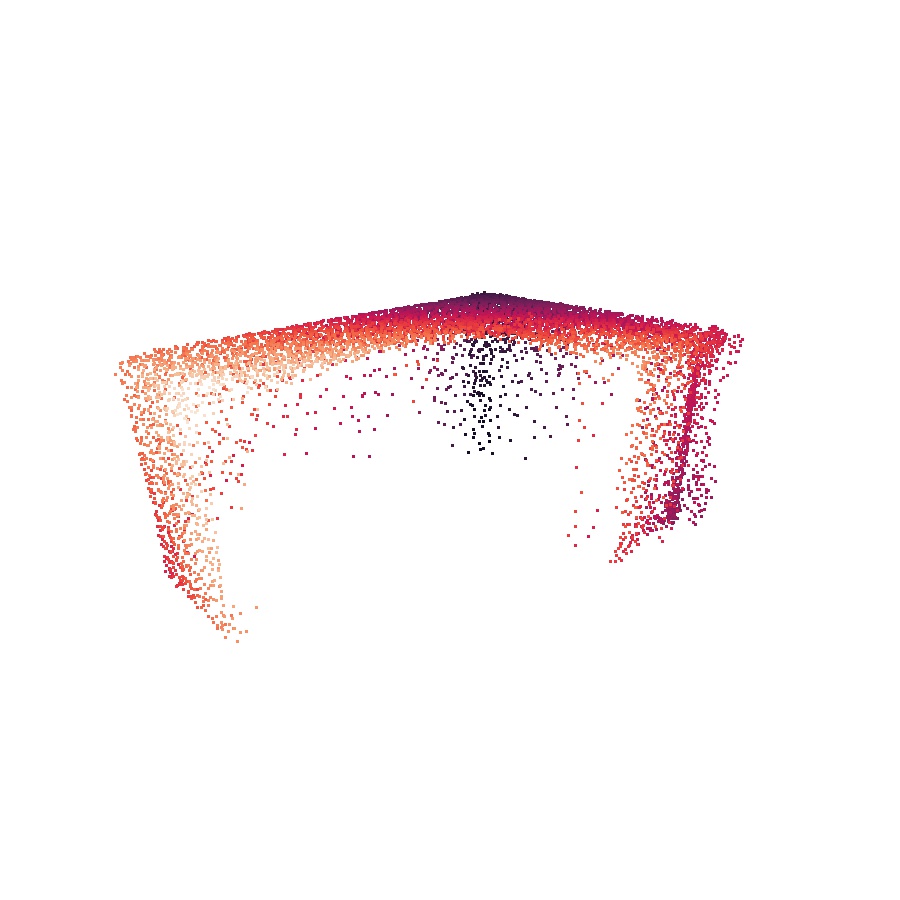}\end{subfigure} \\
GRNet & \begin{subfigure}{0.065\textwidth}\centering\includegraphics[trim=300 150 250 150,clip,width=\textwidth]{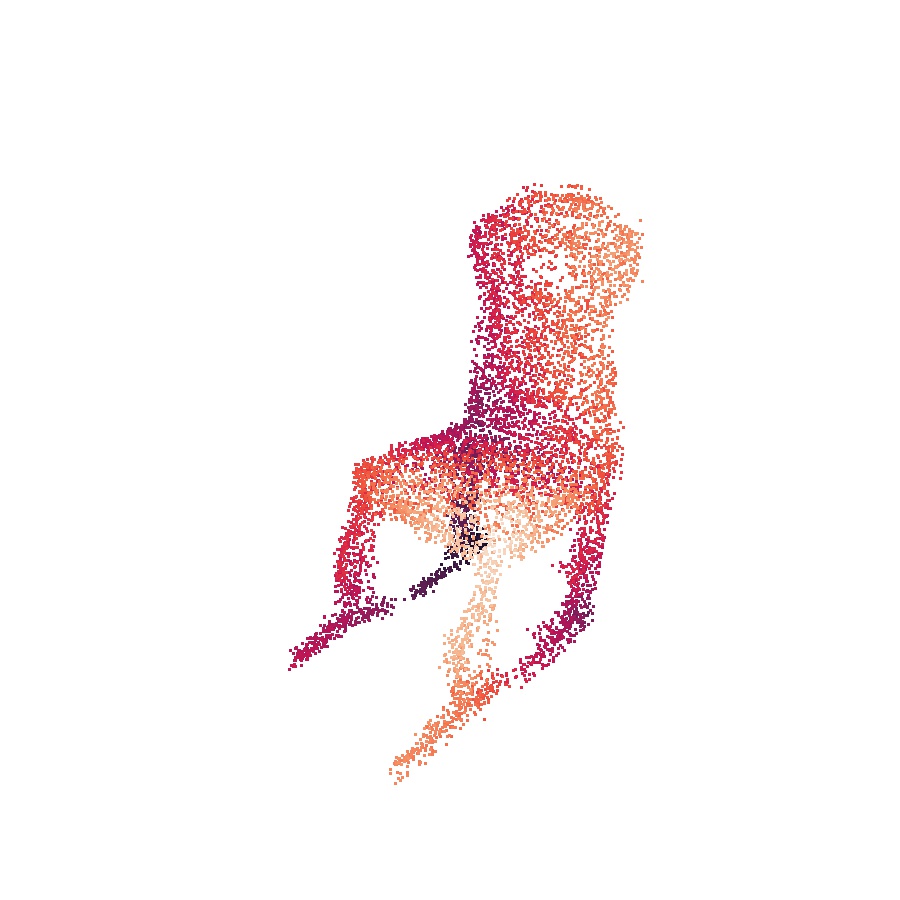}\end{subfigure} & \begin{subfigure}{0.032\textwidth}\centering\includegraphics[trim=380 230 380 190,clip,width=\textwidth]{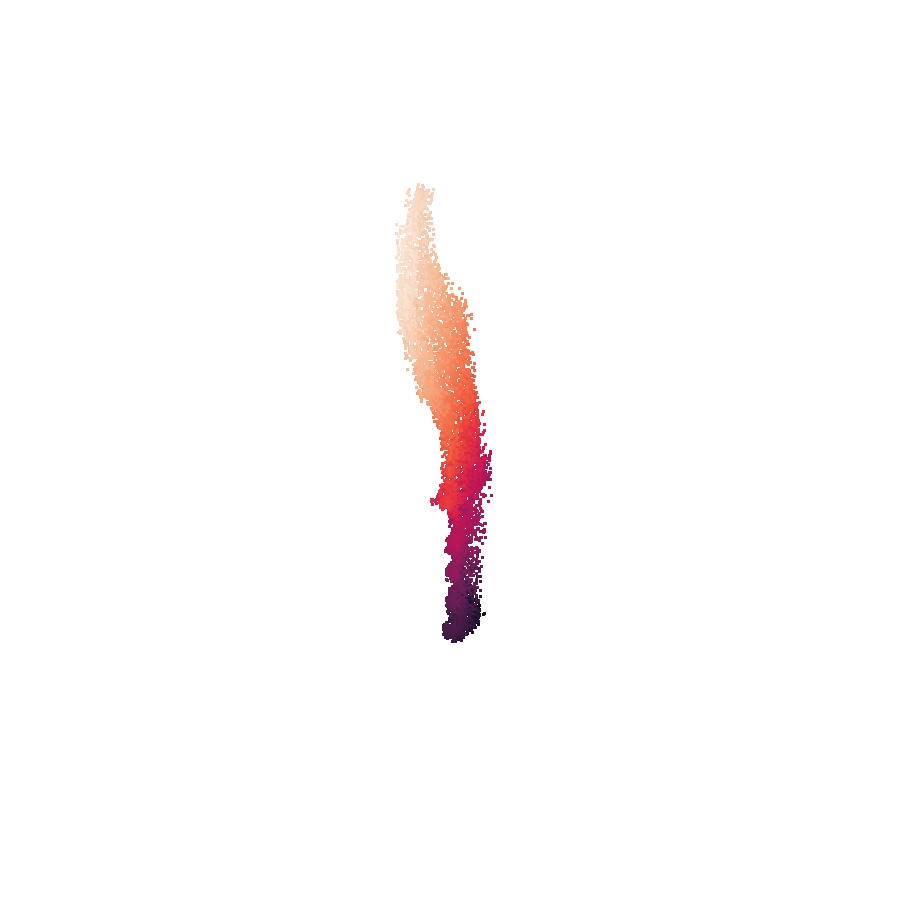}\end{subfigure} & \begin{subfigure}{0.09\textwidth}\centering\includegraphics[trim=200 170 220 100,clip,width=\textwidth]{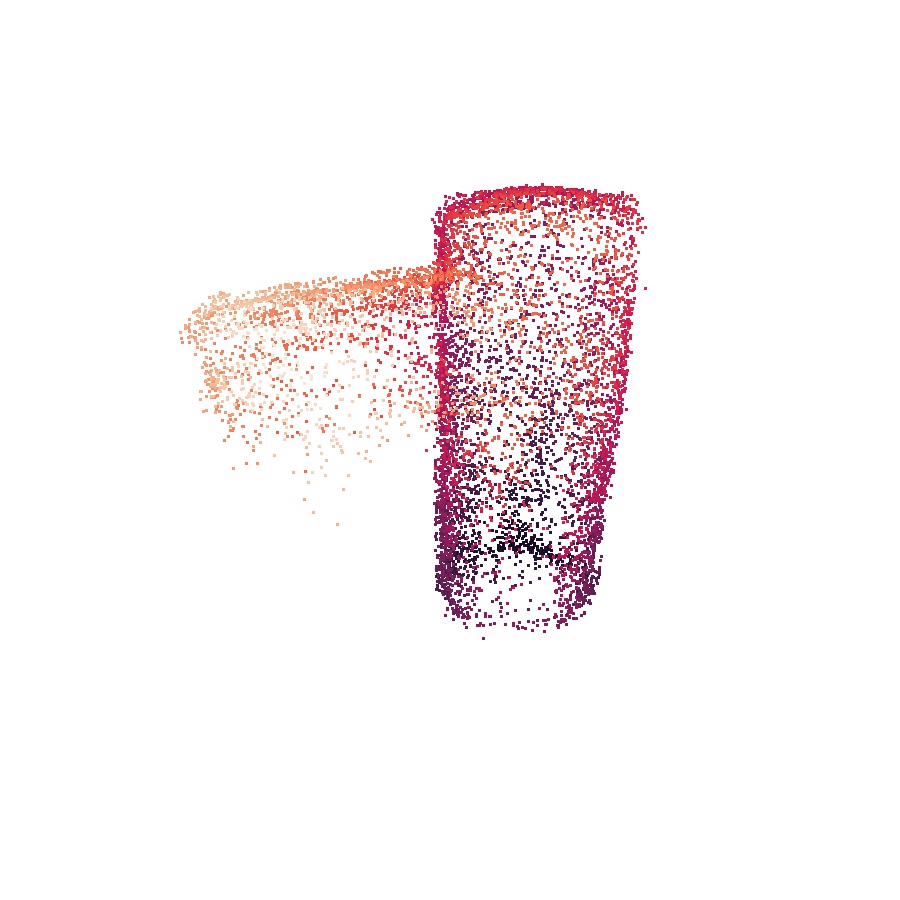}\end{subfigure} & \begin{subfigure}{0.1\textwidth}\centering\includegraphics[trim=100 170 120 110,clip,width=\textwidth]{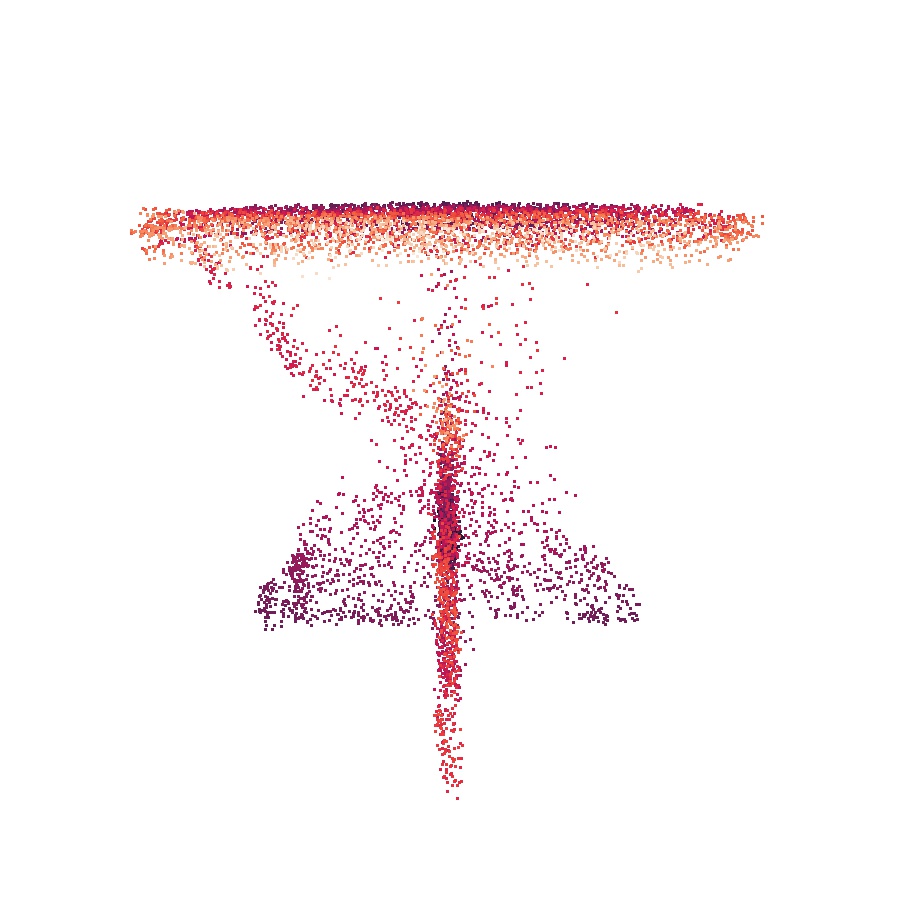}\end{subfigure} & \begin{subfigure}{0.065\textwidth}\centering\includegraphics[trim=250 120 230 100,clip,width=\textwidth]{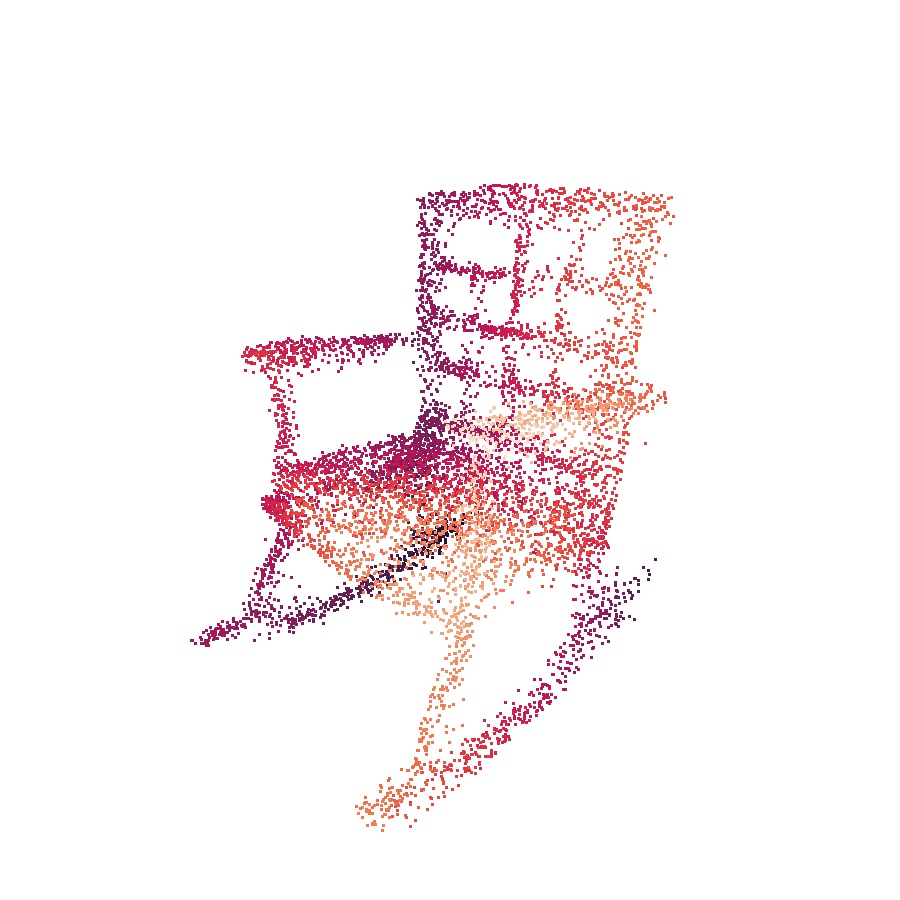}\end{subfigure} & \begin{subfigure}{0.14\textwidth}\centering\includegraphics[trim=120 250 230 190,clip,width=\textwidth]{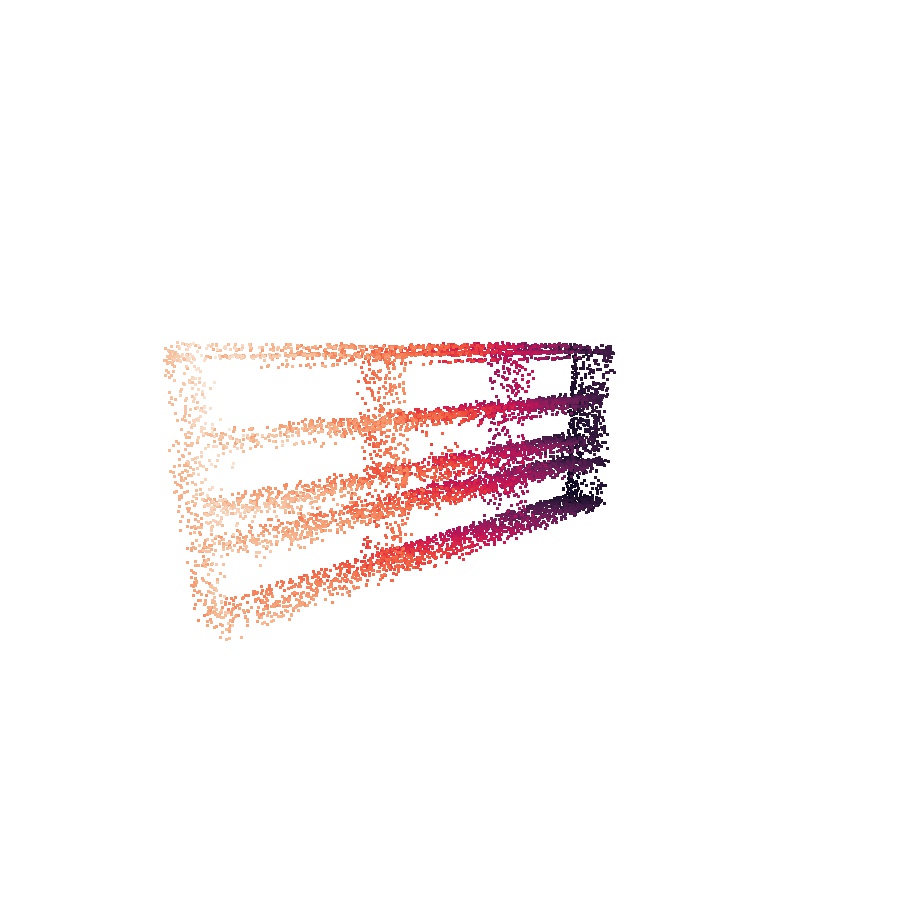}\end{subfigure} & \begin{subfigure}{0.04\textwidth}\centering\includegraphics[trim=370 220 340 190,clip,width=\textwidth]{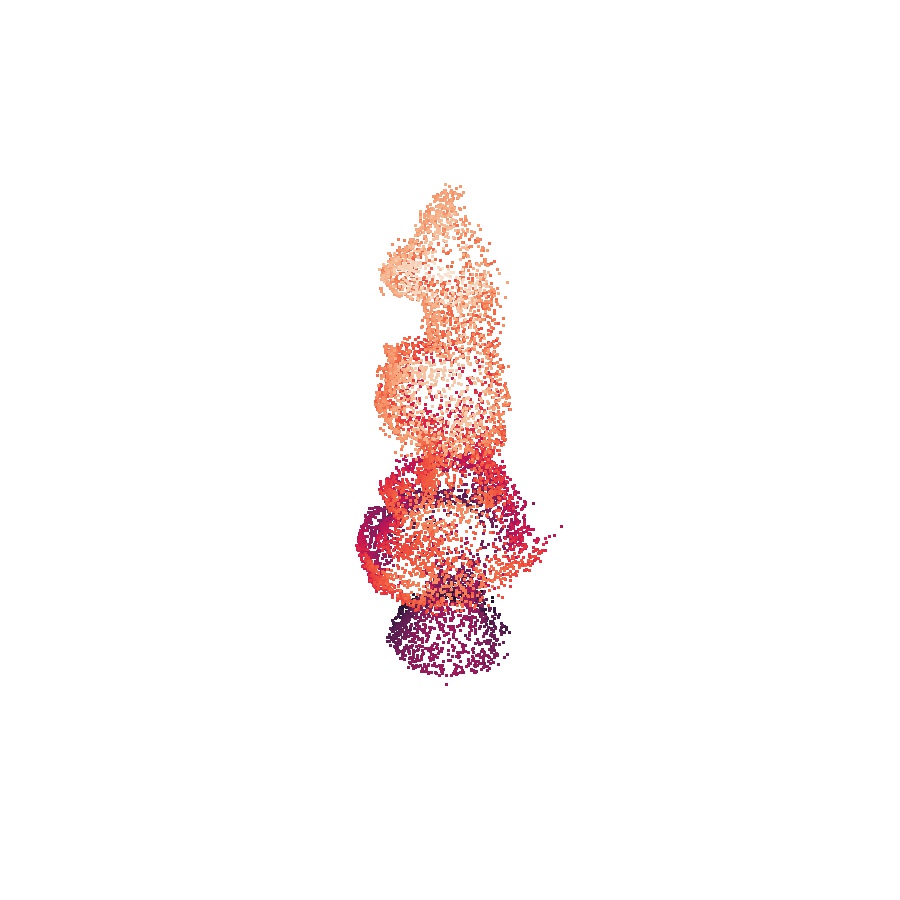}\end{subfigure} & \begin{subfigure}{0.065\textwidth}\centering\includegraphics[trim=260 260 320 160,clip,width=\textwidth]{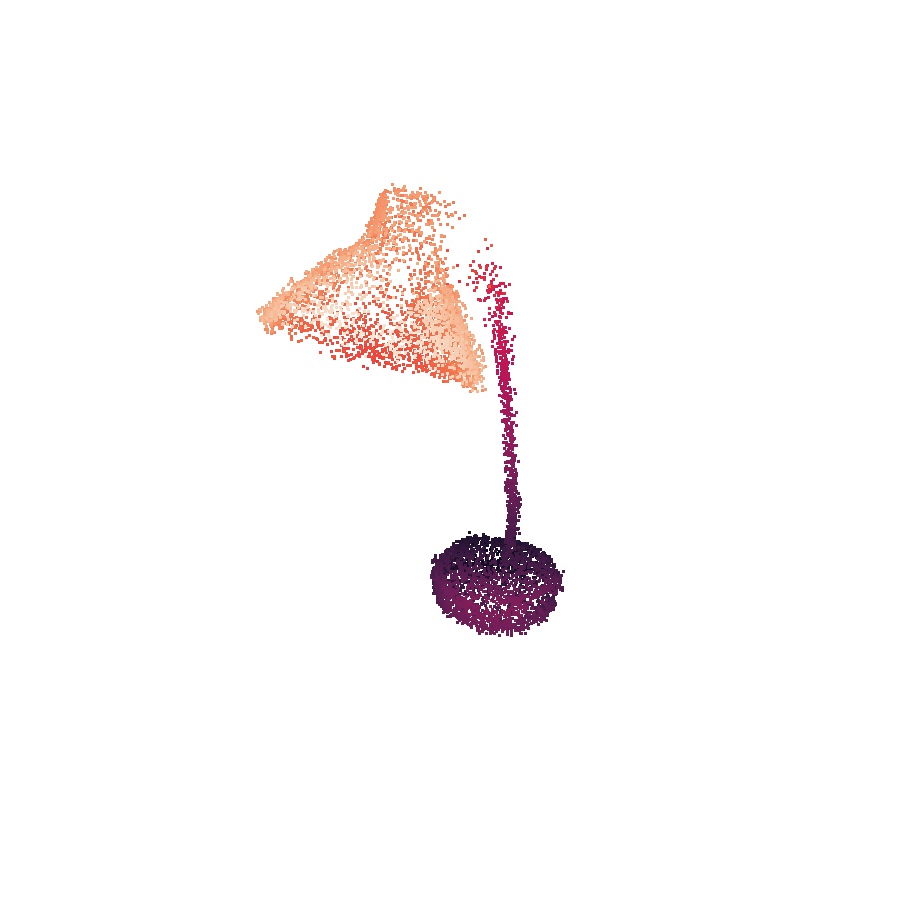}\end{subfigure} & \begin{subfigure}{0.13\textwidth}\centering\includegraphics[trim=100 200 150 180,clip,width=\textwidth]{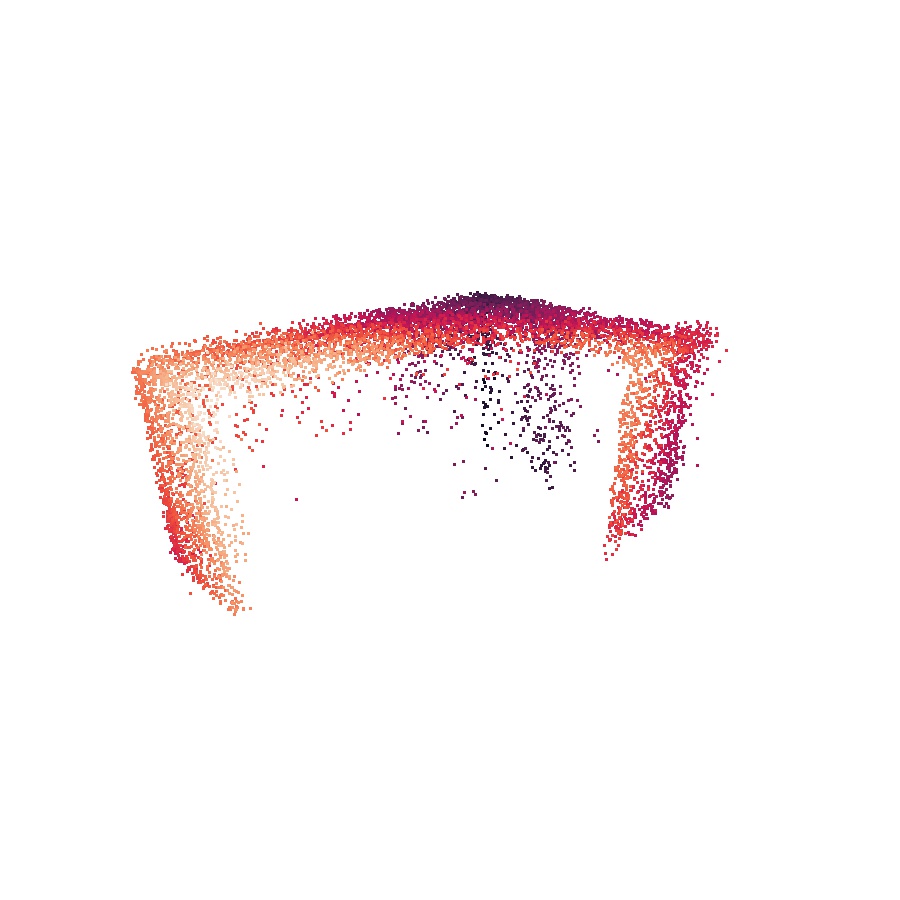}\end{subfigure} \\
Ours & \begin{subfigure}{0.065\textwidth}\centering\includegraphics[trim=300 150 250 150,clip,width=\textwidth]{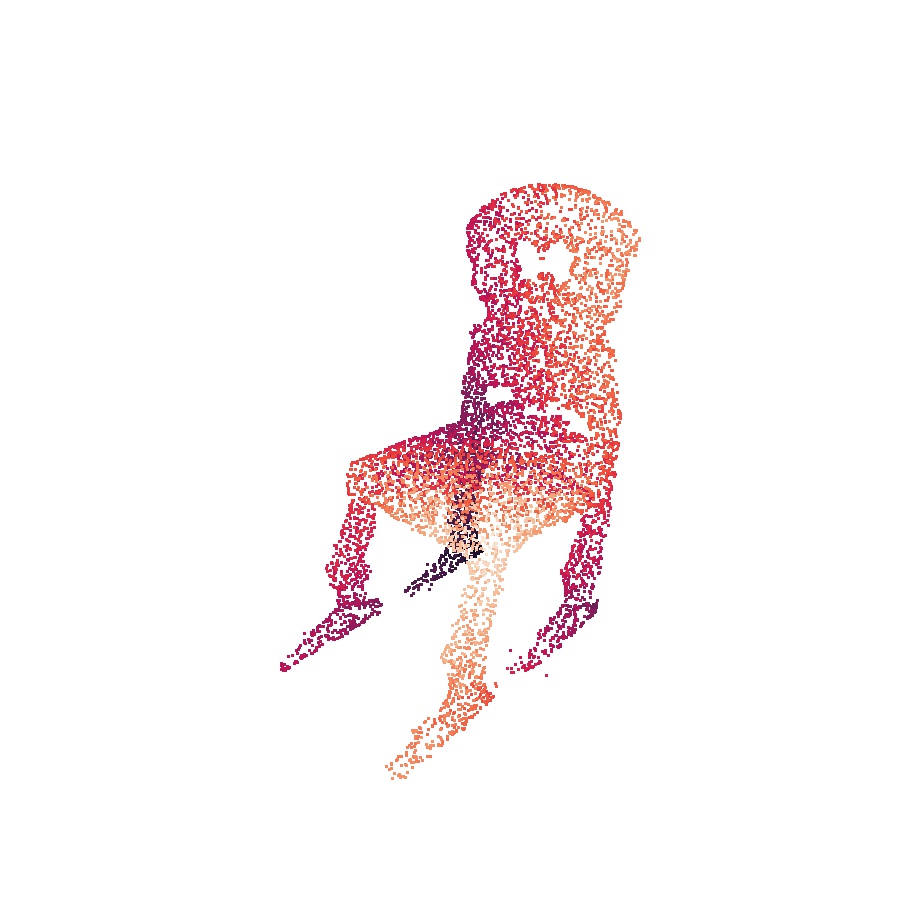}\end{subfigure} & \begin{subfigure}{0.03\textwidth}\centering\includegraphics[trim=380 230 400 190,clip,width=\textwidth]{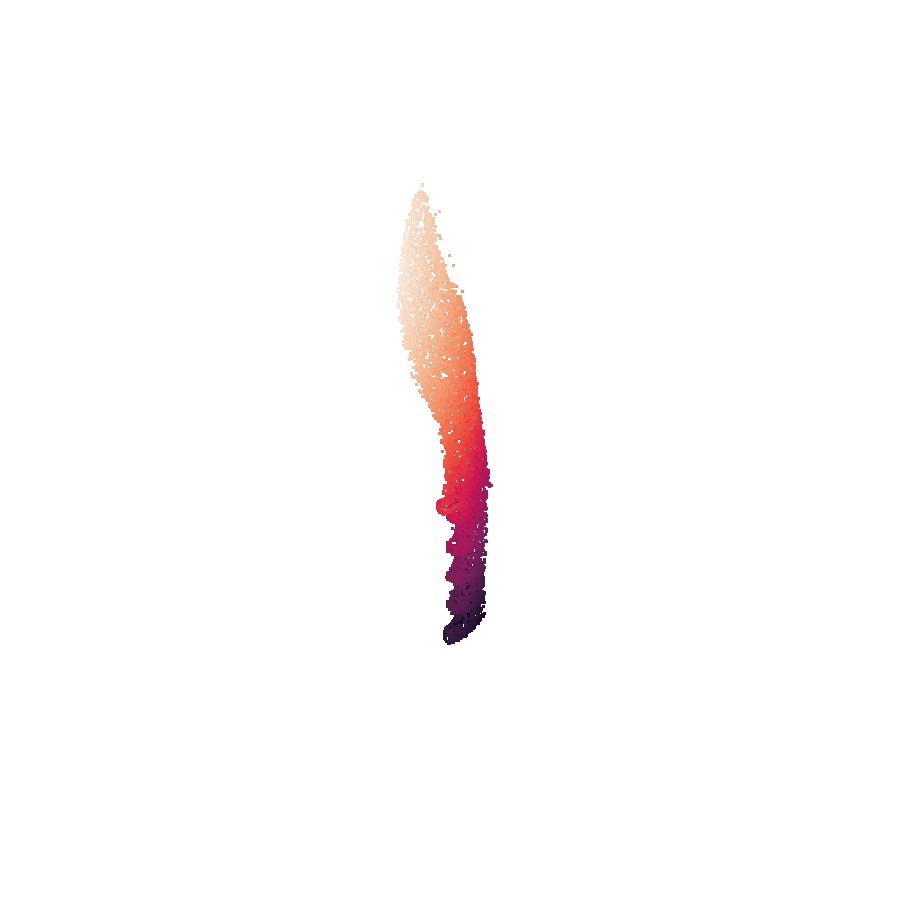}\end{subfigure} & \begin{subfigure}{0.09\textwidth}\centering\includegraphics[trim=200 170 220 100,clip,width=\textwidth]{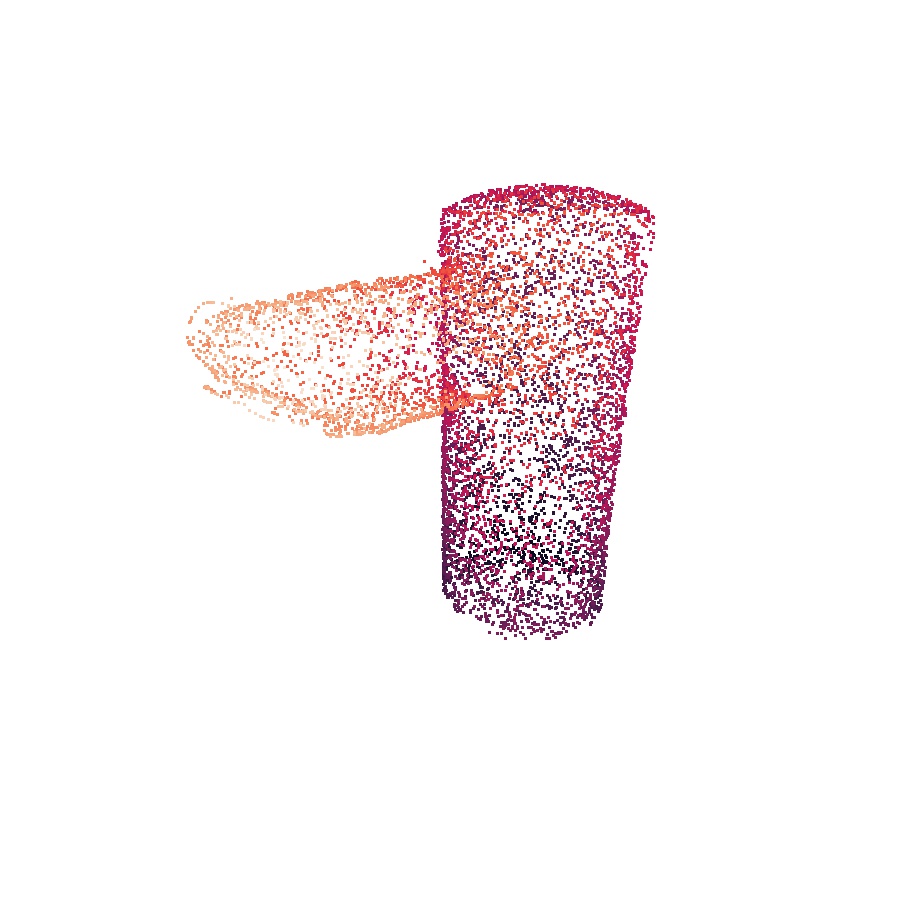}\end{subfigure} & \begin{subfigure}{0.1\textwidth}\centering\includegraphics[trim=100 170 120 110,clip,width=\textwidth]{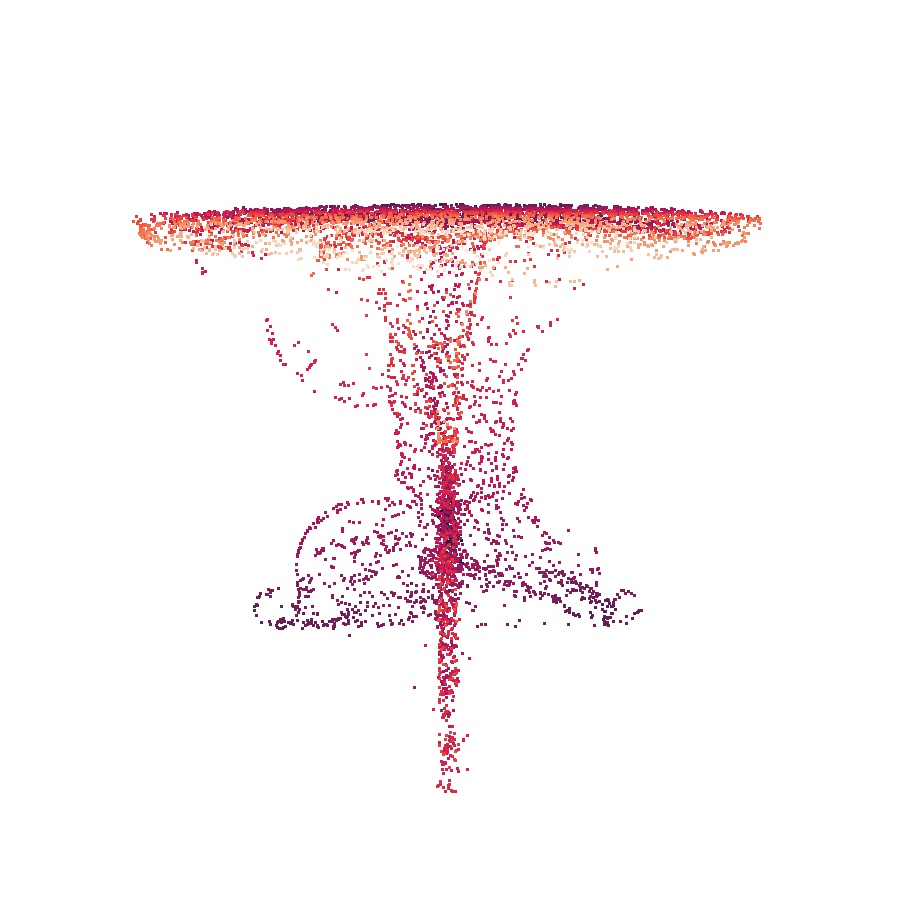}\end{subfigure} & \begin{subfigure}{0.065\textwidth}\centering\includegraphics[trim=250 120 230 100,clip,width=\textwidth]{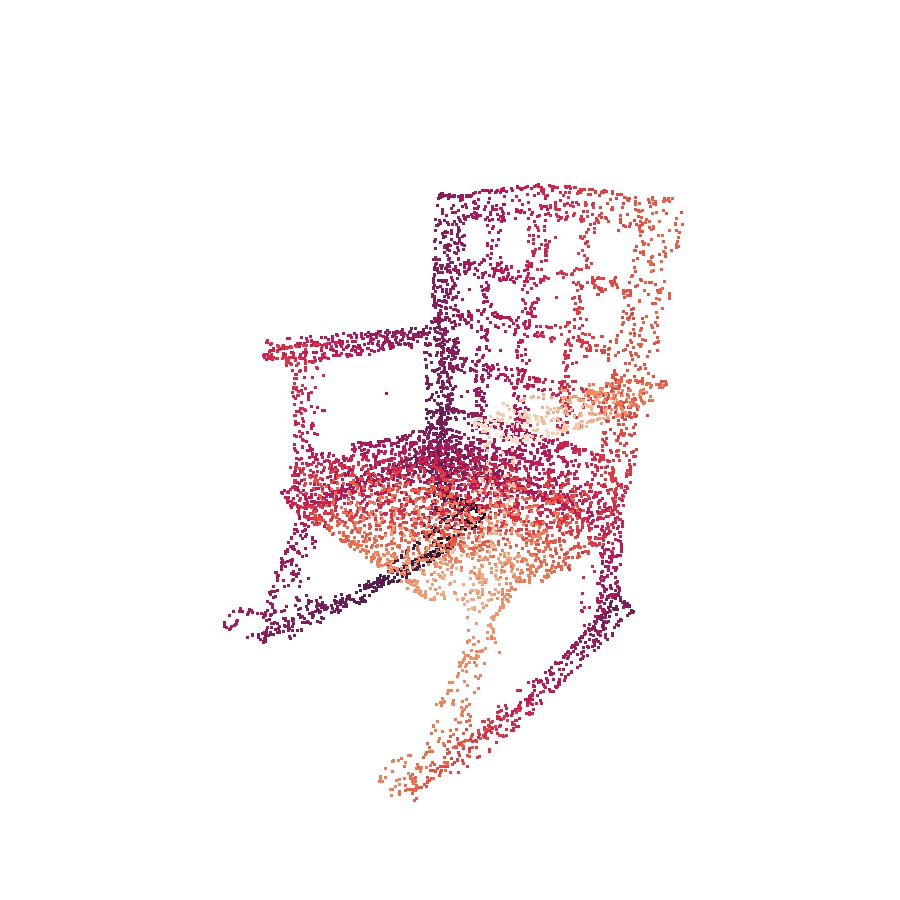}\end{subfigure} & \begin{subfigure}{0.14\textwidth}\centering\includegraphics[trim=120 250 230 190,clip,width=\textwidth]{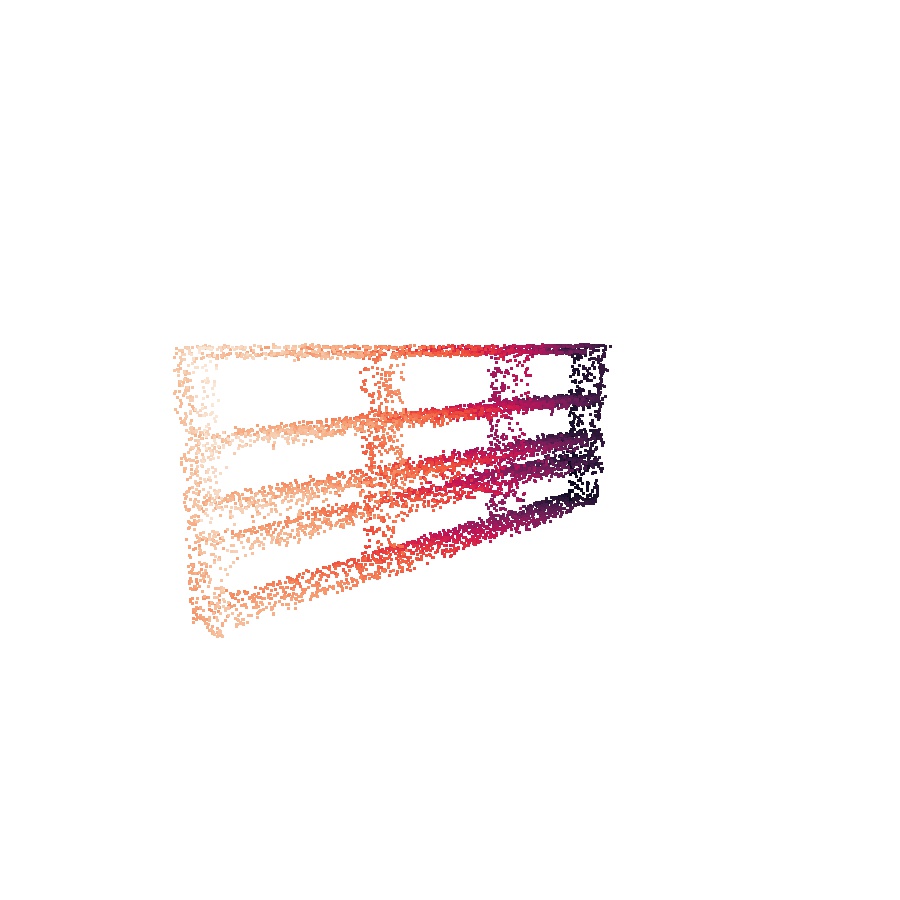}\end{subfigure}&
 \begin{subfigure}{0.04\textwidth}\centering\includegraphics[trim=370 220 340 190,clip,width=\textwidth]{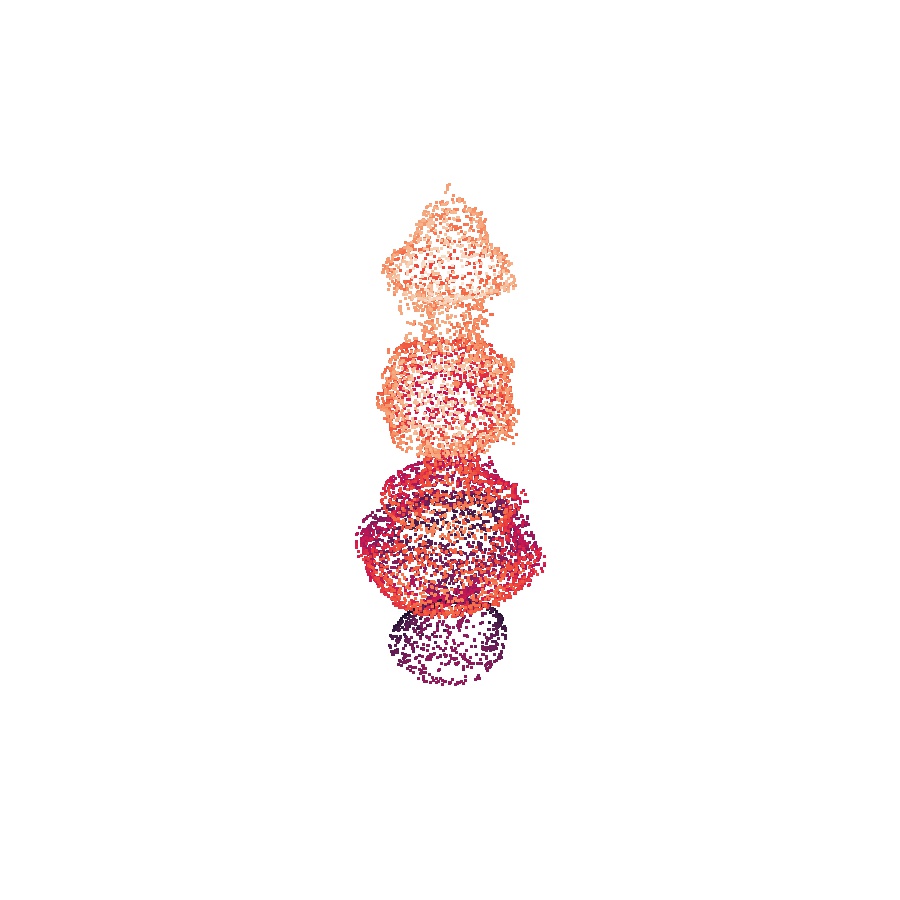}\end{subfigure} & \begin{subfigure}{0.055\textwidth}\centering\includegraphics[trim=260 240 320 160,clip,width=\textwidth]{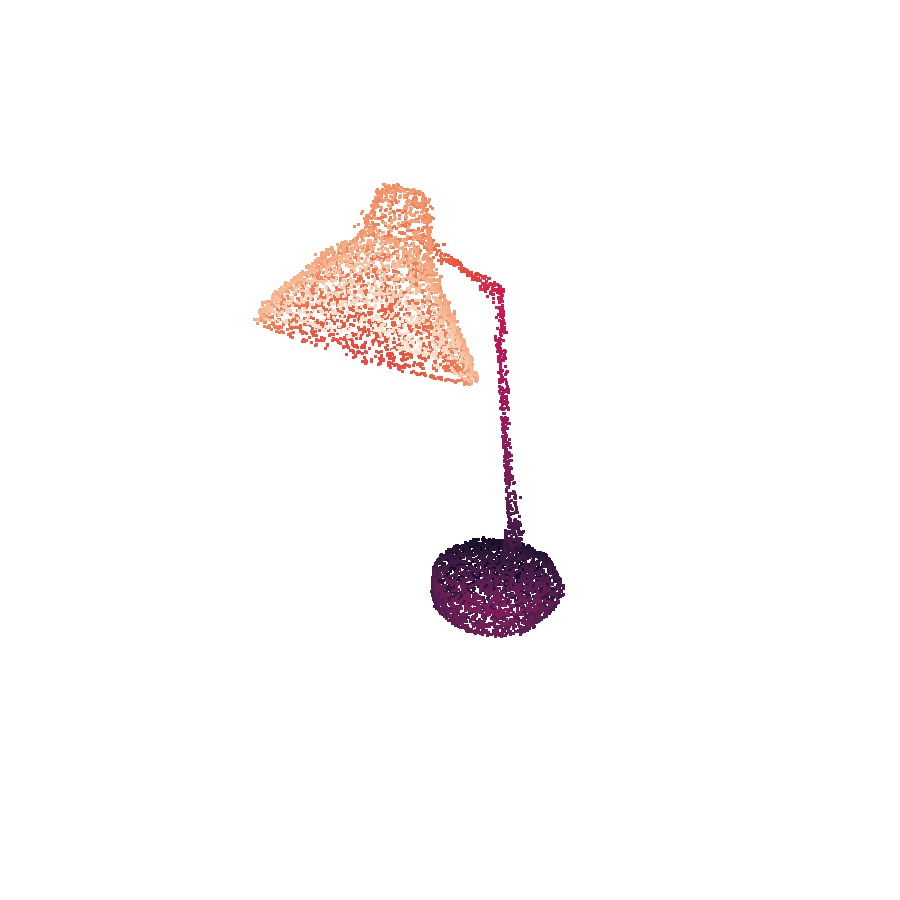}\end{subfigure} & \begin{subfigure}{0.13\textwidth}\centering\includegraphics[trim=100 200 150 190,clip,width=\textwidth]{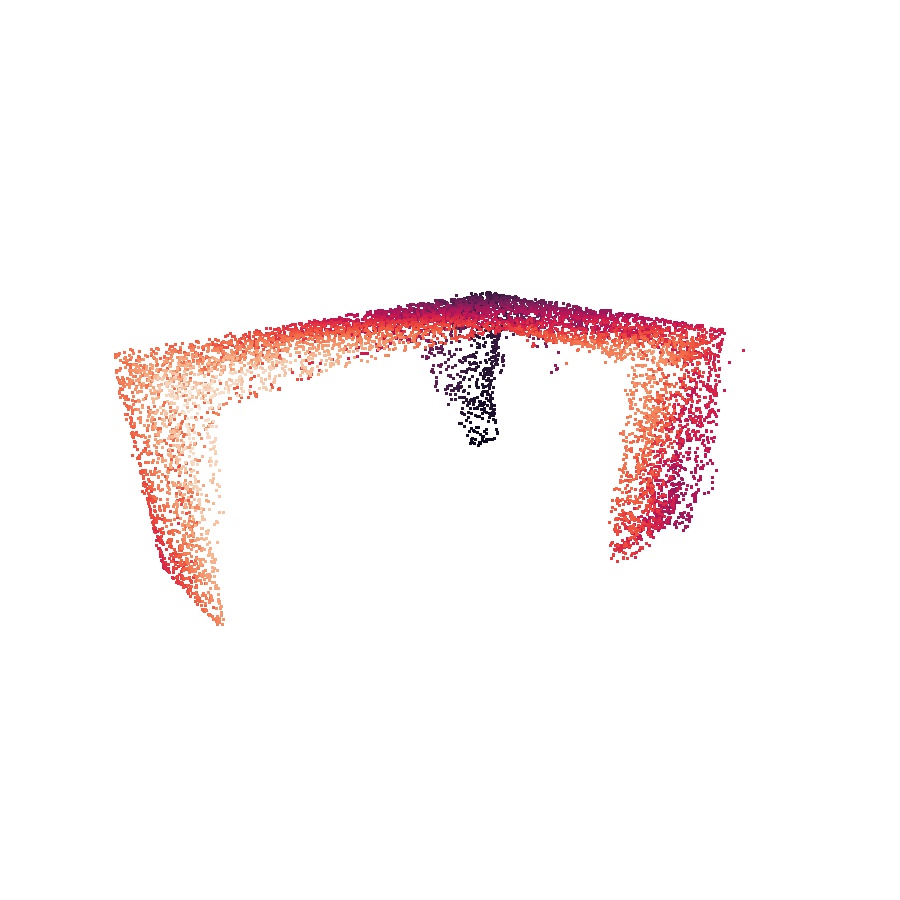}\end{subfigure} \\
GT & \begin{subfigure}{0.065\textwidth}\centering\includegraphics[trim=300 150 250 150,clip,width=\textwidth]{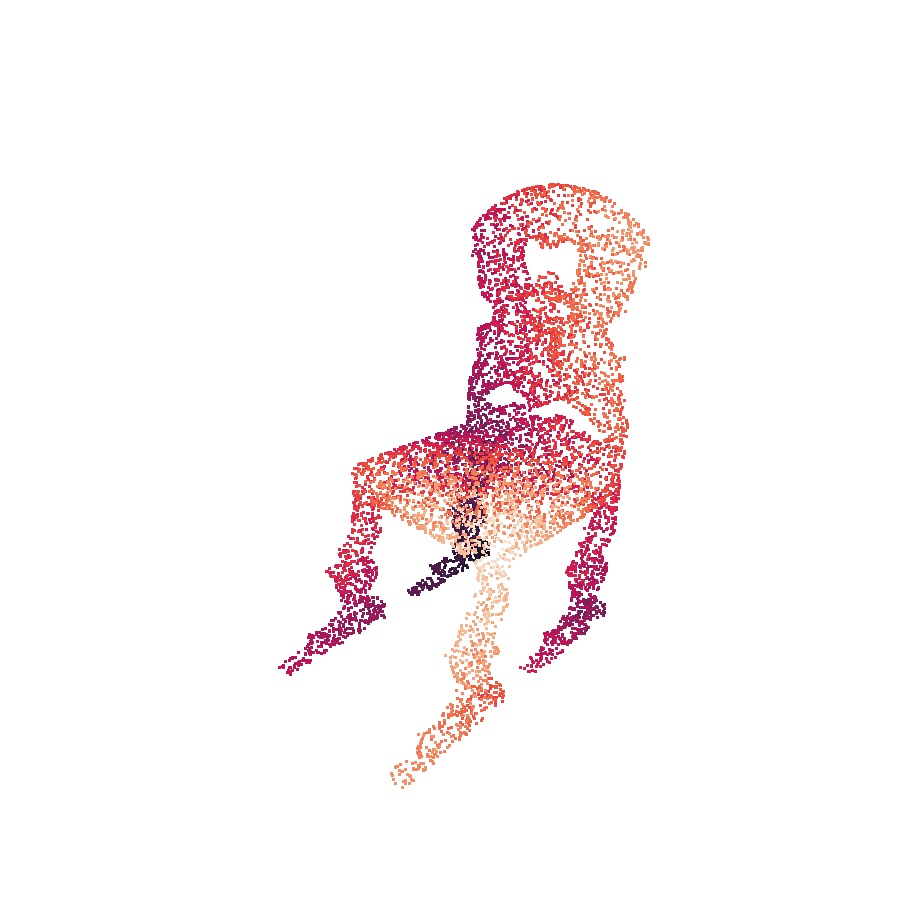}\end{subfigure} & \begin{subfigure}{0.03\textwidth}\centering\includegraphics[trim=380 230 400 190,clip,width=\textwidth]{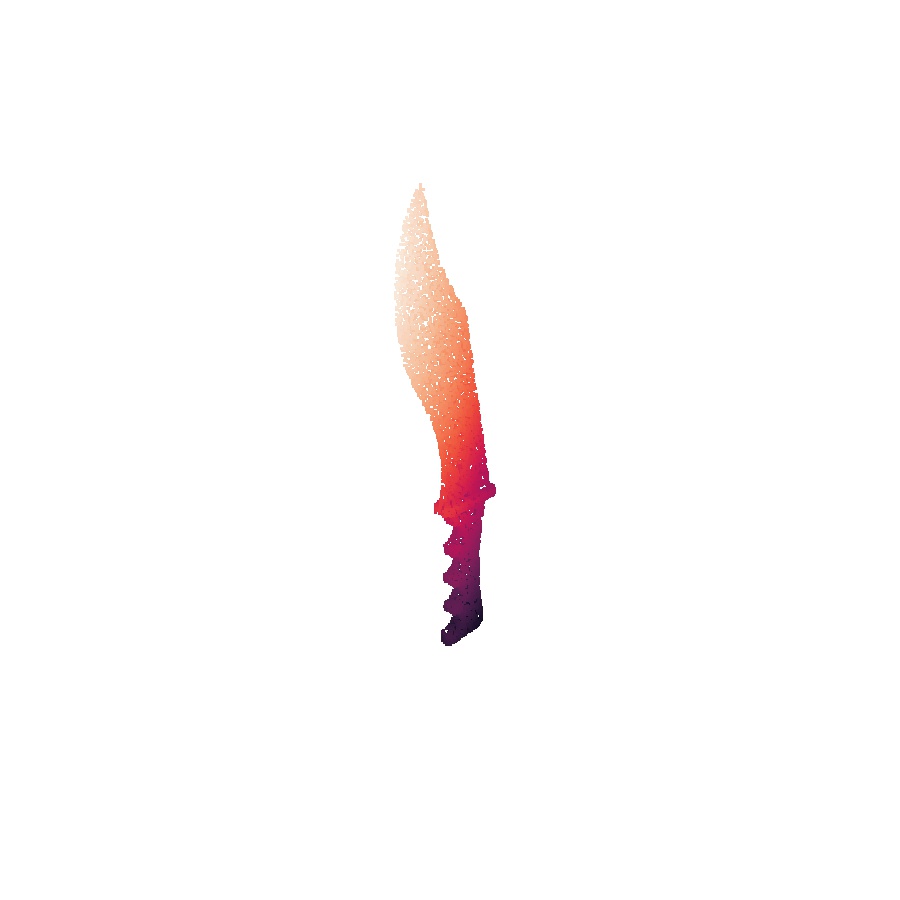}\end{subfigure} & \begin{subfigure}{0.09\textwidth}\centering\includegraphics[trim=200 170 220 100,clip,width=\textwidth]{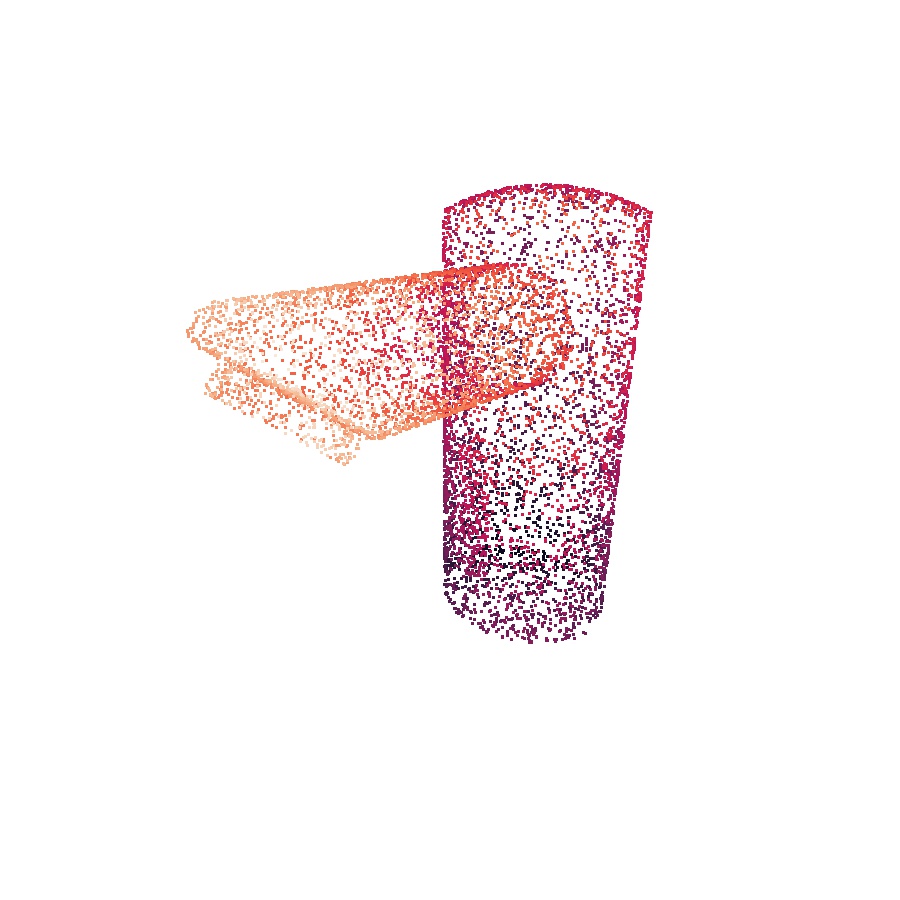}\end{subfigure} & \begin{subfigure}{0.1\textwidth}\centering\includegraphics[trim=100 170 120 110,clip,width=\textwidth]{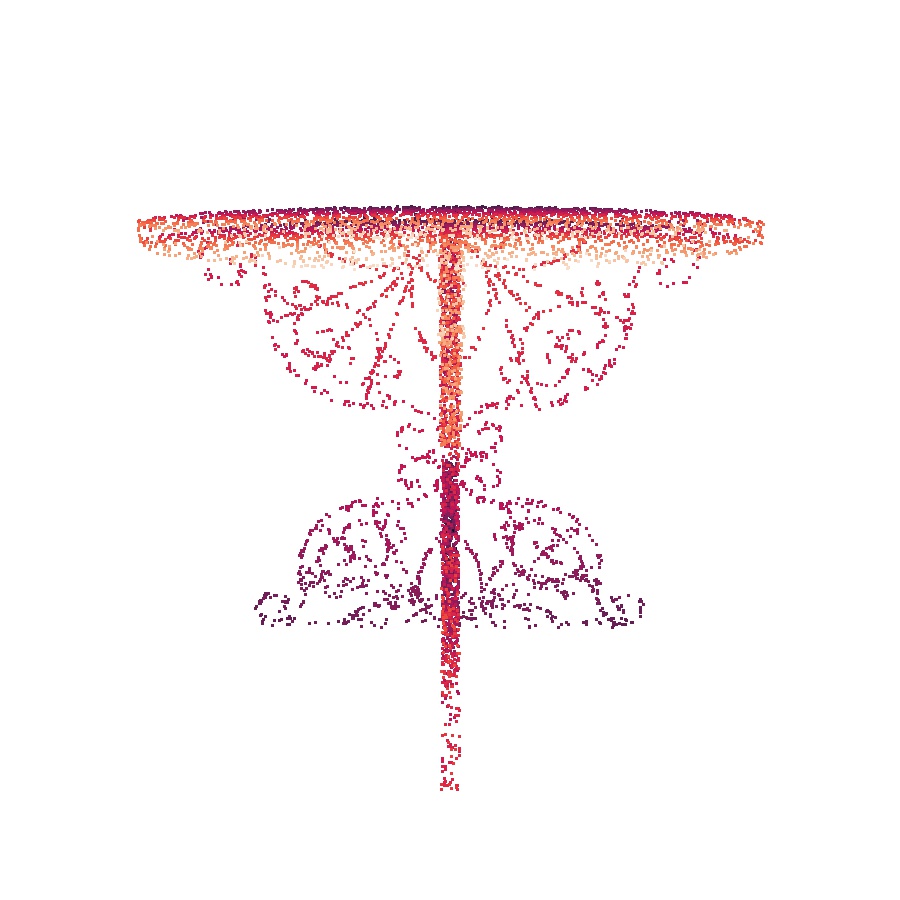}\end{subfigure} & \begin{subfigure}{0.065\textwidth}\centering\includegraphics[trim=250 120 250 100,clip,width=\textwidth]{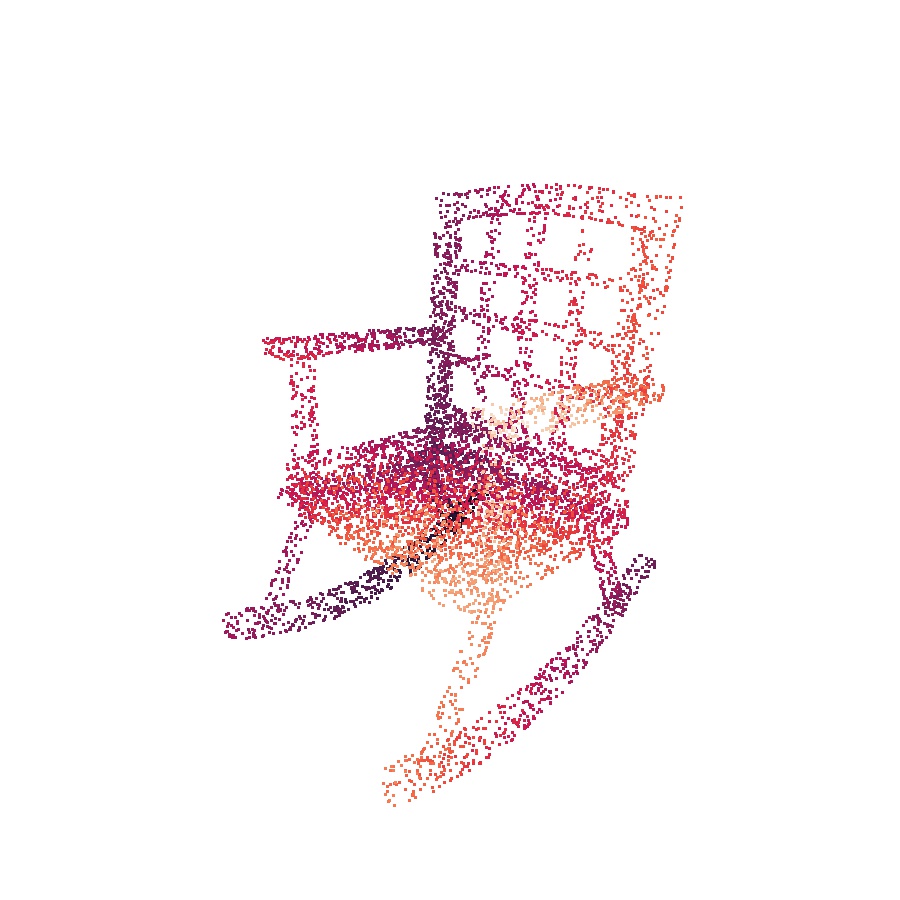}\end{subfigure} & \begin{subfigure}{0.14\textwidth}\centering\includegraphics[trim=100 250 230 190,clip,width=\textwidth]{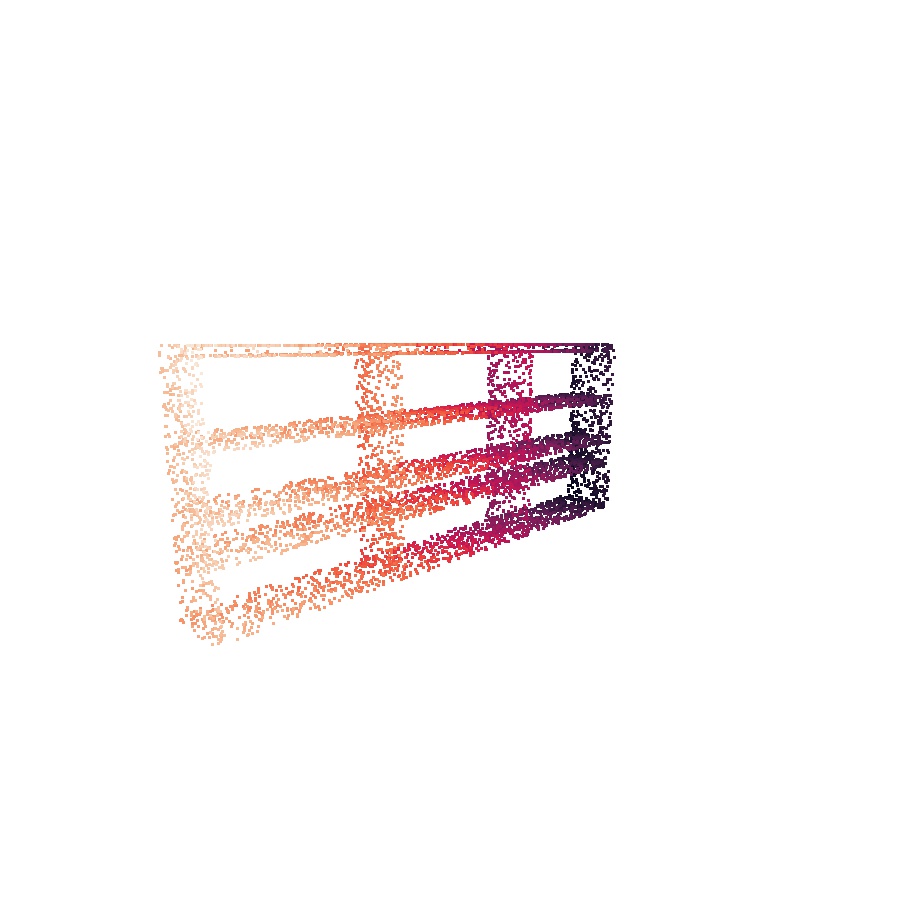}\end{subfigure}&
 \begin{subfigure}{0.04\textwidth}\centering\includegraphics[trim=370 220 340 190,clip,width=\textwidth]{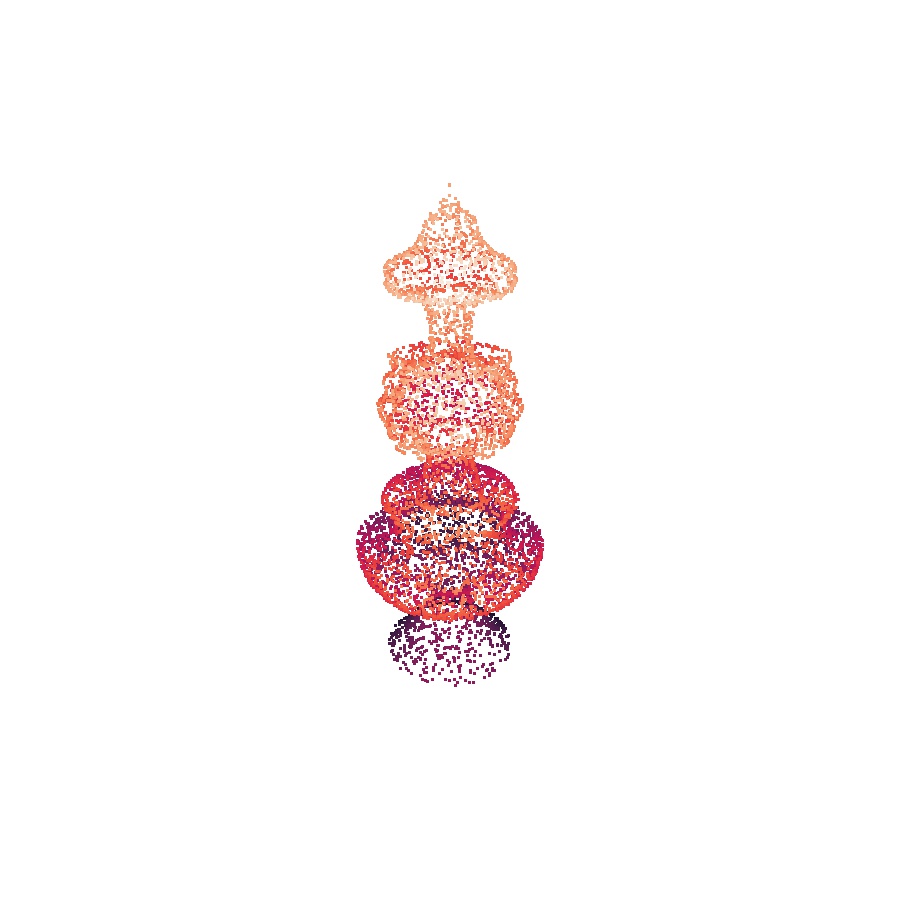}\end{subfigure} & \begin{subfigure}{0.065\textwidth}\centering\includegraphics[trim=260 260 320 160,clip,width=\textwidth]{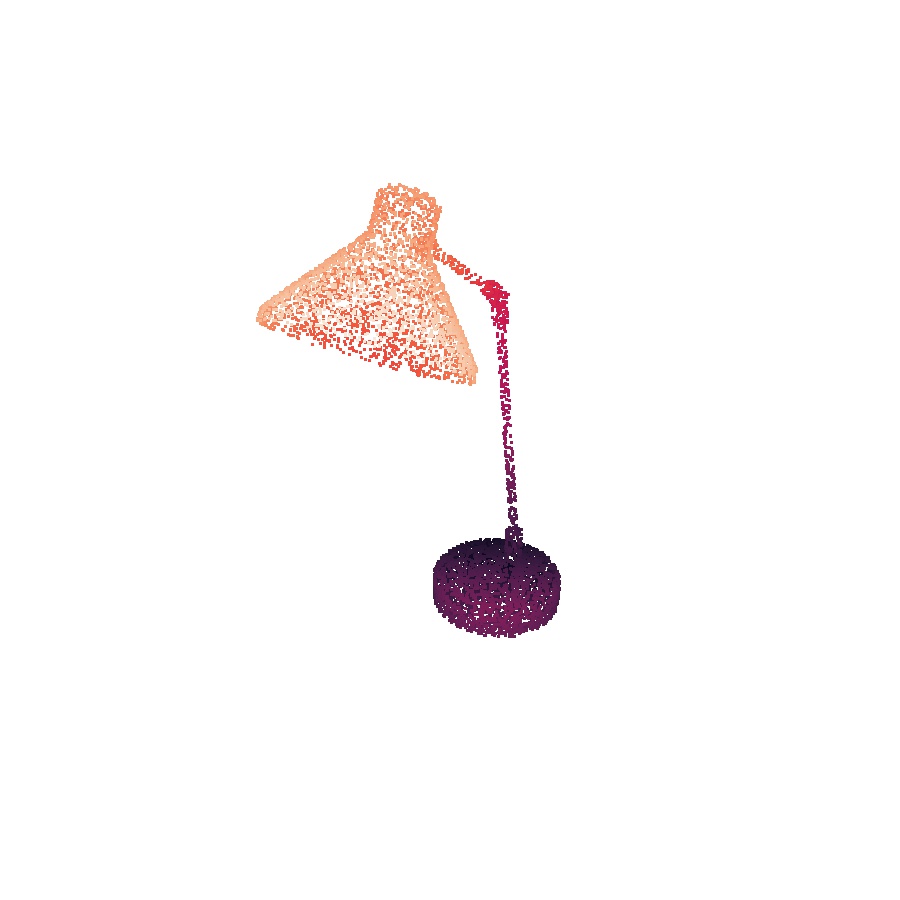}\end{subfigure} & \begin{subfigure}{0.13\textwidth}\centering\includegraphics[trim=100 250 150 190,clip,width=\textwidth]{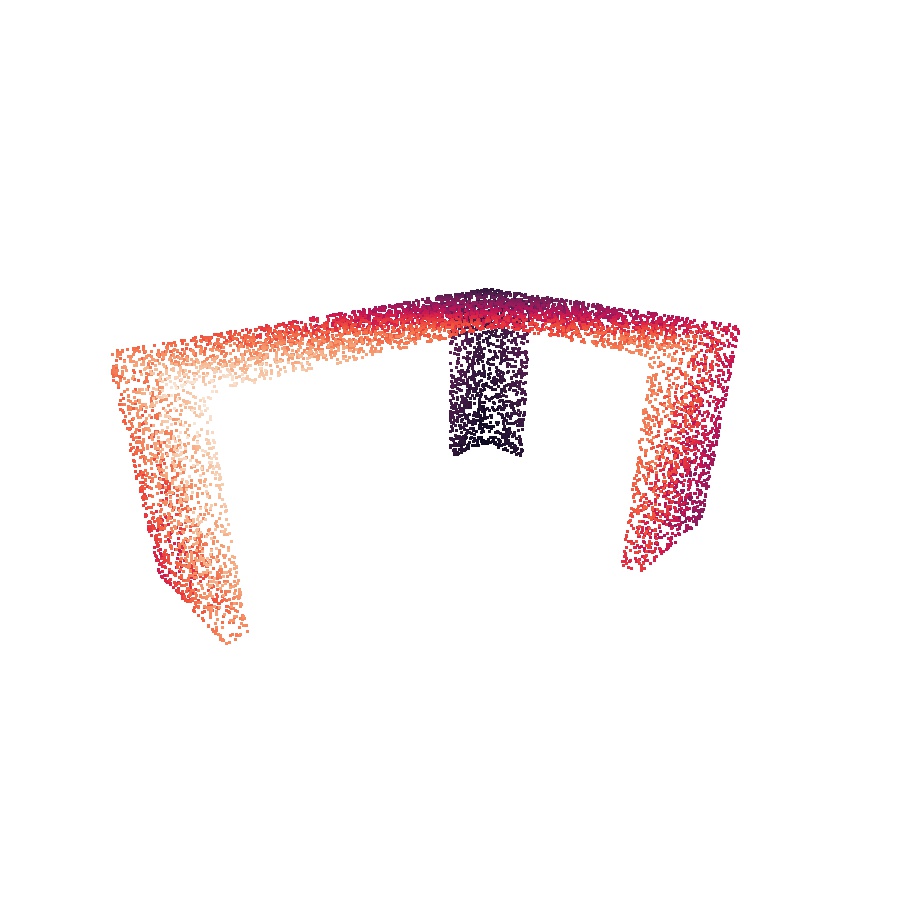}\end{subfigure} \\
\end{tabular}
}
	\caption{Comparisons of different methods on point cloud completion. Note that 8,192 points are exported from each method for comparison, except SoftPoolNet (4,096 points) due to its network specification.}
	\label{fig:sota_fig_big_supp}
\end{figure*}

We list more qualitative comparisons with previous methods in Figure~\ref{fig:sota_fig_big_supp}.


\section{More Quantitative Comparisons}
\label{sec:quantitative}

\subsection{Comparison with Existing Methods}
In this section, we report our quantitative evaluation on all 14 categories:
faucet, cabinet, table, chair, vase, lamp, bottle, clock, display, knife, mug, fridge, scissors and trashcan. Some methods (PCN~\cite{yuan2018pcn}, CRN~\cite{wang2020cascaded}, GR-Net~\cite{xie2020grnet}, and MSN~\cite{liu2020morphing}) support upsampling to recover higher resolution of outputs, we compare our methods with them under the resolution of 8,192. We report the comparisons using EMD \cite{liu2020morphing} and CD scores \cite{fan2017point} in Table~\ref{tab:sota-table-emd-8k-supp} and Table~\ref{tab:sota-table-cd-8k-supp} respectively. For the other methods that do not support upsampling (including PF-Net~\cite{huang2020pf}, P2P-Net\cite{yin2018p2p}, SoftPoolNet\cite{wang2020softpoolnet}), we downsample the output of all the methods to the resolution of 2,048 to enable a fair comparison. The quantitative scores on EMD and CD are respectively listed in Table~\ref{tab:sota-table-emd-2k-supp} and Table~\ref{tab:sota-table-cd-2k-supp}.


\subsection{More Ablation Studies on Ray Encoding}

In our network, empty rays are feed into network together with real/coarse points. The most significant difference between rays and points is that each ray has a directional vector and an offset vector to a neighboring point. In encoder-decoder stage, the input rays are:

\begin{equation}
	\mathcal{R}^{*}_{ray}=\{\{p^{e}_{k}\}, \{D_{k}\},\{v_{k}\}\}\in\mathbf{R}^{M\times K\times9}
\end{equation}

While in the refining stage, we revisit the emptiness information by two set of rays:

\begin{align}
	\mathcal{R}^{d}_{ray1}&=\{\{p^{e}_{k}\}, \{D_{k}\},\{v_{k}\}\}\in\mathbf{R}^{N_{c}\times K\times9}.
 \\
	\mathcal{R}^{d}_{ray2}&=\{\{p^{c,e}_{k}\}, \{D^{c}_{k}\},\{\mathbf{r}^{c}\}\} \in \mathbf{R}^{N_{c}\times K\times 15}
	\label{eq:sampled_rays_supp}
\end{align}

We design an ablation study to explore the effectiveness of directional vector and offset vector in ray representation. In the ablated version, we remove $D_{k}$, $v_{k}$ and $\mathbf{r}^{c}$ in ray representation. As a result, the rays inputted to our network only contains $p^{e}_{k}$ or $p^{c,e}_{k}$. The quantitative results is shown in Table \ref{tab:ablation-table} as `\textbf{only pts in rays}'. The results demonstrate that $D_{k}$, $v_{k}$ and $\mathbf{r}^{c}$ are significant for our method to learn emptiness.

In our paper, the first $\mathcal{R}^{d}_{ray1}$ informs our decoder with the emptiness information, telling our decoder \textit{`whether the coarse points are in empty
regions'.} The second $\mathcal{R}^{d}_{ray2}$ informs our decoder with
the shape information, telling our decoder \textit{`what the real surface looks like'}. To exam the necessity of the two rays in the refining stage, we design two additional ablation studies: `\textbf{remove $\mathcal{R}^{d}_{ray2}$}' and `\textbf{remove $\mathcal{R}^{d}_{ray1}$}'. The first one means the local feature vector is computed only from $\mathcal{R}^{d}_{ray1}$. Similarly, the second one means the local feature vector is computed only from $\mathcal{R}^{d}_{ray2}$. The quantitative results are shown in Table \ref{tab:ablation-table}.  The results prove that both $\mathcal{R}^{d}_{ray1}$ and $\mathcal{R}^{d}_{ray2}$ are important in refining stage.

{\small
	\bibliographystyle{ieee_fullname}
	\bibliography{egbib}
}

\end{document}